\documentclass{article}
\usepackage{amsfonts,amsmath,amsthm,amssymb}

\date{}

\usepackage{algorithm}
\usepackage{algorithmic}
\usepackage{subfigure}
\usepackage{epsfig}
\usepackage{graphicx}
\usepackage{url}
\usepackage{multirow}
\usepackage{amssymb}
\usepackage{mathrsfs}
\usepackage{epstopdf}
\usepackage{multicol}
\usepackage{amsmath}
\usepackage{mathtools}
\usepackage{xcolor}

\usepackage{multicol}
\usepackage{wrapfig}
\usepackage{wrapfig,lipsum,booktabs}

\usepackage{multirow}

\usepackage{makecell}

\usepackage{pifont}

\allowdisplaybreaks[4]

\newtheorem{theorem}{\textbf{Theorem}}
\newtheorem{assumption}{\textbf{Assumption}}
\newtheorem{lemma}{\textbf{Lemma}}
\newtheorem{corollary}{\textbf{Corollary}}
\newtheorem{remark}{\textbf{Remark}}

\numberwithin{equation}{section}

\theoremstyle{plain}

\theoremstyle{definition}

\title{On the Communication Complexity of Decentralized Stochastic Bilevel Optimization}

\author{
	Yihan Zhang \and My T. Thai \and Jie Wu \and 
	Hongchang Gao\thanks{Temple University, {\tt hongchang.gao@temple.edu}} 
	}
\begin{document}
\maketitle

\begin{abstract}
Stochastic bilevel optimization finds widespread applications in machine learning, including meta-learning, hyperparameter optimization, and neural architecture search. To extend stochastic bilevel optimization to distributed data, several decentralized stochastic bilevel optimization algorithms have been developed. However, existing methods often suffer from slow convergence rates and high communication costs in heterogeneous settings, limiting their applicability to real-world tasks. To address these issues, we propose two novel decentralized stochastic bilevel gradient descent algorithms based on \textit{simultaneous} and \textit{alternating} update strategies. Our algorithms can achieve faster convergence rates and lower communication costs than existing methods. Importantly, our convergence analyses do not rely on strong assumptions regarding heterogeneity. More importantly, our theoretical analyses clearly disclose how the computation and communication regarding the Hessian-inverse-vector product under the heterogeneous setting affects the convergence rate.  To the best of our knowledge, this is the first time such favorable theoretical results have been achieved with mild assumptions in the heterogeneous setting. Furthermore, we demonstrate how to establish the convergence rate for the alternating update strategy when combined with the variance-reduced gradient. Finally, experimental results confirm the efficacy of our algorithms.
\end{abstract}

\section{Introduction}
Bilevel optimization has been  used in a wide range of machine learning models. For instance, the hyperparameter optimization \cite{feurer2019hyperparameter,franceschi2017forward}, meta-learning \cite{franceschi2018bilevel,rajeswaran2019meta},  and neural architecture search \cite{liu2018darts} can be formulated as a bilevel optimization problem. Considering its importance in machine learning, bilevel optimization has been attracting significant attention in recent years. Particularly,  to handle the distributed data, parallel bilevel optimization has been actively studied in the past few years. In this paper, we are interested in an important class of parallel bilevel optimization: decentralized bilevel optimization, where all workers perform peer-to-peer communication to collaboratively optimize a bilevel optimization problem. Specifically, the loss function is defined as follows:
\begin{equation} \label{loss_bilevel}
	\begin{aligned}
		& \min_{x\in \mathbb{R}^{d_x}} \frac{1}{K}\sum_{k=1}^{K} f^{(k)}(x, y^*(x)), 
		& s.t. \quad   \ y^*(x) =\arg\min_{y\in \mathbb{R}^{d_y}} \frac{1}{K}\sum_{k=1}^{K} g^{(k)}(x, y) \ , 
	\end{aligned}
\end{equation}
where $k$ is the index of workers,  $g^{(k)}(x, y)=\mathbb{E}_{\zeta\sim \mathcal{D}_g^{(k)} }[g^{(k)}(x, y; \zeta)]$ is the lower-level loss function of the $k$-th worker, $f^{(k)}(x, y)=\mathbb{E}_{\xi\sim \mathcal{D}_f^{(k)}}[f^{(k)}(x, y; \xi)]$  is the upper-level loss function of the $k$-th worker. Here, $\mathcal{D}_g^{(k)}$ and $\mathcal{D}_f^{(k)}$ denote the data distributions of the $k$-th worker. Throughout this paper, it is assumed that different workers have different data distributions. 

In the past few years, a series of decentralized optimization algorithms have been proposed to solve Eq.~(\ref{loss_bilevel}). For instance, \cite{gao2023convergence} developed two decentralized  stochastic bilevel gradient descent  algorithms based on the momentum and variance reduction techniques: MDBO and VRDBO. Specifically, \cite{gao2023convergence}  demonstrates that the variance-reduction based algorithm VRDBO is able to achieve the $O\left(\frac{1}{K\epsilon^{3/2}}\log\frac{1}{\epsilon}\right)$ convergence rate.  More specifically,   it is a double-loop algorithm, where there exists an inner loop to compute the Hessian inverse matrix for estimating stochastic hypergradient and the length of the inner loop is on the order of $O\left(\log \frac{1}{\epsilon}\right)$.  \cite{gao2023convergence} shows that the inner loop does not require communication under the homogeneous setting. As such,  its communication complexity \footnote{In the introduction, we ignore the spectral gap and the communication cost in each communication round to make it clear. The detailed communication complexity can be found in Table~\ref{comparison}.} is on the order of $O\left(\frac{1}{K\epsilon^{3/2}}\right)$. 

However,  it is more challenging to solve Eq.~(\ref{loss_bilevel}) under the heterogeneous setting.  Specifically, unlike the homogeneous setting where the local Jacobian and Hessian matrices are the same as the global ones, each worker under the heterogeneous setting has to pay extra efforts to estimate the  global  Jacobian and Hessian matrices to compute the hypergradient, which causes the following unique challenges for   algorithmic design and theoretical analysis.

\textbf{Heterogeneity causes challenges for communication.}
Under the heterogeneous setting,  each worker has to estimate the global Jacobian and Hessian matrices. This can incur a large communication cost, e.g., \textit{a large number of communication rounds} and \textit{a high communication cost per round}.  
For instance, existing methods \cite{chen2022decentralized,chen2022decentralized2,yang2022decentralized} suffer from a large number of communication rounds.  Specifically, \cite{chen2022decentralized} developed a decentralized stochastic bilevel gradient descent algorithm: DSBO. It is  a double-loop algorithm as VRDBO. However, DSBO requires communication in the inner loop. As such, the number of communication rounds is as large as its iteration complexity (i.e., convergence rate) $O\left(\frac{1}{\epsilon^{2}}\log \frac{1}{\epsilon}\right)$.  \cite{yang2022decentralized} proposed a decentralized  stochastic bilevel gradient descent with momentum algorithm: Gossip-DSBO, which is also a double-loop algorithm and requires communication in the inner loop. The number of communication rounds and the iteration complexity can be slightly improved to $O\left(\frac{1}{K\epsilon^{2}}\log\frac{1}{\epsilon}\right)$.   \cite{chen2022decentralized2} proposed MA-DSBO, which is also a double-loop algorithm and shares the same  number of communication rounds $O\left(\frac{1}{\epsilon^{2}}\log \frac{1}{\epsilon}\right)$ with DSBO.  Moreover, most existing methods suffer from a high communication cost per communication round. 
Specifically, Gossip-DSBO and DSBO  require to communicate the  Hessian matrix or Jacobian matrix, which incurs a high communication cost  $O(d_y^2)$ or $O(d_xd_y)$ in each  round. As a result, the total communication complexity under the heterogeneous setting  is much larger than the homogeneous setting.

It can be observed that  all the aforementioned existing algorithms under the heterogeneous setting suffer from higher communication complexity compared to VRDBO under the homogeneous setting.  This observation naturally leads to the first question: \textit{Can we have a decentralized stochastic bilevel optimization algorithm under the heterogeneous setting to enjoy both a smaller number of communication rounds and a low communication cost per communication round?}  Therefore, the first goal of this paper is to develop a communication-efficient decentralized stochastic bilevel optimization algorithm to tackle this challenge in algorithmic design.

\textbf{Heterogeneity causes challenges for convergence analysis.} 
When establishing the theoretical convergence rate, the aforementioned existing methods \cite{chen2022decentralized,yang2022decentralized,chen2022decentralized2} require strong assumptions to bound the heterogeneity. Specifically, DSBO \cite{chen2022decentralized} requires a Lipschitz continuous upper-level loss function regarding $x$, i.e., $\|\nabla_{1} f^{(k)}(x, y)\|\leq c_{f_x}$, where $c_{f_x}>0$ is a constant.  Gossip-DSBO \cite{yang2022decentralized} also requires this assumption. Meanwhile, it requires that the lower-level loss function is Lipschitz continuous with respect to $y$, i.e., $\|\nabla_{2} g^{(k)}(x, y)\|\leq c_{g_y}$,  where $c_{g_y}>0$ is a constant.  With these strong assumptions, it is easy to bound the heterogeneity when establishing the convergence rate. However, it is worth noting that some of these strong assumptions might not hold. MA-DSBO  \cite{chen2022decentralized2} does not require these strong assumptions at the cost of introducing the bounded heterogeneity assumption, i.e., $\|\nabla_{2} g^{(k)}(x, y)- \nabla_{2} g(x, y)\|\leq \delta$, where $\delta>0$ is a constant.  However, this assumption is also too strong to hold in practice (See Remark~\ref{remark-2}). 

Moreover,  communication under the heterogeneity setting introduces new challenges for convergence analysis. As discussed previously,  additional communication is required to estimate the hypergradient compared to the homogeneous setting. Then, it is important to disclose how this additional communication operation affects the convergence rate. However, existing methods, including DSBO and MA-DSBO, fail to address this aspect, while the convergence rate of Gossip-DSBO is problematic due to  contradictory  assumptions (See Remark~\ref{remark-1}).

Then, the second natural question arises:  \textit{Can we establish the convergence rate of a communication-efficient decentralized stochastic bilevel optimization algorithm without strong assumptions and disclose how the additional communication operation affects the convergence rate under the heterogeneous setting?} Hence, our second goal is to establish the convergence rate under mild assumptions to address these theoretical analysis challenges. 

\begin{table*}[t] 
	\small
	\setlength{\tabcolsep}{1pt}
	\begin{center}
		\caption{ The comparison of the communication complexity between different algorithms under the homogeneous (IID) and heterogeneous (Non-IID) settings.  \textbf{R/IT} denotes the number of communication rounds in each iteration. \textbf{Cost/R} means the communication cost in each round. \textbf{IT} represents the iteration complexity. \textbf{Communication} denotes the communication complexity.    \#a: DSBO and MA-DSBO fail to provide the dependence on the spectral gap.   \#b: Gossip-DSBO assumes all gradients are upper bounded so that it eliminates the dependence on the spectral gap. \#c: $\|\nabla_1 f^{(k)}\|\leq c_{f_x}$.  \#d: $\|\nabla_2 g^{(k)}\|\leq c_{g_y}$. \#e: $\|\nabla_2 g^{(k)}-\nabla_2 g\|\leq \delta$. 
			The limitations of the heterogeneity assumption of DSBO, Gossip-DSBO, and MA-DSBO are discussed in Remark~\ref{remark-1} and Remark~\ref{remark-2}.}
		\begin{tabular}{l|c|c|c|c|c}
			\toprule
			{\textbf{Algorithms}} &  \textbf{R/IT}   & \textbf{ Cost/R} & \textbf{IT} & \textbf{Communication}  & \textbf{Heterogeneity}\\
			\hline
			\makecell{MDBO \\ \cite{gao2023convergence}}  & $O(1)$ & $O(d_x+d_y)$& $O\left(\frac{1}{\epsilon^2(1-\lambda)^2}\right)$ &   $O\left(\frac{d_x+d_y}{\epsilon^2(1-\lambda)^2}\right)$  & i.i.d \\
			\hline
			\makecell{VRDBO \\ \cite{gao2023convergence}}  & $O(1)$  & $O(d_x +d_y)$ &  $O\left(\frac{1}{K\epsilon^{3/2}(1-\lambda)^2}\right)$ &     $O\left(\frac{d_x+d_y}{K\epsilon^{3/2}(1-\lambda)^2}\right)$  & i.i.d \\
			\hline
			\makecell{DSBO \\ \cite{chen2022decentralized}}  & $O\left(\log\frac{1}{\epsilon} \right)$  & $O\left(d_xd_y\right)$ & $O\left(\frac{1}{\epsilon^2(1-\lambda)^{?}}\right)$\tiny{\# a} &   $O\left(\frac{d_xd_y}{\epsilon^2(1-\lambda)^{?}}\log\frac{1}{\epsilon}\right)$   &  \#c \\
			\hline
			\makecell{Gossip-DSBO \\  \cite{yang2022decentralized}}  & $O\left(\log\frac{1}{\epsilon} \right)$ & $O(d_xd_y+d_y^2)$ & $O\left(\frac{1}{K\epsilon^2}\right)$\tiny{\#b} & $O\left(\frac{d_xd_y+d_y^2}{K\epsilon^2}\log\frac{1}{\epsilon}\right)$    & \#d \\
			\hline
			\makecell{	MA-DSBO \\ \cite{chen2022decentralized2}} & $O\left(\log\frac{1}{\epsilon} \right)$  & $O(d_x+d_y)$  &  $O\left(\frac{1}{\epsilon^2(1-\lambda)^{?}}\right)$\tiny{\#a}  &  $O\left(\frac{d_x+d_y}{\epsilon^2(1-\lambda)^{?}}\log\frac{1}{\epsilon}\right)$  &\#e  \\
			\hline
			\makecell{DSVRBGD-S \\ (Ours)}  & $O\left(1\right)$  & $O(d_x+d_y)$  &  $O\left(\frac{1}{K\epsilon^{3/2}(1-\lambda)^{4}}\right)$ & $O\left(\frac{d_x+d_y}{K\epsilon^{3/2}(1-\lambda)^{4}}\right)$ & - \\
			\hline
			\makecell{DSVRBGD-A  \\ (Ours)} & $O\left(1\right)$  & $O(d_x+d_y)$  &  $O\left(\frac{1}{K\epsilon^{3/2}(1-\lambda)^{4}}\right)$ & $O\left(\frac{d_x+d_y}{K\epsilon^{3/2}(1-\lambda)^{4}}\right)$ & - \\
			\bottomrule 
		\end{tabular}
	\end{center}
	
	\label{comparison}
\end{table*}

\textbf{Heterogeneity causes challenges for the variable update strategy.} Since there are two variables in stochastic bilevel optimization, there exist two strategies for updating variables: simultaneous update and alternating update. The former one updates $x$ and $y$ simultaneously, while the latter one updates $x$ and $y$ sequentially.   The heterogeneity introduces more challenges for the alternating strategy. Specifically,  assuming that the variables on different workers are the same in the $t$-th iteration, when  updating $x_t$ and $y_t$ with  the alternating strategy, $y_t$ is updated to $y_{t+1}$ first, and then $x_t$ is updated with the gradient computed on $y_{t+1}$, rather than $y_{t}$ as the simultaneous strategy.  As a result, this gradient computed on $y_{t+1}$ can be more heterogeneous than that computed on $y_{t}$ in the simultaneous strategy, causing new challenges for convergence analysis. Although DSBO \cite{chen2022decentralized} and MA-DSBO \cite{chen2022decentralized2} established the convergence rate for the alternating  strategy, they restrict their focus on the unbiased stochastic gradient. As a result, it is unclear whether the alternating algorithm will converge when employing a biased   variance-reduced gradient estimator \cite{cutkosky2019momentum}.  In particular, it can make the gradient across workers more heterogeneous due to the additional bias caused by this kind of gradient estimator. Specifically, the gradient estimation error regarding one variable can be affected by all the others, which is discussed in Section~\ref{sec:challenges}.  All these issues make the convergence analysis more challenging for the alternating update strategy. 

Then, the third natural question arises: \textit{Can we establish the convergence rate of a communication-efficient decentralized bilevel optimization algorithm under mild conditions when employing the alternating update strategy and the variance-reduced gradient estimator?} Therefore,  our third goal is to establish the convergence rate for the alternating update strategy to tackle these theoretical challenges.

In summary, to address the aforementioned challenges, this paper makes the following contributions.
\begin{itemize}
	\item We have developed a novel decentralized  stochastic bilevel gradient descent algorithm with a faster convergence rate based on \textit{the simultaneous update strategy} and \textit{variance-reduced gradient estimators}:  DSVRBGD-S,  which can  reduce both the number of communication rounds and the cost in each communication round.  In particular,  the communication cost per round is only $O(d_x + d_y)$, and the number of communication rounds can be as small as  $O(\frac{1}{K\epsilon^{3/2}})$, which can match the complexity of VRDBO under the homogeneous setting when ignoring other factors.  To the best of our knowledge, \textbf{this is the first method to achieve such a small communication complexity for decentralized stochastic bilevel optimization}. 
	\item We have established the theoretical convergence rate of our algorithm \textbf{without relying on any strong heterogeneity assumptions}. Furthermore, we have disclosed \textbf{how the computation and communication regarding the Hessian-inverse-vector product affects the dependence of the convergence rate on the spectral gap of the communication topology}. To the best of our knowledge, this is the first time such favorable theoretical results have been achieved.  The detailed comparison between our algorithm and existing ones can be found in Table~\ref{comparison}. 
	\item We have  developed a novel decentralized  stochastic bilevel gradient descent algorithm, DSVRBGD-A,  based on \textit{the alternating update strategy} and \textit{variance-reduced gradient estimators}. Similar to our first algorithm, this algorithm also enjoys a small communication complexity. Moreover, we have established its theoretical convergence rate. Remarkably, \textbf{this marks the first time the convergence rate of the \textit{alternating variance-reduced gradient descent} method for stochastic bilevel optimization has been established}. Notably, even in the single-machine setting, such theoretical results do not exist. Therefore, our theoretical analysis strategy can also be extended to the single-machine setting, bridging an existing gap in the literature. 
\end{itemize}
Finally, we conducted extensive experiments, and the experimental results confirm the efficacy of our proposed algorithms.

\section{Related Works}
\subsection{Stochastic Bilevel Optimization}

The main challenge in bilevel optimization lies in the computation of the hypergradient since it involves the Hessian inverse matrix. To address this issue, a commonly used approach is to leverage the Neumann series expansion technique to approximately compute Hessian inverse. Based on the first approach, \cite{ghadimi2018approximation} developed the  bilevel stochastic approximation algorithm, where the lower-level  problem is solved by stochastic gradient descent, and the upper-level problem is solved by stochastic hypergradient. As for the  nonconvex-strongly-convex bilevel optimization problem, this algorithm achieves $O(\frac{1}{\epsilon^2})$ sample complexity (i.e., the gradient evaluation) for the upper-level problem and $O(\frac{1}{\epsilon^3})$ sample complexity for the lower-level problem. Later, \cite{hong2020two} developed a  two-timescale stochastic approximation algorithm where different time scales are used for the upper-level  and lower-level step sizes, whose sample complexities is $O(\frac{1}{\epsilon^{5/2}})$. \cite{ji2021bilevel} proposed to employ the mini-batch stochastic gradient to improve both sample complexities to $O(\frac{1}{\epsilon^2})$. \cite{chen2021tighter} proposed an alternating stochastic bilevel gradient descent algorithm, which can also improve both sample complexities to $O(\frac{1}{\epsilon^2})$. \cite{yang2021provably,khanduri2021near} leveraged the variance-reduced gradient estimators STORM \cite{cutkosky2019momentum} or SPIDER \cite{fang2018spider} to further improve the sample complexity to $O(\frac{1}{\epsilon^{3/2}})$. However, the Neumann series expansion based algorithm requires an inner loop to estimate  Hessian inverse. As such, this class of algorithms suffers from a large Hessian-vector-product complexity. 

Another commonly used approach for estimating Hessian inverse is to directly estimate the Hessian-inverse-vector product in the hypergradient. Specifically, it views the Hessian-inverse-vector product as the solution of a quadratic optimization problem and then employs the gradient descent algorithm to estimate it. For instance, under the finite-sum setting where the number of samples is finite, \cite{dagreou2022framework} leveraged the variance-reduced gradient estimator SAGA \cite{defazio2014saga} to update the estimation of Hessian-inverse-vector product and two variables, providing the $O(\frac{(n+m)^{2/3}}{\epsilon})$ sample complexity where $n$ and $m$ are the number of samples in the upper-level and lower-level problems. Additionally, \cite{dagreou2023lower} employs a SPIDER-like \cite{fang2018spider} gradient estimator to improve the sample complexity to $O(\frac{(n+m)^{1/2}}{\epsilon})$.  Compared with the Neumann series expansion-based algorithm, this class of bilevel optimization algorithms does not need to use an inner loop to estimate Hessian inverse. Thus, they are more efficient in each iteration.  

\subsection{Decentralized Stochastic Bilevel Optimization}
The decentralized bilevel optimization has been actively studied in the past few years. A series of algorithms have been proposed. For instance, under the homogeneous setting, \cite{gao2023convergence} developed a decentralized bilevel stochastic gradient descent with momentum algorithm, which can achieve $O(\frac{1}{\epsilon^2})$ communication complexity and can be improved to $O(\frac{1}{K\epsilon^2})$ when all gradients are  upper bounded. Additionally, \cite{gao2023convergence}  proposed a bilevel stochastic gradient descent algorithm based on the STORM \cite{cutkosky2019momentum} gradient estimator, which can achieve $O(\frac{1}{K\epsilon^{3/2}})$ communication complexity, even though not all  gradients are upper bounded.  \cite{chen2022decentralized} developed the decentralized bilevel full gradient descent and decentralized bilevel stochastic gradient descent algorithms under both homogeneous and heterogeneous settings. \cite{yang2022decentralized} introduced the decentralized bilevel stochastic gradient descent with momentum algorithm under the heterogeneous setting, whose communication complexity  can achieve linear speedup with respect to the number of workers. All the aforementioned algorithms under the heterogeneous setting employ the Neumann series expansion approach to estimate Hessian inverse. As such, they suffer from a large communication complexity caused by the Neumann series expansion. Recently, \cite{chen2022decentralized2} estimate the Hessian-inverse-vector product under the decentralized setting. However, it uses the standard stochastic gradient so that it needs to use an inner loop to estimate Hessian-inverse-vector product to reduce the estimation error. Thus, it still suffers from a large communication complexity and fails to achieve linear speedup. 

Other than the decentralized bilevel optimization problem defined in Eq.~(\ref{loss_bilevel}), there exists another class of decentralized bilevel optimization problems, where $y^*(x)$ only depends on each local lower-level optimization problem rather than the global one.  Without the global dependence, the hypergradient is much easier to estimate than that in Eq.~(\ref{loss_bilevel}).  To address this class of  decentralized bilevel optimization problems, \cite{lu2022decentralized} developed a stochastic gradient-based algorithm, and \cite{liu2022interact} leveraged the SPIDER \cite{fang2018spider} gradient estimator to update variables.  Moreover, these also exist distributed bilevel optimization algorithms  \cite{gao2022convergence,tarzanagh2022fednest,huang2022fast,li2022local}  under the centralized setting, which are orthogonal to the decentralized setting.  Recently, there appears a concurrent work \cite{dong2023single}, which focuses on {the full gradient}  rather than the stochastic gradient.  


\section{Preliminaries}

\begin{assumption} \label{assumption_bi_strong}
	For $\forall k$, $g^{(k)}(x, y)$ is $\mu$-strongly convex with respect to  $y$ for fixed $x\in \mathbb{R}^{d_x}$ where $\mu>0$ is a constant. 
\end{assumption}
\begin{assumption} \label{assumption_upper_smooth_vr}
	For $\forall k$, $\forall (x, y)\in\mathbb{R}^{d_x}\times \mathbb{R}^{d_y} $, $\nabla_1 f^{(k)}(x, y; \xi)$ is $\ell_{f_x}$-Lipschitz continuous  where $\ell_{f_x}>0$ is a constant, 	$\nabla_2 f^{(k)}(x, y; \xi)$  is $\ell_{f_y}$-Lipschitz continuous where $\ell_{f_y}>0$ is a constant, $\|\nabla_2 f^{(k)}(x, y; \xi)\|\leq c_{f_y}$ where $c_{f_y}>0$ is a constant. 
\end{assumption}
\begin{assumption} \label{assumption_lower_smooth_vr}
	For $\forall k$, $\forall (x, y)\in\mathbb{R}^{d_x}\times \mathbb{R}^{d_y} $, $\nabla_2 g^{(k)}(x, y; \zeta)$ is  $\ell_{g_y}$-Lipschitz continuous where $\ell_{g_y}>0$ is a constant, $\nabla_{12}^2 g^{(k)}(x, y; \zeta)$ is $\ell_{g_{xy}}$-Lipschitz continuous where $\ell_{g_{xy}}>0$ is a constant,  $\nabla_{22}^2 g^{(k)}(x, y; \zeta)$  is $\ell_{g_{yy}}$-Lipschitz continuous where $\ell_{g_{yy}}>0$ is a constant, $\|\nabla_{12}^2 g^{(k)}(x, y; \zeta)\|\leq c_{g_{xy}}$ where $c_{g_{xy}}>0$ is a constant,   and $\mu \mathbf{I} \preceq \nabla_{22}^2 g^{(k)}(x, y; \zeta)\preceq  \ell_{g_y} \mathbf{I} $. 
\end{assumption}
\begin{assumption} \label{assumption_variance}
	All stochastic gradients have bounded variance $\sigma^2$ where $\sigma>0$ is a constant.
\end{assumption}

\begin{remark}\label{remark-1}
	Our assumptions regarding the gradient are milder than existing heterogeneous decentralized bilevel optimization algorithms \cite{chen2022decentralized,yang2022decentralized}. In particular, they have additional assumptions:  $\|\nabla_1 f^{(k)}(x, y)\|\leq c_{f_x}$ and $\|\nabla_2 g^{(k)}(x, y)\|\leq c_{g_y}$. The latter on does not hold for a strongly convex function as discussed in C.1 of  \cite{chen2022decentralized2}.
\end{remark}

\begin{remark}\label{remark-2}
	\cite{chen2022decentralized2} introduces the explicit heterogeneity assumption: $\|\nabla_2 g^{(k)}(x, y) - \nabla_2 g(x, y)\|\leq \delta$ where $\delta>0$ is a constant. In fact, a  quadratic function $g^{(k)}(x, y)=\frac{1}{2}y^TA_ky$ does not satisfy this assumption when $A_k\neq A_{k'}$ for $k\neq k'$ because $\|(A_k-\frac{1}{K}\sum_{k'=1}^{K}A_{k'})y\|^2$ is unbounded for $y\in\mathbb{R}^{d_y}$. 
\end{remark}

\begin{assumption} \label{assumption_graph}
	The adjacency matrix $E$ of the communication graph is symmetric and doubly stochastic, whose eigenvalues  satisfy $|\lambda_n|\leq \cdots \leq |\lambda_2|< |\lambda_1|=1$.
\end{assumption} 
In this paper, we denote $\lambda\triangleq|\lambda_2|$ so that the spectral gap of the communication graph can be represented as $1-\lambda$. Additionally, Assumptions~\ref{assumption_upper_smooth_vr} and~\ref{assumption_lower_smooth_vr} also hold for the full gradient. 
Throughout this paper, we use $a_t^{(k)}$ to denote the variable $a$ on the $k$-th worker in the $t$-th iteration and use $\bar{a}_t=\frac{1}{K}\sum_{k=1}^{K}a_t^{(k)}$ to denote the averaged variables.  Moreover, for a function $f(\cdot, \cdot)$, we use $\nabla_i f(\cdot, \cdot)$ to denote the gradient with respect to the $i$-th argument where $i\in\{1, 2\}$.

\section{Decentralized Stochastic  Bilevel Gradient Descent}
\subsection{Estimation of Stochastic Hypergradient}
In this paper, we denote ${ F^{(k)}(x)}  \triangleq  { f^{(k)}(x, y^*(x))}$, $F(x) \triangleq \frac{1}{K}\sum_{k=1}^{K} F^{(k)}(x) $,  and $ f(x, y^*(x))\triangleq\frac{1}{K}\sum_{k=1}^{K} f^{(k)}(x, y^*(x))$. 
Then, the global hypergradient is defined as follows:
\begin{equation}
	\begin{aligned}
		\nabla{ F(x)}  & = \nabla_1 { f(x, y^*(x))}  \notag \\
		&\quad  -\left[\frac{1}{K}\sum_{k=1}^{K}\nabla_{12}^2 g^{(k)}(x, y^*(x))\right]  \left[\frac{1}{K}\sum_{k=1}^{K}\nabla_{22}^2g^{(k)}(x, y^*(x))\right]^{-1} \nabla_2{ f(x, y^*(x))}  \ . \\
	\end{aligned}
\end{equation}
Here, following \cite{li2022fully}, the Hessian-inverse-vector product $\left[\frac{1}{K}\sum_{k=1}^{K}\nabla_{22}^2g^{(k)}(x, y^*(x))\right]^{-1} \nabla_2{ f(x, y^*(x))} $ can be viewed as the optimal solution of the following constrained strongly-convex quadratic optimization problem:
\begin{equation} \label{loss_z}
	\begin{aligned}
		& \min_{z\in \mathbb{R}^{d_y}}  h(x, z) \triangleq \frac{1}{K}\sum_{k=1}^{K} h^{(k)}(x, z)   \ , \quad  s.t. \quad  \|z\|\leq \frac{c_{f_y}}{\mu} \ , 
	\end{aligned}
\end{equation}
where 
\begin{align}
    & h^{(k)}(x, z)  = \frac{1}{2}z^T\nabla_{22}^2g^{(k)}(x, y^*(x))z - z^T\nabla_{2} f^{(k)}(x, y^*(x)) \ , \notag \\
    & z^*(x) = \left[\frac{1}{K}\sum_{k=1}^{K}\nabla_{22}^2g^{(k)}(x, y^*(x))\right]^{-1} \frac{1}{K}\sum_{k=1}^{K}\nabla_2{ f^{(k)}(x, y^*(x))} \ . \notag
\end{align}
It is easy to know that  $\|z^*(x)\| \leq \frac{c_{f_y}}{\mu}$ based on the aforementioned assumptions. Therefore, we have the constraint $\|z\|\leq \frac{c_{f_y}}{\mu}$ in Eq.~(\ref{loss_z}). Otherwise, the solution of Eq.~(\ref{loss_z}) may not be a good approximation for $z^*(x)$.  Moreover,  a favorable property of Eq.~(\ref{loss_z}) is that the gradient $\nabla_{z} h(x, z^*(x))=0$ even though there exists a constraint.

In terms of $z^*(x)$, the global hypergradient can be represented as 
\begin{align}
    & \nabla{ F(x)}  = \nabla_1 { f(x, y^*(x))}  -\left[\frac{1}{K}\sum_{k=1}^{K}\nabla_{12}^2 g^{(k)}(x, y^*(x))\right]z^*(x) \ . \notag 
\end{align}
With this reformulation, we can use the approximated solution of Eq.~(\ref{loss_z}) to approximate the Hessian-inverse-vector product  without computing Hessian inverse explicitly.

On the other hand, the hypergradient of the $k$-th worker is defined as follows:
\begin{equation}
	\begin{aligned}
		& 	\nabla{ F^{(k)}(x)}  = \nabla_1 { f^{(k)}(x, y^*(x))}  -\nabla_{12}^2 g(x, y^*(x))  \left[\nabla_{22}^2g(x, y^*(x))\right]^{-1} \nabla_2{ f^{(k)}(x, y^*(x))} \ .  \\
	\end{aligned}
\end{equation}
Obviously, it depends on the global Jacobian matrix $\nabla_{12}^2 g(x, y^*(x))$ and Hessian matrix $\nabla_{22}^2g(x, y^*(x))$, which are expensive to obtain in each iteration. Moreover,  $y^*(x)$ and $z^*(x)$ are also expensive to obtain in each iteration of the stochastic gradient based algorithm. Therefore, we proposed the following \textit{biased} gradient estimators to approximate $\nabla{ F^{(k)}(x)}$  and $\nabla_{2}h^{(k)}(x, z)  $ on the $k$-th worker for updating $y$ and $z$:
\begin{equation}
	\begin{aligned}
		& \hat{\mathcal{G}}_{F}^{(k)}(x, y, z)  \triangleq \nabla_1 { f^{(k)}(x, y)}   -\nabla_{12}^2 g^{(k)}(x, y)z^{(k)} ,  \\ &  \hat{\mathcal{G}}_{h}^{(k)}(x, y, z) \triangleq\nabla_{22}^2g^{(k)}(x, y) z^{(k)} -  \nabla_2{ f^{(k)}(x, y)}  \ ,  \\
	\end{aligned}
\end{equation}
where $z^{(k)}$ is the approximated solution of the optimization problem: $\min_{z: \|z\|\leq \frac{c_{f_y}}{\mu} } h^{(k)}(x, z)$, and $y$ is an approximation of $ y^*(x)$.  In other words, we can leverage $\hat{\mathcal{G}}_{h}^{(k)}(x, y, z)$ to update $z^{(k)}$, which will be used to construct the approximated hypergradient $\hat{\mathcal{G}}_{F}^{(k)}(x, y, z)$.  Correspondingly, we can define the stochastic gradient as follows:
\begin{equation} \label{eq_stochastic_hypergradient}
	\begin{aligned}
		& 	\hat{\mathcal{G}}_{F}^{(k)}(x, y, z; \hat{\xi})  \triangleq \nabla_1 { f^{(k)}(x, y; \xi)}   -\nabla_{12}^2 g^{(k)}(x, y; \zeta)z^{(k)} \ ,  \\
		& \hat{\mathcal{G}}_{h}^{(k)}(x, y, z; \hat{\xi}) \triangleq\nabla_{22}^2g^{(k)}(x, y; \zeta) z^{(k)} -  \nabla_2{ f^{(k)}(x, y; \xi)}  \ ,  \\
	\end{aligned}
\end{equation}
where $\hat{\xi}\triangleq\{\xi, \zeta\}$.

To present our algorithms, we introduce the following terminologies:  
\begin{equation}
	\begin{aligned}
		& X_t=[x_t^{(1)}, x_t^{(2)},  \cdots, x_t^{(K)}] \ ,  Y_t=[y_t^{(1)}, y_t^{(2)},  \cdots, y_t^{(K)}] \ ,   Z_t=[z_t^{(1)}, z_t^{(2)},  \cdots, z_t^{(K)}]  \ , \\
		&  \delta_t^{\hat{\mathcal{G}}_{F}} (X_t, Y_t, Z_t; \hat{\xi}_t) \triangleq [\hat{\mathcal{G}}_{F}^{(1)}(x_t^{(1)}, y_t^{(1)}, z_t^{(1)}; \hat{\xi}_t^{(1)}) ,  \cdots,  \hat{\mathcal{G}}_{F}^{(K)}(x_t^{(K)}, y_t^{(K)}, z_t^{(K)}; \hat{\xi}_t^{(K)}) ]  \ , \\
		&  \delta_t^{\hat{\mathcal{G}}_{h}} (X_t, Y_t, Z_t; \hat{\xi}_t)  \triangleq [\hat{\mathcal{G}}_{h}^{(1)}(x_t^{(1)}, y_t^{(1)}, z_t^{(1)}; \hat{\xi}_t^{(1)}) ,   \cdots,  \hat{\mathcal{G}}_{h}^{(K)}(x_t^{(K)}, y_t^{(K)}, z_t^{(K)}; \hat{\xi}_t^{(K)}) ]  \ , \\
		&  \delta_t^{g}  (X_t, Y_t; {\zeta}_t) \triangleq [\nabla_{y} g^{(1)}(x_t^{(1)}, y_t^{(1)}; {\zeta}_t^{(1)}) ,   \cdots, \nabla_{y} g^{(K)}(x_t^{(K)}, y_t^{(K)}; {\zeta}_t^{(K)})]  \  .  
	\end{aligned}
\end{equation}
Then, we use $\delta_t^{\hat{\mathcal{G}}_{F}} (X_t, Y_t, Z_t)$, $\delta_t^{\hat{\mathcal{G}}_{h}} (X_t, Y_t, Z_t)$, and $\delta_t^{g}  (X_t, Y_t)$ to denote the corresponding full gradient. Moreover, we use $\bar{A}_t=[\bar{a}_t, \cdots, \bar{a}_t]$ to denote the matrix which is composed of $K$ mean variables $\bar{a}_t$, where $a$ can be any variable in this paper.

\begin{algorithm}[h]
	\small
	\caption{Decentralized Stochastic  Variance-Reduced Bilevel Gradient Descent Algorithm with  \textcolor{blue}{Simultaneous} (corresponds to Option-I) and  \textcolor{red}{Alternating} (corresponds to Option-II) update. }
	\label{alg_VRDBO}
	\begin{algorithmic}[1]
		\REQUIRE ${x}_{0}^{(k)}={x}_{0}$, ${y}_{0}^{(k)}={y}_{0}$, ${z}_{0}^{(k)}={z}_{0}$, $\eta>0$, $\alpha_1>0$, $\alpha_2>0$, $\alpha_3>0$, $\beta_1>0$, $\beta_2>0$, $\beta_3>0$,  $\alpha_1\eta^2<1$,    $\alpha_2\eta^2<1$,   $\alpha_3\eta^2<1$. 
		$P_{-1} = 0$, $Q_{-1} = 0$, $R_{-1} = 0$, $U_{-1} = 0$, $V_{-1} = 0$, $W_{-1} = 0$. 
		\FOR{$t=0,\cdots, T-1$} 
		\STATE 
		\textbf{Option-I \& II: Compute variance-reduced gradient $V_t$ for simultaneous/alternating update} \\
		$V_{t} =   \begin{cases}\delta_{t}^{g}(X_{t}, Y_{t}; {\zeta}_{t}), & t=0 \\
			(1-\alpha_2\eta^2)(V_{t-1} - \delta_{t}^{g}(X_{t-1}, Y_{t-1}; {\zeta}_{t}))+\delta_{t}^{g}(X_{t}, Y_{t}; {\zeta}_{t}), & t>0
		\end{cases}$ \ , \\
		\textbf{Update} $Y$:  \\
		$Q_{t} = Q_{t-1}E + V_{t} - V_{t-1}$ ,  
		$Y_{t+\frac{1}{2}} = Y_{t}E - \beta_2Q_{t}$,  \ $Y_{t+1}=Y_{t} + \eta (Y_{t+\frac{1}{2}}- Y_{t} )$, \\
		\STATE  
		\textbf{Option-I: Compute variance-reduced gradient $W_t$ for \textcolor{blue}{simultaneous} update} \\
		$W_{t}  =\begin{cases}
			\delta_{t}^{\hat{\mathcal{G}}_{h}}(X_{t}, Y_{\textcolor{blue}{t}}, Z_{t}; \hat{\xi}_{t}) , & t=0 \\
			(1-\alpha_3\eta^2)(W_{t-1} - \delta_{t}^{\hat{\mathcal{G}}_{h}}(X_{t-1}, Y_{\textcolor{blue}{t-1}}, Z_{t-1}; \hat{\xi}_{t}))+  \delta_{t}^{\hat{\mathcal{G}}_{h}}(X_{t}, Y_{\textcolor{blue}{t}}, Z_{t}; \hat{\xi}_{t}) , & t>0 \\
		\end{cases}$ \ , \\
		\textbf{Option-II: Compute variance-reduced gradient $W_t$ for \textcolor{red}{alternating} update} \\
		$W_{t}  =\begin{cases}
			\delta_{t}^{\hat{\mathcal{G}}_{h}}(X_{t}, Y_{\textcolor{red}{t+1}}, Z_{t}; \hat{\xi}_{t}) , & t=0 \\
			(1-\alpha_3\eta^2)(W_{t-1} - \delta_{t}^{\hat{\mathcal{G}}_{h}}(X_{t-1}, Y_{\textcolor{red}{t}}, Z_{t-1}; \hat{\xi}_{t}))+  \delta_{t}^{\hat{\mathcal{G}}_{h}}(X_{t}, Y_{\textcolor{red}{t+1}}, Z_{t}; \hat{\xi}_{t}) , & t>0 \\
		\end{cases}$ \ , \\
		\textbf{Update $Z$: }\\
		$R_{t} = R_{t-1}E + W_{t} - W_{t-1}$ ,  
		$Z_{t+\frac{1}{2}} = \mathcal{P}(Z_{t}E -\beta_3 R_{t})$,  \ $Z_{t+1}=Z_{t} + \eta (Z_{t+\frac{1}{2}}- Z_{t} )$, \\
		
		\STATE  
		\textbf{Option-I: Compute variance-reduced gradient $U_t$ for  \textcolor{blue}{simultaneous} update} \\
		$U_{t}  = \begin{cases}
			\delta_{t}^{\hat{\mathcal{G}}_{F}}(X_{t}, Y_{\textcolor{blue}{t}}, Z_{\textcolor{blue}{t}}; \hat{\xi}_{t}), & t=0 \\
			(1-\alpha_1\eta^2)(U_{t-1} - \delta_{t}^{\hat{\mathcal{G}}_{F}}(X_{t-1}, Y_{\textcolor{blue}{t-1}}, Z_{\textcolor{blue}{t-1}}; \hat{\xi}_{t}))+  \delta_{t}^{\hat{\mathcal{G}}_{F}}(X_{t}, Y_{\textcolor{blue}{t}}, Z_{\textcolor{blue}{t}}; \hat{\xi}_{t}) , & t>0 \\
		\end{cases}$ \ , \\ 
		\textbf{Option-II: Compute variance-reduced gradient $U_t$ for \textcolor{red}{alternating} update} \\
		$U_{t}  = \begin{cases}
			\delta_{t}^{\hat{\mathcal{G}}_{F}}(X_{t}, Y_{\textcolor{red}{t+1}}, Z_{\textcolor{red}{t+1}}; \hat{\xi}_{t}), & t=0 \\
			(1-\alpha_1\eta^2)(U_{t-1} - \delta_{t}^{\hat{\mathcal{G}}_{F}}(X_{t-1}, Y_{\textcolor{red}{t}}, Z_{\textcolor{red}{t}}; \hat{\xi}_{t}))+  \delta_{t}^{\hat{\mathcal{G}}_{F}}(X_{t}, Y_{\textcolor{red}{t+1}}, Z_{\textcolor{red}{t+1}}; \hat{\xi}_{t}) , & t>0 \\
		\end{cases}$ \ , \\ 
		\textbf{Update} $X$: \\
		$P_{t} = P_{t-1}E + U_{t} - U_{t-1}$ ,  
		$X_{t+\frac{1}{2}} = X_{t}E - \beta_1P_{t}$,  \ $X_{t+1}=X_{t} + \eta (X_{t+\frac{1}{2}}- X_{t} )$, \\
		
		\ENDFOR
	\end{algorithmic}
\end{algorithm}

\subsection{Decentralized Stochastic  Variance-Reduced Bilevel Gradient Descent Algorithm}
In Algorithm~\ref{alg_VRDBO}, we present two novel decentralized stochastic  variance-reduced bilevel gradient descent algorithms   based on the simultaneous update (\textbf{DSVRBGD-S})   and alternating update  (\textbf{DSVRBGD-A})   strategies, respectively.  Generally speaking, for \textit{the computation on each worker}, we use the variance-reduced gradient estimator \cite{cutkosky2019momentum}, which is a biased gradient estimator,  to solve the lower-level optimization problem and the upper-level optimization problem in Eq.~(\ref{loss_bilevel}), as well as the additional constrained quadratic optimization problem in Eq.~(\ref{loss_z}). For \textit{the communication across workers}, we leverage the gradient tracking communication strategy to communicate three variables and the corresponding gradient estimators between neighboring workers. 

\textbf{DSVRBGD-S Algorithm.} DSVRBGD-S employs the \textbf{simultaneous} update strategy.  Specifically,  as shown in Option-I of each step in Algorithm~\ref{alg_VRDBO},    DSVRBGD-S constructs the variance-reduced gradient estimator with   the stochastic gradient \footnote{Throughout this paper, we ignore the discussion of the stochastic gradient computed on the variable in the $(t-1)$-th iteration for simplicity.}: $\delta_{t}^{g}(X_{t}, Y_{t}; {\zeta}_{t})$, $\delta_{t}^{\hat{\mathcal{G}}_{h}}(X_{t}, Y_{\textcolor{blue}{t}}, Z_{t}; \hat{\xi}_{t})$, and $\delta_{t}^{\hat{\mathcal{G}}_{F}}(X_{t}, Y_{\textcolor{blue}{t}}, Z_{\textcolor{blue}{t}}; \hat{\xi}_{t})$, which are computed on the variable in the ${t}$-th iteration: $\{X_t, Y_t, Z_t\}$.  These three stochastic gradients can be computed simultaneously, and the update of three variables can also be completed simultaneously. 

\textbf{DSVRBGD-A Algorithm.} DSVRBGD-A leverages the \textbf{alternating} update strategy, which is to  update three variables sequentially. Specifically, as shown in Step 2 of Algorithm~\ref{alg_VRDBO}, DSVRBGD-A first computes the variance-reduced gradient estimator $V_t$ based on the variable $\{X_t, Y_t, Z_t\}$, with which the variable $Y_t$ is updated to $Y_{t+1}$.  After that, as shown in Option-II of Step 3 in Algorithm~\ref{alg_VRDBO},  DSVRBGD-A computes the variance-reduced gradient estimator $W_t$ based on the variable $\{X_t, Y_{\textcolor{red}{t+1}}, Z_t\}$ as:
\begin{equation}
	\begin{aligned}
		W_{t} & =  (1-\alpha_3\eta^2)(W_{t-1} - \delta_{t}^{\hat{\mathcal{G}}_{h}}(X_{t-1}, Y_{\textcolor{red}{t}}, Z_{t-1}; \hat{\xi}_{t}))  +  \delta_{t}^{\hat{\mathcal{G}}_{h}}(X_{t}, Y_{\textcolor{red}{t+1}}, Z_{t}; \hat{\xi}_{t})  \  ,
	\end{aligned}
\end{equation}
where $\alpha_3>0$, $\eta>0$ and $\alpha_3\eta^2<1$.  It can be observed the new update $Y_{\textcolor{red}{t+1}}$ is used for computing $\delta_{t}^{\hat{\mathcal{G}}_{h}}(X_{t}, Y_{\textcolor{red}{t+1}}, Z_{t}; \hat{\xi}_{t})$, rather than using the prior update $Y_{t}$ to compute $\delta_{t}^{\hat{\mathcal{G}}_{h}}(X_{t}, Y_{\textcolor{blue}{t}}, Z_{t}; \hat{\xi}_{t})$ as Option-I of DSVRBGD-S.  Then, DSVRBGD-A employs the following gradient-tracking approach to update and communicate $Z_{t}$:
\begin{equation}
	\begin{aligned}
		& R_t = R_{t-1}E + W_{t} - W_{t-1},   Z_{t+\frac{1}{2}} = \mathcal{P}(Z_{t}E -\beta_3 R_{t}) \ ,     Z_{t+1}=Z_{t} + \eta (Z_{t+\frac{1}{2}}- Z_{t} ) \  , 
	\end{aligned}
\end{equation}
where  $\beta_3>0$, $R_t$ can be viewed as the estimation of the global $\bar{W}_t$, $R_{t-1}E$ and $Z_{t}E$ denote the peer-to-peer communication to communicate $R$ and $Z$ in terms of the adjacency matrix $E$. Since Eq.~(\ref{loss_z}) is a constrained optimization problem, we apply the projection operator $\mathcal{P}(\cdot)$ to the intermediate variable  $Z_{t+\frac{1}{2}}$ such that the intermediate variable on all workers always satisfies that constraint.  It is worth noting that $Z_{t+1}$ also satisfies that constraint because it is  a convex combination between $Z_{t}$ and $Z_{t+\frac{1}{2}}$.  After obtaining $Z_{t+1}$, DSVRBGD-A constructs the stochastic hypergradient $\delta_{t}^{\hat{\mathcal{G}}_{F}}(X_{t}, Y_{\textcolor{red}{t+1}}, Z_{\textcolor{red}{t+1}}; \hat{\xi}_{t})$ based on $\{X_{t}, Y_{\textcolor{red}{t+1}}, Z_{\textcolor{red}{t+1}}\}$ in Option-II of Step 4 in Algorithm~\ref{alg_VRDBO}. Then, the variance-reduced gradient estimator is computed for local update and the gradient tracking communication strategy is used for communication to obtain the new update  $X_{t+1}$. 

Both the computation overhead  and communication cost for estimating the Hessian-inverse-vector product in our two algorithms are small. In particular,  estimating this product on each worker only requires a gradient descent update, whose cost is on the order of $O(d_y)$ in each iteration. As for the communication, our algorithms only need to communicate the auxiliary optimization variable and its gradient estimator, with a cost of $O(d_y)$ in each iteration. Therefore, the computation and communication costs are small compared to those of the double-loop baselines shown in Table 1.

\textbf{Key Features.}  
Our two algorithms have the following favorable features. 
\begin{itemize}
	\item  The communication cost of our two algorithms in each communication round is just $O(d_x + d_y)$ because only $x\in \mathbb{R}^{d_x}$, $y\in \mathbb{R}^{d_x}$, $z\in \mathbb{R}^{d_y}$, and their gradient estimators are communicated. It is smaller than  $O(d_y^2+d_xd_y)$ of Gossip-DSBO \cite{yang2022decentralized} and  $O(d_xd_y)$ of DSBO \cite{chen2022decentralized}. 
	\item  There is only one loop in our two algorithms. On the contrary, the existing methods, including DSBO \cite{chen2022decentralized}, Gossip-DSBO \cite{yang2022decentralized}, and MA-DSBO \cite{yang2022decentralized}, are double-loop algorithms. As a result, the number of communication rounds of our two algorithms in each iteration is just $O(1)$, which is  smaller than $O(\log\frac{1}{\epsilon})$  of those three existing methods. 
	\item  We have developed algorithms based on both the simultaneous and alternating update strategies. Notably,  this is the first time applying the alternating update strategy to the variance-reduced gradient for bilevel optimization. 
\end{itemize}
\vspace{-5pt}
In summary, our two algorithms are communication-efficient due to the low communication cost in each round and the smaller number of communication rounds.

\section{Convergence Analysis}
In this section, we present the theoretical convergence rate of our algorithms and discuss the unique challenges and novelties in the convergence analysis. The proof sketch can be found in Appendix~\ref{sec:proof-sketch}, and the detailed proof can be found in Appendix~\ref{sec:proof-sim} and Appendix~\ref{sec:proof-alt}.

\subsection{Convergence Rate}
\begin{theorem} \label{theorem-sim}
	Under Assumptions~\ref{assumption_bi_strong}-\ref{assumption_graph}, by letting $\eta$, $\beta_1$, $\beta_2$, and $\beta_3$ satisfy Eq.~(\ref{eq:hyper}), and setting   $\alpha_1=O(\frac{1}{K})$,  $\alpha_2=O(\frac{1}{K})$, and $\alpha_3=O(\frac{1}{K})$, for a sufficiently large $T\geq \frac{K^{2}}{(1-\lambda)^{4}}$, DSVRBGD-S has the following convergence rate:
	\begin{equation} 
		\small
		\begin{aligned}
			&   \frac{1}{T}\sum_{t=0}^{T-1}\mathbb{E} [ \| \nabla F(\bar{x}_{t})\|^2 ] \leq   O\left(\frac{1}{\eta T}\right)+ O\left(\frac{1}{\eta TB_0} \right)   +  O\left( \frac{1}{\beta_1\eta T}\right)+  O\left( \frac{1}{\beta_2\eta T}\right) + O\left( \frac{1}{\beta_3\eta T}\right)  \\
			&   + O\left(\frac{1}{\alpha_1 \eta^2 T KB_0}\right)  + O\left( \frac{1}{\alpha_2 \eta^2 TKB_0}\right)  +  O\left(\frac{1}{\alpha_3  \eta^2 TKB_0} \right)  +O\left( \frac{\alpha_1\eta^2}{ K}\right)  + O\left( \frac{\alpha_2\eta^2 }{ K} \right)   \\
			&   +O\left(\frac{\alpha_3\eta^2}{ K } \right)  + O\left(\alpha_1^2\eta^3\right)  +  O\left(\alpha_2^2\eta^3\right)+ O\left(\alpha_3^2\eta^3\right)   \  ,  \\
		\end{aligned}
	\end{equation}
	where $B_0$ is the batch size in the first iteration. 
\end{theorem}
\begin{corollary}\label{corollary-sim}
	Under Assumptions~\ref{assumption_bi_strong}-\ref{assumption_graph}, by setting  $\alpha_1=O({1}/{K})$,  $\alpha_2=O({1}/{K})$,  $\alpha_3=O({1}/{K})$,  $\beta_1 = O((1-\lambda)^4)$, $\beta_2 = O((1-\lambda)^2)$, $\beta_3 = O((1-\lambda)^4)$, 
	$\eta=O(K\epsilon^{1/2})$, the batch size in the first iteration as $B_0=O(1/\epsilon^{1/2})$,  the batch size in other iterations as $O(1)$,   and  $T=O\left(\frac{1}{K(1-\lambda)^4\epsilon^{3/2}}\right)$,  DSVRBGD-S can  achieve the $\epsilon$-accuracy solution: $\frac{1}{T}\sum_{t=0}^{T-1}\mathbb{E}[\| \nabla F(\bar{x}_{t})\|^2] \leq \epsilon$. 
\end{corollary}

\begin{remark}
	1) Our convergence rate  does not depend on any  assumptions regarding the heterogeneity. On the contrary, existing methods \cite{chen2022decentralized,yang2022decentralized,chen2022decentralized2} require strong assumptions, which are shown in Table~\ref{comparison}.  2) Our algorithm is more communication-efficient than them \cite{chen2022decentralized,yang2022decentralized,chen2022decentralized2} because  DSVRBGD-S has a smaller number of communication rounds and the low cost in each round.  3) DSVRBGD-S has a worse dependence on the spectral gap than  the homogeneous method VRDBO \cite{gao2023convergence}, i.e., $1/(1-\lambda)^4$ versus $1/(1-\lambda)^2$. This is caused by the projection operation for the estimator of the Hessian-inverse-vector product as discussed in Section~\ref{sec:challenges}. 
\end{remark}

\begin{theorem} \label{theorem-alt}
	Under Assumptions~\ref{assumption_bi_strong}-\ref{assumption_graph}, by letting $\eta$, $\beta_1$, $\beta_2$, and $\beta_3$ satisfy Eq.~(\ref{eq:hyper-alt}), and setting   $\alpha_1=O(\frac{1}{K})$,  $\alpha_2=O(\frac{1}{K})$, and $\alpha_3=O(\frac{1}{K})$, for a sufficiently large $T\geq \frac{K^{2}}{(1-\lambda)^{4}}$, DSVRBGD-S has the following convergence rate:
	\begin{equation}
		\small
		\begin{aligned}
			&  \frac{1}{T}\sum_{t=0}^{T-1}\mathbb{E}[\| \nabla F(\bar{x}_{t})\|^2] \leq   O\left(\frac{1}{\eta T} \right)  + O\left(\frac{1}{\eta TB_0} \right)   + O\left(\frac{1}{\beta_1\eta T}\right)+    O\left(\frac{1}{\beta_2\eta T}\right) +  O\left(\frac{1}{\beta_3\eta T}\right) \\
			&  + O\left(\frac{1}{\alpha_1\eta^2 TKB_0}\right) + O\left(\frac{1}{\alpha_2\eta^2 TKB_0}\right)   +  O\left(\frac{\alpha_1\eta^2}{K}\right)   + O\left(\frac{\alpha_2\eta^2}{K}\right)   + O\left(\frac{\alpha_3\eta^2}{K}\right)   \\
			&  + O\left(\alpha_3^2\eta^3\right) + O\left(\alpha_1^2\eta^3\right)   + O\left(\alpha_2^2\eta^3\right) + O\left(\alpha^2_2\eta^4\right)       + O\left(\frac{\eta\beta_2^2}{T}\right) + O\left(\frac{\eta\beta_3^2}{T}\right)  +   O\left(\frac{\eta^3\beta_2^2\beta_3^2}{T}\right)   \\
			&    + O\left( \frac{\beta_2^2}{ \alpha_1T} \right)   + O\left(\frac{\beta_3^2}{\alpha_1 T}\right)+  O\left( \frac{ \eta^2\beta_2^2\beta_3^2}{ \alpha_1T} \right)   + O\left(\frac{\eta\beta_2^2}{TB_0}\right)  + O\left( \frac{\eta^3\beta_2^2\beta_3^2}{ TB_0}\right)  +  O\left( \frac{ \beta_2^2}{ \alpha_1TB_0} \right)  \\
			&  + O\left(\frac{1}{\alpha_3\eta^2 TKB_0}\right) +  O\left( \frac{ \eta^2\beta_2^2\beta_3^2}{ \alpha_1TB_0} \right)   \ ,   \\
		\end{aligned}
	\end{equation}
	where $B_0$ is the batch size in the first iteration. 
\end{theorem}

\begin{corollary}\label{corollary-alt}
	Under Assumptions~\ref{assumption_bi_strong}-\ref{assumption_graph}, by setting  $\alpha_1=O({1}/{K})$,  $\alpha_2=O({1}/{K})$,  $\alpha_3=O({1}/{K})$,  $\beta_1 = O((1-\lambda)^4)$, $\beta_2 = O((1-\lambda)^2)$, $\beta_3 = O((1-\lambda)^4)$, 
	$\eta=O(K\epsilon^{1/2})$, the batch size in the first iteration as $B_0=O(1/\epsilon^{1/2})$,  the batch size in other iterations as $O(1)$,   and  $T=O\left(\frac{1}{K(1-\lambda)^4\epsilon^{3/2}}\right)$,  DSVRBGD-A can  achieve the $\epsilon$-accuracy solution: $\frac{1}{T}\sum_{t=0}^{T-1}\mathbb{E}[\| \nabla F(\bar{x}_{t})\|^2] \leq \epsilon$. 
\end{corollary}
\begin{remark}
	According to Corollary~\ref{corollary-sim} and Corollary~\ref{corollary-alt}, the convergence rate of DSVRBGD-A is in the same order as that of DSVRBGD-S.  
	From Theorem~\ref{theorem-sim} and Theorem~\ref{theorem-alt},	it can be observed  that DSVRBGD-A has some additional terms. These additional terms  are only related to the  higher order  of $\epsilon$, which can be found in Eq.~(\ref{eq:convergence-upper-bound-alt}).
\end{remark}
\begin{remark}
	The communication complexity of our two algorithms can be further improved from two perspectives. On the one hand, we can improve the dependence on $\epsilon$ by using the SPIDER gradient estimator \cite{fang2018spider}.   In particular, by using SPIDER, we can use a large batch size in each iteration, which  leads to a smaller number of iterations. As a result, the communication complexity can be reduced.  In fact, in the single-machine setting, Corollary 2 in  \cite{yang2021provably} has demonstrated the iteration complexity is $O(\epsilon^{-1})$ when using the SPIDER gradient estimator. On the other hand, we can improve the dependence on the spectral gap $1-\lambda$ by using other communication strategies.  For example, we can use the multi-round communication, i.e.,  performing multiple rounds of communication in each iteration as in \cite{xin2021stochastic} to improve the dependence on the spectral gap. 
\end{remark}

\subsection{Key Challenges and Novelties} \label{sec:challenges}

\textbf{The projection operation on the communicated variable $z$ introduces new challenges for convergence analysis.}  In Algorithm~\ref{alg_VRDBO}, to solve the constrained quadratic optimization problem in Eq.~(\ref{loss_z}), we employ the projection operation to ensure the new update $z_{t}^{(k)}$ to satisfy the constraint $\|z_{t}^{(k)}\|\leq \frac{c_{f_y}}{\mu}$, which incurs unique challenges  for decentralized optimization. Specifically, since  the projection operator is NOT a linear operator,  it leads to $\sum_{k=1}^{K} \mathcal{P}(z_{t}^{(k)}) \neq \mathcal{P}(\sum_{k=1}^{K} z_{t}^{(k)})$, which makes it challenging to bound the optimization error $\mathbb{E}[\|\bar{ z}_{t} -    {z}^{*}(\bar{   {x}}_{t})\| ^2 ]$.  

First, due to the non-linear projection projection, we cannot bound $\mathbb{E}[\|\bar{ z}_{t} -    {z}^{*}(\bar{   {x}}_{t})\| ^2 ]$ as $\mathbb{E}[\|\bar{y}_{t} -    {y}^{*}(\bar{   {x}}_{t})\| ^2 ]$. In particular, when there is no projection, the averaged variable $\bar{y}_{t}$ behaves like the variable under the single-machine setting. Therefore, it is easy to bound $\mathbb{E}[\|\bar{y}_{t} -    {y}^{*}(\bar{   {x}}_{t})\| ^2 ]$ by following Lemma 28 in \cite{huang2020accelerated} under the single-machine setting. However, due to the non-linear projection operation, Eq.~(83) in \cite{huang2020accelerated} only holds for the variable on each worker $z_{t}^{(k)}$, rather than the averaged variable $\bar{z}_{t}$.  Therefore, the approach  \cite{huang2020accelerated}  designed for the single-machine setting cannot be applied to our algorithms. 

Second, although there exists an alternative approach, i.e., Lemma 2 in \cite{zhang2024jointly},   to bound $\mathbb{E}[\|\bar{y}_{t} -    {y}^{*}(\bar{   {x}}_{t})\| ^2 ]$ under the decentralized setting, it cannot be directly applied to $\mathbb{E}[\|\bar{ z}_{t} -    {z}^{*}(\bar{   {x}}_{t})\| ^2 ]$ due to the projection operation. In particular, just as shown in Eq.~(\ref{eq:z-bar-z-star-0}), we cannot  obtain $ \mathbb{E}[\left\|\bar{z}_{ t+1} - z^*(\bar{x}_{t})\right\|^2]=\mathbb{E}[\left\| \bar{z}_t  - \eta\beta_3  \bar{r}_t -  z^*(\bar{x}_{t})\right\|^2]$ as Eq.~(23) in \cite{zhang2024jointly} due to $\sum_{k=1}^{K} \mathcal{P}(z_{t}^{(k)}) \neq \mathcal{P}(\sum_{k=1}^{K} z_{t}^{(k)})$. Therefore, we establish a new upper bound for $\mathbb{E}[\left\|\bar{z}_{ t+1} - z^*(\bar{x}_{t})\right\|^2]$, which has two additional terms as shown in  Eq.~(\ref{eq:z-bar-z-star-0}). Especially, our theoretical analysis shows that the additional term $\mathbb{E}[\|   Z_{t}  - \bar{Z}_t  \|_F^2]$ in Eq.~(\ref{eq:z-bar-z-star-0}) makes the convergence rate  have a worse dependence on the spectral gap. Specifically, it makes the coefficient $c_4$ of $\mathbb{E}[\|{Z}_t- \bar{Z}_{t}\|_F^2]$ in the potential function $\mathcal{L}_t$ in Eq.~(\ref{eq:potential-function}) have an additional term compared with $c_3$ of $\mathbb{E}[\|{Y}_t- \bar{Y}_{t}\|_F^2]$, i.e.,  $c_4=\frac{2\beta_1}{(1-\lambda^2)}\tilde{c}_4+  \frac{5\beta_1}{\beta_3\eta(1-\lambda^2)}   \Tilde{c}_1$ (See Eq.~(\ref{eq:c-4})),  where the second term has a factor $\frac{1}{\beta_3}$.  Such a factor makes $\beta_3$ have to be as small as $O((1-\lambda)^4)$. Specifically, as shown in Eq.~(\ref{eq:beta-3-inequalities}), that factor leads to the second inequality, which further leads to  Eq.~(\ref{eq:beta3-upper-2}), resulting in a worse dependence  $O((1-\lambda)^4)$. Obviously, if there were no such a factor in $c_4$, the upper bound of $\beta_3$ could  be improved to $O((1-\lambda)^2)$, which will lead to a better convergence rate. 

In summary, the additional communication required for estimating hypergradient makes the convergence analysis more challenging due to the involved projection operation. Our theoretical analysis clearly discloses how it affects the convergence rate.

\textbf{The alternating update introduces more challenges for convergence analysis.} Specifically, considering the $t$-th iteration, the alternating update introduces new challenges when bounding the gradient estimation error: $\mathbb{E} [ \|\delta^{\hat{\mathcal{G}}_{F}}(X_{t}, Y_{t+1}, Z_{t+1})  -   U_{t} \|_F^2 ]$ and $\mathbb{E} [ \|\delta^{\hat{\mathcal{G}}_{h}}(X_{t}, Y_{t+1}, Z_{t}) -   W_{t}\|_F^2 ]$. Taking the former one as an example, as shown in Option-II of Step 4 in Algorithm~\ref{alg_VRDBO},  $U_{t}$ should use the new update $Y_{t+1}$ and $Z_{t+1}$.  As a result, its upper bound depends on $\mathbb{E}[\|Y_{t+1}-Y_{t}\|_F^2]$ and  $\mathbb{E}[\|Z_{t+1}-Z_{t}\|_F^2]$ as shown in Lemma~\ref{lemma_hyper_storm_var_mean-alt}. Furthermore, as shown in Lemma~\ref{lemma:y-incremental-2-alt} and Lemma~\ref{lemma:z-incremental-2-alt}, those two terms depend on the gradient estimation error $\mathbb{E}[\| V_{t-1}- \delta^{g}(X_{t-1}, Y_{t-1})   \|_F^2]$ and $\mathbb{E}[\| W_{t-1}- \delta^{\hat{\mathcal{G}}_{h}}(X_{t-1}, Y_{t}, Z_{t-1})   \|_F^2]$, respectively.  As a result, the upper bound of  $\mathbb{E} [ \|\delta^{\hat{\mathcal{G}}_{F}}(X_{t}, Y_{t+1}, Z_{t+1})  -   U_{t+1} \|_F^2 ]$ depends on all three gradient estimation errors in the $(t-1)$-th iteration, i.e., $\mathbb{E} [ \|\delta^{\hat{\mathcal{G}}_{F}}(X_{t-1}, Y_{t}, Z_{t})  -   U_{t-1} \|_F^2 ]$,   $\mathbb{E}[\| V_{t-1}- \delta^{g}(X_{t-1}, Y_{t-1})   \|_F^2]$,  and $\mathbb{E}[\| W_{t-1}- \delta^{\hat{\mathcal{G}}_{h}}(X_{t-1}, Y_{t}, Z_{t-1})   \|_F^2]$, which can be found in Eq.~(\ref{eq:expand-u-est-error}).  On the contrary, for the simultaneous update, as shown in Lemma~\ref{lemma_h_storm_var_mean},   it only depends on its own estimation error in the $t$-th iteration. Therefore, it is much more challenging to establish the convergence rate for DSVRBGD-A compared with DSVRBGD-S. To address this challenge, we design a new potential function to handle the challenging dependence between those gradient estimation errors. In particular, as shown in Eq.~(\ref{eq:coefficient-alt}), we introduce additional terms for the coefficient $c_{11}$ and $c_{13}$ of   $\mathbb{E}[\| V_{t}- \delta^{g}(X_{t}, Y_{t})   \|_F^2]$ and $\mathbb{E}[\| W_{t}- \delta^{\hat{\mathcal{G}}_{h}}(X_{t}, Y_{t+1}, Z_{t})   \|_F^2]$ compared with those in Eq.~(\ref{eq:coeff}) of the simultaneous update.

\section{Experiment}
In our experiments, we focus on the hyperparameter optimization problem and the hyper-representation learning problem.

\subsection{Hyperparameter Optimization}
In our experiments, we focus on the following hyperparameter optimization problem:
\begin{equation} \label{eq:exp-hyperparameter-opt}
	\begin{aligned}
		& \min_{x\in \mathbb{R}^d} \frac{1}{K} \sum_{k=1}^{K}\frac{1}{n^{(k)}}\sum_{i=1}^{n^{(k)}}\ell(y^*(x)^Ta_{v, i}^{(k)}, b_{v,i}^{(k)}) \\
		& s.t. \  y^*(x) =  \arg\min_{y\in \mathbb{R}^{d\times c}}\frac{1}{K} \sum_{k=1}^{K} \frac{1}{m^{(k)}}\sum_{i=1}^{m^{(k)}}  \ell(y^Ta_{t, i}^{(k)}, b_{t, i}^{(k)})  + \frac{1}{cd}\sum_{p=1}^{c}\sum_{q=1}^{d} \exp(x_q)y_{pq}^2 \ ,
	\end{aligned}
\end{equation}
where the lower-level optimization problem optimizes the classifier's parameter $y\in \mathbb{R}^{d\times c}$ based on the training set $\{(a_{t,i}^{(k)}, b_{t,i}^{(k)})\}_{i=1}^{m^{(k)}}$, the upper-level optimization problem optimizes the hyperparameter $x\in \mathbb{R}^d$ based on the validation set $\{(a_{v,i}^{(k)}, b_{v,i}^{(k)})\}_{i=1}^{n^{(k)}}$, the loss function is the cross-entropy loss function.

To verify the performance of our algorithm,  we use four real-world LIBSVM datasets \footnote{\url{https://www.csie.ntu.edu.tw/~cjlin/libsvmtools/datasets/}}: w8a, a9a, ijcnn1, covtype. For each dataset, we randomly select $10\%$ of samples as the test set, $70\%$ of the remaining samples as the training set, and the others as the validation set.  Then, to demonstrate the performance of our algorithms under the heterogeneous setting, we construct a heterogeneous variant for each training set.  In detail, we use eight workers in our experiments. We set the imbalance ratio on these workers, i.e., the ratio between the number of samples in the positive class and the total number of samples, as $\{0.1, 0.15, 0.2, 0.25, 0.3, 0.35, 0.4, 0.45\}$ by randomly dropping samples from the positive or negative class. 

We compare  our algorithm with five baseline methods: including MDBO \cite{gao2023convergence},  DSBO \cite{chen2022decentralized},  Gossip-DSBO  \cite{yang2022decentralized},  MA-DSBO \cite{chen2022decentralized2},  and VRDBO \cite{gao2023convergence}. 
As for the hyperparameter, we set the solution accuracy $\epsilon$ to $0.05$ so that the learning rates of the first four methods are set to $\eta=\epsilon^2$ according to their theoretical results, e.g., Theorem 3.3 in \cite{chen2022decentralized2}, and the learning rate of VRDBO and our algorithm is set to $\eta=\epsilon$ in terms of the corresponding theoretical results. Moreover, $\alpha_i$ is set such that $\alpha_i\eta^2=0.9$ and $\beta_i$ is set to $1$ for both VRDBO and our algorithms. As for the double-loop basline methods: DSBO,Gossip-DSBO,   and MA-DSBO, the number of iterations for the lower-level update and the Hessian-inverse-vector product update is set to $10$.  For the single-loop baseline method MDBO, we use a learning rate for the lower-level variable that is ten times larger than those double-loop methods to mitigate the effect of the double loop.
Additionally, the batch size of all algorithms is set to $100$, and the number of iterations is set to $2,000$. 
As for the communication topology, we consider three classes: ring graph, random graph, and torus graph. In particular, the probability for generating the random graph is set to $0.4$ in our experiments.

\begin{figure*}[ht]
	\centering 
	\hspace{-15pt}
	\subfigure[a9a]{
		\includegraphics[scale=0.22]{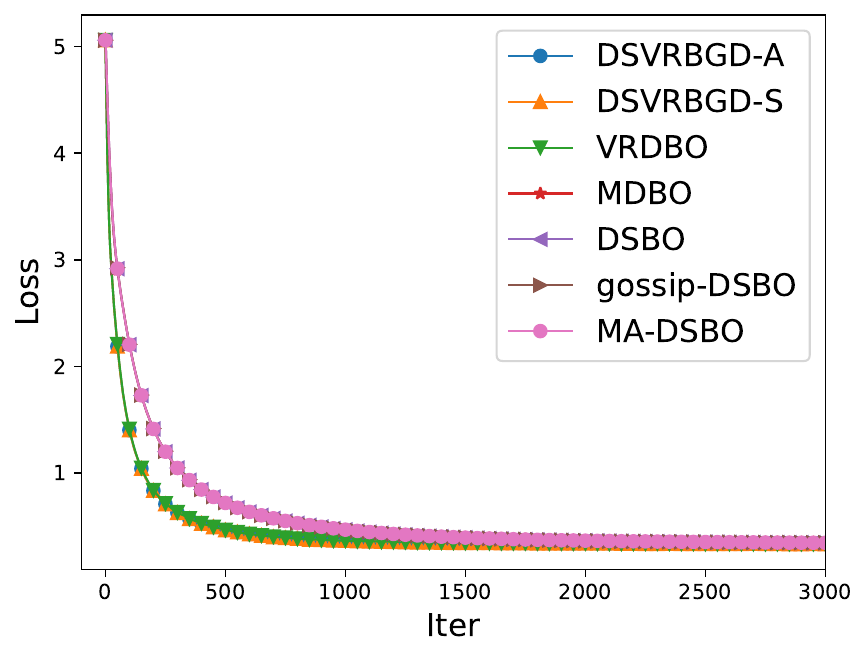}
	}
	\hspace{-11pt}
	\subfigure[w8a]{
		\includegraphics[scale=0.22]{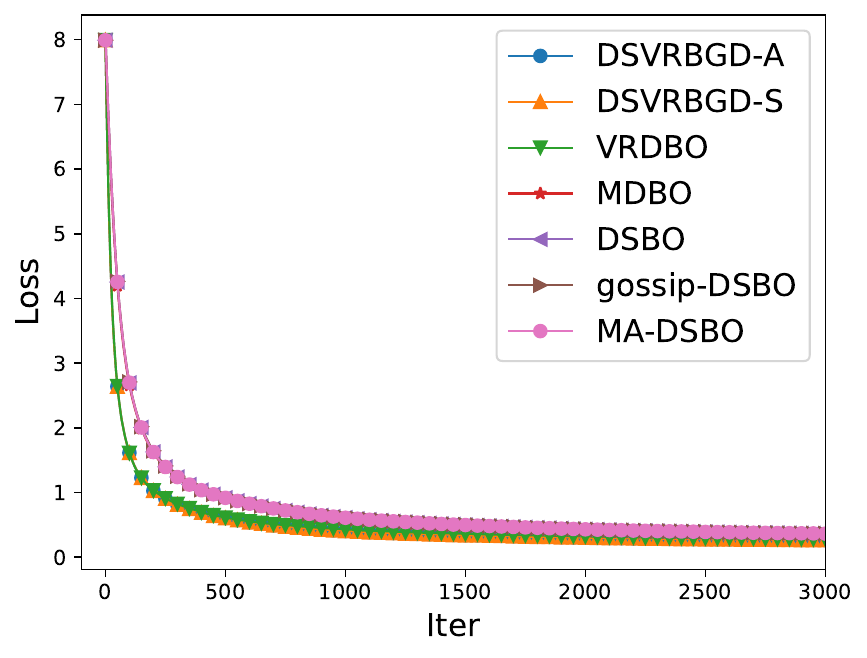}
	}
	\hspace{-11pt}
	\subfigure[ijcnn1]{
		\includegraphics[scale=0.22]{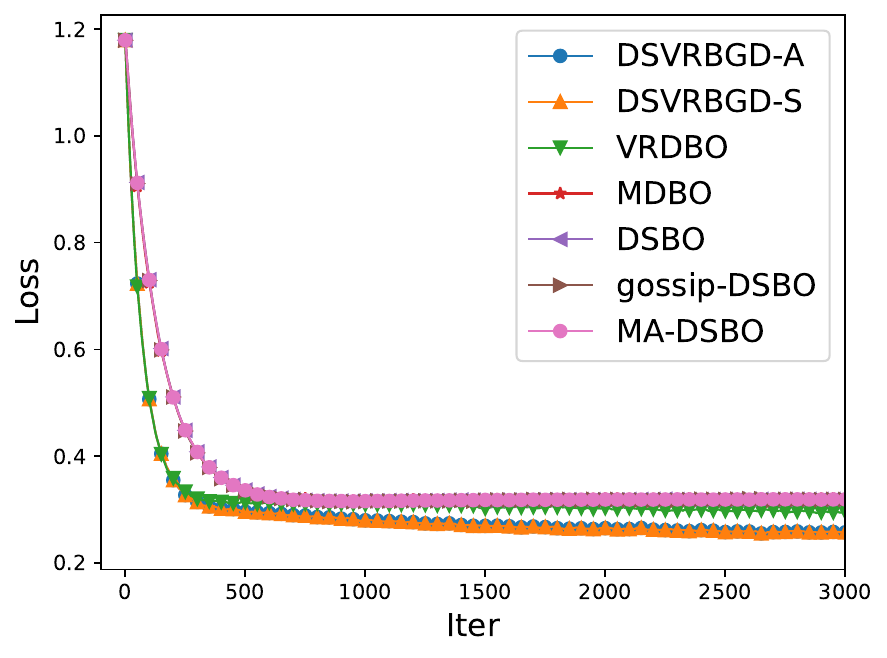}
	}
	\hspace{-11pt}
	\subfigure[covtype]{
		\includegraphics[scale=0.22]{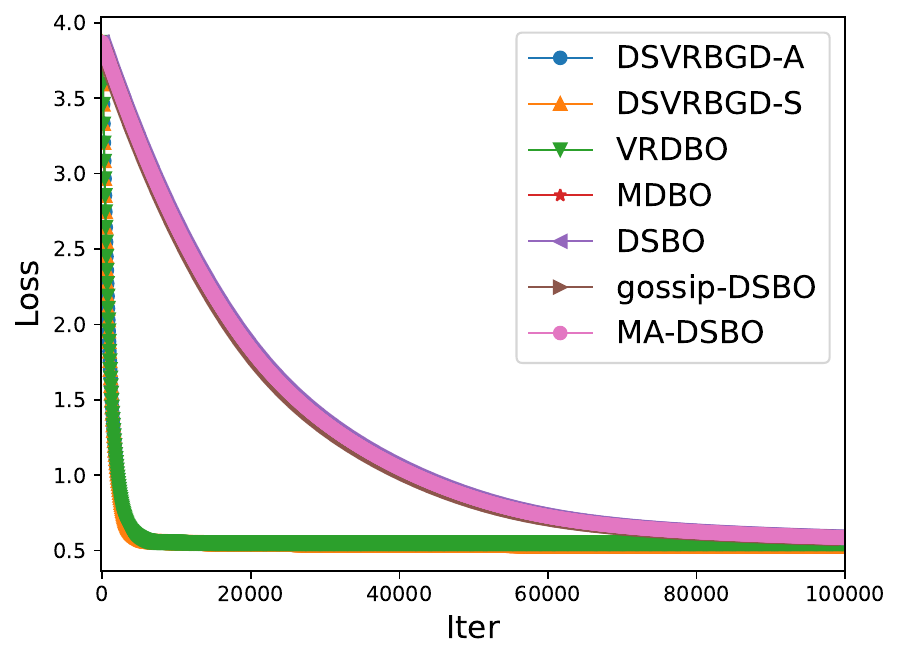}
	}
	\caption{ The upper-level loss function value  versus the number of iterations on a {ring} graph. }
	\label{real-world_loss-iteration-ring}
\end{figure*}

\begin{figure*}[!t]
	\centering 
	\hspace{-15pt}
	\subfigure[a9a]{
		\includegraphics[scale=0.22]{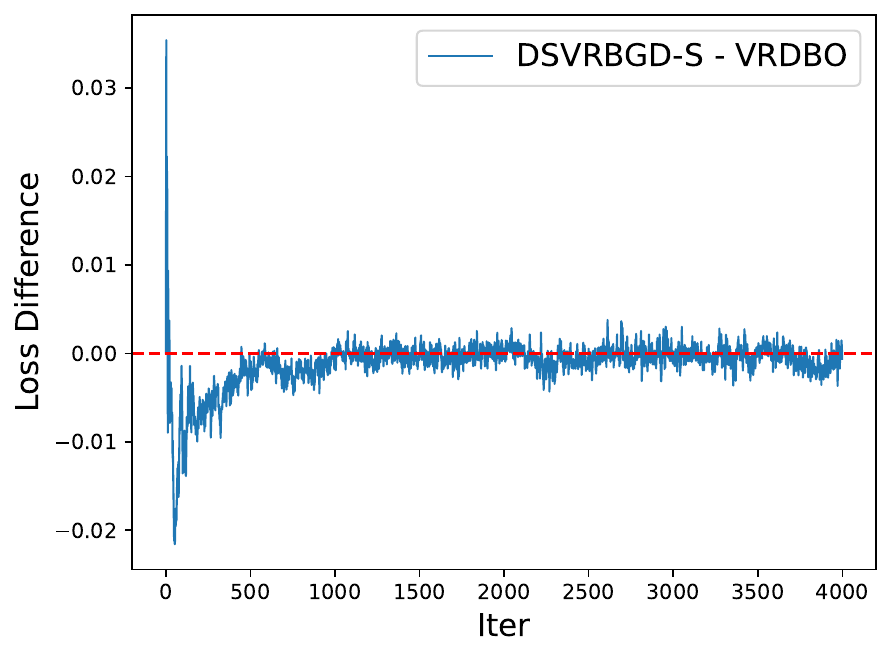}
	}
	\hspace{-13pt}
	\subfigure[w8a]{
		\includegraphics[scale=0.22]{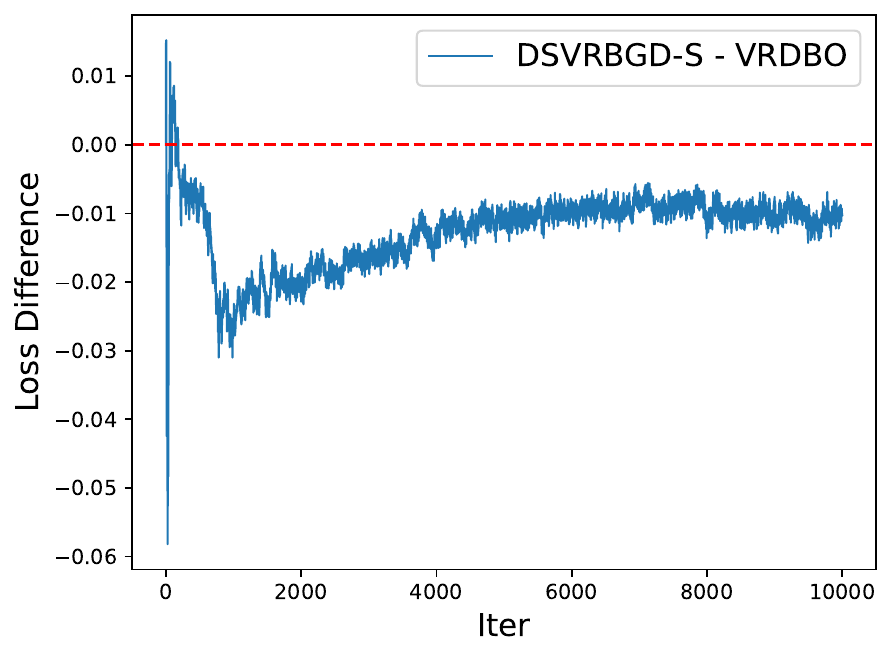}
	}
	\hspace{-13pt}
	\subfigure[ijcnn1]{
		\includegraphics[scale=0.22]{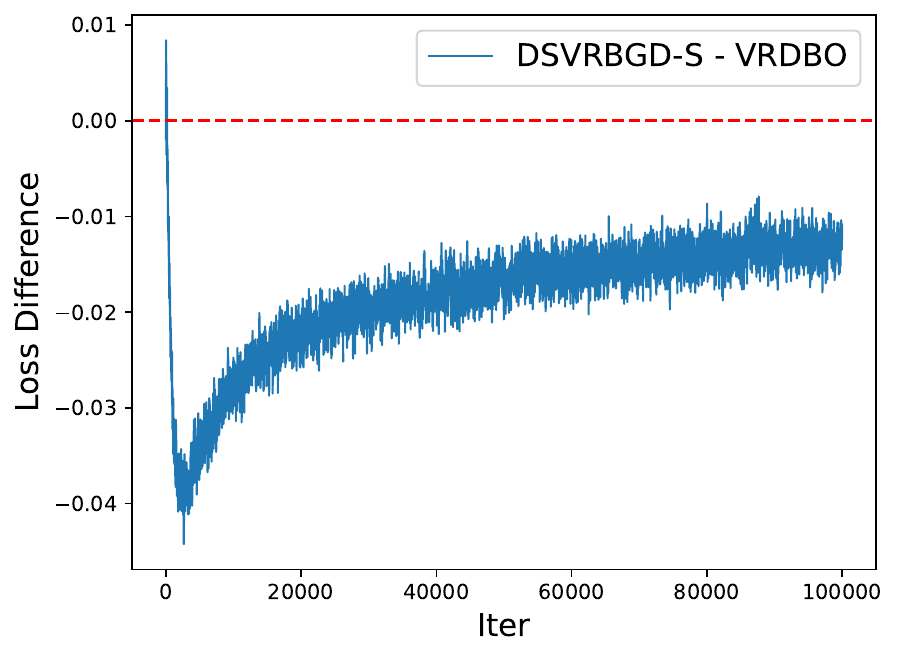}
	}
	\hspace{-13pt}
	\subfigure[covtype]{
		\includegraphics[scale=0.22]{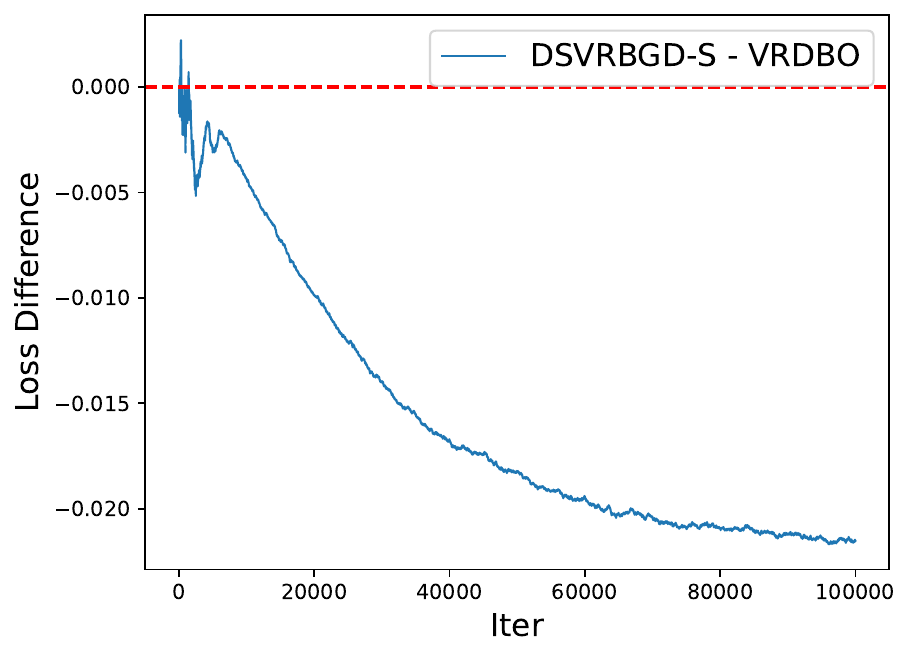}
	}\\
	\hspace{-15.5pt}
	\subfigure[a9a]{
		\includegraphics[scale=0.22]{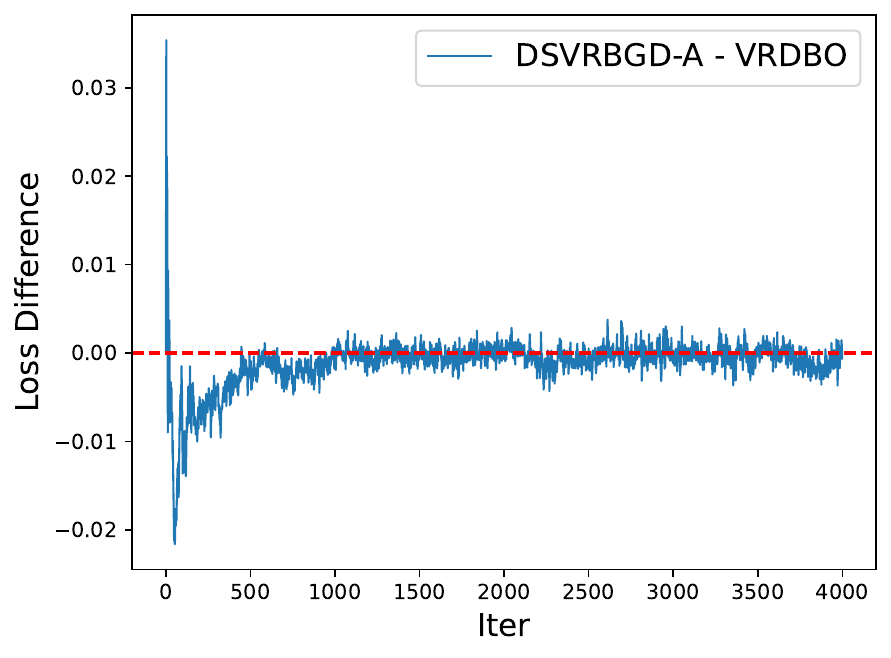}
	}
	\hspace{-13.5pt}
	\subfigure[w8a]{
		\includegraphics[scale=0.22]{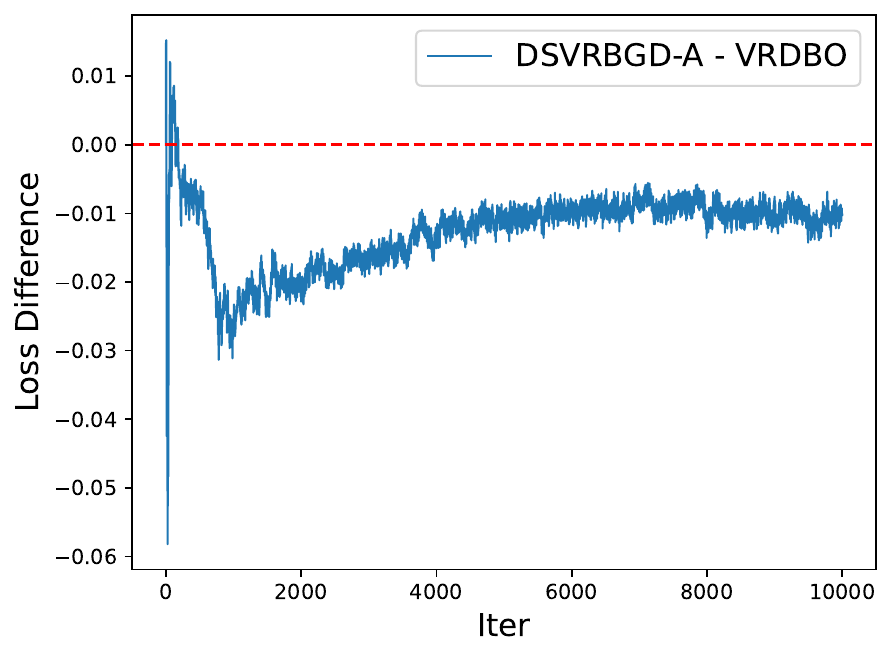}
	}
	\hspace{-13.5pt}
	\subfigure[ijcnn1]{
		\includegraphics[scale=0.22]{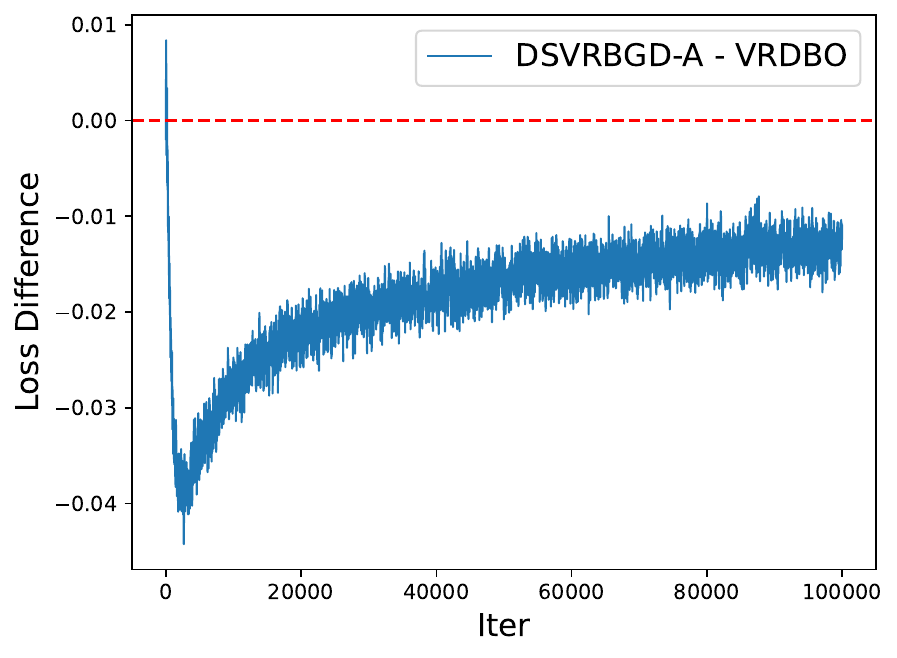}
	}
	\hspace{-14pt}
	\subfigure[covtype]{
		\includegraphics[scale=0.22]{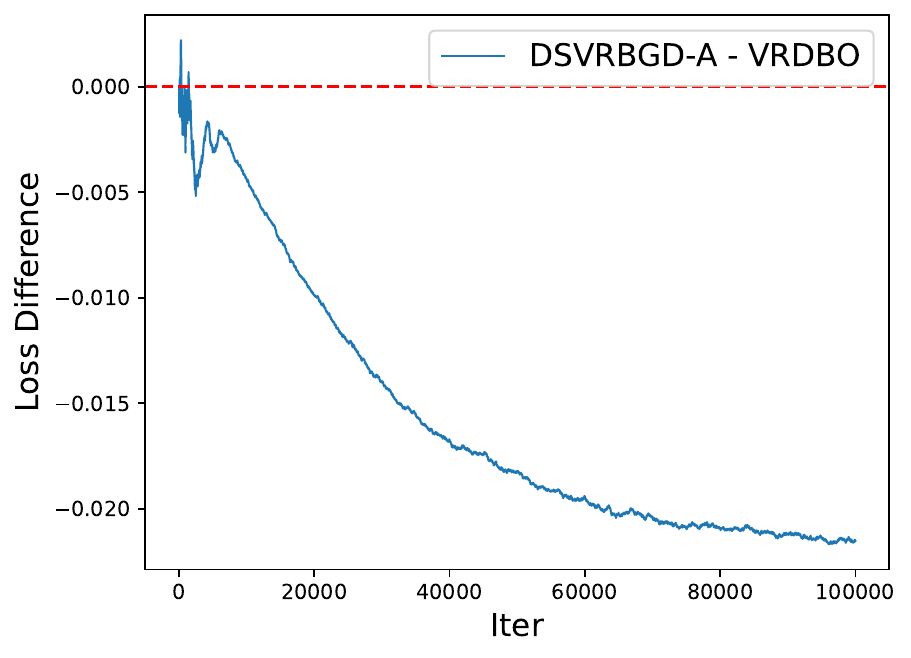}
	}
	\caption{The first row shows the loss function value difference: $\Delta=\text{LOSS}_{\text{DSVRBGD-S}}-\text{LOSS}_{\text{VRDBO}}$. The second row shows the loss function value difference: $\Delta=\text{LOSS}_{\text{DSVRBGD-A}}-\text{LOSS}_{\text{VRDBO}}$. The negative difference value denotes our methods converge faster than VRDBO, as they reduce the loss function value more quickly.}
	\label{real-world_loss-diff-iteration-ring}
\end{figure*}

In Figure~\ref{real-world_loss-iteration-ring}, we plot the upper-level loss function value versus the number of iterations.  It can be observed that our two methods converge faster than the method that does not use variance-reduced gradient, including  Gossip-DSBO, DSBO, MA-DSBO, and  MDBO, and also enjoys a faster convergence speed than the method using variance reduction, i.e., VRDBO. In addition, to better show the improvement of our two methods over VRDBO, we plot the difference between the loss function value obtained from our methods and that obtained from VRDBO in Figure~\ref{real-world_loss-diff-iteration-ring}. In particular, we plot $\Delta=\text{LOSS}_{\text{DSVRBGD-S}}-\text{LOSS}_{\text{VRDBO}}$ in the first row and $\Delta=\text{LOSS}_{\text{DSVRBGD-A}}-\text{LOSS}_{\text{VRDBO}}$ in the second row. When $\Delta<0$, DSVRBGD-S(DSVRBGD-A) decreases the loss function value faster than VRDBO. Otherwise, VRDBO converges faster. In Figure~\ref{real-world_loss-diff-iteration-ring}, it can be seen that the value of the loss function decreases more rapidly with our two methods than with VRDBO, confirming that our two methods outperform VRDBO.

\begin{figure*}[h]
	\centering 
	\hspace{-15pt}
	\subfigure[a9a]{
		\includegraphics[scale=0.22]{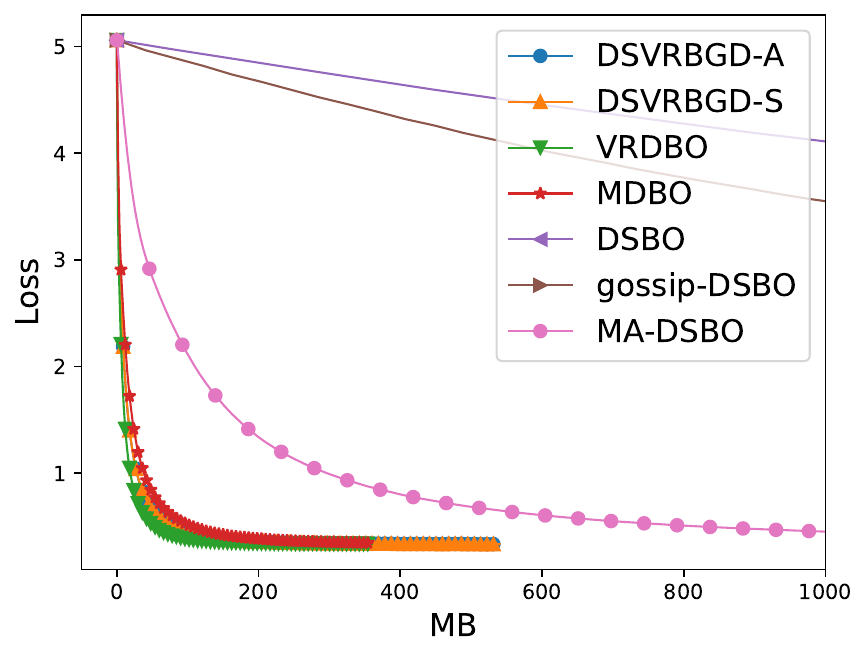}
	}
	\hspace{-11pt}
	\subfigure[w8a]{
		\includegraphics[scale=0.22]{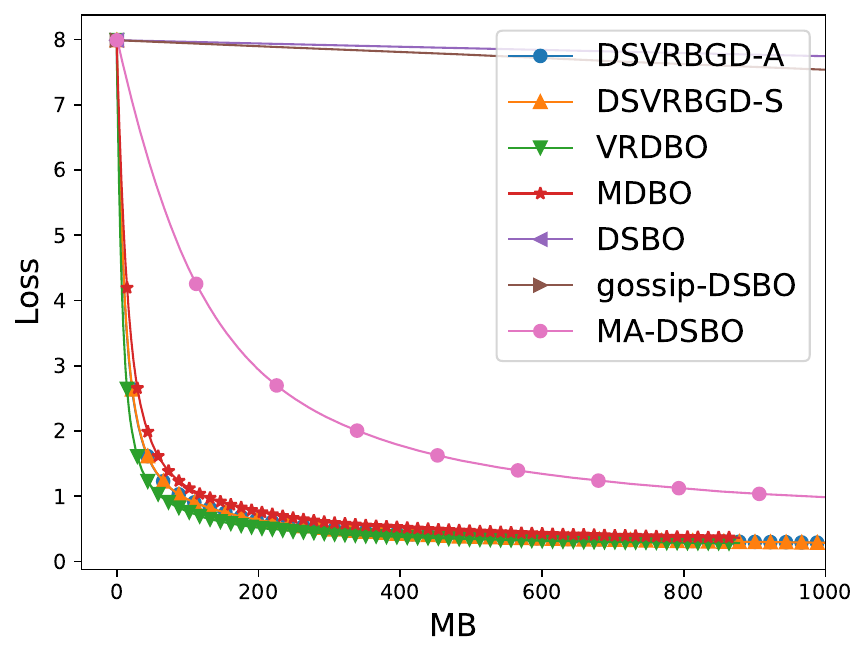}
	}
	\hspace{-11pt}
	\subfigure[ijcnn1]{
		\includegraphics[scale=0.22]{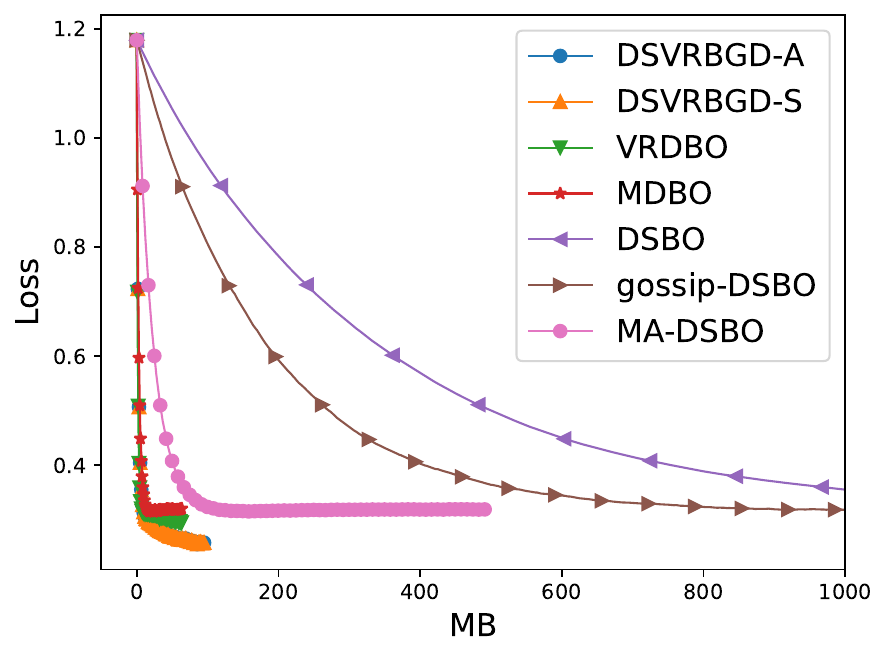}
	}
	\hspace{-11pt}
	\subfigure[covtype]{
		\includegraphics[scale=0.22]{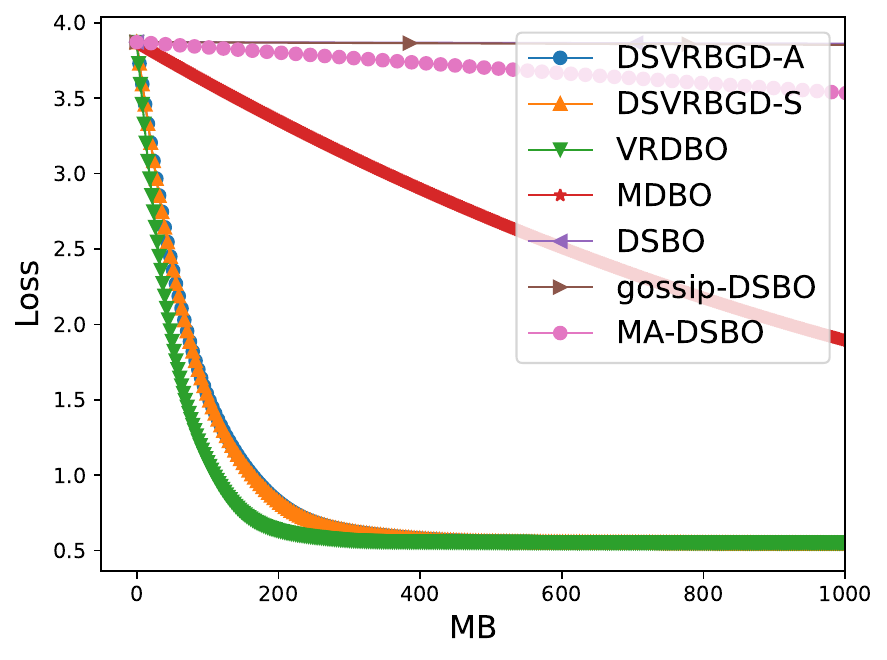}
	}
	\caption{ The upper-level loss function value  versus the communication cost (MB) on a {ring} graph. }
	\label{real-world_loss-communication-ring}
\end{figure*}

In Figure~\ref{real-world_loss-communication-ring}, we plot the upper-level loss function value versus the number of communicated megabytes (MB),  when using the ring graph. Note that we only show the first 100MB to make the comparison clearer.  Compared with the homogeneous baseline methods, we can find  that VRDBO and MDBO have a smaller communication cost. The reason is that they are designed for  homogeneous setting so that they do not need to communicate any information regarding Hessian or Jacobian. On the contrary, our methods, DSVRBGD-S and DSVRBGD-A,  should communicate an additional Hessian-inverse-vector product $z$ compared to these two baseline methods. Therefore, our communication cost is a little larger than them. Compared with the heterogeneous baseline methods, we can find that our two algorithms are much more communication-efficient than all of them, because these baseline methods should communicate the high-dimensional Hessian or Jacobian matrices.  In contrast, our methods only need to communicate the  low-dimensional Hessian-inverse-vector product $z$, saving communication costs significantly.



\begin{figure*}[h]
	\centering 
	\hspace{-15pt}
	\subfigure[a9a]{
		\includegraphics[scale=0.22]{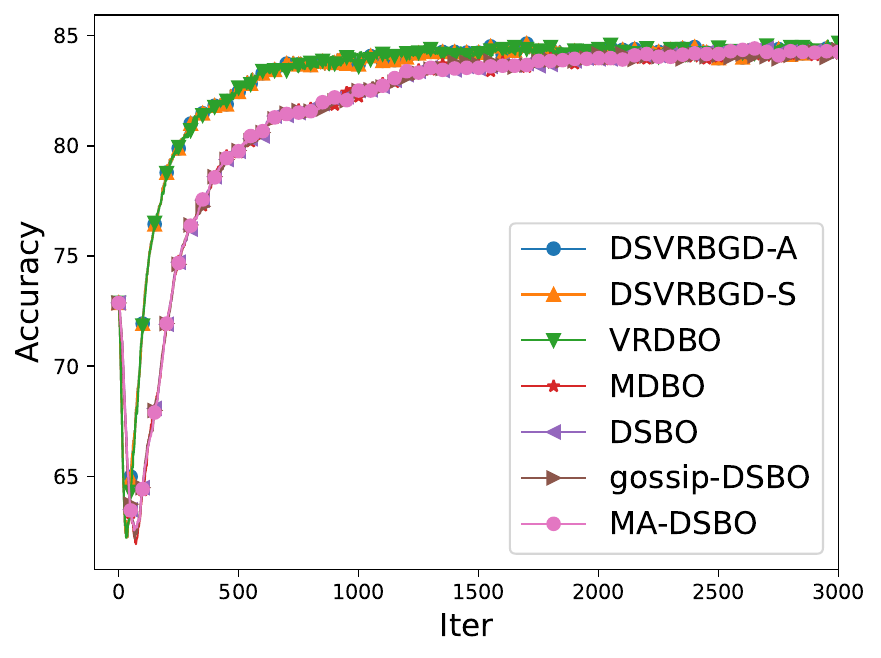}
	}
	\hspace{-12pt}
	\subfigure[w8a]{
		\includegraphics[scale=0.22]{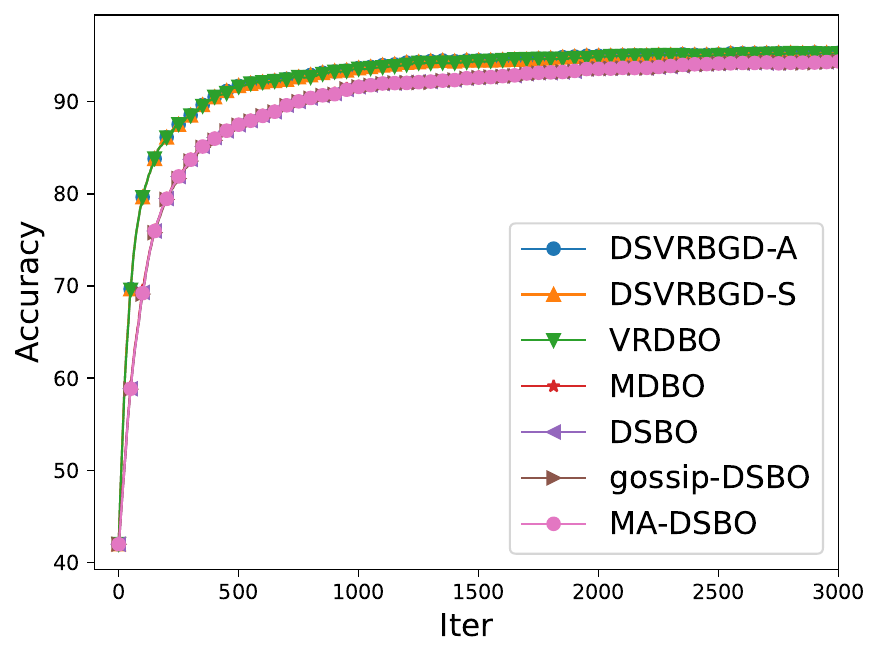}
	}
	\hspace{-12pt}
	\subfigure[ijcnn1]{
		\includegraphics[scale=0.22]{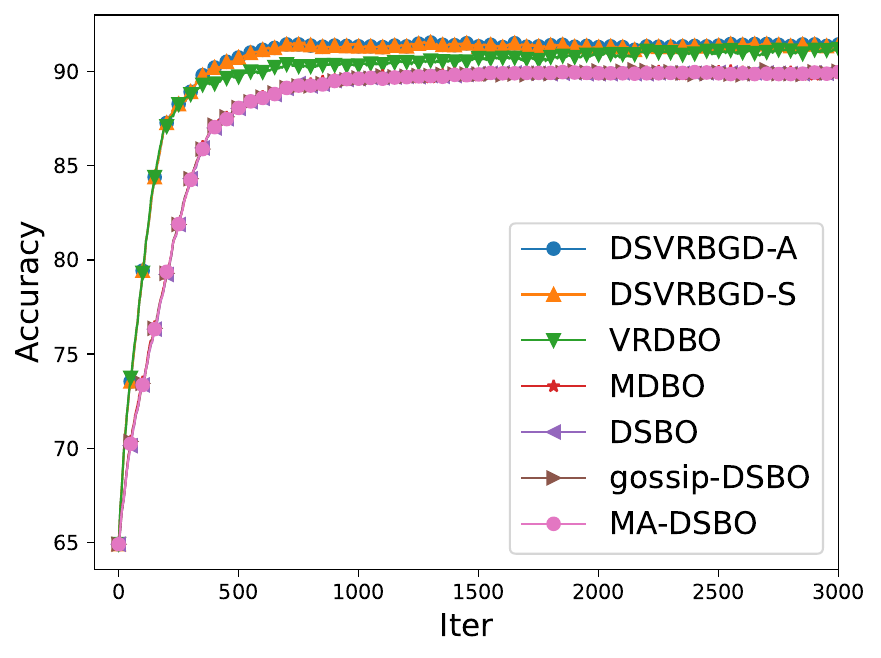}
	}
	\hspace{-12pt}
	\subfigure[covtype]{
		\includegraphics[scale=0.22]{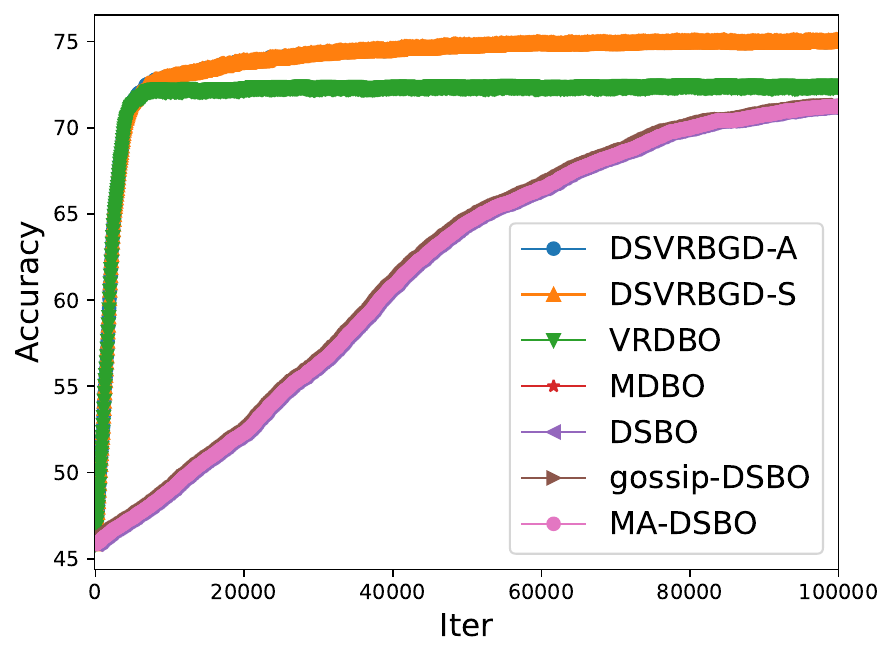}
	}
	\caption{The test accuracy  versus the number of iterations on a {ring} graph. }
	\label{real-world_acc-iteration-ring}
\end{figure*}

\begin{figure*}[h]
	\centering 
	\hspace{-15pt}
	\subfigure[a9a]{
		\includegraphics[scale=0.22]{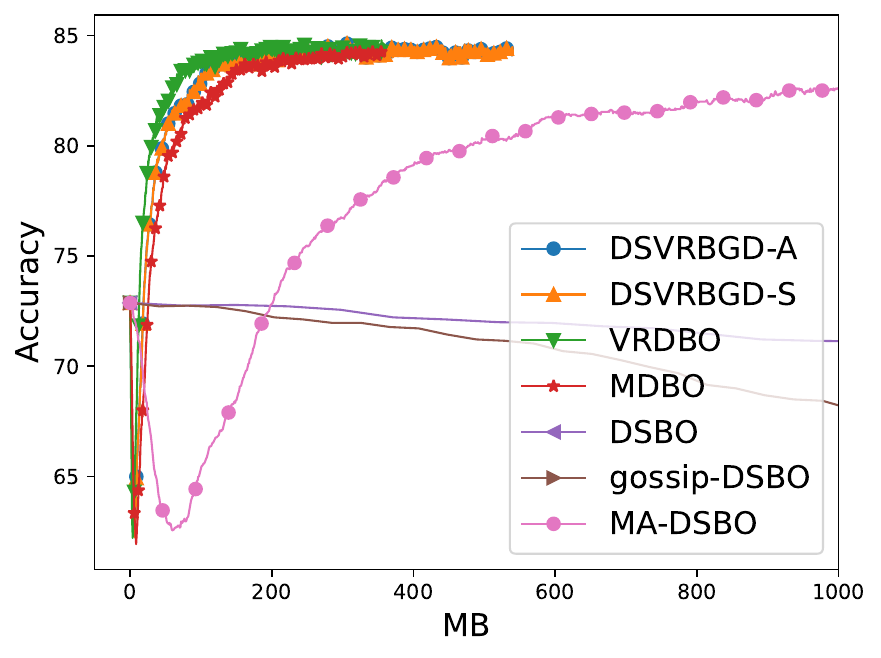}
	}
	\hspace{-11pt}
	\subfigure[w8a]{
		\includegraphics[scale=0.22]{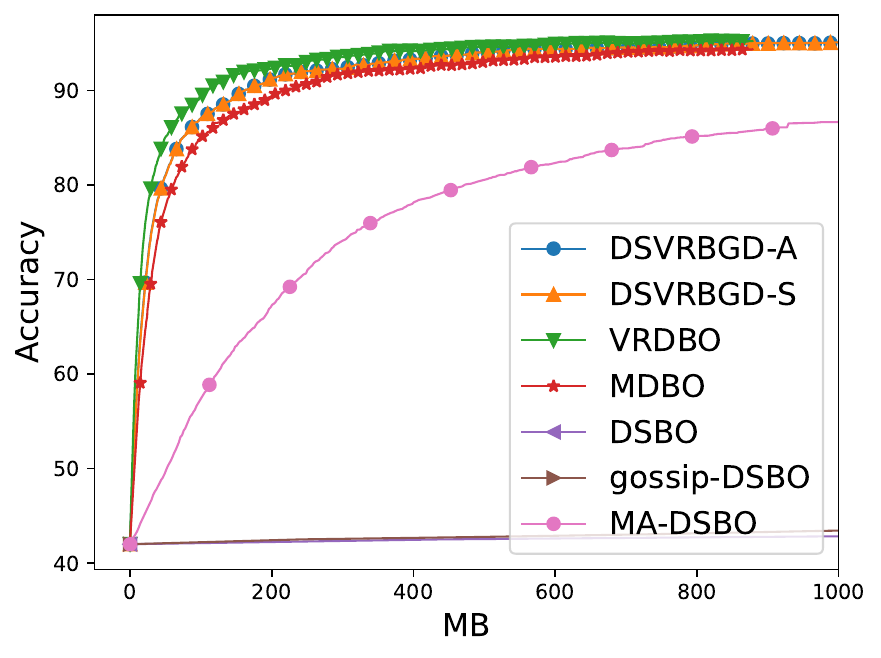}
	}
	\hspace{-11pt}
	\subfigure[ijcnn1]{
		\includegraphics[scale=0.22]{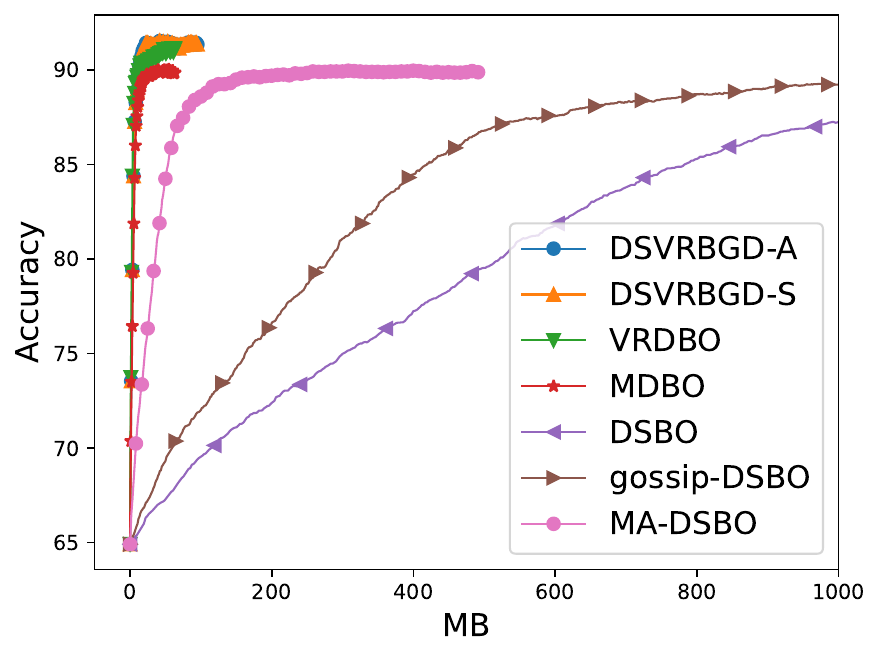}
	}
	\hspace{-11pt}
	\subfigure[covtype]{
		\includegraphics[scale=0.22]{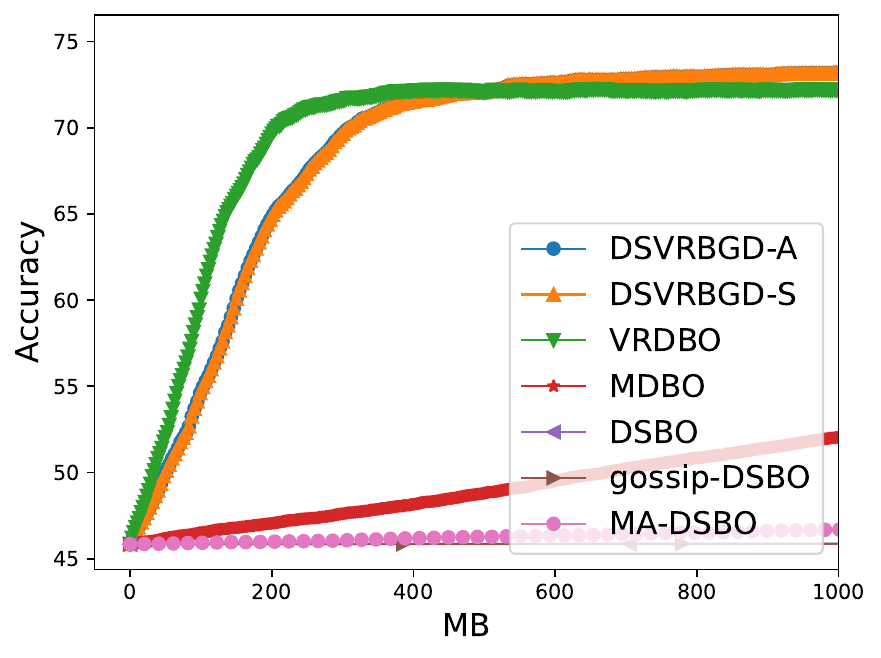}
	}
	\caption{The test accuracy  versus the communication cost (MB) on a {ring} graph. }
	\label{real-world_acc-communication-ring}
\end{figure*}

Moreover, in Figure~\ref{real-world_acc-iteration-ring}  and Figure~\ref{real-world_acc-communication-ring}, we plot the test accuracy versus the number of iterations  and the communication cost, respectively. We can still find that our methods enjoy small communication costs and converge faster than all baselines that are designed for the heterogeneous setting.

\begin{figure*}[!t]
	\centering 
	\hspace{-15pt}
	\subfigure[Loss vs. MB]{
		\includegraphics[scale=0.22]{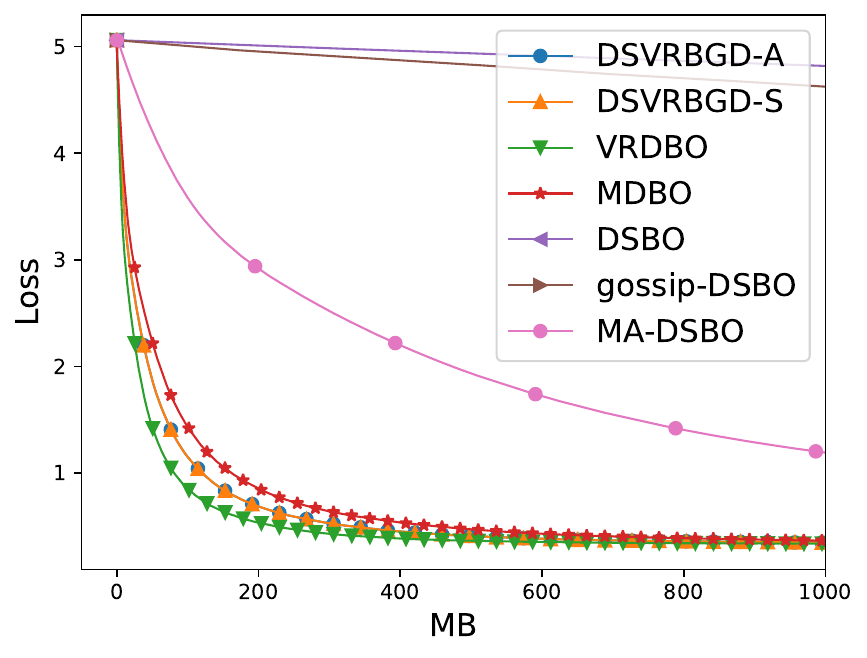}
	}
	\hspace{-11pt}
	\subfigure[Loss vs. Iteration]{
		\includegraphics[scale=0.22]{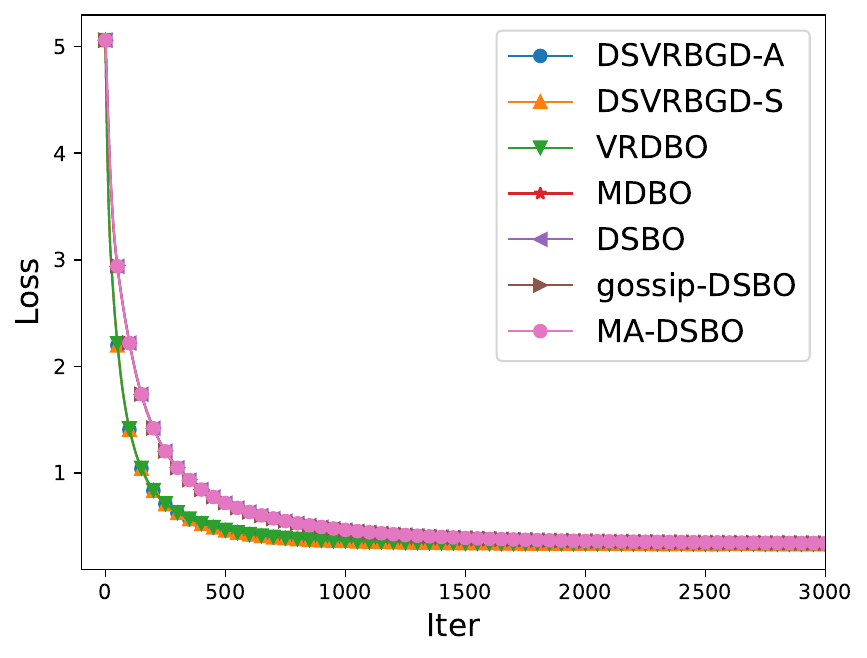}
	}
	\hspace{-11pt}
	\subfigure[Accuracy vs. MB]{
		\includegraphics[scale=0.22]{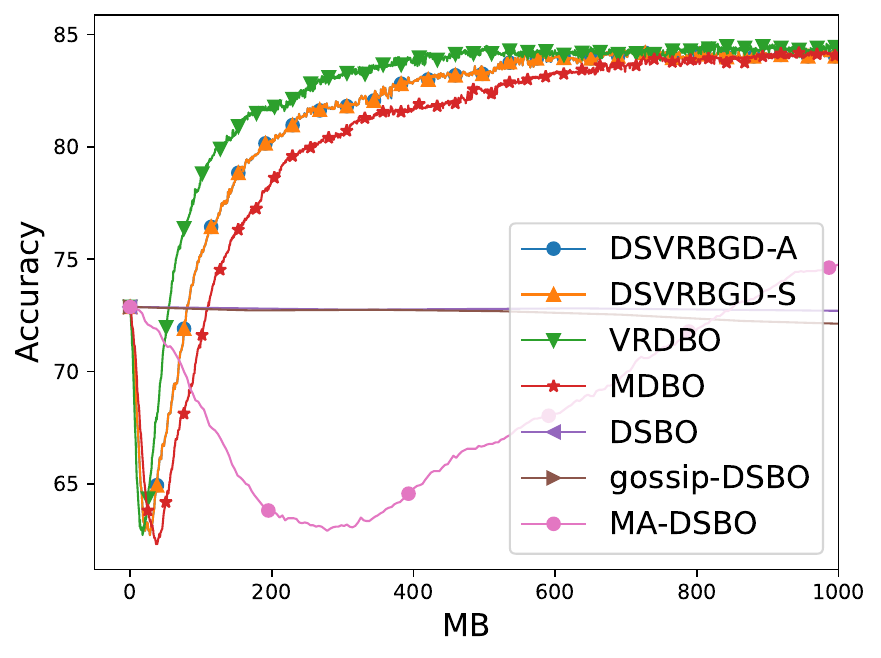}
	}
	\hspace{-11pt}
	\subfigure[Accuracy vs. Iteration]{
		\includegraphics[scale=0.22]{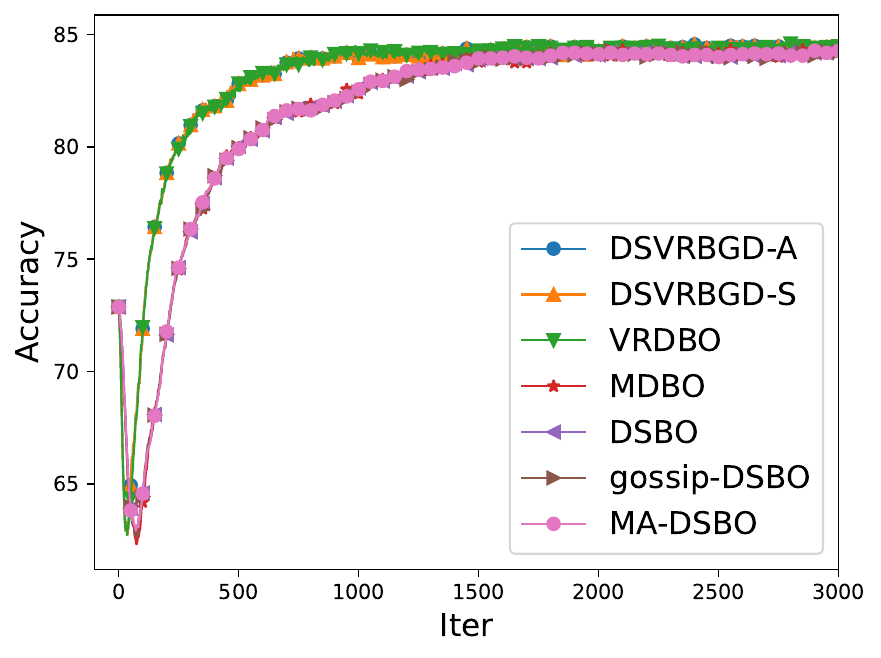}
	}\\
	\hspace{-15pt}
	\subfigure[Loss vs. MB]{
		\includegraphics[scale=0.22]{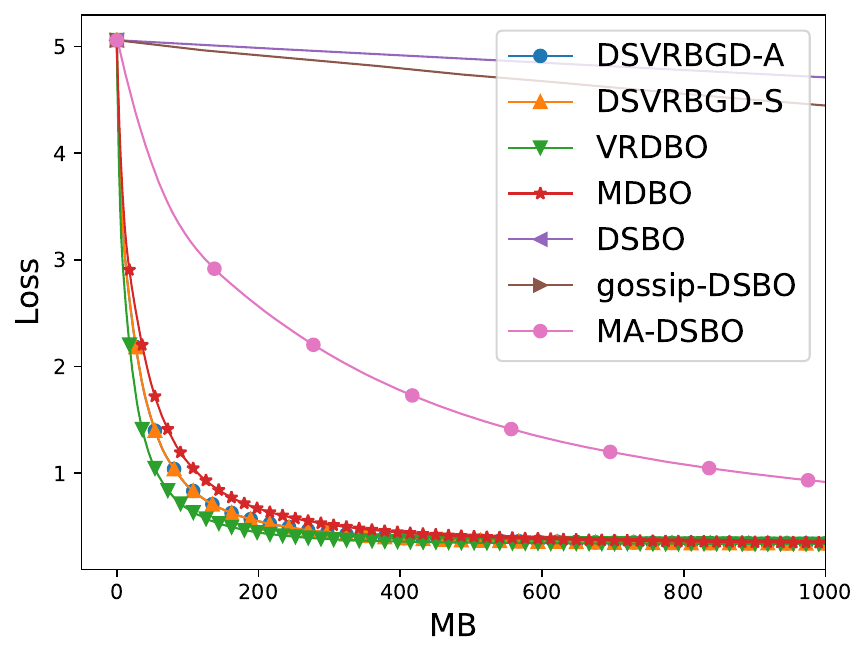}
	}
	\hspace{-11pt}
	\subfigure[Loss vs. Iteration]{
		\includegraphics[scale=0.22]{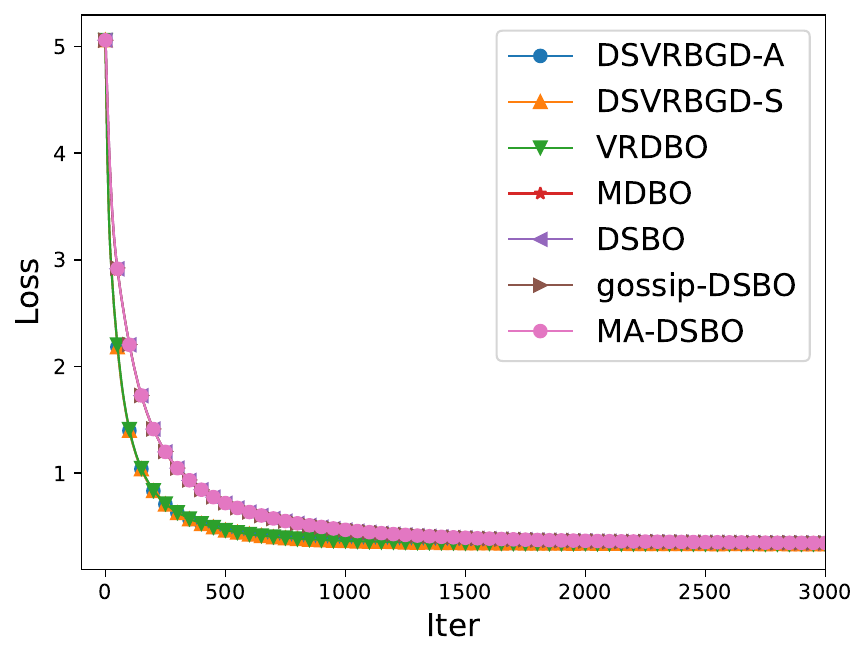}
	}
	\hspace{-11pt}
	\subfigure[Accuracy vs. MB]{
		\includegraphics[scale=0.22]{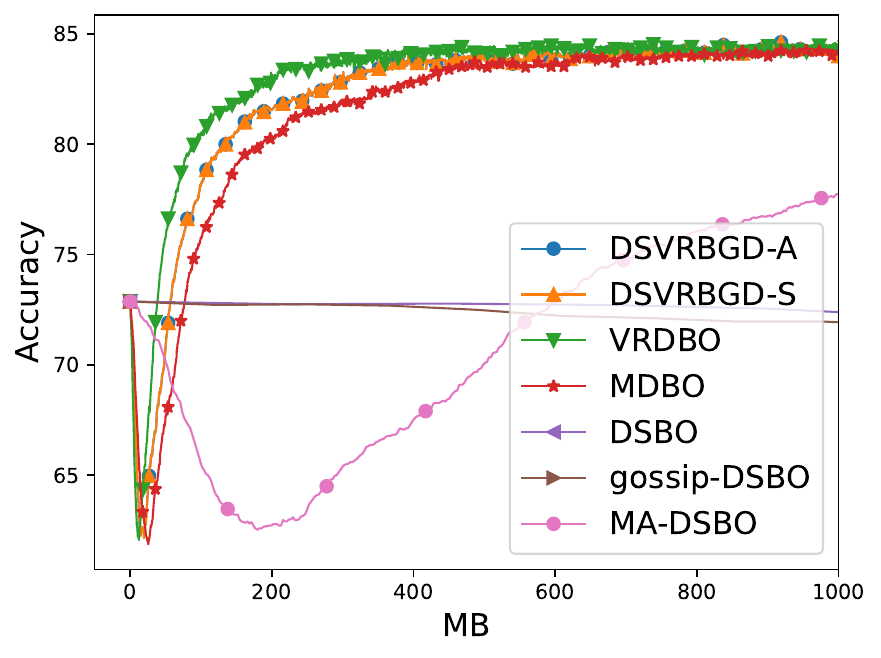}
	}
	\hspace{-11pt}
	\subfigure[Accuracy vs. Iteration]{
		\includegraphics[scale=0.22]{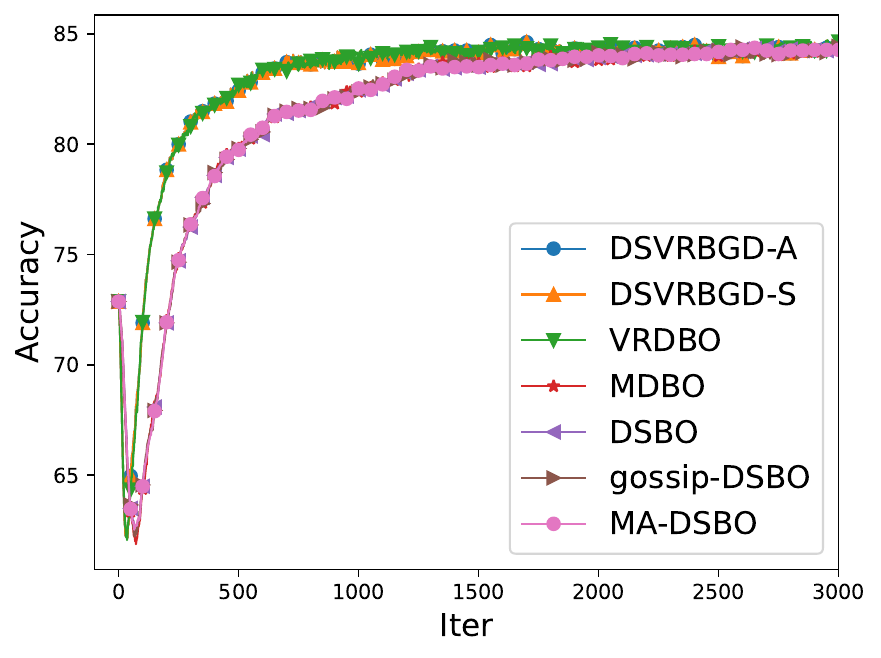}
	}
	\caption{The first row uses a random communication graph. The second row uses a torus communication graph. The datasets is a9a. }
	\label{a9a-topology}
\end{figure*}

\paragraph{Topology.}  In Figure~\ref{a9a-topology}, we plot the loss function value and test accuracy with respect to the communication cost and  the number of iterations on the random communication graph and the torus communication graph. Under these two settings, we can still find that our methods outperform all baseline methods designed for the heterogeneous setting.

\begin{figure*}[ht]
	\centering 
	\hspace{-15pt}
	\subfigure[a9a]{
		\includegraphics[scale=0.22]{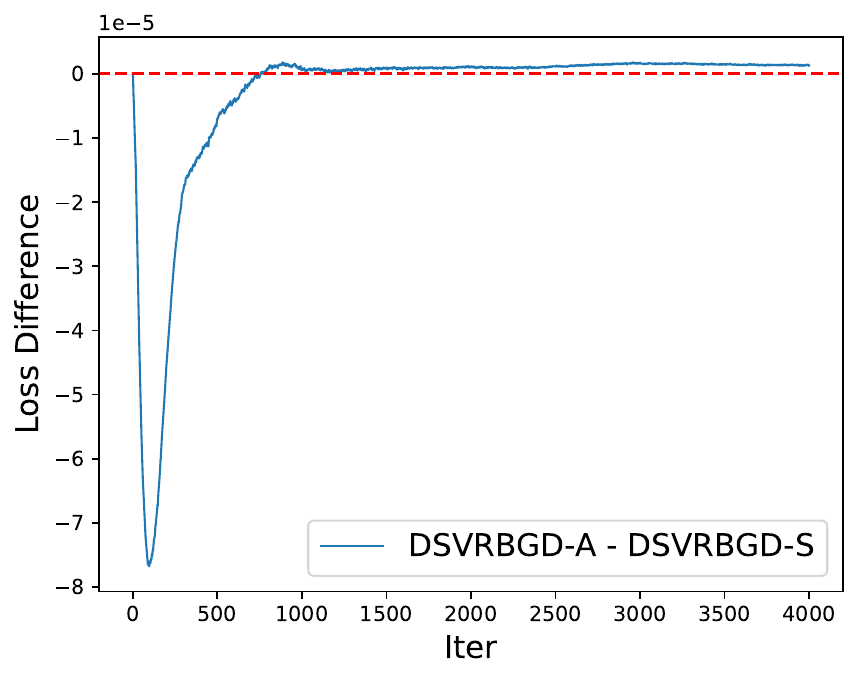}
	}
	\hspace{-11pt}
	\subfigure[w8a]{
		\includegraphics[scale=0.22]{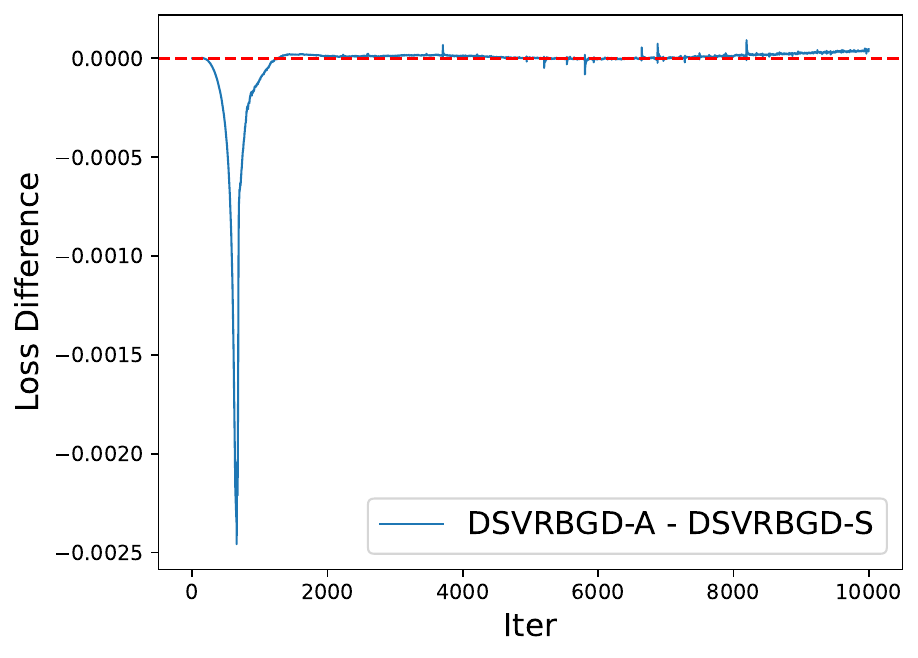}
	}
	\hspace{-11pt}
	\subfigure[ijcnn1]{
		\includegraphics[scale=0.22]{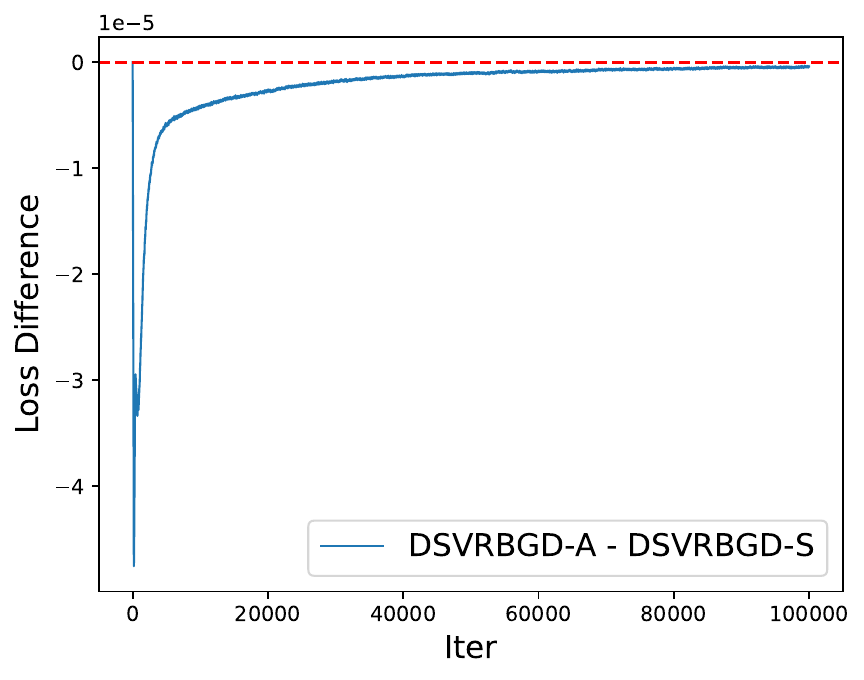}
	}
	\hspace{-11pt}
	\subfigure[covtype]{
		\includegraphics[scale=0.22]{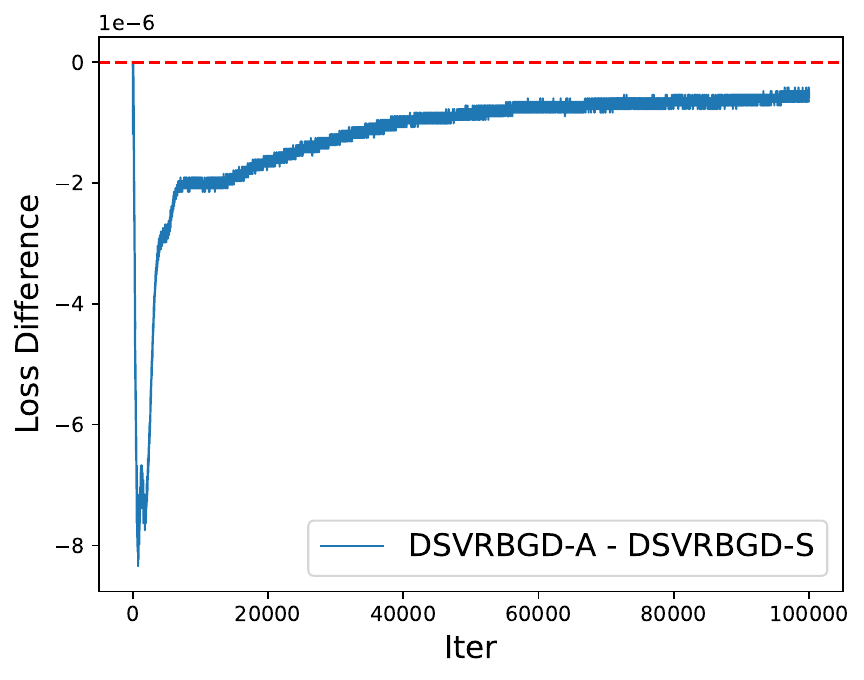}
	}
	\caption{The loss function value difference: $\Delta=\text{LOSS}_{\text{DSVRBGD-A}}-\text{LOSS}_{\text{DSVRBGD-S}}$.  The negative difference value denotes DSVRBGD-A converges faster than DSVRBGD-S, as it reduces the loss function value more quickly.}
	\label{sim-alt}
\end{figure*}

\paragraph{Simultaneous Update vs Alternating Update.}
In Figure~\ref{sim-alt}, we show the difference between the upper-level loss function value obtained from the simultaneous update (DSVRBGD-S) and the alternating update (DSVRBGD-A). In particular, we plot $\Delta=\text{Loss}_{\text{DSVRBGD-A}}-\text{Loss}_{\text{DSVRBGD-S}}$. When $\Delta<0$, DSVRBGD-A decreases the loss function value  faster than DSVRBGD-S. Otherwise, DSVRBGD-S converges faster. From Figure~\ref{sim-alt}, we can find that the alternating update (DSVRBGD-A) converges a  faster than the simultaneous update (DSVRBGD-S)  because $\Delta<0$.

\begin{figure*}[ht]
	\centering 
	\subfigure[Upper-level loss versus MB]{
		\includegraphics[scale=0.37]{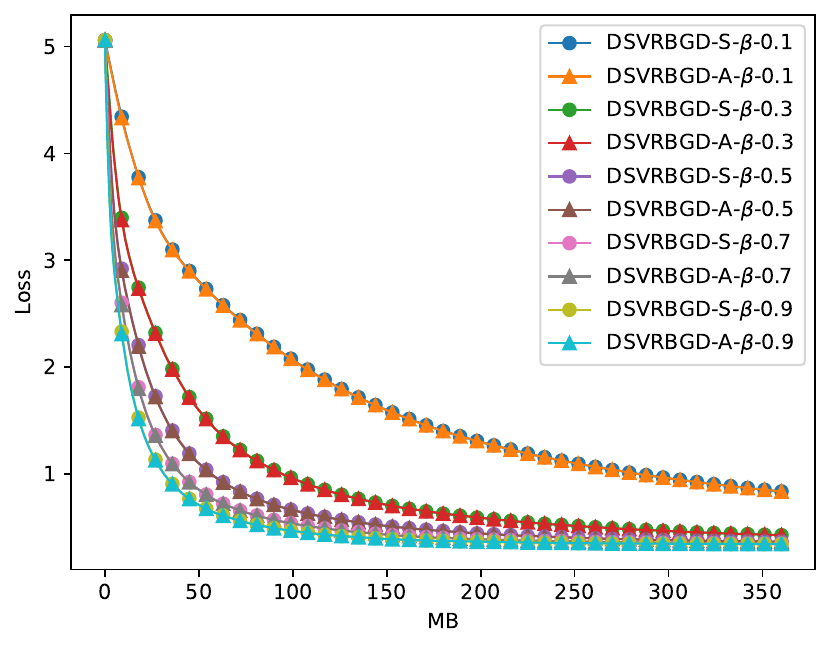}
	}
	\subfigure[Test Accuracy  versus MB]{
		\includegraphics[scale=0.37]{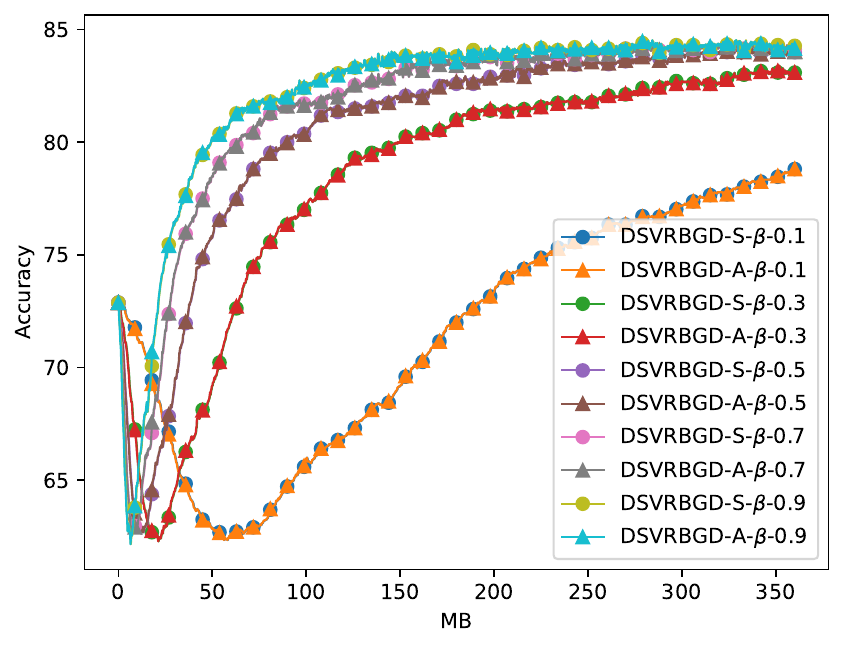}
	}
	\caption{The convergence performance of our  DSVRBGD-S and DSVRBGD-A for different values of $\beta_i$. The dataset is a9a.}
	\label{fig:hyper-beta}
\end{figure*}

\begin{figure*}[h]
	\centering 
	\subfigure[Upper-level loss versus MB]{
		\includegraphics[scale=0.37]{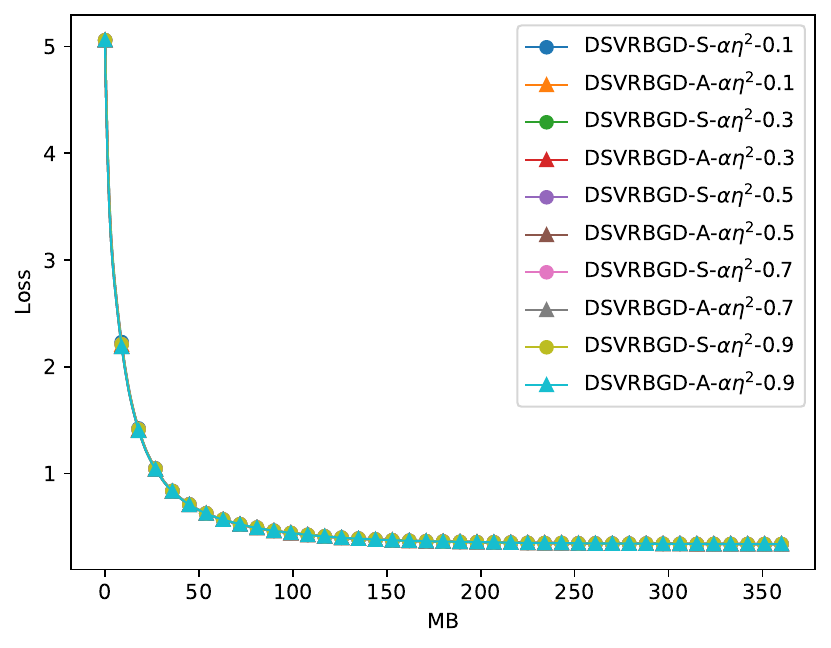}
	}
	\subfigure[Test Accuracy  versus MB]{
		\includegraphics[scale=0.37]{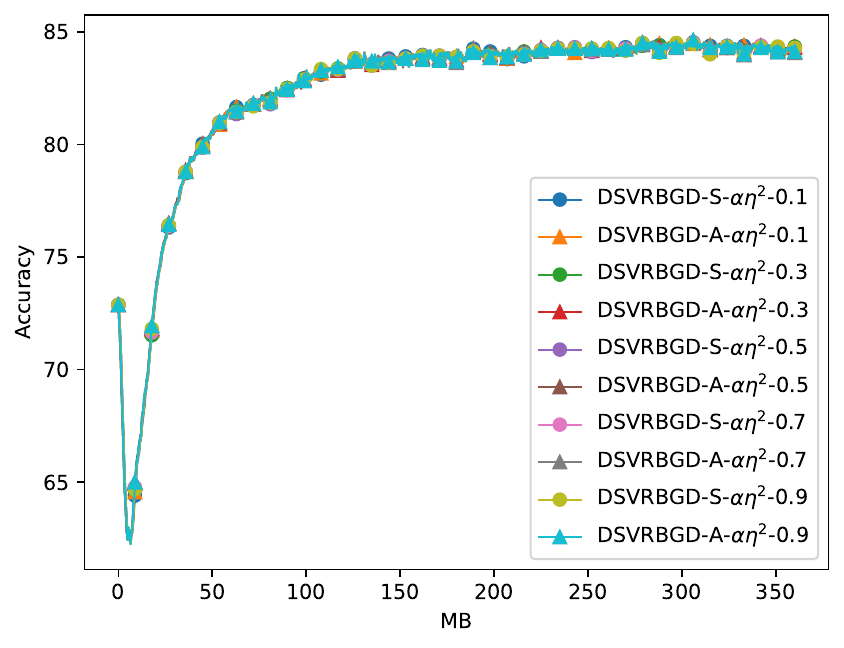}
	}
	\caption{The convergence performance of our  DSVRBGD-S and DSVRBGD-A for different values of $\alpha_i$. The dataset is a9a.}
	\label{fig:hyper-alpha}
\end{figure*}

\paragraph{Ablation Studies for Hyperparameters.}
We conduct experiments to show the performance of our methods with different hyperparameters, where the dataset is a9a.   In Figure~\ref{fig:hyper-beta}, we show the effect of $\beta_i$ (where $i\in\{1, 2, 3\}$) on the convergence performance of our DSVRBGD-S and DSVRBGD-A. Specifically, we set $\beta_i=\{0.1, 0.3, 0.5, 0.7, 0.9\}$ while fixing $\eta=0.05$. Since the actual learning rate is $\beta_i\eta$ from the global view as $\bar{X}_{t+1}=\bar{X}_{t}-\beta_1\eta \bar{P}_{t}$, a smaller $\beta_i$ results in a smaller learning rate. 
Therefore,  a smaller $\beta_i$ leads to a slower convergence rate as shown in Figure~\ref{fig:hyper-beta}. In Figure~\ref{fig:hyper-alpha}, we show the effect of $\alpha_i$ (where $i\in\{1, 2, 3\}$) on the convergence performance of our DSVRBGD-S and DSVRBGD-A. In this experiment, we fix $\eta=0.05$ and then vary $\alpha_i$ such that $\alpha_i\eta^2=\{0.1, 0.3, 0.5, 0.7, 0.9\}$. It can be observed that it does not affect the convergence rate significantly.

\begin{figure*}[h]
	\centering 
	\subfigure[Upper-level loss vs. MB]{
		\includegraphics[scale=0.37]{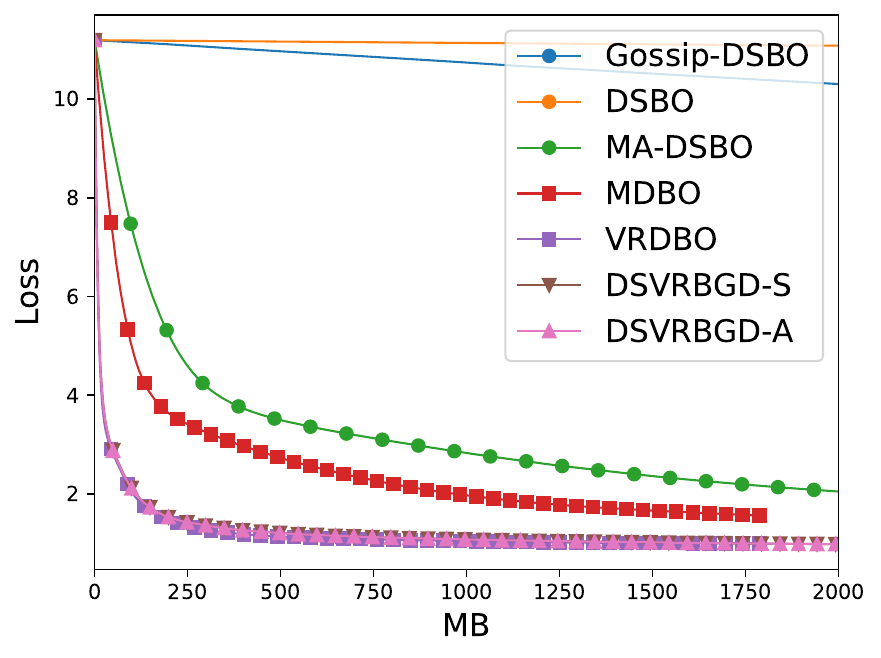}
	}
	\subfigure[Test Accuracy  vs. MB]{
		\includegraphics[scale=0.37]{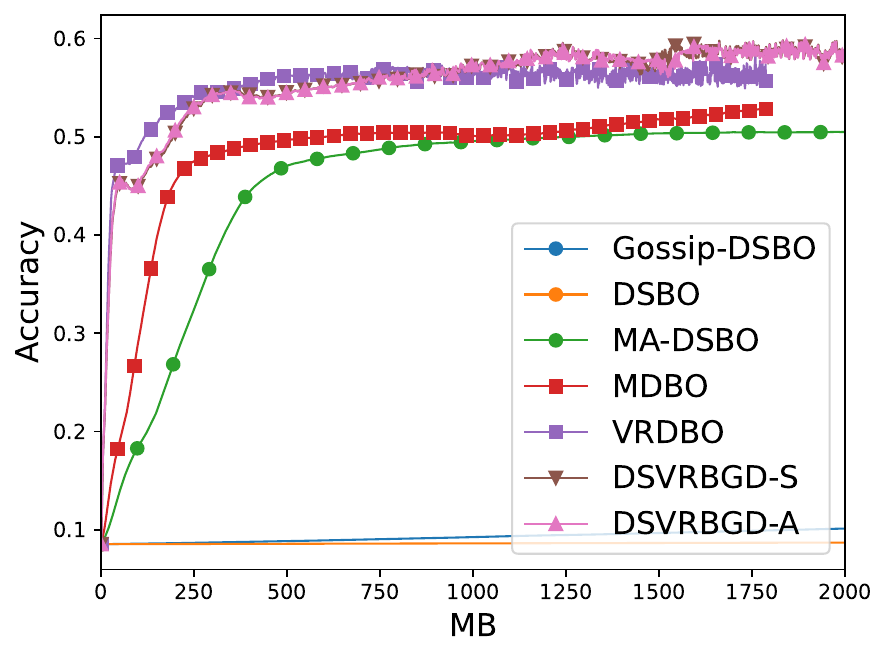}
	}\\
	\subfigure[Upper-level loss  vs. iteration]{
		\includegraphics[scale=0.37]{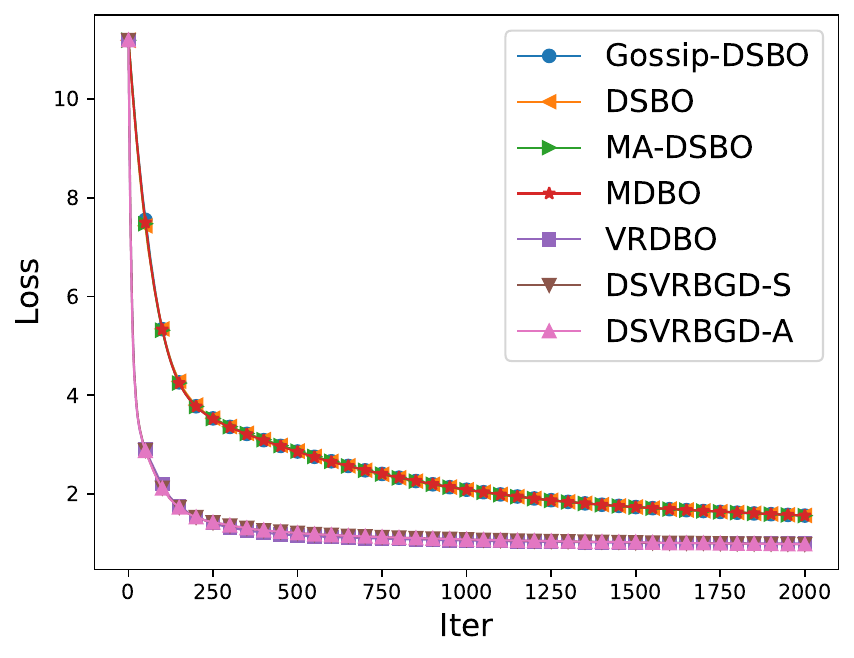}
	}
	\subfigure[Test Accuracy vs. iteration]{
		\includegraphics[scale=0.37]{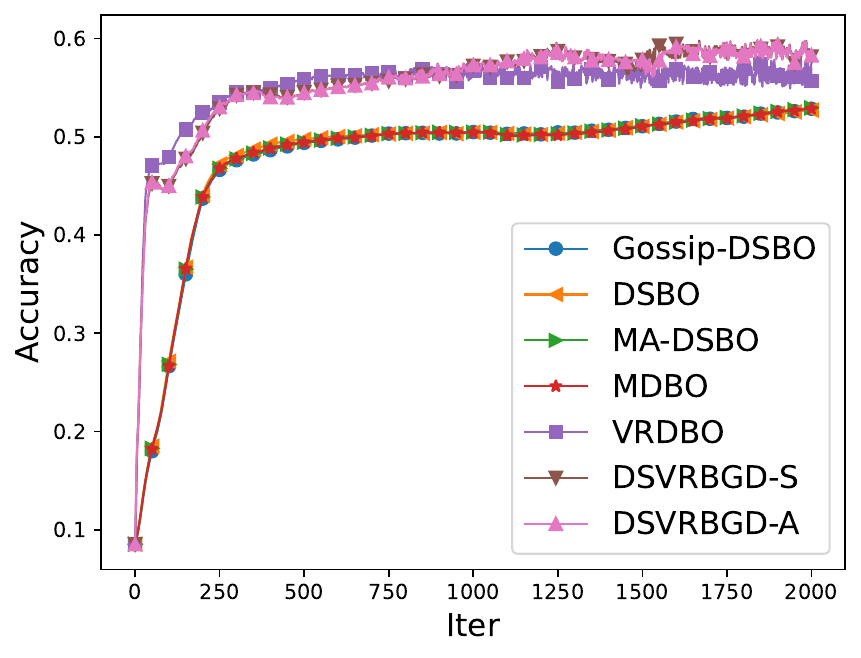}
	}
	\caption{The comparison of different methods for hyper-representation learning using the (multi-class) covtype dataset and a ring graph.}
	\label{real-world_loss-communication-ring-hyper-representation}
\end{figure*}

\subsection{Hyper-representation Learning}
In this additional experiment, by following FedNEST \cite{tarzanagh2022fednest}, we focus on the  hyper-representation optimization problem. Specifically, the upper-level optimization problem is to optimize the backbone neural network's weight to obtain better  features, while the lower-level optimization problem is to optimize the classifier's weight given the backbone neural network. In our experiment, we use a two-layer fully-connected neural network, whose hidden layer has 30 neurons. In this way, the upper-level optimization problem learns the weight of the first layer, which is nonconvex optimization problem. The lower-level optimization problem learns the weight of the last layer, which is actually a logistic regression problem  when fixing the first layer. It is a strongly convex problem when  adding a regularization term. In this experiment, we consider the multi-class classification dataset from LIBSVM \footnote{\url{https://www.csie.ntu.edu.tw/~cjlin/libsvmtools/datasets/multiclass.html}}: covtype, which has 7 classes, 581,012 samples, and 54 features. In this experiment, we still randomly select $10\%$ of samples as the test set, $70\%$ of the remaining samples as the training set, and the others as the validation set. In addition, the  solution accuracy $\epsilon$ is set to $0.02$. The other settings are set in the same way as our first experiment.

In Figure~\ref{real-world_loss-communication-ring-hyper-representation}, we plot the upper-level loss function value and test accuracy versus the communication cost and the number of iterations, respectively. It can be observed that our two methods converge much faster than all  state-of-the-art heterogeneous methods  in terms of both communication costs and iterations,  which confirms the effectiveness of our proposed methods. Note that the homogeneous methods, VRDBO and MDBO, have a smaller communication cost because they do not communicate Hessian or Jacobian matrices.

\section{Future Work}
	In our algorithms, we use the Gradient Tracking (GT) communication strategy. In fact, there are communication strategies other than GT. For example, SPARKLE \cite{zhu2024sparkle} studies the EXTRA and ED communication strategies for bilevel optimization, showing that they enjoy better transient iteration complexity than GT. Specifically, as shown in Table 1 of that paper, when the spectral gap satisfies $1-\rho < \frac{1}{n^3}$, i.e., when the communication graph is very sparse, EXTRA and ED perform better than GT in terms of transient complexity. However, applying these two communication strategies to our algorithms introduces significant challenges, and it is unclear whether they would lead to similar conclusions, as discussed below.

	SPARKLE \cite{zhu2024sparkle} algorithm uses  stochastic gradient for updating $y$ and $z$ and uses  momentum to update $x$. As a result, the dominating term $\frac{1}{\sqrt{nT}}$ (or $\frac{1}{n\epsilon^2}$ for achieving $\epsilon$-accuracy solution) in its convergence upper bound does not rely on the spectral gap $1-\rho$.  This leads to the notion of transient iteration complexity, which is defined as:  \textit{those iterations before an algorithm reaches its linear-speedup stage} \cite{chen2021accelerating}. However, our algorithms use the variance-reduced gradient  estimator STORM to update all variables. As a result, the dominating term in the convergence upper bound depends on the spectral gap, i.e., $O\left(\frac{1}{K(1-\lambda)^4\epsilon^{3/2}}\right)$. Therefore, according to \cite{chen2021accelerating}, the transient iteration complexity is not well defined when there exists a spectral gap in the dominating term. Consequently, the challenge for studying the influence of different communication topology  is to establish  a convergence rate whose dominating term is independent of the spectral gap. In fact, this is still an open problem even for the traditional single-level optimization algorithm.  For example, as shown in Table 1 of \cite{xin2021hybrid}, which  uses the STORM gradient estimator for decentralized minimization problems, the iteration complexity relies on the spectral gap under the worst case. In summary, the advantage of EXTRA and ED over GT is reflected by the transient iteration complexity. However, this notion is not well defined for algorithms that rely on variance-reduced gradient estimators, since the dominating term in their convergence upper bounds depends on the spectral gap. 

Since comparing different communication strategies for variance-reduced gradient–based methods involves an open problem and is nontrivial to address, we leave this for future work.

\section{Conclusion}

In this paper, we analyzed the convergence rate of decentralized stochastic bilevel optimization algorithms under the heterogeneous setting. In particular, to reduce the communication rounds in each iteration, we proposed to employ the variance-reduced gradient descent on each worker to estimate the Hessian-inverse-vector product. As a result, our algorithm can achieve a small number of iterations. Meanwhile, it can reduce the communication rounds in each iteration and also reduce the cost in each round. In addition, we develop a new algorithm based on the alternating update strategy, which also enjoy these nice empirical and theoretical properties. The extensive experimental results confirm the effectiveness of our two algorithms.

\bibliographystyle{abbrv}
\bibliography{sample-base}

\appendix

\section{Useful Lemmas}

\begin{lemma} \cite{ghadimi2018approximation} \label{lemma_hypergrad_smooth_optimal_var}
	Under Assumptions~\ref{assumption_bi_strong}-\ref{assumption_graph}, we have
	\begin{equation}
		\begin{aligned}
			& \| \nabla F^{(k)}(x_1) - {\nabla} F^{(k)}(x_2)    \| \leq  {L}_{F}\|x_1 - x_2\|  \ ,  \\
			& \|y^*(x_1)-y^*(x_2)\| \leq L_y \|x_1 - x_2\| \ , 
		\end{aligned}
	\end{equation}
	where ${L}_F=(1+\frac{c_{g_{xy}}}{\mu})(\ell_{f_x}+\frac{\ell_{f_y}c_{g_{xy}}}{\mu}+\frac{c_{f_y}\ell_{g_{xy}}}{\mu}+\frac{\ell_{g_{yy}}c_{f_{y}}c_{g_{xy}}}{\mu^2})$, $L_y=\frac{c_{g_{xy}}}{\mu}$.
\end{lemma}

\begin{lemma} \label{lemma:z-star-upper-bound}
	Under Assumptions~\ref{assumption_bi_strong}-\ref{assumption_graph}, we have
	\begin{equation}
		\begin{aligned}
			& \|z^*(x)\| \leq \frac{c_{f_y}}{\mu}  \ , 
			& \|z^*(x_1)-z^*(x_2)\| \leq L_z \|x_1 - x_2\| \ , 
		\end{aligned}
	\end{equation}
	where $L_z=(\frac{\ell_{f_y}}{\mu} +\frac{c_{f_y}\ell_{g_{yy}}}{\mu^2})(1+L_y)$. 
\end{lemma}
\begin{proof}
	According to the definition of $z^*(x)$, it is easy to know
	\begin{equation}
		\begin{aligned}
			& \|z^*(x)\| = \left\|\left[\nabla_{22}^2g(x, y^*(x))\right]^{-1} \nabla_2{ f(x, y^*(x))}\right\| \leq \frac{c_{f_y}}{\mu} \ . 
		\end{aligned}
	\end{equation}
	As for the second inequality, we have
	\begin{equation}
		\begin{aligned}
			& \quad \|z^*(x_1)-z^*(x_2)\| \\
			& = \|\left[\nabla_{22}^2g(x_1, y^*(x_1))\right]^{-1} \nabla_2{ f(x_1, y^*(x_1))}- \left[\nabla_{22}^2g(x_2, y^*(x_2))\right]^{-1} \nabla_2{ f(x_2, y^*(x_2))}\| \\
			& \leq \|\left[\nabla_{22}^2g(x_1, y^*(x_1))\right]^{-1} \nabla_2{ f(x_1, y^*(x_1))} - \left[\nabla_{22}^2g(x_1, y^*(x_1))\right]^{-1} \nabla_2{ f(x_2, y^*(x_2))} \|\\
			& \quad + \|\left[\nabla_{22}^2g(x_1, y^*(x_1))\right]^{-1} \nabla_2{ f(x_2, y^*(x_2))}- \left[\nabla_{22}^2g(x_2, y^*(x_2))\right]^{-1} \nabla_2{ f(x_2, y^*(x_2))}\| \\
			& \leq \frac{\ell_{f_y}}{\mu} (1+L_y)\|x_1 - x_2\| + \frac{c_{f_y}\ell_{g_{yy}}}{\mu^2} (1+L_y)\|x_1 - x_2\|\\
			& = (\frac{\ell_{f_y}}{\mu} +\frac{c_{f_y}\ell_{g_{yy}}}{\mu^2})(1+L_y)\|x_1 - x_2\|  \ . 
		\end{aligned}
	\end{equation}
\end{proof}

\section{Proof Sketch} \label{sec:proof-sketch}
In this section, we provide the proof sketch of Theorem~\ref{theorem-sim} and Theorem~\ref{theorem-alt}. In particular, we first show  the hypergradient bias and then demonstrate how to control this bias by different classes of estimation errors. 

\subsection{Proof Sketch of Theorem~\ref{theorem-sim}}

At first, since we employ a biased hypergradient on each worker to do updates, we establish the upper bound of the hypergradient's estimation bias   in Lemma~\ref{lemma_hypergrad_bias-sim}. In particular, we have 
		\begin{align}\label{lemma_bias-sim}
			&\quad \mathbb{E}\left[\left\|\nabla F(\bar{x}_{t}) -    \frac{1}{K}\delta^{\hat{\mathcal{G}}_{F}} (X_t, Y_t, Z_t) \mathbf{1}  \right\|^2\right]  \notag \\
			& \leq 6 (\ell_{f_x}^2 + \frac{c_{f_y}^2\ell_{g_{xy}}^2}{\mu^2})\mathbb{E}\left[\left\| y^*(\bar{x}_t) - \bar{y}_t\right\|^2\right]   + 6c_{g_{xy}}^2 \mathbb{E}\left[\left\|z^*(\bar{x}_t) - \bar{z}_t \right\|^2\right] \notag \\
			&  \quad + 6(\ell_{f_x}^2+\frac{c_{f_y}^2\ell_{g_{xy}}^2}{\mu^2} )\frac{1}{K}\mathbb{E}\left[\left\|X_{t} - \bar{X}_{t}\right\|_F^2\right] + 6(\ell_{f_x}^2+\frac{c_{f_y}^2\ell_{g_{xy}}^2}{\mu^2} ) \frac{1}{K}\mathbb{E}\left[\left\|Y_{t} - \bar{Y}_{t}\right\|_F^2\right]  \notag \\
			& \quad  + 6c_{g_{xy}}^2\frac{1}{K}\mathbb{E}\left[\left\|Z_{t}- \bar{Z}_{t}\right\|_F^2\right]  \ . 
		\end{align}
It can be observed that the  hypergradient's estimation bias  depends on the optimization errors: $\| y^*(\bar{x}_t) - y_t\|^2$ and $\|z^*(\bar{x}_t) - \bar{z}_t \|^2$, and the consensus errors: $\left\|X_{t} - \bar{X}_{t}\right\|_F^2$, $\left\|Y_{t} - \bar{Y}_{t}\right\|_F^2$, and $\left\|Z_{t}- \bar{Z}_{t}\right\|_F^2$.  Based on this observation, we will bound these two classes of errors. In the following, we will just show one example for each error to save space.

\paragraph{Step 1: Bounding Optimization Error for $z$ and $y$.}
We bound the estimation error about $ \|\bar{z}_{t}   - z^{*}(\bar{{x}}_t)\| ^2$ in Lemma~\ref{lemma_z_opt} and $ \|\bar{y}_{t}   - y^{*}(\bar{{x}}_t)\| ^2$ in Lemma~\ref{lemma_y_opt}. 

 For instance, $ \|\bar{z}_{t+1}   - z^{*}(\bar{{x}}_{t+1})\| ^2$  can be bounded as follows:
		\begin{align}\label{lemma_z_opt-sim-main}
			&   \mathbb{E} [ \|\bar{z}_{t+1} - z^{*}(\bar{{x}}_{t+1})\| ^2]  \leq  (1-\frac{\eta\beta_3\mu}{8}) \mathbb{E} [ \|\bar{z}_t - z^*(\bar{x}_{t})\|^2 ]+ \frac{9\eta\beta_1^2L_z^2}{\beta_3\mu} \mathbb{E} [ \|\bar{u}_{t}\|^2]  \notag \\
			& \quad + \frac{9\eta\beta_3}{\mu}\Big(\frac{c_{f_y}^2\ell_{g_{yy}}^2}{\mu^2}+\ell_{f_y}^2\Big) \mathbb{E} [ \| \bar{y}_t  - y^*(\bar{x}_t)\|^2]  + \frac{70\eta\beta_3}{\mu}\Big(\frac{c_{f_y}^2\ell_{g_{yy}}^2}{\mu^2} + \ell_{f_y}^2\Big)\frac{1}{K}\mathbb{E} [ \|\bar{X}_{t} - {X}_{t}\|_F^2]  \notag \\
			& \quad  + \frac{70\eta\beta_3}{\mu}\Big(\frac{c_{f_y}^2\ell_{g_{yy}}^2}{\mu^2}+ \ell_{f_y}^2\Big)\frac{1}{K}\mathbb{E} [ \| \bar{Y}_t - {Y}_{t}\|_F^2] +\Big(\frac{70\eta\beta_3}{\mu}\ell_{g_y}^2+ \textcolor{red}{ \frac{9}{4}}\Big)\frac{1}{K}\mathbb{E} [ \|{Z}_t- \bar{Z}_{t}\|_F^2]    \notag \\
			& \quad + \frac{9\eta\beta_3^2}{4}\frac{1}{K}\mathbb{E} [ \|   R_t -   \bar{R}_t \|_F^2] +\frac{25\eta\beta_3}{\mu} \mathbb{E} [ \| \frac{1}{K}\delta^{\hat{\mathcal{G}}_{h}}({X}_{t}, {Y}_{t},  {Z}_t)\mathbf{1}-   \frac{1}{K}{W}_{t}\mathbf{1}\|^2]   \ . 
		\end{align}
Here, it can be observed that the optimization error $\|\bar{z}_{t+1} - z^{*}(\bar{{x}}_{t+1})\| ^2$ depends on the optimization error itself in the prior iteration and the optimization error regarding $y$: $\| y^*(\bar{x}_t)-  \bar{y}_t \|^2 $, which makes the final convergence analysis challenging. 
Moreover, it  also depends on the gradient estimation error and consensus error. Thus, we should also bound these two  classes of  errors. 
It is worth noting that the consensus error $\textcolor{red}{ \frac{9}{4}}\mathbb{E} [ \|{Z}_t- \bar{Z}_{t}\|_F^2]$ is caused by the projection optimization as discussed in Section~\ref{sec:challenges}, which will lead to a high order dependence on the spectral gap for $\beta_3$ and the final convergence rate. 

\paragraph{Step 2: Bounding Estimation Error  for  Gradient.}  We bound the gradient estimation error for $U$, $V$, $W$ in Lemma~\ref{lemma_hyper_storm_var_mean}, Lemma~\ref{lemma_g_storm_var_mean}, and Lemma~\ref{lemma_h_storm_var_mean}, respectively. 

For instance, we bound $\mathbb{E} [ \|\frac{1}{K}\delta^{\hat{\mathcal{G}}_{F}}(X_t, Y_t, Z_t) \mathbf{1} -  \frac{1}{K} U_t \mathbf{1} \|^2 ]$ as follows:
	\begin{align}\label{eq:var-u-mean-main-sim}
	& \quad\frac{1}{K}\mathbb{E} \Big[ \left\|\delta^{\hat{\mathcal{G}}_{F}}(X_t, Y_t, Z_t) -   U_t  \right\|_F^2 \Big]  \notag \\
	& \leq  (1-\alpha_1 \eta^2) \frac{1}{K}\mathbb{E} \Big[ \left\| \delta^{\hat{\mathcal{G}}_{F}}(X_{t-1}, Y_{t-1}, Z_{t-1}) - U_{t-1}  \right\|_F^2 \Big]\notag \\
	& \quad + 6(\ell_{f_x}^2+\frac{c_{f_y}^2\ell_{g_{xy}}^2}{\mu^2})\frac{1}{K}\mathbb{E}[\left\|X_{t} - X_{t-1}\right\|_F^2] + 6(\ell_{f_x}^2+\frac{c_{f_y}^2\ell_{g_{xy}}^2}{\mu^2})\frac{1}{K}\mathbb{E}[\left\|Y_{t} - Y_{t-1}\right\|_F^2]\notag \\
	& \quad + 6c_{g_{xy}}^2\frac{1}{K}\mathbb{E}[\left\|Z_{t} - Z_{t-1}\right\|_F^2] + 4\alpha_1^2\eta^4 (1 + \frac{c_{f_y}^2}{\mu^2})\sigma^2 \ . 
\end{align}

\paragraph{Step 3: Bounding  Consensus Error for Variable and Gradient.} 
We bound the consensus error for the variable $x$, $y$, and $z$ in Lemma~\ref{lemma_consensus_x} and the consensus error for the gradient $P$, $Q$, and $R$ in Lemma~\ref{lemma_consensus_p}, Lemma~\ref{lemma_consensus_q}, and Lemma~\ref{lemma_consensus_r}, respectively. 

For instance, we bound $\mathbb{E}[\|X_{t+1} - \bar{X}_{t+1}\|_F^2]$ as follows:
		\begin{align}
			& \quad\mathbb{E}[\|X_{t+1} - \bar{X}_{t+1}\|_F^2] \leq  \Big(1-\frac{\eta(1-\lambda^2)}{2}\Big)\mathbb{E}[\|X_t -\bar{X}_t\|_F^2]+ \frac{2\eta \beta_1^2}{1-\lambda^2}\mathbb{E}[\|P_t- \bar{ P}_t\|_F^2]  \ .
		\end{align}

This upper bound indicates we should bound the consensus error regarding the gradient $\|P_t- \bar{ P}_t\|_F^2$, which is shown below:
		\begin{align}
			& \quad  \frac{1}{K}\mathbb{E}[\|P_{t} - \bar{P}_{t}\|_F^2] \notag \\
			&  \leq  \lambda\frac{1}{K}\mathbb{E}[\| P_{t-1} -  \bar{P}_{t-1} \|_F^2] +  \frac{3\alpha_1^2\eta^4}{1-\lambda}\frac{1}{K}\mathbb{E}[\| U_{t-1}- \delta^{\hat{\mathcal{G}}_{F}}(X_{t-1}, Y_{t-1}, Z_{t-1})   \|_F^2]\notag \\
			& \quad  + (\ell_{f_x}^2+\frac{c_{f_y}^2\ell_{g_{xy}}^2}{\mu^2})\frac{9}{1-\lambda}\frac{1}{K} \mathbb{E}[\|X_{t} - X_{t-1}\|_F^2]\notag \\
			& \quad + (\ell_{f_x}^2+\frac{c_{f_y}^2\ell_{g_{xy}}^2}{\mu^2})\frac{9}{1-\lambda}\frac{1}{K} \mathbb{E}[\|Y_{t} - Y_{t-1}\|_F^2]\notag \\
			& \quad + c_{g_{xy}}^2\frac{9}{1-\lambda}\frac{1}{K} \mathbb{E}[\|Z_{t} - Z_{t-1}\|_F^2] +  \frac{6\alpha_1^2\eta^4}{1-\lambda}(1 + \frac{c_{f_y}^2}{\mu^2})\sigma^2 \ .  
		\end{align}

\paragraph{Step 4: Construct Potential Function.}
According to different classes of  errors mentioned above, we can observe that they are dependent on each other, which makes the convergence analysis challenging. To address this challenge, we construct a novel potential function as follows:
	\begin{align}\label{eq:potential-function}
		& \mathcal{L}_{t} = {\mathbb{E}}[F(x_{t})] +  c_0\mathbb{E}[\|\bar{   {y}}_{t} -    {y}^{*}(\bar{   {x}}_{t})\| ^2 ]  +  c_1\mathbb{E}[\|\bar{ z}_{t} -    {z}^{*}(\bar{   {x}}_{t})\| ^2 ]   \notag \\
		& \quad + c_2\frac{1}{K} \mathbb{E}[\|X_{t} - \bar{X}_{t}\|_F^2 ] + c_3 \frac{1}{K}\mathbb{E}[\|Y_{t} - \bar{Y}_{t}\|_F^2 ]  +c_4\frac{1}{K}\mathbb{E}[\| Z_{t}-\bar{Z}_{t} \|_F^2] \notag \\
		& \quad + c_5\frac{1}{K} \mathbb{E}[\|P_{t} - \bar{P}_{t}\|_F^2 ] + c_6 \frac{1}{K}\mathbb{E}[\|Q_{t} - \bar{Q}_{t}\|_F^2 ]  +c_7\frac{1}{K}\mathbb{E}[\| R_{t}-\bar{R}_{t} \|_F^2] \notag \\
		& \quad + c_8 \mathbb{E} [ \|\frac{1}{K}\delta^{\hat{\mathcal{G}}_{F}}(X_{t}, Y_{t}, Z_{t}) \mathbf{1}  -  \frac{1}{K} U_{t} \mathbf{1} \|^2 ]+ c_9 \frac{1}{K}\mathbb{E} [ \|\delta^{\hat{\mathcal{G}}_{F}}(X_{t}, Y_{t}, Z_{t}) -  U_{t} \|_F^2 ] \notag \\
		& \quad  + c_{10}\mathbb{E} [ \|\frac{1}{K}\delta^{g}(X_{t}, Y_{t}) \mathbf{1} -  \frac{1}{K} V_{t} \mathbf{1} \|^2 ] + c_{11} \frac{1}{K}\mathbb{E} [ \|\delta^{g}(X_{t}, Y_{t}) - V_{t}  \|_F^2 ]\notag \\
		& \quad  + c_{12} \mathbb{E} [ \|\frac{1}{K}\delta^{\hat{\mathcal{G}}_{h}}(X_{t}, Y_{t}, Z_{t})\mathbf{1}  -  \frac{1}{K} W_{t} \mathbf{1} \|^2 ] +   c_{13} \frac{1}{K}\mathbb{E} [ \|\delta^{\hat{\mathcal{G}}_{h}}(X_{t}, Y_{t}, Z_{t}) -   W_{t}  \|_F^2 ] \ ,
	\end{align}
where $\{c_i\}_{i=0}^{13}$ are constant values, which will be determined in the main proof. 
Then, we study how this potential function evolves across iterations. Specifically, we have
	\begin{align}
		&   \mathcal{L}_{t+1} -  \mathcal{L}_{t} \leq - \frac{\beta_1\eta}{2}\mathbb{E} [ \| \nabla F(\bar{x}_{t})\|^2 ]+  6\beta_1\alpha_1^2\eta^4(1 + \frac{c_{f_y}^2}{\mu^2})\sigma^2 +  6\beta_1\alpha_2^2\eta^4\sigma^2 \notag \\
		&+ 4\alpha_1^2\eta^3 \frac{2\beta_1}{\alpha_1 K}(1 + \frac{c_{f_y}^2}{\mu^2})\sigma^2 + 12\beta_1\alpha_1^2\eta^4 (1 + \frac{c_{f_y}^2}{\mu^2})\sigma^2  + 2\alpha_2^2\eta^3 \sigma^2\frac{25\beta_2}{3\alpha_2 K\mu}c_0 + 6\beta_1\alpha_2^2\eta^4 \sigma^2  \notag \\
		& + 4\alpha_3^2\eta^3(1 + \frac{c_{f_y}^2}{\mu^2})\sigma^2 \frac{100\beta_3}{\alpha_3 K \mu}c_1  + 12\beta_1\alpha_3^2\eta^4(1 + \frac{c_{f_y}^2}{\mu^2})\sigma^2+  6\beta_1\alpha_3^2\eta^4(1 + \frac{c_{f_y}^2}{\mu^2})\sigma^2   \ . 
	\end{align}
With this critical inequality,  we are able to finish the proof of Theorem~\ref{theorem-sim}.

\subsection{Proof Sketch of Theorem~\ref{theorem-alt}}
Similarly,  we establish the upper bound of the hypergradient's estimation bias, which is shown as follows:
		\begin{align}
			& \quad \mathbb{E}\left[\left\|\nabla F(\bar{x}_{t}) -    \frac{1}{K}\delta^{\hat{\mathcal{G}}_{F}} (X_{t}, Y_{t+1}, Z_{t+1}) \mathbf{1}  \right\|^2 \right]  \notag \\
			& \leq  9(\ell_{f_x}^2+\frac{c_{f_y}^2\ell_{g_{xy}}^2}{\mu^2})  \mathbb{E}\left[\left\| y^*(\bar{x}_{t}) - \bar{y}_{t}\right\|^2\right] + 9 c_{g_{xy}}^2  \mathbb{E}\left[\left\|z^*(\bar{x}_{t}) - \bar{z}_{t} \right\|^2\right] + 9 \beta_3^2\eta^2c_{g_{xy}}^2   \mathbb{E}\left[ \left\|\bar{w}_{t}\right\|^2\right]   \notag \\
			& \quad  +  9(\ell_{f_x}^2+\frac{c_{f_y}^2\ell_{g_{xy}}^2}{\mu^2})\frac{1}{K}\mathbb{E}\left[\left\|{X}_{t}- \bar{X}_{t}\right\|_F^2\right]  + 9(\ell_{f_x}^2+\frac{c_{f_y}^2\ell_{g_{xy}}^2}{\mu^2}) \frac{1}{K} \mathbb{E}\left[\left\|{Y}_{t+1}- \bar{Y}_{t+1}\right\|_F^2\right]   \notag \\
			& \quad+  9c_{g_{xy}}^2\frac{1}{K} \mathbb{E}\left[\left\|{Z}_{t+1}- \bar{Z}_{t+1}\right\|_F^2\right] + 9\beta_2^2\eta^2(\ell_{f_x}^2+\frac{c_{f_y}^2\ell_{g_{xy}}^2}{\mu^2} )  \mathbb{E}\left[ \left\| \bar{v}_{t}\right\|^2\right] \ . 
		\end{align}
Due to the alternating update, this upper bound of the hypergradient bias has  two additional terms $ \left\| \bar{v}_{t}\right\|^2$ and $ \left\|\bar{w}_{t}\right\|^2$ compared with Eq.~(\ref{lemma_bias-sim}). Meanwhile, it depends on the consensus error regarding the new update: $\left\|{Y}_{t+1}- \bar{Y}_{t+1}\right\|_F^2$ and $\left\|{Z}_{t+1}- \bar{Z}_{t+1}\right\|_F^2$. 

Similarly, based on this gradient bias, we will bound the optimization error and consensus error.

\paragraph{Step 1: Bounding Optimization Error for $z$ and $y$.}
We bound the estimation error about $ \|\bar{z}_{t}   - z^{*}(\bar{{x}}_t)\| ^2$ in Lemma~\ref{lemma_z_opt-alt} and $ \|\bar{y}_{t}   - y^{*}(\bar{{x}}_t)\| ^2$ in Lemma~\ref{lemma_y_opt-alt}.  

For instance, $ \|\bar{z}_{t}   - z^{*}(\bar{{x}}_t)\| ^2$  can be bounded as follows. 
		\begin{align}\label{lemma_z_opt-alt-main}
			&  \quad  \mathbb{E} [ \|\bar{z}_{t+1} - z^{*}(\bar{{x}}_{t+1})\| ^2]  \notag \\
			& \leq  (1-\frac{\eta\beta_3\mu}{8}) \mathbb{E} [ \|\bar{z}_{t} - z^*(\bar{x}_{t})\|^2 ]+ \frac{18\eta\beta_3}{\mu}\Big(\frac{c_{f_y}^2\ell_{g_{yy}}^2}{\mu^2}+\ell_{f_y}^2\Big) \mathbb{E} [ \| \bar{y}_{t}  - y^*(\bar{x}_{t})\|^2] \notag \\
			& \quad+ \frac{70\eta\beta_3}{\mu}\Big(\frac{c_{f_y}^2\ell_{g_{yy}}^2}{\mu^2} + \ell_{f_y}^2\Big)\frac{1}{K}\mathbb{E} [ \|\bar{X}_{t} - {X}_{t}\|_F^2]   \notag \\
			& \quad + \frac{70\eta\beta_3}{\mu}\Big(\frac{c_{f_y}^2\ell_{g_{yy}}^2}{\mu^2}+ \ell_{f_y}^2\Big)\frac{1}{K}\mathbb{E} [ \| \bar{Y}_{t+1} - {Y}_{t+1}\|_F^2] \notag \\
			& \quad +\Big(\frac{70\eta\beta_3}{\mu}\ell_{g_y}^2+  \frac{9}{4}\Big)\frac{1}{K}\mathbb{E} [ \|{Z}_{t}- \bar{Z}_{t}\|_F^2]   + \frac{9\eta\beta_3^2}{4}\frac{1}{K}\mathbb{E} [ \|   R_{t} -   \bar{R}_{t} \|_F^2]  \notag \\
			& \quad  +\frac{25\eta\beta_3}{\mu} \mathbb{E} [ \| \frac{1}{K}\delta^{\hat{\mathcal{G}}_{h}}({X}_{t}, {Y}_{t+1},  {Z}_{t})\mathbf{1}-   \frac{1}{K}{W}_{t}\mathbf{1}\|^2]+  \frac{9\eta\beta_1^2L_z^2}{\beta_3\mu} \mathbb{E} [ \|\bar{u}_{t}\|^2]  \notag \\
			& \quad +  \frac{18\eta^3\beta_2^2\beta_3}{\mu}\Big(\frac{c_{f_y}^2\ell_{g_{yy}}^2}{\mu^2}+\ell_{f_y}^2\Big)  \mathbb{E}[\|  \bar{v}_{t}\|^2] \ . 
		\end{align}
Compared with Eq.~(\ref{lemma_z_opt-sim-main}),there is an additional term regarding  $\|  \bar{v}_{t}\|^2$ in Eq.~(\ref{lemma_z_opt-alt-main}),  since $y$ is updated before $z$ in DSVRBGD-A.

\paragraph{Step 2: Bounding Estimation Error  for  Gradient.}  We bound the gradient estimation error for $U$, $V$, $W$ in Lemma~\ref{lemma_hyper_storm_var_mean-alt}, Lemma~\ref{lemma_g_storm_var_mean-alt}, and Lemma~\ref{lemma_h_storm_var_mean-alt}, respectively. 

For instance, we bound $\mathbb{E} [ \|\frac{1}{K}\delta^{\hat{\mathcal{G}}_{F}}(X_t, Y_{t+1}, Z_{t+1}) \mathbf{1} -  \frac{1}{K} U_t \mathbf{1} \|^2 ]$ as follows. 
	\begin{align} \label{eq:var-u-mean-main-alt}
		& \frac{1}{K}\mathbb{E} [ \|\delta^{\hat{\mathcal{G}}_{F}}(X_{t}, Y_{t+1}, Z_{t+1}) -   U_{t}  \|_F^2 ]  \leq  (1-\alpha_1 \eta^2) \frac{1}{K}\mathbb{E} [ \| \delta^{\hat{\mathcal{G}}_{F}}(X_{t-1}, Y_{t}, Z_{t}) - U_{t-1}  \|_F^2 ]\notag \\
		& \quad + 6(\ell_{f_x}^2+\frac{c_{f_y}^2\ell_{g_{xy}}^2}{\mu^2})\frac{1}{K}\mathbb{E}[\|X_{t} - X_{t-1}\|_F^2]  + 6(\ell_{f_x}^2+\frac{c_{f_y}^2\ell_{g_{xy}}^2}{\mu^2})\frac{1}{K}\mathbb{E}[\|Y_{t+1} - Y_{t}\|_F^2]\notag \\
		& \quad + 6c_{g_{xy}}^2\frac{1}{K}\mathbb{E}[\|Z_{t+1} - Z_{t}\|_F^2] + 4\alpha_1^2\eta^4 (1 + \frac{c_{f_y}^2}{\mu^2})\sigma^2 \ . 
	\end{align}
The gradient estimation error in Eq.~(\ref{eq:var-u-mean-main-alt}) depends on $\|Y_{t+1} - Y_{t}\|_F^2$ and $\|Z_{t+1} - Z_{t}\|_F^2$, 
while that in Eq.~(\ref{eq:var-u-mean-main-sim}) depends on  $\|Y_{t} - Y_{t-1}\|_F^2$ and $\|Z_{t} - Z_{t-1}\|_F^2$. This difference makes the proof of Theorem~\ref{theorem-alt} significantly different from that of Theorem~\ref{theorem-sim}. Specifically, we use Lemma~\ref{lemma:y-incremental-2-alt} and Lemma~\ref{lemma:z-incremental-2-alt} to replace $\|Y_{t+1} - Y_{t}\|_F^2$ and $\|Z_{t+1} - Z_{t}\|_F^2$ in Eq.~(\ref{eq:var-u-mean-main-alt}). Then, we can obtain
	\begin{align}\label{eq:expand-u-est-error}
		& \frac{1}{K}\mathbb{E} [ \|\delta^{\hat{\mathcal{G}}_{F}}(X_{t}, Y_{t+1}, Z_{t+1}) -   U_{t}  \|_F^2 ]  \leq  (1-\alpha_1 \eta^2) \frac{1}{K}\mathbb{E} [ \| \delta^{\hat{\mathcal{G}}_{F}}(X_{t-1}, Y_{t}, Z_{t}) - U_{t-1}  \|_F^2 ]  \notag \\
		& \quad +\left[ \frac{216\alpha^2_2\beta_2^2\eta^6}{1-\lambda}\left(\ell_{f_x}^2+\frac{c_{f_y}^2\ell_{g_{xy}}^2}{\mu^2}\right)+ \frac{36\alpha^2_2\beta_2^2\eta^6}{1-\lambda}\frac{648\eta^2\beta_3^2c_{g_{xy}}^2}{1-\lambda}\left(\ell_{f_y}^2+\frac{c_{f_y}^2\ell_{g_{yy}}^2}{\mu^2}\right)\right] \notag \\
		& \qquad \times \frac{1}{K}\mathbb{E}[\| V_{t-1}- \delta^{g}(X_{t-1}, Y_{t-1})   \|_F^2] \notag \\
		& \quad + \left[6c_{g_{xy}}^2 \frac{36\beta_3^2\alpha^2_3\eta^6}{1-\lambda}\right] \frac{1}{K}\mathbb{E}[\| W_{t-1}- \delta^{\hat{\mathcal{G}}_{h}}(X_{t-1}, Y_{t}, Z_{t-1})   \|_F^2]  \notag \\
		& \quad + \left[48\eta^2\left(\ell_{f_x}^2+\frac{c_{f_y}^2\ell_{g_{xy}}^2}{\mu^2}\right)+  \frac{5184\eta^4\beta_3^2c_{g_{xy}}^2}{1-\lambda}\left(\ell_{f_y}^2+\frac{c_{f_y}^2\ell_{g_{yy}}^2}{\mu^2}\right) \right] \frac{1}{K}\mathbb{E}[\|Y_{t-1} -\bar{Y}_{t-1}\|_F^2] \notag \\
		& \quad + \left[48c_{g_{xy}}^2\eta^2\right]\frac{1}{K}\mathbb{E}[\|Z_{t-1} -\bar{Z}_{t-1}\|_F^2] \notag \\
		& \quad +\left[\frac{120\eta^2 \beta_2^2}{1-\lambda^2}\left(\ell_{f_x}^2+\frac{c_{f_y}^2\ell_{g_{xy}}^2}{\mu^2}\right)+\frac{20\eta^2 \beta_2^2}{1-\lambda^2}\frac{648\eta^2\beta_3^2c_{g_{xy}}^2}{1-\lambda}\left(\ell_{f_y}^2+\frac{c_{f_y}^2\ell_{g_{yy}}^2}{\mu^2}\right)\right] \notag \\
		& \qquad \times \frac{1}{K}\mathbb{E}[\|Q_{t-1}- \bar{ Q}_{t-1}\|_F^2]     \notag \\
		& \quad + \left[ \frac{120\eta^3 \beta_3^2c_{g_{xy}}^2}{1-\lambda^2}\right]\frac{1}{K}\mathbb{E}[\|R_{t-1}- \bar{R}_{t-1}\|_F^2] \notag \\
		& \quad + \Bigg[\left(6+\frac{360 \beta_2^2\eta^2\ell_{g_y}^2}{1-\lambda}\right)\left(\ell_{f_x}^2+\frac{c_{f_y}^2\ell_{g_{xy}}^2}{\mu^2}\right) +\frac{648\eta^2\beta_3^2c_{g_{xy}}^2}{1-\lambda}\left(\ell_{f_y}^2+\frac{c_{f_y}^2\ell_{g_{yy}}^2}{\mu^2}\right)\notag \\
		& \quad \quad  +\frac{60 \beta_2^2\eta^2\ell_{g_y}^2}{1-\lambda}\frac{648\eta^2\beta_3^2c_{g_{xy}}^2}{1-\lambda}\left(\ell_{f_y}^2+\frac{c_{f_y}^2\ell_{g_{yy}}^2}{\mu^2}\right) \Bigg]\frac{1}{K}\mathbb{E}[\|X_{t} - X_{t-1}\|_F^2] \notag \\
		& \quad + \left[\frac{360 \beta_2^2\eta^2\ell_{g_y}^2}{1-\lambda}\left(\ell_{f_x}^2+\frac{c_{f_y}^2\ell_{g_{xy}}^2}{\mu^2}\right)+ \frac{60 \beta_2^2\eta^2\ell_{g_y}^2}{1-\lambda}\frac{648\eta^2\beta_3^2c_{g_{xy}}^2}{1-\lambda}\left(\ell_{f_y}^2+\frac{c_{f_y}^2\ell_{g_{yy}}^2}{\mu^2}\right)\right] \notag \\
		& \qquad \times  \frac{1}{K} \mathbb{E}[\|Y_{t} - Y_{t-1}\|_F^2] \notag \\
		& \quad +\left[6c_{g_{xy}}^2\frac{ 108\eta^2\beta_3^2\ell_{g_y}^2 }{1-\lambda}\right] \frac{1}{K} \mathbb{E}[\|Z_{t} - Z_{t-1}\|_F^2]   \notag \\
		& \quad + \left[48\eta^2\beta_2^2\left(\ell_{f_x}^2+\frac{c_{f_y}^2\ell_{g_{xy}}^2}{\mu^2}\right)+ 8\eta^2\beta_2^2\frac{648\eta^2\beta_3^2c_{g_{xy}}^2}{1-\lambda}\left(\ell_{f_y}^2+\frac{c_{f_y}^2\ell_{g_{yy}}^2}{\mu^2}\right) \right]\frac{1}{K}\mathbb{E}[\|\bar{Q}_{t-1}\|_F^2] \notag \\
		& \quad +\left[6c_{g_{xy}}^28\eta^2\beta_3^2\right]\frac{1}{K}\mathbb{E}[\| \bar{R}_{t-1}\|_F^2]  \notag \\
		& \quad  + 4\alpha_1^2\eta^4 \left(1 + \frac{c_{f_y}^2}{\mu^2}\right)\sigma^2 +  \frac{288\alpha^2_2\beta_2^2\eta^6}{1-\lambda}\left(\ell_{f_x}^2+\frac{c_{f_y}^2\ell_{g_{xy}}^2}{\mu^2}\right)\sigma^2  \notag \\
		& \quad + 6c_{g_{xy}}^2 \frac{72\alpha^2_3\beta_3^2\eta^6}{1-\lambda}\left(1 + \frac{c_{f_y}^2}{\mu^2}\right)\sigma^2  + \frac{648\eta^2\beta_3^2c_{g_{xy}}^2}{1-\lambda}\left(\ell_{f_y}^2+\frac{c_{f_y}^2\ell_{g_{yy}}^2}{\mu^2}\right) \frac{48\alpha^2_2\beta_2^2\eta^6}{1-\lambda}\sigma^2 \ .  
	\end{align}
Compared with Eq.~(\ref{eq:var-u-mean-main-sim}),  the gradient estimation error $\mathbb{E} [ \|\delta^{\hat{\mathcal{G}}_{F}}(X_{t}, Y_{t+1}, Z_{t+1}) -   U_{t}  \|_F^2 ]$ of DSVRBGD-A depends on ALL  gradient estimation errors in the $(t-1)$-th iteration:  $\mathbb{E} [ \| \delta^{\hat{\mathcal{G}}_{F}}(X_{t-1}, Y_{t}, Z_{t}) - U_{t-1}  \|_F^2 ]$, $\mathbb{E}[\| V_{t-1}- \delta^{g}(X_{t-1}, Y_{t-1})   \|_F^2]$, and $\mathbb{E}[\| W_{t-1}- \delta^{\hat{\mathcal{G}}_{h}}(X_{t-1}, Y_{t}, Z_{t-1})   \|_F^2]$. Meanwhile, it has additional terms regarding the consensus error: $\mathbb{E}[\|Y_{t-1} -\bar{Y}_{t-1}\|_F^2] $,  $\mathbb{E}[\|Z_{t-1} -\bar{Z}_{t-1}\|_F^2] $, $\mathbb{E}[\|Q_{t-1}- \bar{ Q}_{t-1}\|_F^2]$, and $\mathbb{E}[\|R_{t-1}- \bar{R}_{t-1}\|_F^2]$. The reason is that the gradient $\delta^{\hat{\mathcal{G}}_{F}}(X_{t}, Y_{t+1}, Z_{t+1}) $ is computed on $\{Y_{t+1}, Z_{t+1}\}$, instead of $\{Y_{t}, Z_{t}\}$.

\paragraph{Step 3: Bounding  Consensus Error for Variable and Gradient.} 
We bound the consensus error for the variable $x$, $y$, and $z$ in Lemma~\ref{lemma_consensus_x-alt} and the consensus error for the gradient $P$, $Q$, and $R$ in Lemma~\ref{lemma_consensus_p-alt}, Lemma~\ref{lemma_consensus_q-alt}, and Lemma~\ref{lemma_consensus_r-alt}, respectively.

\paragraph{Step 4: Construct Potential Function.}
Due to the alternating update strategy, the potential function in Eq.~(\ref{eq:potential-function}) does not work. Therefore, we proposed a new potential function to handle the alternating update as follows.
	\begin{align}\label{eq:potential-function-alt}
		& \mathcal{L}_{t} = {\mathbb{E}}[F(x_{t})] +  c_0\mathbb{E}[\|\bar{   {y}}_{t} -    {y}^{*}(\bar{   {x}}_{t})\| ^2 ]  +  c_1\mathbb{E}[\|\bar{ z}_{t} -    {z}^{*}(\bar{   {x}}_{t})\| ^2 ]   \notag \\
		& \quad + c_2\frac{1}{K} \mathbb{E}[\|X_{t} - \bar{X}_{t}\|_F^2 ] + c_3 \frac{1}{K}\mathbb{E}[\|Y_{t} - \bar{Y}_{t}\|_F^2 ]  +c_4\frac{1}{K}\mathbb{E}[\| Z_{t}-\bar{Z}_{t} \|_F^2] \notag \\
		& \quad + c_5\frac{1}{K} \mathbb{E}[\|P_{t} - \bar{P}_{t}\|_F^2 ] + c_6 \frac{1}{K}\mathbb{E}[\|Q_{t} - \bar{Q}_{t}\|_F^2 ]  +c_7\frac{1}{K}\mathbb{E}[\| R_{t}-\bar{R}_{t} \|_F^2] \notag \\
		& \quad + c_8 \mathbb{E} [ \|\frac{1}{K}\delta^{\hat{\mathcal{G}}_{F}}(X_{t}, Y_{t+1}, Z_{t+1}) \mathbf{1}  -  \frac{1}{K} U_{t} \mathbf{1} \|^2 ]+ c_9 \frac{1}{K}\mathbb{E} [ \|\delta^{\hat{\mathcal{G}}_{F}}(X_{t}, Y_{t+1}, Z_{t+1}) -  U_{t} \|_F^2 ] \notag \\
		& \quad  + c_{10}\mathbb{E} [ \|\frac{1}{K}\delta^{g}(X_{t}, Y_{t}) \mathbf{1} -  \frac{1}{K} V_{t} \mathbf{1} \|^2 ] + c_{11} \frac{1}{K}\mathbb{E} [ \|\delta^{g}(X_{t}, Y_{t}) - V_{t}  \|_F^2 ]\notag \\
		& \quad  + c_{12} \mathbb{E} [ \|\frac{1}{K}\delta^{\hat{\mathcal{G}}_{h}}(X_{t}, Y_{t+1}, Z_{t})\mathbf{1}  -  \frac{1}{K} W_{t} \mathbf{1} \|^2 ] +   c_{13} \frac{1}{K}\mathbb{E} [ \|\delta^{\hat{\mathcal{G}}_{h}}(X_{t}, Y_{t+1}, Z_{t}) -   W_{t}  \|_F^2 ] \ . 
	\end{align}
where $\{c_i\}_{i=0}^{13}$ are constant values.  Here, \textbf{it is more challenging to determine $\{c_i\}_{i=0}^{13}$ for Theorem~\ref{theorem-alt}. The reason is that these constants have to accommodate those additional terms in different optimization errors and estimation errors due to the alternating update. }

After obtaining $\{c_i\}_{i=0}^{13}$, we have
	\begin{align}
		&   \mathcal{L}_{t+1} -  \mathcal{L}_{t} \leq   - \frac{\beta_1\eta}{2}\mathbb{E}[\| \nabla F(\bar{x}_{t})\|^2]  \notag \\
		& \quad + \left( \tilde{D}_y+ \frac{2916c_{g_{xy}}^2\beta_3^2}{1-\lambda}\left(1+ \frac{1}{\alpha_1 K}\right)  \left(\ell_{f_y}^2+\frac{c_{f_y}^2\ell_{g_{yy}}^2}{\mu^2}\right)\right) \frac{48\beta_1\alpha^2_2\beta_2^2\eta^5}{1-\lambda}\sigma^2 \notag \\
		& \quad  +  6\beta_1\alpha_3^2\eta^4\left(1 + \frac{c_{f_y}^2}{\mu^2}\right)\sigma^2 +   \frac{1944\beta_1\alpha^2_3\beta_3^2\eta^5c_{g_{xy}}^2}{1-\lambda}\left(1+ \frac{1}{\alpha_1 K}\right) \left(1 + \frac{c_{f_y}^2}{\mu^2}\right)\sigma^2 \notag \\
		& \quad + 4 \beta_1\alpha_1\eta^3 \left(1 + \frac{c_{f_y}^2}{\mu^2}\right)\frac{\sigma^2}{K}  + 12\beta_1\alpha_1^2\eta^4 \left(1 + \frac{c_{f_y}^2}{\mu^2}\right)\sigma^2  + \tilde{c}_0\frac{50\beta_1 \alpha_2\eta^3}{3 \mu}   \frac{\sigma^2}{K}  \notag \\
		& \quad + 2\beta_1\alpha_2^2\eta^4  \left( 3   + \frac{36\beta_2^2}{1-\lambda}\left(\tilde{D}_y+  2916 c_{g_{xy}}^2\left(1+ \frac{1}{\alpha_1 K}\right)   \left(\ell_{f_y}^2+\frac{c_{f_y}^2\ell_{g_{yy}}^2}{\mu^2}\right)\right)\right)\sigma^2\notag \\
		& \quad  + \tilde{c}_1\frac{400\beta_1\alpha_3\eta^3}{  \mu}  \left(1 + \frac{c_{f_y}^2}{\mu^2}\right)\frac{\sigma^2}{K} + 4\beta_1\alpha_3^2\eta^4\left(1 + \frac{c_{f_y}^2}{\mu^2}\right)\left(3 +  972\beta_3 c_{g_{xy}}^2\left(1+\frac{1}{\alpha_1 K} \right) \right)\sigma^2 \notag \\
		& \quad + 6\beta_1\alpha_2^2\eta^4\sigma^2 +  6\beta_1\alpha_1^2\eta^4\left(1 + \frac{c_{f_y}^2}{\mu^2}\right)\sigma^2 \ . 
	\end{align}
Based on these critical steps,  we are able to finish the proof of Theorem~\ref{theorem-alt}.

\section{Proof of Theorem~\ref{theorem-sim}} \label{sec:proof-sim}

\begin{lemma} \label{lemma_hypergrad_bias-sim}
	Under Assumptions~\ref{assumption_bi_strong}-\ref{assumption_graph}, we have
\begin{align} 
		&\mathbb{E}\left[\left\|\nabla F(\bar{x}_{t}) -    \frac{1}{K}\delta^{\hat{\mathcal{G}}_{F}} (X_t, Y_t, Z_t) \mathbf{1}  \right\|^2\right]  \leq 6 (\ell_{f_x}^2 + \frac{c_{f_y}^2\ell_{g_{xy}}^2}{\mu^2})\mathbb{E}\left[\left\| y^*(\bar{x}_t) - \bar{y}_t\right\|^2\right]   \notag \\
		& \quad + 6c_{g_{xy}}^2 \mathbb{E}\left[\left\|z^*(\bar{x}_t) - \bar{z}_t \right\|^2\right]  + 6(\ell_{f_x}^2+\frac{c_{f_y}^2\ell_{g_{xy}}^2}{\mu^2} )\frac{1}{K}\mathbb{E}\left[\left\|X_{t} - \bar{X}_{t}\right\|_F^2\right]  \notag \\
		& \quad + 6(\ell_{f_x}^2+\frac{c_{f_y}^2\ell_{g_{xy}}^2}{\mu^2} ) \frac{1}{K}\mathbb{E}\left[\left\|Y_{t} - \bar{Y}_{t}\right\|_F^2\right]   + 6c_{g_{xy}}^2\frac{1}{K}\mathbb{E}\left[\left\|Z_{t}- \bar{Z}_{t}\right\|_F^2\right]  \ . 
\end{align}
\end{lemma}
\begin{proof}
		\begin{align} 
			& \quad\left\|\nabla F(\bar{x}_{t}) -    \frac{1}{K}\delta^{\hat{\mathcal{G}}_{F}} (X_t, Y_t, Z_t) \mathbf{1}  \right\|^2 \notag \\
			& = \Bigg\|\frac{1}{K}\sum_{k=1}^{K}\nabla_1 f^{(k)}(\bar{x}_t, y^*(\bar{x}_t))  - \frac{1}{K}\sum_{k=1}^{K} \nabla_{12}^2 g^{(k)}(\bar{x}_t, y^*(\bar{x}_t))   z^*(\bar{x}_t) \notag \\
			& \quad   -  \frac{1}{K}\sum_{k=1}^{K}\nabla_1 { f^{(k)}({x}_{t}^{(k)}, {y}_{t}^{(k)})}   +\frac{1}{K}\sum_{k=1}^{K}\nabla_{12}^2 g^{(k)}({x}_{t}^{(k)}, {y}_{t}^{(k)})z_t^{(k)} \Bigg\|^2 \notag \\
			& \leq  6\left\|\frac{1}{K}\sum_{k=1}^{K}\nabla_{1} f^{(k)}(\bar{x}_t, y^*(\bar{x}_t)) - \frac{1}{K}\sum_{k=1}^{K}\nabla_{1} f^{(k)}(\bar{x}_t, \bar{y}_t) \right\|^2 \notag \\
			& \quad  +6\left\| \frac{1}{K}\sum_{k=1}^{K}\nabla_{1} f^{(k)}(\bar{x}_t, \bar{y}_t)-  \frac{1}{K}\sum_{k=1}^{K}\nabla_{1} { f^{(k)}({x}_{t}^{(k)}, {y}_{t}^{(k)})} \right\|^2  \notag \\
			& \quad+ 6\left\| - \frac{1}{K}\sum_{k=1}^{K} \nabla_{12}^2 g^{(k)}(\bar{x}_t, y^*(\bar{x}_t))   z^*(\bar{x}_t) + \frac{1}{K}\sum_{k=1}^{K} \nabla_{12}^2 g^{(k)}(\bar{x}_t, y^*(\bar{x}_t)) \bar{z}_t  \right\|^2\notag \\
			& \quad + 6\left\|- \frac{1}{K}\sum_{k=1}^{K} \nabla_{12}^2 g^{(k)}(\bar{x}_t, y^*(\bar{x}_t))   \bar{z}_t  + \frac{1}{K}\sum_{k=1}^{K} \nabla_{12}^2 g^{(k)}(\bar{x}_t, \bar{y}_t)   \bar{z}_t \right\|^2\notag \\
			& \quad + 6\left\| - \frac{1}{K}\sum_{k=1}^{K} \nabla_{12}^2 g^{(k)}(\bar{x}_t, \bar{y}_t)   \bar{z}_t + \frac{1}{K}\sum_{k=1}^{K} \nabla_{12}^2 g^{(k)}(\bar{x}_t, \bar{y}_t)   z_t^{(k)}  \right\|^2\notag \\
			& \quad + 6\left\|- \frac{1}{K}\sum_{k=1}^{K} \nabla_{12}^2 g^{(k)}(\bar{x}_t, \bar{y}_t)  z_t^{(k)}  +\frac{1}{K}\sum_{k=1}^{K}\nabla_{12}^2 g^{(k)}({x}_{t}^{(k)}, {y}_{t}^{(k)})z_t^{(k)} \right\|^2 \notag \\
			& \leq 6 (\ell_{f_x}^2 + \frac{c_{f_y}^2\ell_{g_{xy}}^2}{\mu^2})\left\| y^*(\bar{x}_t) - \bar{y}_t\right\|^2  + 6c_{g_{xy}}^2 \left\|z^*(\bar{x}_t) - \bar{z}_t \right\|^2  \notag \\
			& \quad + 6(\ell_{f_x}^2+\frac{c_{f_y}^2\ell_{g_{xy}}^2}{\mu^2} )\frac{1}{K}\sum_{k=1}^{K}\left\|{x}_{t}^{(k)} - \bar{x}_{t}\right\|^2+ 6(\ell_{f_x}^2+\frac{c_{f_y}^2\ell_{g_{xy}}^2}{\mu^2} ) \frac{1}{K}\sum_{k=1}^{K}\left\|{y}_{t}^{(k)} - \bar{y}_{t}\right\|^2 \notag \\
			& \quad + 6c_{g_{xy}}^2\frac{1}{K}\sum_{k=1}^{K}\left\|{z}_{t}^{(k)} - \bar{z}_{t}\right\|^2   \ ,  
	\end{align}
	where the last step follows from Assumptions~\ref{assumption_upper_smooth_vr},~\ref{assumption_lower_smooth_vr}, and  the projection operation of our algorithm.

\end{proof}

\begin{lemma} \label{lemma_F_iter}
	Under Assumptions~\ref{assumption_bi_strong}-\ref{assumption_graph}, if $\eta\leq\frac{1}{2\beta_1 L_F}$, we have
		\begin{align}
			& \mathbb{E}[F(\bar{x}_{t+1})] \leq \mathbb{E}[F(\bar{x}_{t}) ]- \frac{\beta_1\eta}{2}\mathbb{E}[\left\| \nabla F(\bar{x}_{t})\right\|^2]  -  \frac{\beta_1\eta}{4}\mathbb{E}[\left\|\bar{u} _{t}\right\|^2] \notag \\
			& \quad  +  12\beta_1\eta \Big(\ell_{f_x}^2 +\frac{c_{f_y}^2\ell_{g_{xy}}^2}{\mu^2} \Big)\mathbb{E}[\left\| y^*(\bar{x}_t) - \bar{y}_t\right\|^2] + 12\beta_1\eta c_{g_{xy}}^2\mathbb{E}[ \left\|z^*(\bar{x}_t) -  \bar{z}_t \right\|^2]\notag \\
			& \quad + 12\beta_1\eta(\ell_{f_x}^2+\frac{c_{f_y}^2\ell_{g_{xy}}^2}{\mu^2} )\frac{1}{K}\mathbb{E}\left[\left\|{X}_{t} - \bar{X}_{t}\right\|_F^2\right] + 12\beta_1\eta(\ell_{f_x}^2+ \frac{c_{f_y}^2\ell_{g_{xy}}^2}{\mu^2})\frac{1}{K}\mathbb{E}\left[\left\|{Y}_{t}- \bar{Y}_{t}\right\|_F^2 \right]\notag \\
			& \quad + 12\beta_1\eta c_{g_{xy}}^2\frac{1}{K}\mathbb{E}\left[\left\|{Z}_{t} - \bar{Z}_{t}\right\|_F^2 \right]+ 2\beta_1\eta\mathbb{E}\left[\left\|\frac{1}{K}\delta^{\hat{\mathcal{G}}_{F}} (X_t, Y_t, Z_t) \mathbf{1} - \frac{1}{K}{U}_t \mathbf{1}\right\|^2 \right]\ . 
	\end{align}
\end{lemma}

\begin{proof}
	Because $F(\cdot)$ is $L_F$-smooth, we can obtain  
	\begin{align}\label{eq:f_inc1}
			& F(\bar{x}_{t+1}) \leq F(\bar{x}_{t}) + \langle \nabla F(\bar{x}_{t}), \bar{x}_{t+1} - \bar{x}_{t}\rangle + \frac{L_{F}}{2}\left\|\bar{x}_{t+1} - \bar{x}_{t}\right\|^2 \notag \\
			& = F(\bar{x}_{t}) - \beta_1\eta\langle \nabla F(\bar{x}_{t}), {u} _{t}\rangle + \frac{\beta_1^2\eta^2L_{F}}{2}\left\|\bar{u} _{t}\right\|^2 \notag \\
			& = F(\bar{x}_{t}) - \frac{\beta_1\eta}{2}\left\| \nabla F(\bar{x}_{t})\right\|^2  -  \frac{\beta_1\eta}{2}\left\|\bar{u} _{t}\right\|^2 +  \frac{\beta_1\eta}{2}\left\|\nabla F(\bar{x}_{t}) - \bar{u} _{t}\right\|^2 + \frac{\beta_1^2\eta^2L_{F}}{2}\left\|\bar{u} _{t}\right\|^2 \notag \\
			& \leq F(\bar{x}_{t}) - \frac{\beta_1\eta}{2}\left\| \nabla F(\bar{x}_{t})\right\|^2  -  \frac{\beta_1\eta}{4}\left\|\bar{u} _{t}\right\|^2 +  \frac{\beta_1\eta}{2}\left\|\nabla F(\bar{x}_{t}) - \bar{u} _{t}\right\|^2 \notag \\
			& \leq F(\bar{x}_{t}) - \frac{\beta_1\eta}{2}\left\| \nabla F(\bar{x}_{t})\right\|^2  -  \frac{\beta_1\eta}{4}\left\|\bar{u} _{t}\right\|^2 +\beta_1\eta\left\|\frac{1}{K}\delta^{\hat{\mathcal{G}}_{F}} (X_t, Y_t, Z_t) \mathbf{1} - \frac{1}{K}{U}_t \mathbf{1}\right\|^2 \notag \\
			& \quad +  \beta_1\eta\left\|\nabla F(\bar{x}_{t}) -  \frac{1}{K}\delta^{\hat{\mathcal{G}}_{F}} (X_t, Y_t, Z_t) \mathbf{1}  \right\|^2 \  ,   
	\end{align}
where the fourth step follows from $\eta\leq\frac{1}{2\beta_1 L_F}$.  Then, by plugging Lemma~\ref{lemma_hypergrad_bias-sim} into the above inequality, we complete the proof. 

\end{proof}

\begin{lemma} \label{lemma_z_opt}
 	Under Assumptions~\ref{assumption_bi_strong}-\ref{assumption_graph}, if  $\eta \leq \frac{1}{\beta_3\mu}$, we have
\begin{align}
			&  \quad  \mathbb{E} [ \|\bar{z}_{t+1} - z^{*}(\bar{{x}}_{t+1})\| ^2] \notag \\
			&  \leq  (1-\frac{\eta\beta_3\mu}{8}) \mathbb{E} [ \|\bar{z}_t - z^*(\bar{x}_{t})\|^2 ]+ \frac{9\eta\beta_3}{\mu}\Big(\frac{c_{f_y}^2\ell_{g_{yy}}^2}{\mu^2}+\ell_{f_y}^2\Big) \mathbb{E} [ \| \bar{y}_t  - y^*(\bar{x}_t)\|^2]  \notag \\
			& \quad + \frac{70\eta\beta_3}{\mu}\Big(\frac{c_{f_y}^2\ell_{g_{yy}}^2}{\mu^2} + \ell_{f_y}^2\Big)\frac{1}{K}\mathbb{E} [ \|\bar{X}_{t} - {X}_{t}\|_F^2]  + \frac{70\eta\beta_3}{\mu}\Big(\frac{c_{f_y}^2\ell_{g_{yy}}^2}{\mu^2}+ \ell_{f_y}^2\Big)\frac{1}{K}\mathbb{E} [ \| \bar{Y}_t - {Y}_{t}\|_F^2] \notag \\
			& \quad +\Big(\frac{70\eta\beta_3}{\mu}\ell_{g_y}^2+ \textcolor{red}{ \frac{9}{4}}\Big)\frac{1}{K}\mathbb{E} [ \|{Z}_t- \bar{Z}_{t}\|_F^2]  + \frac{9\eta\beta_3^2}{4}\frac{1}{K}\mathbb{E} [ \|   R_t -   \bar{R}_t \|_F^2]  \notag \\
			& \quad +\frac{25\eta\beta_3}{\mu} \mathbb{E} [ \| \frac{1}{K}\delta^{\hat{\mathcal{G}}_{h}}({X}_{t}, {Y}_{t},  {Z}_t)\mathbf{1}-   \frac{1}{K}{W}_{t}\mathbf{1}\|^2] + \frac{9\eta\beta_1^2L_z^2}{\beta_3\mu} \mathbb{E} [ \|\bar{u}_{t}\|^2]   \ . 
	\end{align}
	
\end{lemma}

\begin{proof}

	At first, we have
		\begin{align}\label{eq:z-bar-z-star-2}
			& \quad \left\| \bar{z}_t  -\eta\beta_3 \hat{\mathcal{G}}_{h}(\bar{x}_t, \bar{y}_t, \bar{z}_t) -  z^*(\bar{x}_{t}) + \eta\beta_3 \nabla_z h(\bar{x}_t, z^*(\bar{x}_{t}))\right\|^2  \notag \\
			& =  \Big\| \bar{z}_t  -\eta\beta_3 (\nabla_{22}^2g(\bar{x}_t, \bar{y}_t) \bar{z}_{t}   -  \nabla_2 f(\bar{x}_t,  \bar{y}_t)) \notag \\
			& \quad  -  z^*(\bar{x}_{t}) + \eta\beta_3 (\nabla_{22}^2g(\bar{x}_t, y^*(\bar{x}_t)) z^*(\bar{x}_{t})   -  \nabla_2{ f(\bar{x}_t, y^*(\bar{x}_t))})\Big\|^2  \notag \\
			& \leq (1+a)\left\|(I-\eta\beta_3\nabla_{22}^2g(\bar{x}_t, \bar{y}_t) )\bar{z}_t - (I-\eta\beta_3\nabla_{22}^2g(\bar{x}_t, \bar{y}_t) )z^*(\bar{x}_{t})\right\|^2\notag \\
			& \quad + 2(1+1/a)\left\|(I-\eta\beta_3\nabla_{22}^2g(\bar{x}_t, \bar{y}_t) )z^*(\bar{x}_{t})  - (I-\eta\beta_3\nabla_{22}^2g(\bar{x}_t, y^*(\bar{x}_t)) )z^*(\bar{x}_{t})\right\|^2\notag \\
			& \quad +  2(1+1/a)\left\|\eta\beta_3(\nabla_2 f(\bar{x}_t,  \bar{y}_t)-\nabla_2{ f(\bar{x}_t, y^*(\bar{x}_t))})\right\|^2 \notag \\
			& \leq (1+a) (1-\eta\beta_3\mu)^2 \left\|\bar{z}_t - z^*(\bar{x}_{t})\right\|^2 + 2 \eta^2\beta_3^2\frac{c_{f_y}^2\ell_{g_{yy}}^2}{\mu^2}(1+1/a) \left\| \bar{y}_t  - y^*(\bar{x}_t)\right\|^2 \notag \\
			& \quad + 2\eta^2\beta_3^2\ell_{f_y}^2(1+1/a)\left\| \bar{y}_t- y^*(\bar{x}_t)\right\|^2 \notag \\
			& \leq  (1+a) (1-\eta\beta_3\mu) \left\|\bar{z}_t - z^*(\bar{x}_{t})\right\|^2 + 2 \eta^2\beta_3^2\Big(\frac{c_{f_y}^2\ell_{g_{yy}}^2}{\mu^2}+\ell_{f_y}^2\Big)(1+1/a) \left\| \bar{y}_t  - y^*(\bar{x}_t)\right\|^2 \notag \notag \\
			& \leq   (1-\frac{\eta\beta_3\mu}{2}) \left\|\bar{z}_t - z^*(\bar{x}_{t})\right\|^2 + \frac{6\eta\beta_3}{\mu}\Big(\frac{c_{f_y}^2\ell_{g_{yy}}^2}{\mu^2}+\ell_{f_y}^2\Big) \left\| \bar{y}_t  - y^*(\bar{x}_t)\right\|^2  \ , 
	\end{align}
	where the third step follows from Assumptions~\ref{assumption_bi_strong}-\ref{assumption_lower_smooth_vr}, the last step follows from $a=\frac{\eta\beta_3\mu}{2}$ and $\eta \leq \frac{1}{\beta_3\mu}$. 
	
	Then, we have
		\begin{align} \label{eq:z-bar-z-star-3}
			& \quad \left\| \bar{z}_t  - \eta\beta_3  \bar{r}_t -  z^*(\bar{x}_{t})\right\|^2  = \left\| \bar{z}_t  - \eta\beta_3  \bar{w}_t -  z^*(\bar{x}_{t})\right\|^2 \notag \\
			& = \left\| \bar{z}_t  -\eta\beta_3 \hat{\mathcal{G}}_{h}(\bar{x}_t, \bar{y}_t, \bar{z}_t)+ \eta\beta_3 \hat{\mathcal{G}}_{h}(\bar{x}_t, \bar{y}_t, \bar{z}_t) - \eta\beta_3  \bar{w}_t -  z^*(\bar{x}_{t})\right\|^2 \notag \\
			& \leq (1+a)\left\| \bar{z}_t  -\eta\beta_3 \hat{\mathcal{G}}_{h}(\bar{x}_t, \bar{y}_t, \bar{z}_t)-  z^*(\bar{x}_{t})\right\|^2  +(1+1/a)\eta^2\beta_3^2 \left\|  \hat{\mathcal{G}}_{h}(\bar{x}_t, \bar{y}_t, \bar{z}_t) -  \frac{1}{K}{W}_{t}\mathbf{1}\right\|^2\notag \\
			& \leq  (1+a)\left\| \bar{z}_t  -\eta\beta_3 \hat{\mathcal{G}}_{h}(\bar{x}_t, \bar{y}_t, \bar{z}_t) -  z^*(\bar{x}_{t}) + \eta\beta_3 \nabla_z h(\bar{x}_t, z^*(\bar{x}_{t}))\right\|^2  \notag \\
			& \quad +(1+1/a)\eta^2\beta_3^2 \left\|  \hat{\mathcal{G}}_{h}(\bar{x}_t, \bar{y}_t, \bar{z}_t) -  \frac{1}{K}{W}_{t}\mathbf{1}\right\|^2\notag \\
			& \leq (1+a) (1-\frac{\eta\beta_3\mu}{2}) \left\|\bar{z}_t - z^*(\bar{x}_{t})\right\|^2 + (1+a)\frac{6\eta\beta_3}{\mu}\Big(\frac{c_{f_y}^2\ell_{g_{yy}}^2}{\mu^2}+\ell_{f_y}^2\Big) \left\| \bar{y}_t  - y^*(\bar{x}_t)\right\|^2  \notag \\
			& \quad +2(1+1/a)\eta^2\beta_3^2 \left\|  \hat{\mathcal{G}}_{h}(\bar{x}_t, \bar{y}_t, \bar{z}_t) -  \frac{1}{K}\delta^{\hat{\mathcal{G}}_{h}}({X}_{t}, {Y}_{t},  {Z}_t)\mathbf{1}\right\|^2\notag \\
			& \quad +2(1+1/a)\eta^2\beta_3^2 \left\| \frac{1}{K}\delta^{\hat{\mathcal{G}}_{h}}({X}_{t}, {Y}_{t},  {Z}_t)\mathbf{1} -  \frac{1}{K}{W}_{t}\mathbf{1}\right\|^2\notag \\
			& \leq  (1-\frac{\eta\beta_3\mu}{4}) \left\|\bar{z}_t - z^*(\bar{x}_{t})\right\|^2 + \frac{8\eta\beta_3}{\mu}\Big(\frac{c_{f_y}^2\ell_{g_{yy}}^2}{\mu^2}+\ell_{f_y}^2\Big) \left\| \bar{y}_t  - y^*(\bar{x}_t)\right\|^2  \notag \\
			& \quad +\frac{20\eta\beta_3}{\mu}\left\| \hat{\mathcal{G}}_{h}(\bar{x}_t, \bar{y}_t, \bar{z}_t) - \frac{1}{K}\delta^{\hat{\mathcal{G}}_{h}}({X}_{t}, {Y}_{t},  {Z}_t)\mathbf{1}\right\|^2 +\frac{20\eta\beta_3}{\mu} \left\| \frac{1}{K}\delta^{\hat{\mathcal{G}}_{h}}({X}_{t}, {Y}_{t},  {Z}_t)\mathbf{1}-   \frac{1}{K}{W}_{t}\mathbf{1}\right\|^2\notag \\
			& \leq  (1-\frac{\eta\beta_3\mu}{4}) \left\|\bar{z}_t - z^*(\bar{x}_{t})\right\|^2 + \frac{8\eta\beta_3}{\mu}\Big(\frac{c_{f_y}^2\ell_{g_{yy}}^2}{\mu^2}+\ell_{f_y}^2\Big) \left\| \bar{y}_t  - y^*(\bar{x}_t)\right\|^2  \notag \\
			& \quad +\frac{60\eta\beta_3}{\mu}\ell_{g_y}^2\frac{1}{K}\left\|{Z}_t- \bar{Z}_{t}\right\|_F^2  + \frac{60\eta\beta_3}{\mu}\Big(\frac{c_{f_y}^2\ell_{g_{yy}}^2}{\mu^2} + \ell_{f_y}^2\Big)\frac{1}{K}\left\|\bar{X}_{t} - {X}_{t}\right\|_F^2 \notag \\
			& \quad +  \frac{60\eta\beta_3}{\mu}\Big(\frac{c_{f_y}^2\ell_{g_{yy}}^2}{\mu^2}+ \ell_{f_y}^2\Big)\frac{1}{K}\left\| \bar{Y}_t - {Y}_{t}\right\|_F^2 +\frac{20\eta\beta_3}{\mu} \left\| \frac{1}{K}\delta^{\hat{\mathcal{G}}_{h}}({X}_{t}, {Y}_{t},  {Z}_t)\mathbf{1} -   \frac{1}{K}{W}_{t}\mathbf{1}\right\|^2 \ , 
	\end{align}
	where the third step follows from $\nabla_z h(\bar{x}_t, z^*(\bar{x}_{t})) = 0$, the fourth step follows from Eq.~(\ref{eq:z-bar-z-star-2}), the fifth step follows from  $a=\frac{\eta\beta_3\mu}{4}$ and $\eta\leq \frac{1}{\beta_3\mu}$, and the last step follows from the following inequality.
		\begin{align} \label{eq:h-grad-consensus}
			&\quad  \left\|  \hat{\mathcal{G}}_{h}(\bar{x}_t, \bar{y}_t, \bar{z}_t) -  \frac{1}{K}\delta^{\hat{\mathcal{G}}_{h}}({X}_{t}, {Y}_{t},  {Z}_t)\mathbf{1}\right\|^2\notag \\
			&  \leq  \frac{1}{K}\sum_{k=1}^{K}\left\|\nabla_{22}^2g^{(k)}(\bar{x}_{t}, \bar{y}_t) \bar{z}_t-  \nabla_{2}{ f^{(k)}(\bar{x}_{t}, \bar{y}_t)}  - \nabla_{22}^2g^{(k)}({x}_{t}^{(k)}, {y}_{t}^{(k)}) {z}_t^{(k)} +  \nabla_{2}{ f^{(k)}({x}_{t}^{(k)}, {y}_{t}^{(k)})}\right\|^2  \notag \\
			& \leq  3\frac{1}{K}\sum_{k=1}^{K}\left\|\nabla_{22}^2g^{(k)}(\bar{x}_{t}, \bar{y}_t) \bar{z}_t - \nabla_{22}^2g^{(k)}(\bar{x}_{t}, \bar{y}_t) {z}_t^{(k)}  \right\|^2 \notag \\
   & \quad + 3\frac{1}{K}\sum_{k=1}^{K}\left\|\nabla_{22}^2g^{(k)}(\bar{x}_{t}, \bar{y}_t) {z}_t^{(k)}  - \nabla_{22}^2g^{(k)}({x}_{t}^{(k)}, {y}_{t}^{(k)}) {z}_t^{(k)} \right\|^2  \notag \\
			& \quad + 3\frac{1}{K}\sum_{k=1}^{K}\left\|- \nabla_{2}{ f^{(k)}(\bar{x}_{t}, \bar{y}_t)} +  \nabla_{2}{ f^{(k)}({x}_{t}^{(k)}, {y}_{t}^{(k)})}\right\|^2  \notag \\
			& \leq   3\Big(\frac{c_{f_y}^2\ell_{g_{yy}}^2}{\mu^2} + \ell_{f_y}^2\Big)\frac{1}{K}\left\|\bar{X}_{t} - {X}_{t}\right\|_F^2 +  3\Big(\frac{c_{f_y}^2\ell_{g_{yy}}^2}{\mu^2}+ \ell_{f_y}^2\Big)\frac{1}{K}\left\| \bar{Y}_t - {Y}_{t}\right\|_F^2 \notag \\
			& \quad + 3\ell_{g_y}^2\frac{1}{K}\left\|{Z}_t- \bar{Z}_{t}\right\|_F^2  \ , 
	\end{align}
	where the last step follows from Assumptions~\ref{assumption_bi_strong}-\ref{assumption_lower_smooth_vr}. 
Then, we have
\begin{align}
			&\quad   \left\|\bar{z}_{ t+1} - z^*(\bar{x}_{t+1})\right\|^2  \leq (1+a)  \left\|\bar{z}_{ t+1} - z^*(\bar{x}_{t})\right\|^2 + (1+1/a)  \left\|z^*(\bar{x}_{t+1})- z^*(\bar{x}_{t})\right\|^2 \notag \\
			& \leq (1+a)  \frac{2}{K}\left\|  Z_{t}  - \bar{Z}_t  \right\|_F^2+ (1+a)2\eta\beta_3^2\frac{1}{K}\left\|   R_t -   \bar{R}_t \right\|_F^2\notag \\
			& \quad  +  (1+a)\left\| \bar{z}_t  - \eta\beta_3  \bar{r}_t -  z^*(\bar{x}_{t})\right\|^2  + (1+1/a)  \left\|z^*(\bar{x}_{t+1})- z^*(\bar{x}_{t})\right\|^2 \notag \\
			& \leq (1+a)  \frac{2}{K}\left\|  Z_{t}  - \bar{Z}_t  \right\|_F^2+ (1+a)2\eta\beta_3^2\frac{1}{K}\left\|   R_t -   \bar{R}_t \right\|_F^2 + (1+1/a)  L_z^2\left\|\bar{x}_{t+1}- \bar{x}_{t}\right\|^2 \notag \\
			& \quad   +  (1+a)(1-\frac{\eta\beta_3\mu}{4}) \left\|\bar{z}_t - z^*(\bar{x}_{t})\right\|^2 +(1+a) \frac{8\eta\beta_3}{\mu}\Big(\frac{c_{f_y}^2\ell_{g_{yy}}^2}{\mu^2}+\ell_{f_y}^2\Big) \left\| \bar{y}_t  - y^*(\bar{x}_t)\right\|^2  \notag \\
			& \quad +(1+a)\frac{60\eta\beta_3}{\mu}\ell_{g_y}^2\frac{1}{K}\left\|{Z}_t- \bar{Z}_{t}\right\|_F^2  +(1+a) \frac{60\eta\beta_3}{\mu}\Big(\frac{c_{f_y}^2\ell_{g_{yy}}^2}{\mu^2} + \ell_{f_y}^2\Big)\frac{1}{K}\left\|\bar{X}_{t} - {X}_{t}\right\|_F^2 \notag \\
			& \quad +  (1+a)\frac{60\eta\beta_3}{\mu}\Big(\frac{c_{f_y}^2\ell_{g_{yy}}^2}{\mu^2}+ \ell_{f_y}^2\Big)\frac{1}{K}\left\| \bar{Y}_t - {Y}_{t}\right\|_F^2 \notag \\
			& \quad  +(1+a)\frac{20\eta\beta_3}{\mu} \left\| \frac{1}{K}\sum_{k=1}^{K}\hat{\mathcal{G}}_{h}^{(k)}({x}_{t}^{(k)}, {y}_{t}^{(k)}, {z}_{t}^{(k)})-   \frac{1}{K}{W}_{t}\mathbf{1}\right\|^2\notag \\
			& \leq  (1-\frac{\eta\beta_3\mu}{8}) \left\|\bar{z}_t - z^*(\bar{x}_{t})\right\|^2 + \frac{9\eta\beta_3}{\mu}\Big(\frac{c_{f_y}^2\ell_{g_{yy}}^2}{\mu^2}+\ell_{f_y}^2\Big) \left\| \bar{y}_t  - y^*(\bar{x}_t)\right\|^2 + \frac{9\eta\beta_1^2L_z^2}{\beta_3\mu} \left\|\bar{u}_{t}\right\|^2  \notag \\
			& \quad + \frac{9\eta\beta_3^2}{4}\frac{1}{K}\left\|   R_t -   \bar{R}_t \right\|_F^2 +\frac{25\eta\beta_3}{\mu} \left\| \frac{1}{K}\sum_{k=1}^{K}\hat{\mathcal{G}}_{h}^{(k)}({x}_{t}^{(k)}, {y}_{t}^{(k)}, {z}_{t}^{(k)})-   \frac{1}{K}{W}_{t}\mathbf{1}\right\|^2\notag \\
			& \quad   + \frac{70\eta\beta_3}{\mu}\Big(\frac{c_{f_y}^2\ell_{g_{yy}}^2}{\mu^2} + \ell_{f_y}^2\Big)\frac{1}{K}\left\|\bar{X}_{t} - {X}_{t}\right\|_F^2+ \frac{70\eta\beta_3}{\mu}\Big(\frac{c_{f_y}^2\ell_{g_{yy}}^2}{\mu^2}+ \ell_{f_y}^2\Big)\frac{1}{K}\left\| \bar{Y}_t - {Y}_{t}\right\|_F^2 \notag \\
			& \quad+\Big(\frac{70\eta\beta_3}{\mu}\ell_{g_y}^2+  \frac{9}{4}\Big)\frac{1}{K}\left\|{Z}_t- \bar{Z}_{t}\right\|_F^2  \ , 
	\end{align}
	where the second step follows from Eq.~(\ref{eq:z-bar-z-star-0}),  the third step follows from Eq.~(\ref{eq:z-bar-z-star-2}), the last step follows from $a = \frac{\eta\beta_3\mu}{8}$ and $\eta\leq\frac{1}{\beta_3\mu}$. 
\begin{align} \label{eq:z-bar-z-star-0}
		&   \quad \left\|\bar{z}_{ t+1} - z^*(\bar{x}_{t})\right\|^2  \notag \\
  &\leq \frac{1}{K}\sum_{k=1}^{K} \left\| z_{ t+1}^{(k)}  -  z^*(\bar{x}_{t})\right\|^2 \notag \\
  & = \frac{1}{K}\sum_{k=1}^{K} \left\| (1-\eta) z_{ t}^{(k)}  + \eta \mathcal{P}[\sum_{j\in\mathcal{N}_{k}}w_{kj}z_{t}^{(j)} -\beta_3  r^{(k)}_t]-  z^*(\bar{x}_{t})\right\|^2 \notag \\
		& \leq \frac{1}{K}\sum_{k=1}^{K} \left\| \eta\sum_{j\in\mathcal{N}_{k}}w_{kj}z_{t}^{(j)} -\eta\beta_3  r^{(k)}_t+(1-\eta) z_{ t}^{(k)}  -  z^*(\bar{x}_{t})\right\|^2 \notag \\
		& \leq \frac{1}{K}\sum_{k=1}^{K} \Big\| \eta\sum_{j\in\mathcal{N}_{k}}w_{kj}z_{t}^{(j)}  -\eta \bar{z}_t + \eta \bar{z}_t -\eta\beta_3  r^{(k)}_t+ \eta\beta_3  \bar{r}_t- \eta\beta_3  \bar{r}_t  +(1-\eta) z_{ t}^{(k)} - (1-\eta) \bar{z}_t\notag \\
		& \quad + (1-\eta) \bar{z}_t  -  z^*(\bar{x}_{t})\Big\|^2 \notag \\
		& = \frac{1}{K}\sum_{k=1}^{K} \left\| \eta\sum_{j\in\mathcal{N}_{k}}w_{kj}z_{t}^{(j)}  -\eta \bar{z}_t -\eta\beta_3  r^{(k)}_t+ \eta\beta_3  \bar{r}_t +(1-\eta) z_{ t}^{(k)} - (1-\eta) \bar{z}_t\right\|^2\notag \\
		& \quad +  \left\| \bar{z}_t  - \eta\beta_3  \bar{r}_t -  z^*(\bar{x}_{t})\right\|^2 \notag \\
		& = \frac{1}{K}\left\| \eta (W Z_{t}  - \bar{Z}_t -\beta_3  R_t+ \beta_3  \bar{R}_t )+(1-\eta) (Z_{ t}- \bar{Z}_t) \right\|_F^2+   \left\| \bar{z}_t  - \eta\beta_3  \bar{r}_t -  z^*(\bar{x}_{t})\right\|^2 \notag \\
		& \leq  \frac{1}{K}\eta\left\|  W Z_{t}  - \bar{Z}_t -\beta_3  R_t+ \beta_3  \bar{R}_t \right\|_F^2+(1-\eta)\frac{1}{K}\left\| Z_{ t}- \bar{Z}_t \right\|_F^2+\left\| \bar{z}_t  - \eta\beta_3  \bar{r}_t -  z^*(\bar{x}_{t})\right\|^2 \notag \\
  		& \leq  2\eta\frac{1}{K}\left\|  W Z_{t}  - \bar{Z}_t  \right\|_F^2+ 2\eta\beta_3^2\frac{1}{K}\left\|   R_t -   \bar{R}_t \right\|_F^2+(1-\eta)\frac{1}{K}\left\| Z_{ t}- \bar{Z}_t \right\|_F^2 \notag \\
  		& \quad +\left\| \bar{z}_t  - \eta\beta_3  \bar{r}_t -  z^*(\bar{x}_{t})\right\|^2 \notag \\
    & \leq  2\eta\lambda^2\frac{1}{K}\left\|   Z_{t}  - \bar{Z}_t  \right\|_F^2+ 2\eta\beta_3^2\frac{1}{K}\left\|   R_t -   \bar{R}_t \right\|_F^2+(1-\eta)\frac{1}{K}\left\| Z_{ t}- \bar{Z}_t \right\|_F^2 \notag \\
    & \quad +\left\| \bar{z}_t  - \eta\beta_3  \bar{r}_t -  z^*(\bar{x}_{t})\right\|^2 \notag \\
    & \leq  2\frac{1}{K}\left\|   Z_{t}  - \bar{Z}_t  \right\|_F^2+ 2\eta\beta_3^2\frac{1}{K}\left\|   R_t -   \bar{R}_t \right\|_F^2+\left\| \bar{z}_t  - \eta\beta_3  \bar{r}_t -  z^*(\bar{x}_{t})\right\|^2  \ , 
\end{align}
where the third step follows from that $z_t^{(k)}$ and $z^*(\bar{x}_t)$ satisfy the constraint and the projection is non-expansive, the fifth step follows from $\frac{1}{K}\sum_{k=1}^{K} (\eta\sum_{j\in\mathcal{N}_{k}}w_{kj}z_{t}^{(j)}  -\eta \bar{z}_t -\eta\beta_3  r^{(k)}_t+ \eta\beta_3  \bar{r}_t +(1-\eta) z_{ t}^{(k)} - (1-\eta) \bar{z}_t) =0$, 
the fourth  to last step follows from  $\left\|x+y\right\|^2 \leq (1+a)\left\|x\right\|^2 + (1+1/a)\left\|y\right\|^2$ with $a=\frac{\eta}{1-\eta}$, and the last step follows from $\eta<1$ and $\lambda<1$.

\end{proof}

\begin{lemma} \label{lemma_y_opt}
		Under Assumptions~\ref{assumption_bi_strong}-\ref{assumption_graph}, if  $\beta_2\leq \frac{1}{6\ell_{g_y}}$, we have
\begin{align}
		&   \mathbb{E}[\|\bar{   {y}}_{t+1} -    {y}^{*}(\bar{   {x}}_{t+1})\| ^2 ] \leq  (1-\frac{\beta_2\eta\mu}{4}) \mathbb{E}[\|\bar{   {y}}_{t}   -    {y}^{*}(\bar{   {x}}_t)\| ^2] - \frac{3\eta\beta_2^2}{4} \mathbb{E}[\|\bar{v}_{t}  \|^2]   \notag \\
		& \quad  +\frac{25\eta\beta_1^2L_{y}^2 }{6\beta_2\mu} \mathbb{E}[\|\bar{u}_{t}\| ^2 ] +  \frac{25\beta_2 \eta \ell_{g_y}^2}{3\mu}  \frac{1}{K}\mathbb{E}[\| \bar{X}_t - {X}_t\|_F^2] +  \frac{25\beta_2 \eta \ell_{g_y}^2}{3\mu}  \frac{1}{K}\mathbb{E}[\|\bar{Y}_t -  {Y}_t\|_F^2]\notag \\
		& \quad  +  \frac{25\beta_2 \eta}{3\mu}  \mathbb{E}[\|\frac{1}{K} \delta^{g}  ({X}_t, {Y}_t) \mathbf{1} - \frac{1}{K}{V}_t\mathbf{1} \|^2]  \ . 
\end{align}
\end{lemma}
This lemma can be proved by following Eq.~(52) in \cite{gao2023convergence}.

\begin{lemma} \label{lemma_h_storm_var_mean}
	Under Assumptions~\ref{assumption_bi_strong}-\ref{assumption_graph} and $\eta \leq \frac{1}{\sqrt{\alpha_3}}$,  we have the following two inequalities:
	\begin{align} \label{eq:h_storm_var_mean}
			&\quad  \mathbb{E} \left[ \left\|(\delta^{\hat{\mathcal{G}}_{h}}(X_t, Y_t, Z_t) -   W_t) \frac{1}{K} \mathbf{1} \right\|^2 \right] \notag \\
			& \leq   (1-\alpha_3 \eta^2) \mathbb{E} \left[ \left\| (\delta^{\hat{\mathcal{G}}_{h}}(X_{t-1}, Y_{t-1}, Z_{t-1}) -  W_{t-1}) \frac{1}{K} \mathbf{1}  \right\|^2 \right]\notag \\
			& \quad +6(\frac{c_{f_y}^2\ell_{g_{yy}}^2}{\mu^2}+\ell_{f_y}^2)\frac{1}{K^2}\mathbb{E}\left[\left\|X_{t} - X_{t-1}\right\|_F^2\right]  + 6(\frac{c_{f_y}^2\ell_{g_{yy}}^2}{\mu^2}+\ell_{f_y}^2)\frac{1}{K^2}\mathbb{E}\left[\left\|Y_{t} - Y_{t-1}\right\|_F^2\right] \notag \\
			& \quad + 6\ell_{g_y}^2 \frac{1}{K^2}\mathbb{E}\left[\left\|Z_{t} - Z_{t-1}\right\|_F^2\right] + 4\alpha_3^2\eta^4(1 + \frac{c_{f_y}^2}{\mu^2})\sigma^2\frac{1}{K} \ , 
	\end{align}
	and
	\begin{align} \label{eq:h_storm_var_individual}
			& \quad \frac{1}{K}\mathbb{E} \left[ \left\|\delta^{\hat{\mathcal{G}}_{h}}(X_t, Y_t, Z_t) -   W_t\right\|^2 \right]\notag \\
			&  \leq   (1-\alpha_3 \eta^2) \frac{1}{K}\mathbb{E} \left[ \left\| \delta^{\hat{\mathcal{G}}_{h}}(X_{t-1}, Y_{t-1}, Z_{t-1}) -  W_{t-1}  \right\|^2 \right]\notag \\
			& \quad +6(\frac{c_{f_y}^2\ell_{g_{yy}}^2}{\mu^2}+\ell_{f_y}^2)\frac{1}{K}\mathbb{E}\left[\left\|X_{t} - X_{t-1}\right\|_F^2\right]  + 6(\frac{c_{f_y}^2\ell_{g_{yy}}^2}{\mu^2}+\ell_{f_y}^2)\frac{1}{K}\mathbb{E}\left[\left\|Y_{t} - Y_{t-1}\right\|_F^2\right] \notag \\
			& \quad + 6\ell_{g_y}^2 \frac{1}{K}\mathbb{E}\left[\left\|Z_{t} - Z_{t-1}\right\|_F^2\right] + 4\alpha_3^2\eta^4(1 + \frac{c_{f_y}^2}{\mu^2})\sigma^2 \ . 
	\end{align}
\end{lemma}

\begin{proof}
	Eq.~(\ref{eq:h_storm_var_mean}) can be proved as follows:
	\begin{align}
			& \quad\mathbb{E} \left[ \left\|(\delta^{\hat{\mathcal{G}}_{h}}(X_t, Y_t, Z_t) -   W_t) \frac{1}{K} \mathbf{1} \right\|^2 \right]\notag \\
			& = \mathbb{E} \Big[ \Big\| \Big( (1-\alpha_3 \eta^2)(\delta^{\hat{\mathcal{G}}_{h}}(X_{t-1}, Y_{t-1}, Z_{t-1}) -  W_{t-1}) \notag \\
			& \quad +(1-\alpha_3 \eta^2)(\delta^{\hat{\mathcal{G}}_{h}}(X_t, Y_t, Z_t; \hat{\xi}_t)  - \delta^{\hat{\mathcal{G}}_{h}}(X_t, Y_t, Z_t; \hat{\xi}_t)   \notag \\
			& \qquad - \delta^{\hat{\mathcal{G}}_{h}}(X_t, Y_t, Z_t)  + \delta^{\hat{\mathcal{G}}_{h}}(X_{t-1}, Y_{t-1}, Z_{t-1}) ) \notag \\
			& \quad + \alpha_3\eta^2(\delta^{\hat{\mathcal{G}}_{h}}(X_t, Y_t, Z_t; \hat{\xi}_t)- \delta^{\hat{\mathcal{G}}_{h}}(X_t, Y_t, Z_t))\Big )\frac{1}{K} \mathbf{1}  \Big\|^2 \Big]\notag \\
			& \leq   (1-\alpha_3 \eta^2)^2 \mathbb{E} \left[ \left\| (\delta^{\hat{\mathcal{G}}_{h}}(X_{t-1}, Y_{t-1}, Z_{t-1}) -  W_{t-1}) \frac{1}{K} \mathbf{1}  \right\|^2 \right]\notag \\
			& \quad +2(1-\alpha_3 \eta^2)^2\frac{1}{K^2}\mathbb{E} \Big[ \Big\|  \delta^{\hat{\mathcal{G}}_{h}}(X_t, Y_t, Z_t; \hat{\xi}_t)  - \delta^{\hat{\mathcal{G}}_{h}}(X_{t-1}, Y_{t-1}, Z_{t-1}; \hat{\xi}_t)  \notag \\
			& \quad \quad - \delta^{\hat{\mathcal{G}}_{h}}(X_t, Y_t, Z_t)  + \delta^{\hat{\mathcal{G}}_{h}}(X_{t-1}, Y_{t-1}, Z_{t-1})    \Big\|_F^2 \Big]\notag \\
			& \quad + 2\alpha_3^2\eta^4\frac{1}{K^2}\mathbb{E} \Big[ \left\|\delta^{\hat{\mathcal{G}}_{h}}(X_t, Y_t, Z_t; \hat{\xi}_t)- \delta^{\hat{\mathcal{G}}_{h}}(X_t, Y_t, Z_t) \right\|_F^2 \Big]\notag \\
			& \leq   (1-\alpha_3 \eta^2) \mathbb{E} \left[ \left\| (\delta^{\hat{\mathcal{G}}_{h}}(X_{t-1}, Y_{t-1}, Z_{t-1}) -  W_{t-1}) \frac{1}{K} \mathbf{1}  \right\|^2 \right]\notag \\
			& \quad +2\frac{1}{K^2} \mathbb{E} \left[ \left\|  \delta^{\hat{\mathcal{G}}_{h}}(X_t, Y_t, Z_t; \hat{\xi}_t)  - \delta^{\hat{\mathcal{G}}_{h}}(X_{t-1}, Y_{t-1}, Z_{t-1}; \hat{\xi}_t)   \right\|_F^2 \right]  \notag \\
			& \quad + 2\alpha_3^2\eta^4\frac{1}{K^2}\mathbb{E} \left[ \left\|\delta^{\hat{\mathcal{G}}_{h}}(X_t, Y_t, Z_t; \hat{\xi}_t)- \delta^{\hat{\mathcal{G}}_{h}}(X_t, Y_t, Z_t) \right\|_F^2 \right]\notag \\
			& \leq   (1-\alpha_3 \eta^2) \mathbb{E} \left[ \left\| (\delta^{\hat{\mathcal{G}}_{h}}(X_{t-1}, Y_{t-1}, Z_{t-1}) -  W_{t-1}) \frac{1}{K} \mathbf{1}  \right\|^2 \right]\notag \\
			& \quad +6(\frac{c_{f_y}^2\ell_{g_{yy}}^2}{\mu^2}+\ell_{f_y}^2)\frac{1}{K^2}\mathbb{E}\left[\left\|X_{t} - X_{t-1}\right\|_F^2\right]  + 6(\frac{c_{f_y}^2\ell_{g_{yy}}^2}{\mu^2}+\ell_{f_y}^2)\frac{1}{K^2}\mathbb{E}\left[\left\|Y_{t} - Y_{t-1}\right\|_F^2\right] \notag \\
			& \quad + 6\ell_{g_y}^2 \frac{1}{K^2}\mathbb{E}\left[\left\|Z_{t} - Z_{t-1}\right\|_F^2\right] + 4\alpha_3^2\eta^4(1 + \frac{c_{f_y}^2}{\mu^2})\sigma^2\frac{1}{K} \ , 
	\end{align}
	where the last step follows from the following two inequalities:
	\begin{align} \label{eq:h-var}
			& \quad \mathbb{E} \left[ \left\|  \delta^{\hat{\mathcal{G}}_{h}}(X_t, Y_t, Z_t; \hat{\xi}_t)  - \delta^{\hat{\mathcal{G}}_{h}}(X_{t-1}, Y_{t-1}, Z_{t-1}; \hat{\xi}_t)   \right\|_F^2 \right]\notag \\
			& = \sum_{k=1}^{K}\mathbb{E} \Big[ \Big\| \nabla_{22}^2g^{(k)}(x_{t}^{(k)}, y_{t}^{(k)}; \zeta_{t}^{(k)}) z_{t}^{(k)}-  \nabla_{2}{ f^{(k)}(x_{t}^{(k)}, y_{t}^{(k)}; \xi_{t}^{(k)})} \notag \\
			& \quad  - \nabla_{22}^2g^{(k)}(x_{t-1}^{(k)}, y_{t-1}^{(k)}; \zeta_{t}^{(k)}) z_{t-1}^{(k)} + \nabla_{2}{ f^{(k)}(x_{t-1}^{(k)}, y_{t-1}^{(k)}; \xi_{t}^{(k)})} \Big\|^2 \Big]\notag \\
			& = \sum_{k=1}^{K}\mathbb{E} \Big[ \Big\| \nabla_{22}^2g^{(k)}(x_{t}^{(k)}, y_{t}^{(k)}; \zeta_{t}^{(k)}) z_{t}^{(k)} - \nabla_{22}^2g^{(k)}(x_{t-1}^{(k)}, y_{t-1}^{(k)}; \zeta_{t}^{(k)}) z_{t}^{(k)} \notag \\
			& \quad + \nabla_{22}^2g^{(k)}(x_{t-1}^{(k)}, y_{t-1}^{(k)}; \zeta_{t}^{(k)}) z_{t}^{(k)} - \nabla_{22}^2g^{(k)}(x_{t-1}^{(k)}, y_{t-1}^{(k)}; \zeta_{t}^{(k)}) z_{t-1}^{(k)} \notag \\
			& \quad -  \nabla_{2}{ f^{(k)}(x_{t}^{(k)}, y_{t}^{(k)}; \xi_{t}^{(k)})} + \nabla_{2}{ f^{(k)}(x_{t-1}^{(k)}, y_{t-1}^{(k)}; \xi_{t}^{(k)})} \Big\|^2 \Big]\notag \\
			& \leq 3(\frac{c_{f_y}^2\ell_{g_{yy}}^2}{\mu^2}+\ell_{f_y}^2)\mathbb{E}\left[\left\|X_{t} - X_{t-1}\right\|_F^2\right]  + 3(\frac{c_{f_y}^2\ell_{g_{yy}}^2}{\mu^2}+\ell_{f_y}^2)\mathbb{E}\left[\left\|Y_{t} - Y_{t-1}\right\|_F^2\right] \notag \\
			& \quad + 3\ell_{g_y}^2 \mathbb{E}\left[\left\|Z_{t} - Z_{t-1}\right\|_F^2\right]  \ , 
	\end{align}
	and 
	\begin{align}\label{eq:stochastic-grad-h-incremental}
			& \quad \mathbb{E} \left[ \left\|\delta^{\hat{\mathcal{G}}_{h}}(X_t, Y_t, Z_t; \hat{\xi}_t)- \delta^{\hat{\mathcal{G}}_{h}}(X_t, Y_t, Z_t) \right\|_F^2 \right]\notag \\
			& =\sum_{k=1}^{K} \mathbb{E} \Big[ \Big\|\nabla_{22}^2g^{(k)}(x_{t}^{(k)}, y_{t}^{(k)}; \zeta_{t}^{(k)}) z_{t}^{(k)}-  \nabla_2{ f^{(k)}(x_{t}^{(k)}, y_{t}^{(k)}; \xi_{t}^{(k)})}  \notag \\
			& \quad - \nabla_{22}^2g^{(k)}(x_{t}^{(k)}, y_{t}^{(k)}) z_{t}^{(k)} +  \nabla_2{ f^{(k)}(x_{t}^{(k)}, y_{t}^{(k)})}  \Big\|^2 \Big]\notag \\
			& \leq 2(1 + \frac{c_{f_y}^2}{\mu^2})\sigma^2K \ . 
	\end{align}
	Eq.~(\ref{eq:h_storm_var_individual}) can be proved by following the above proof so that we omit the detailed steps. 
	
\end{proof}

\begin{lemma} \label{lemma_hyper_storm_var_mean}
	Under Assumptions~\ref{assumption_bi_strong}-\ref{assumption_graph} and $\eta \leq \frac{1}{\sqrt{\alpha_1}}$, we have the following two inequalities:
		\begin{align} \label{eq:hyper_storm_var_mean}
			&  \mathbb{E} \left[ \left\|\frac{1}{K}\delta^{\hat{\mathcal{G}}_{F}}(X_t, Y_t, Z_t) \mathbf{1} -  \frac{1}{K} U_t \mathbf{1} \right\|^2 \right] \notag \\
			&  \leq  (1-\alpha_1 \eta^2) \mathbb{E} \left[ \left\| \frac{1}{K}\delta^{\hat{\mathcal{G}}_{F}}(X_{t-1}, Y_{t-1}, Z_{t-1}) \mathbf{1} -  \frac{1}{K}U_{t-1} \mathbf{1}  \right\|^2 \right]\notag \\
			& \quad + 6(\ell_{f_x}^2+\frac{c_{f_y}^2\ell_{g_{xy}}^2}{\mu^2})\frac{1}{K^2}\mathbb{E}[\left\|X_{t} - X_{t-1}\right\|_F^2]  + 6(\ell_{f_x}^2+\frac{c_{f_y}^2\ell_{g_{xy}}^2}{\mu^2})\frac{1}{K^2}\mathbb{E}[\left\|Y_{t} - Y_{t-1}\right\|_F^2]\notag \\
			& \quad + 6c_{g_{xy}}^2\frac{1}{K^2}\mathbb{E}[\left\|Z_{t} - Z_{t-1}\right\|_F^2] + 4\alpha_1^2\eta^4 (1 + \frac{c_{f_y}^2}{\mu^2})\sigma^2\frac{1}{K} \ ,  
	\end{align}
	and 
\begin{align}\label{eq:hyper_storm_var_individual}
		& \quad\frac{1}{K}\mathbb{E} \Big[ \left\|\delta^{\hat{\mathcal{G}}_{F}}(X_t, Y_t, Z_t) -   U_t  \right\|_F^2 \Big] \notag \\
		&   \leq  (1-\alpha_1 \eta^2) \frac{1}{K}\mathbb{E} \Big[ \left\| \delta^{\hat{\mathcal{G}}_{F}}(X_{t-1}, Y_{t-1}, Z_{t-1}) - U_{t-1}  \right\|_F^2 \Big]\notag \\
		& \quad + 6(\ell_{f_x}^2+\frac{c_{f_y}^2\ell_{g_{xy}}^2}{\mu^2})\frac{1}{K}\mathbb{E}[\left\|X_{t} - X_{t-1}\right\|_F^2] + 6(\ell_{f_x}^2+\frac{c_{f_y}^2\ell_{g_{xy}}^2}{\mu^2})\frac{1}{K}\mathbb{E}[\left\|Y_{t} - Y_{t-1}\right\|_F^2]\notag \\
		& \quad + 6c_{g_{xy}}^2\frac{1}{K}\mathbb{E}[\left\|Z_{t} - Z_{t-1}\right\|_F^2] + 4\alpha_1^2\eta^4 (1 + \frac{c_{f_y}^2}{\mu^2})\sigma^2 \ . 
\end{align}
	
\end{lemma}

\begin{proof}
	Eq.~(\ref{eq:hyper_storm_var_mean}) can be proved as follows:
	\begin{align}
			& \quad\mathbb{E} \left[ \left\|(\delta^{\hat{\mathcal{G}}_{F}}(X_t, Y_t, Z_t) -   U_t) \frac{1}{K} \mathbf{1} \right\|^2 \right]\notag \\
			& \leq   (1-\alpha_1 \eta^2) \mathbb{E} \left[ \left\| (\delta^{\hat{\mathcal{G}}_{F}}(X_{t-1}, Y_{t-1}, Z_{t-1}) -  U_{t-1}) \frac{1}{K} \mathbf{1}  \right\|^2 \right]\notag \\
			& \quad +2\frac{1}{K^2}\mathbb{E} \left[ \left\|  \delta^{\hat{\mathcal{G}}_{F}}(X_t, Y_t, Z_t; \hat{\xi}_t)  - \delta^{\hat{\mathcal{G}}_{F}}(X_{t-1}, Y_{t-1}, Z_{t-1}; \hat{\xi}_t)   \right\|_F^2 \right]\notag \\
			& \quad + 2\alpha_1^2\eta^4\frac{1}{K^2}\mathbb{E}\left [ \left\|\delta^{\hat{\mathcal{G}}_{F}}(X_t, Y_t, Z_t; \hat{\xi}_t)- \delta^{\hat{\mathcal{G}}_{F}}(X_t, Y_t, Z_t) \right\|_F^2 \right]\notag \\
			& \leq  (1-\alpha_1 \eta^2) \mathbb{E} \left[ \left\| (\delta^{\hat{\mathcal{G}}_{F}}(X_{t-1}, Y_{t-1}, Z_{t-1}) -  U_{t-1}) \frac{1}{K} \mathbf{1}  \right\|^2 \right]\notag \\
			& \quad + 6(\ell_{f_x}^2+\frac{c_{f_y}^2\ell_{g_{xy}}^2}{\mu^2})\frac{1}{K^2}\mathbb{E}[\left\|X_{t} - X_{t-1}\right\|_F^2] + 6(\ell_{f_x}^2+\frac{c_{f_y}^2\ell_{g_{xy}}^2}{\mu^2})\frac{1}{K^2}\mathbb{E}[\left\|Y_{t} - Y_{t-1}\right\|_F^2]\notag \\
			& \quad + 6c_{g_{xy}}^2\frac{1}{K^2}\mathbb{E}[\left\|Z_{t} - Z_{t-1}\right\|_F^2] + 4\alpha_1^2\eta^4 (1 + \frac{c_{f_y}^2}{\mu^2})\sigma^2\frac{1}{K} \ , 
	\end{align}
where the first step can be obtained by following the proof of Lemma~\ref{lemma_h_storm_var_mean}, the last step follows from the following two inequalities:
	\begin{align} \label{eq:F-var}
			& \quad \mathbb{E} \left[ \left\|\delta^{\hat{\mathcal{G}}_{F}}(X_t, Y_t, Z_t; \hat{\xi}_t)- \delta^{\hat{\mathcal{G}}_{F}}(X_t, Y_t, Z_t)\right\|_F^2 \right]\notag \\
			& = \sum_{k=1}^{K}\mathbb{E} \left[ \left\|\hat{\mathcal{G}}_{F}^{(k)}(x_{t}^{(k)}, y_{t}^{(k)}, z_{t}^{(k)};  \hat{\xi}_{t}^{(k)}) - \hat{\mathcal{G}}_{F}^{(k)}(x_{t}^{(k)}, y_{t}^{(k)}, z_{t}^{(k)}) \right\|^2 \right]\notag \\
			& =\sum_{k=1}^{K} \mathbb{E} \Big[ \Big\|\nabla_{1} { f^{(k)}(x_{t}^{(k)}, y_{t}^{(k)}; {\xi}_{t}^{(k)})} - \nabla_{12}^2 g^{(k)}(x_{t}^{(k)}, y_{t}^{(k)}; {\zeta}_{t}^{(k)})z_t^{(k)} \notag \\
			& \quad - \nabla_{1} { f^{(k)}(x_{t}^{(k)}, y_{t}^{(k)})} + \nabla_{12}^2 g^{(k)}(x_{t}^{(k)}, y_{t}^{(k)})z_t^{(k)}  \Big\|^2 \Big]\notag \\
			& \leq 2(1 + \frac{c_{f_y}^2}{\mu^2})\sigma^2K \ , 
	\end{align}
and 
\begin{align}\label{eq:stochastic-grad-F-incremental}
		&\quad  \mathbb{E} \left[ \left\|  \delta^{\hat{\mathcal{G}}_{F}}(X_t, Y_t, Z_t; \hat{\xi}_t)  - \delta^{\hat{\mathcal{G}}_{F}}(X_{t-1}, Y_{t-1}, Z_{t-1}; \hat{\xi}_t) \right\|_F^2 \right]\notag \\
		&   = \sum_{k=1}^{K} \mathbb{E} \Big[ \Big\|\nabla_{1} { f^{(k)}(x_{t}^{(k)}, y_{t}^{(k)}; {\xi}_{t}^{(k)})}  -  \nabla_{1} { f^{(k)}(x_{t-1}^{(k)}, y_{t-1}^{(k)}; {\xi}_{t}^{(k)})}  \notag \\
		& \quad - \nabla_{12}^2 g^{(k)}(x_{t}^{(k)}, y_{t}^{(k)}; {\zeta}_{t}^{(k)})z_t^{(k)}  + \nabla_{12}^2 g^{(k)}(x_{t}^{(k)}, y_{t}^{(k)}; {\zeta}_{t}^{(k)})z_{t-1}^{(k)} \notag \\
		& \quad - \nabla_{12}^2 g^{(k)}(x_{t}^{(k)}, y_{t}^{(k)}; {\zeta}_{t}^{(k)})z_{t-1}^{(k)} + \nabla_{12}^2 g^{(k)}(x_{t-1}^{(k)}, y_{t-1}^{(k)}; {\zeta}_{t}^{(k)})z_{t-1}^{(k)} \Big\|^2 \Big]\notag \\
		& \leq 3(\ell_{f_x}^2+\frac{c_{f_y}^2\ell_{g_{xy}}^2}{\mu^2})\mathbb{E}[\left\|X_{t} - X_{t-1}\right\|_F^2]+ 3(\ell_{f_x}^2+\frac{c_{f_y}^2\ell_{g_{xy}}^2}{\mu^2})\mathbb{E}[\left\|Y_{t} - Y_{t-1}\right\|_F^2]\notag \\
		& \quad + 3c_{g_{xy}}^2\mathbb{E}[\left\|Z_{t} - Z_{t-1}\right\|_F^2] \ . 
\end{align}
Eq.~(\ref{eq:hyper_storm_var_individual}) can be proved by following the above proof so that we omit the detailed steps. 

\end{proof}

\begin{lemma}  \label{lemma_g_storm_var_mean}
		Under Assumptions~\ref{assumption_bi_strong}-\ref{assumption_graph} and  $\eta \leq \frac{1}{\sqrt{\alpha_2}}$,  we have the following two inequalities:
	\begin{align} \label{eq:g_storm_var_mean}
			& \mathbb{E} \left[ \left\|(\delta^{g}(X_t, Y_t ) -   V_t) \frac{1}{K} \mathbf{1} \right\|^2 \right]\leq  (1-\alpha_2 \eta^2) \mathbb{E} \left[ \left\| (\delta^{g}(X_{t-1}, Y_{t-1} ) -  V_{t-1}) \frac{1}{K} \mathbf{1}  \right\|^2 \right]\notag \\
			& \quad + 2\ell_{g_y}^2\frac{1}{K^2}\mathbb{E}\left[\left\|X_{t} - X_{t-1}\right\|_F^2\right] + 2\ell_{g_y}^2\frac{1}{K^2}\mathbb{E}\left[\left\|Y_{t} - Y_{t-1}\right\|_F^2\right] + 2\alpha_2^2\eta^4 \sigma^2\frac{1}{K} \ , 
	\end{align}
	and
\begin{align} \label{eq:g_storm_var_individual}
		& \frac{1}{K} \mathbb{E} \left[ \left\|\delta^{g}(X_t, Y_t ) -   V_t\right\|_F^2 \right] \leq  (1-\alpha_2 \eta^2) \frac{1}{K} \mathbb{E} \left[ \left\| \delta^{g}(X_{t-1}, Y_{t-1} ) -  V_{t-1} \right\|_F^2 \right]\notag \\
		& \quad + 2\ell_{g_y}^2\frac{1}{K}\mathbb{E}\left[\left\|X_{t} - X_{t-1}\right\|_F^2\right] + 2\ell_{g_y}^2\frac{1}{K}\mathbb{E}\left[\left\|Y_{t} - Y_{t-1}\right\|_F^2\right] + 2\alpha_2^2\eta^4 \sigma^2 \ . 
\end{align}
\end{lemma}

\begin{proof}
	Eq.~(\ref{eq:g_storm_var_mean}) can be proved as follows:
		\begin{align}
			& \quad\mathbb{E} \left[ \left\|(\delta^{g}(X_t, Y_t ) -   V_t) \frac{1}{K} \mathbf{1} \right\|^2 \right]\notag \\
			& \leq   (1-\alpha_2 \eta^2) \mathbb{E} \left[ \left\| (\delta^{g}(X_{t-1}, Y_{t-1} ) -  V_{t-1}) \frac{1}{K} \mathbf{1}  \right\|^2 \right]\notag \\
			& \quad +2\frac{1}{K^2}\mathbb{E} \left[ \left\|  \delta^{g}(X_t, Y_t ; {\zeta}_t)  - \delta^{g}(X_{t-1}, Y_{t-1} ; {\zeta}_t)   \right\|_F^2 \right] \notag \\
			& \quad + 2\alpha_2^2\eta^4\frac{1}{K^2}\mathbb{E} \left[ \left\|\delta^{g}(X_t, Y_t ; {\zeta}_t)- \delta^{g}(X_t, Y_t ) \right\|_F^2 \right]\notag \\
			& \leq  (1-\alpha_2 \eta^2) \mathbb{E} \left[ \left\| (\delta^{g}(X_{t-1}, Y_{t-1} ) -  V_{t-1}) \frac{1}{K} \mathbf{1}  \right\|^2 \right]\notag \\
			& \quad + 2\ell_{g_y}^2\frac{1}{K^2}\mathbb{E}\left[\left\|X_{t} - X_{t-1}\right\|_F^2\right] + 2\ell_{g_y}^2\frac{1}{K^2}\mathbb{E}\left[\left\|Y_{t} - Y_{t-1}\right\|_F^2\right] + 2\alpha_2^2\eta^4 \sigma^2\frac{1}{K} \  , 
	\end{align}
	where  the first step can be obtained by following the proof of Lemma~\ref{lemma_h_storm_var_mean}, the last step follows from Assumption~\ref{assumption_lower_smooth_vr} and Assumption~\ref{assumption_variance}.  
	
Eq.~(\ref{eq:g_storm_var_individual}) can be proved by following the above proof so that we omit the detailed steps. 
\end{proof}

\begin{lemma} \label{lemma_consensus_r}
	Under Assumptions~\ref{assumption_bi_strong}-\ref{assumption_graph}, we have 
	\begin{align}
			& \quad \frac{1}{K}\mathbb{E}\left[\left\|R_{t} - \bar{R}_{t}\right\|_F^2\right] \notag \\
			& \leq  \lambda\frac{1}{K}\mathbb{E}\left[\left\| R_{t-1} -  \bar{R}_{t-1} \right\|_F^2\right] +  \frac{3\alpha_3^2\eta^4}{1-\lambda}\frac{1}{K}\mathbb{E}\left[\left\| W_{t-1}- \delta^{\hat{\mathcal{G}}_{h}}(X_{t-1}, Y_{t-1}, Z_{t-1})   \right\|_F^2\right]\notag \\
			& \quad  + (\ell_{f_y}^2+\frac{c_{f_y}^2\ell_{g_{yy}}^2}{\mu^2})\frac{9}{1-\lambda}\frac{1}{K} \mathbb{E}\left[\left\|X_{t} - X_{t-1}\right\|_F^2\right]\notag \\
			& \quad + (\ell_{f_y}^2+\frac{c_{f_y}^2\ell_{g_{yy}}^2}{\mu^2})\frac{9}{1-\lambda}\frac{1}{K} \mathbb{E}\left[\left\|Y_{t} - Y_{t-1}\right\|_F^2\right]\notag \\
			& \quad + \ell_{g_y}^2\frac{9}{1-\lambda}\frac{1}{K} \mathbb{E}\left[\left\|Z_{t} - Z_{t-1}\right\|_F^2\right] +  \frac{6\alpha_3^2\eta^4}{1-\lambda}(1 + \frac{c_{f_y}^2}{\mu^2})\sigma^2 \ . 
	\end{align}
\end{lemma}

\begin{proof}
	\begin{align}
			& \quad \frac{1}{K}\mathbb{E}\left[\left\|R_{t} - \bar{R}_{t}\right\|_F^2\right]  \notag \\
			& = \frac{1}{K}\mathbb{E}\left[\left\| R_{t-1}E +W_{t} - W_{t-1} -  \bar{R}_{t-1}- \bar{W}_{t} +\bar{W}_{t-1}\right\|_F^2\right] \notag \\
			&\leq \frac{1}{K}(1+a)\mathbb{E}\left[\left\| R_{t-1}E -  \bar{R}_{t-1} \right\|_F^2\right]  + (1+\frac{1}{a})\frac{1}{K}\mathbb{E}\left[\left\|W_{t} - W_{t-1} - \bar{W}_{t} +\bar{W}_{t-1}\right\|_F^2\right]\notag \\
			& \leq \lambda\frac{1}{K}\mathbb{E}\left[\left\| R_{t-1} -  \bar{R}_{t-1} \right\|_F^2\right]  + \frac{1}{1-\lambda}\frac{1}{K}\mathbb{E}\left[\left\|W_{t} - W_{t-1} \right\|_F^2\right]\notag \\
			& = \lambda\frac{1}{K}\mathbb{E}\left[\left\| R_{t-1} -  \bar{R}_{t-1} \right\|_F^2\right]   \notag \\
			& \quad + \frac{1}{1-\lambda}\frac{1}{K}\mathbb{E}\left[\left\|(1-\alpha_3\eta^2)(W_{t-1} - \delta^{\hat{\mathcal{G}}_{h}}(X_{t-1}, Y_{t-1}, Z_{t-1}; \hat{\xi}_t))  + \delta^{\hat{\mathcal{G}}_{h}}(X_t, Y_t, Z_t; \hat{\xi}_t)- W_{t-1} \right\|_F^2\right]\notag \\
			& \leq  \lambda\frac{1}{K}\mathbb{E}\left[\left\| R_{t-1} -  \bar{R}_{t-1} \right\|_F^2\right] + \frac{3}{1-\lambda}\frac{1}{K}\mathbb{E}\left[\left\|\delta^{\hat{\mathcal{G}}_{h}}(X_{t}, Y_{t}, Z_{t}; \hat{\xi}_t) - \delta^{\hat{\mathcal{G}}_{h}}(X_{t-1}, Y_{t-1}, Z_{t-1}; \hat{\xi}_t)    \right\|_F^2\right]\notag \\
			& \quad +  \frac{3\alpha_3^2\eta^4}{1-\lambda}\frac{1}{K}\mathbb{E}\left[\left\| W_{t-1}- \delta^{\hat{\mathcal{G}}_{h}}(X_{t-1}, Y_{t-1}, Z_{t-1})   \right\|_F^2\right] \notag \\
			& \quad +  \frac{3\alpha_3^2\eta^4}{1-\lambda}\frac{1}{K}\mathbb{E}\left[\left\|\delta^{\hat{\mathcal{G}}_{h}}(X_{t-1}, Y_{t-1}, Z_{t-1})  - \delta^{\hat{\mathcal{G}}_{h}}(X_{t-1}, Y_{t-1}, Z_{t-1}; \hat{\xi}_t)   \right\|_F^2\right]\notag \\
			& \leq  \lambda\frac{1}{K}\mathbb{E}\left[\left\| R_{t-1} -  \bar{R}_{t-1} \right\|_F^2\right] +  \frac{3\alpha_3^2\eta^4}{1-\lambda}\frac{1}{K}\mathbb{E}\left[\left\| W_{t-1}- \delta^{\hat{\mathcal{G}}_{h}}(X_{t-1}, Y_{t-1}, Z_{t-1})   \right\|_F^2\right]\notag \\
			& \quad  + (\ell_{f_y}^2+\frac{c_{f_y}^2\ell_{g_{yy}}^2}{\mu^2})\frac{9}{1-\lambda}\frac{1}{K} \mathbb{E}\left[\left\|X_{t} - X_{t-1}\right\|_F^2\right]+ (\ell_{f_y}^2+\frac{c_{f_y}^2\ell_{g_{yy}}^2}{\mu^2})\frac{9}{1-\lambda}\frac{1}{K} \mathbb{E}\left[\left\|Y_{t} - Y_{t-1}\right\|_F^2\right]\notag \\
			& \quad + \ell_{g_y}^2\frac{9}{1-\lambda}\frac{1}{K} \mathbb{E}\left[\left\|Z_{t} - Z_{t-1}\right\|_F^2\right] +  \frac{6\alpha_3^2\eta^4}{1-\lambda}(1 + \frac{c_{f_y}^2}{\mu^2})\sigma^2  \  ,  
	\end{align}
	where the third step follows from $a=\frac{1-\lambda}{\lambda}$ and  $\left\| R_{t-1}E -  \bar{R}_{t-1} \right\|_F^2\leq \lambda^2 \left\| R_{t-1} -  \bar{R}_{t-1} \right\|_F^2$, the last step follows from Eqs.~(\ref{eq:h-var}-\ref{eq:stochastic-grad-h-incremental}).

\end{proof}

\begin{lemma} \label{lemma_consensus_p}
			Under Assumptions~\ref{assumption_bi_strong}-\ref{assumption_graph}, we have 
		\begin{align}
			&  \frac{1}{K}\mathbb{E}\left[\left\|P_{t} - \bar{P}_{t}\right\|_F^2\right] \notag \\
			& \leq  \lambda\frac{1}{K}\mathbb{E}\left[\left\| P_{t-1} -  \bar{P}_{t-1} \right\|_F^2\right] +  \frac{3\alpha_1^2\eta^4}{1-\lambda}\frac{1}{K}\mathbb{E}\left[\left\| U_{t-1}- \delta^{\hat{\mathcal{G}}_{F}}(X_{t-1}, Y_{t-1}, Z_{t-1})   \right\|_F^2\right]\notag \\
			& \quad  + (\ell_{f_x}^2+\frac{c_{f_y}^2\ell_{g_{xy}}^2}{\mu^2})\frac{9}{1-\lambda}\frac{1}{K} \mathbb{E}\left[\left\|X_{t} - X_{t-1}\right\|_F^2\right]\notag \\
			& \quad + (\ell_{f_x}^2+\frac{c_{f_y}^2\ell_{g_{xy}}^2}{\mu^2})\frac{9}{1-\lambda}\frac{1}{K} \mathbb{E}\left[\left\|Y_{t} - Y_{t-1}\right\|_F^2\right]\notag \\
			& \quad + c_{g_{xy}}^2\frac{9}{1-\lambda}\frac{1}{K} \mathbb{E}\left[\left\|Z_{t} - Z_{t-1}\right\|_F^2\right] +  \frac{6\alpha_1^2\eta^4}{1-\lambda}(1 + \frac{c_{f_y}^2}{\mu^2})\sigma^2\ . 
	\end{align}
\end{lemma}

\begin{proof}
	\begin{align}
			& \quad \frac{1}{K}\mathbb{E}\left[\left\|P_{t} - \bar{P}_{t}\right\|_F^2\right] \notag \\
			& \leq  \lambda\frac{1}{K}\mathbb{E}\left[\left\| P_{t-1} -  \bar{P}_{t-1} \right\|_F^2\right]  + \frac{1}{1-\lambda}\frac{1}{K}\mathbb{E}\Big[\Big\|\delta^{\hat{\mathcal{G}}_{F}}(X_{t}, Y_{t}, Z_{t}; \hat{\xi}_t) - \delta^{\hat{\mathcal{G}}_{F}}(X_{t-1}, Y_{t-1}, Z_{t-1}; \hat{\xi}_t)   \notag \\
			& \quad  -\alpha_1\eta^2 (U_{t-1}- \delta^{\hat{\mathcal{G}}_{F}}(X_{t-1}, Y_{t-1}, Z_{t-1}))  \notag \\
			& \quad   -\alpha_1\eta^2(\delta^{\hat{\mathcal{G}}_{F}}(X_{t-1}, Y_{t-1}, Z_{t-1})  - \delta^{\hat{\mathcal{G}}_{F}}(X_{t-1}, Y_{t-1}, Z_{t-1}; \hat{\xi}_t))     \Big\|_F^2\Big]\notag \\
			& \leq  \lambda\frac{1}{K}\mathbb{E}\left[\left\| P_{t-1} -  \bar{P}_{t-1} \right\|_F^2\right]  + \frac{3}{1-\lambda}\frac{1}{K}\mathbb{E}\left[\left\|\delta^{\hat{\mathcal{G}}_{F}}(X_{t}, Y_{t}, Z_{t}; \hat{\xi}_t) - \delta^{\hat{\mathcal{G}}_{F}}(X_{t-1}, Y_{t-1}, Z_{t-1}; \hat{\xi}_t)    \right\|_F^2\right]\notag \\
			& \quad +  \frac{3\alpha_1^2\eta^4}{1-\lambda}\frac{1}{K}\mathbb{E}\left[\left\| U_{t-1}- \delta^{\hat{\mathcal{G}}_{F}}(X_{t-1}, Y_{t-1}, Z_{t-1})   \right\|_F^2\right]\notag \\
			& \quad  +  \frac{3\alpha_1^2\eta^4}{1-\lambda}\frac{1}{K}\mathbb{E}\left[\left\|\delta^{\hat{\mathcal{G}}_{F}}(X_{t-1}, Y_{t-1}, Z_{t-1})  - \delta^{\hat{\mathcal{G}}_{F}}(X_{t-1}, Y_{t-1}, Z_{t-1}; \hat{\xi}_t)   \right\|_F^2\right]\notag \\
			& \leq  \lambda\frac{1}{K}\mathbb{E}\left[\left\| P_{t-1} -  \bar{P}_{t-1} \right\|_F^2\right] +  \frac{3\alpha_1^2\eta^4}{1-\lambda}\frac{1}{K}\mathbb{E}\left[\left\| U_{t-1}- \delta^{\hat{\mathcal{G}}_{F}}(X_{t-1}, Y_{t-1}, Z_{t-1})   \right\|_F^2\right]\notag \\
			& \quad  + (\ell_{f_x}^2+\frac{c_{f_y}^2\ell_{g_{xy}}^2}{\mu^2})\frac{9}{1-\lambda}\frac{1}{K} \mathbb{E}\left[\left\|X_{t} - X_{t-1}\right\|_F^2\right]+ (\ell_{f_x}^2+\frac{c_{f_y}^2\ell_{g_{xy}}^2}{\mu^2})\frac{9}{1-\lambda}\frac{1}{K} \mathbb{E}\left[\left\|Y_{t} - Y_{t-1}\right\|_F^2\right]\notag \\
			& \quad + c_{g_{xy}}^2\frac{9}{1-\lambda}\frac{1}{K} \mathbb{E}\left[\left\|Z_{t} - Z_{t-1}\right\|_F^2\right] +  \frac{6\alpha_1^2\eta^4}{1-\lambda}(1 + \frac{c_{f_y}^2}{\mu^2})\sigma^2 \  ,   
	\end{align}
	where the first step can be obtained by following the proof of Lemma~\ref{lemma_consensus_r}, the last step can be obtained by following Eqs.~(\ref{eq:F-var}-\ref{eq:stochastic-grad-F-incremental}).

\end{proof}

\begin{lemma} \label{lemma_consensus_q}
	Under Assumptions~\ref{assumption_bi_strong}-\ref{assumption_graph}, we have 
	\begin{align}
			& \quad \frac{1}{K}\mathbb{E}\left[\left\|Q_{t} - \bar{Q}_{t}\right\|_F^2\right] \notag \\
			&  \leq  \lambda\frac{1}{K}\mathbb{E}\left[\left\| Q_{t-1} -  \bar{Q}_{t-1} \right\|_F^2\right] +  \frac{3\alpha_2^2\eta^4}{1-\lambda}\frac{1}{K}\mathbb{E}\left[\left\| V_{t-1}- \delta^{g}(X_{t-1}, Y_{t-1})   \right\|_F^2\right]\notag \\
			& \quad  +\frac{9\ell_{g_y}^2}{1-\lambda}\frac{1}{K} \mathbb{E}\left[\left\|X_{t} - X_{t-1}\right\|_F^2\right]+ \frac{9\ell_{g_y}^2}{1-\lambda}\frac{1}{K} \mathbb{E}\left[\left\|Y_{t} - Y_{t-1}\right\|_F^2\right] +  \frac{6\alpha_2^2\eta^4}{1-\lambda}\sigma^2  \ . 
	\end{align}
\end{lemma}

\begin{proof}
	\begin{align}
			& \quad \frac{1}{K}\mathbb{E}\left[\left\|Q_{t} - \bar{Q}_{t}\right\|_F^2\right] \notag \\
			& \leq  \lambda\frac{1}{K}\mathbb{E}\left[\left\| Q_{t-1} -  \bar{Q}_{t-1} \right\|_F^2\right]   + \frac{1}{1-\lambda}\frac{1}{K}\mathbb{E}\Big[\Big\|\delta^{g}(X_{t}, Y_{t}; \zeta_t) - \delta^{g}(X_{t-1}, Y_{t-1}; \zeta_t)  \notag \\
			& \quad -\alpha_2\eta^2 (V_{t-1}- \delta^{g}(X_{t-1}, Y_{t-1}))   -\alpha_2\eta^2(\delta^{g}(X_{t-1}, Y_{t-1})  - \delta^{g}(X_{t-1}, Y_{t-1}; \zeta_t))     \Big\|_F^2\Big]\notag \\
			& \leq  \lambda\frac{1}{K}\mathbb{E}\left[\left\| Q_{t-1} -  \bar{Q}_{t-1} \right\|_F^2\right]+ \frac{3}{1-\lambda}\frac{1}{K}\mathbb{E}\left[\left\|\delta^{g}(X_{t}, Y_{t}; \zeta_t) - \delta^{g}(X_{t-1}, Y_{t-1}; \zeta_t)    \right\|_F^2\right]\notag \\
			& \quad +  \frac{3\alpha_2^2\eta^4}{1-\lambda}\frac{1}{K}\mathbb{E}\left[\left\| V_{t-1}- \delta^{g}(X_{t-1}, Y_{t-1})   \right\|_F^2\right] \notag \\
			& \quad +  \frac{3\alpha_2^2\eta^4}{1-\lambda}\frac{1}{K}\mathbb{E}\left[\left\|\delta^{g}(X_{t-1}, Y_{t-1})  - \delta^{g}(X_{t-1}, Y_{t-1}; \zeta_t)   \right\|_F^2\right]\notag \\
			& \leq  \lambda\frac{1}{K}\mathbb{E}\left[\left\| Q_{t-1} -  \bar{Q}_{t-1} \right\|_F^2\right] +  \frac{3\alpha_2^2\eta^4}{1-\lambda}\frac{1}{K}\mathbb{E}\left[\left\| V_{t-1}- \delta^{g}(X_{t-1}, Y_{t-1})   \right\|_F^2\right]\notag \\
			& \quad  +\frac{9\ell_{g_y}^2}{1-\lambda}\frac{1}{K} \mathbb{E}\left[\left\|X_{t} - X_{t-1}\right\|_F^2\right]+ \frac{9\ell_{g_y}^2}{1-\lambda}\frac{1}{K} \mathbb{E}\left[\left\|Y_{t} - Y_{t-1}\right\|_F^2\right] +  \frac{6\alpha_2^2\eta^4}{1-\lambda}\sigma^2  \  ,  
	\end{align}
		where the first step can be obtained by following the proof of Lemma~\ref{lemma_consensus_r}, and the last step follows from Assumption~\ref{assumption_lower_smooth_vr} and Assumption~\ref{assumption_variance}. 
\end{proof}


\begin{lemma} \label{lemma_consensus_x}
				Under Assumptions~\ref{assumption_bi_strong}-\ref{assumption_graph}, we have 
	\begin{align}
			& \quad  \frac{1}{K}\mathbb{E}\left[\left\|X_{t+1} - \bar{X}_{t+1}\right\|_F^2\right] \notag \\
			& \leq  \Big(1-\frac{\eta(1-\lambda^2)}{2}\Big)\frac{1}{K}\mathbb{E}\left[\left\|X_t -\bar{X}_t\right\|_F^2\right]+ \frac{2\eta \beta_1^2}{1-\lambda^2}\frac{1}{K}\mathbb{E}\left[\left\|P_t- \bar{ P}_t\right\|_F^2\right]  \ ,\notag \\ 
			&\quad  \frac{1}{K}\mathbb{E}\left[\left\|Y_{t+1} - \bar{Y}_{t+1}\right\|_F^2\right] \notag \\
			& \leq  \Big(1-\frac{\eta(1-\lambda^2)}{2}\Big)\frac{1}{K}\mathbb{E}\left[\left\|Y_t -\bar{Y}_t\right\|_F^2\right]+ \frac{2\eta \beta_2^2}{1-\lambda^2}\frac{1}{K}\mathbb{E}\left[\left\|Q_t- \bar{ Q}_t\right\|_F^2\right]  \ ,\notag \\ 
			& \quad \frac{1}{K}\mathbb{E}\left[\left\|Z_{t+1} - \bar{Z}_{t+1}\right\|_F^2\right] \notag \\
			& \leq  \Big(1-\frac{\eta(1-\lambda^2)}{2}\Big)\frac{1}{K}\mathbb{E}\left[\left\|Z_t -\bar{Z}_t\right\|_F^2\right]+ \frac{2\eta \beta_3^2}{1-\lambda^2}\frac{1}{K}\mathbb{E}\left[\left\|R_t- \bar{R}_t\right\|_F^2\right]  \ .  
	\end{align}
\end{lemma}

\begin{proof}

\begin{align} 
		&   \quad \frac{1}{K}\mathbb{E}\left[\left\|Z_{t+1} - \bar{Z}_{t+1}\right\|_F^2\right] = \frac{1}{K}\sum_{k=1}^{K} \mathbb{E}\left[\left\| z_{ t+1}^{(k)}  -  \bar{z}_{t+1}\right\|^2\right] \notag \\
		& =\frac{1}{K}\sum_{k=1}^{K} \mathbb{E}\Bigg[\Bigg\| (1-\eta) z_{ t}^{(k)}  + \eta \mathcal{P}\Bigg[\sum_{j\in\mathcal{N}_{k}}w_{kj}z_{t}^{(j)} -\beta_3  r^{(k)}_t\Bigg] \notag \\
		& \quad -  (1-\eta) \frac{1}{K} \sum_{k'=1}^{K} z_{ t}^{(k')}  - \frac{1}{K} \sum_{k'=1}^{K} \eta \mathcal{P}\Bigg[\sum_{j\in\mathcal{N}_{k'}}w_{k'j}z_{t}^{(j)} -\beta_3  r^{(k')}_t\Bigg]\Bigg\|^2 \Bigg]\notag \\
		& \leq (1+a)(1-\eta)^2\frac{1}{K}\sum_{k=1}^{K} \mathbb{E}\left[\left\|  z_{ t}^{(k)}  -   \frac{1}{K} \sum_{k'=1}^{K} z_{ t}^{(k')} \right\|^2\right] \notag \\
		& \quad + (1+1/a) \eta^2\frac{1}{K}\sum_{k=1}^{K} \mathbb{E}\left[\left\|  \mathcal{P}\left[\sum_{j\in\mathcal{N}_{k}}w_{kj}z_{t}^{(j)} -\beta_3  r^{(k)}_t\right] - \frac{1}{K} \sum_{k'=1}^{K}  \mathcal{P}\left[\sum_{j\in\mathcal{N}_{k'}}w_{k'j}z_{t}^{(j)} -\beta_3  r^{(k')}_t\right]\right\|^2 \right]\notag \\
		& \leq (1+a)(1-\eta)^2 \frac{1}{K}\mathbb{E}\left[\left\|  Z_{ t} - \bar{Z}_{t}\right\|_F^2\right] \notag \\
		& \quad  + (1+1/a) \eta^2\frac{1}{K} \sum_{k=1}^{K} \mathbb{E}\left[\left\|  \mathcal{P}\left[\sum_{j\in\mathcal{N}_{k}}w_{kj}z_{t}^{(j)} -\beta_3  r^{(k)}_t\right] -  \mathcal{P}\left[\frac{1}{K} \sum_{k'=1}^{K} \sum_{j\in\mathcal{N}_{k'}}w_{k'j}z_{t}^{(j)} -\beta_3  r^{(k')}_t\right]\right\|^2\right] \notag \\
		& \leq (1+a)(1-\eta)^2 \frac{1}{K}\mathbb{E}\left[\left\|  Z_{ t} - \bar{Z}_{t}\right\|_F^2\right]  \notag \\
		& \quad + (1+1/a) \eta^2\frac{1}{K}\sum_{k=1}^{K} \mathbb{E}\left[\left\|  \sum_{j\in\mathcal{N}_{k}}w_{kj}z_{t}^{(j)} -\beta_3  r^{(k)}_t -\frac{1}{K} \sum_{k'=1}^{K} \sum_{j\in\mathcal{N}_{k'}}w_{k'j}z_{t}^{(j)} -\beta_3  r^{(k')}_t\right\|^2 \right]\notag \\
		& \leq (1-\eta) \frac{1}{K}\mathbb{E}\left[ \left\|  Z_{ t} - \bar{Z}_{t}\right\|_F^2\right]  +  \eta \frac{1}{K}\sum_{k=1}^{K} \mathbb{E}\left[\left\|  \sum_{j\in\mathcal{N}_{k}}w_{kj}z_{t}^{(j)}  - \bar{z}_{t}-\beta_3  r^{(k)}_t  + \beta_3  \bar{r}_t\right\|^2\right] \notag \\
		& \leq (1-\eta) \frac{1}{K} \mathbb{E}\left[\left\|  Z_{ t} - \bar{Z}_{t}\right\|_F^2\right]  +  \eta \frac{1}{K}\mathbb{E}\left[\left\|  Z_tW  - \bar{Z}_{t}-\beta_3  R_t  + \beta_3  \bar{R}_t\right\|_F^2\right] \notag \\
		& \leq (1-\eta) \frac{1}{K}\mathbb{E}\left[\left\|  Z_{ t} - \bar{Z}_{t}\right\|_F^2\right]  +  \eta (1+b)\frac{1}{K}\mathbb{E}\left[\left\|  Z_tW  - \bar{Z}_{t} \right\|_F^2\right] +\eta \beta_3^2(1+1/b)\frac{1}{K}\mathbb{E}\left[\left\| R_t  -   \bar{R}_t\right\|_F^2\right] \notag \\
		& \leq  (1-\eta) \frac{1}{K}\mathbb{E}\left[\left\|  Z_{ t} - \bar{Z}_{t}\right\|_F^2\right]   +  \frac{\eta(1+\lambda^2)}{2\lambda^2}\frac{1}{K}\mathbb{E}\left[\left\|  Z_tW  - \bar{Z}_{t} \right\|_F^2\right] +\frac{2\eta \beta_3^2}{1-\lambda^2}\frac{1}{K}\mathbb{E}\left[\left\| R_t  -   \bar{R}_t\right\|_F^2 \right]\notag \\
		& \leq  \Big(1-\frac{\eta(1-\lambda^2)}{2}\Big) \frac{1}{K}\mathbb{E}\left[\left\|  Z_{ t} - \bar{Z}_{t}\right\|_F^2\right]    +\frac{2\eta \beta_3^2}{1-\lambda^2}\frac{1}{K}\mathbb{E}\left[\left\| R_t  -   \bar{R}_t\right\|_F^2\right]  \ , 
\end{align}
where the second inequality follows form Eq.~(16) of \cite{spiridonoff2021communication},  the fourth inequality follows from $a=\frac{\eta}{1-\eta}$,  the second to  last step follows from $b=\frac{1-\lambda^2}{2\lambda^2}$, and the last step follows from $\left\|  Z_tW  - \bar{Z}_{t} \right\|_F^2\leq \lambda^2\left\|  Z_t  - \bar{Z}_{t} \right\|_F^2$.  

The other two inequalities can be proved similarly without the projection operation.

\end{proof}

\begin{lemma} \label{lemma_incremental_x}
					Under Assumptions~\ref{assumption_bi_strong}-\ref{assumption_graph}, we have 
	\begin{align}
			& \quad \mathbb{E}\left[\left\|X_{t+1}-X_{t}\right\|_F^2\right]\leq 8\eta^2\mathbb{E}\left[\left\|X_{t}-\bar{X}_t\right\|_F^2\right] + 4\eta^2\beta_1^2\mathbb{E}\left[\left\|P_t-\bar{P}_t\right\|_F^2\right]+4\eta^2\beta_1^2\mathbb{E}\left[\left\|\bar{P}_t\right\|_F^2\right]  \ , \notag \\
			& \quad \mathbb{E}\left[\left\|Y_{t+1}-Y_{t}\right\|_F^2\right]\leq 8\eta^2\mathbb{E}\left[\left\|Y_{t}-\bar{Y}_t\right\|_F^2\right] + 4\eta^2\beta_2^2\mathbb{E}\left[\left\|Q_t-\bar{Q}_t\right\|_F^2\right]+4\eta^2\beta_2^2\mathbb{E}\left[\left\|\bar{Q}_t\right\|_F^2\right]  \ ,\notag \\
			& \quad \mathbb{E}\left[\left\|Z_{t+1}-Z_{t}\right\|_F^2\right]\leq 8\eta^2\mathbb{E}\left[\left\|Z_{t}-\bar{Z}_t\right\|_F^2\right] + 4\eta^2\beta_3^2\mathbb{E}\left[\left\|R_t-\bar{R}_t\right\|_F^2\right]+4\eta^2\beta_3^2\mathbb{E}\left[\left\|\bar{R}_t\right\|_F^2\right]  \ . 
	\end{align}
\end{lemma}

\begin{proof}
	\begin{align}
			& \quad \mathbb{E}\left[\left\|Z_{t+1}-Z_{t}\right\|_F^2\right] = \sum_{k=1}^{K}\mathbb{E}\left[\left\|z^{(k)}_{t+1}-z^{(k)}_{t}\right\|^2\right] \notag \\
			&  = \sum_{k=1}^{K}\mathbb{E}\left[\left\|(1-\eta) z_{ t}^{(k)}  + \eta \mathcal{P}\left[\sum_{j\in\mathcal{N}_{k}}w_{kj}z_{t}^{(j)} -\beta_3  r^{(k)}_t\right]-z^{(k)}_{t}\right\|^2\right] \notag \\
			&  = \eta^2\sum_{k=1}^{K}\mathbb{E}\left[\left\|- z_{ t}^{(k)}  +  \mathcal{P}\left[\sum_{j\in\mathcal{N}_{k}}w_{kj}z_{t}^{(j)} -\beta_3  r^{(k)}_t\right]\right\|^2\right] \notag \\
			&  \leq  \eta^2\sum_{k=1}^{K}\mathbb{E}\left[\left\|- z_{ t}^{(k)}  +  \sum_{j\in\mathcal{N}_{k}}w_{kj}z_{t}^{(j)} -\beta_3  r^{(k)}_t\right\|^2\right] \notag \\
			& \leq  2\eta^2\mathbb{E}\left[\left\|Z_{t}W - Z_{t}\right\|_F^2\right] + 2\eta^2\beta_3^2 \mathbb{E}\left[\left\| R_t\right\|_F^2\right] \notag \\
			& \leq 2\eta^2\mathbb{E}\left[\left\|(Z_{t}-\bar{Z}_t)(W-I)\right\|_F^2\right] + 2\eta^2\beta_3^2 \mathbb{E}\left[\left\| R_t\right\|_F^2\right] \notag \\
			& \leq  8\eta^2\mathbb{E}\left[\left\|Z_{t}-\bar{Z}_t\right\|_F^2\right] + 4\eta^2\beta_3^2\mathbb{E}\left[\left\|R_t-\bar{R}_t\right\|_F^2\right]+4\eta^2\beta_3^2\mathbb{E}\left[\left\|\bar{R}_t\right\|_F^2\right]   \ ,  
	\end{align}
	where the last step follows from $\left\|(Z_{t}-\bar{Z}_t)(W-I)\right\|_F^2\leq \left\|W-2\right\|^2_2\left\|Z_{t}-\bar{Z}_t\right\|_F^2\leq 4\left\|Z_{t}-\bar{Z}_t\right\|_F^2$.

	The other two inequalities can be proved similarly without the projection operation. 
	
\end{proof}

\begin{lemma}
		Under Assumptions~\ref{assumption_bi_strong}-\ref{assumption_graph}, we have 
		\begin{align}
				& \quad \mathbb{E}\left[\left\|\bar{r}_{t}\right\|^2\right] = \mathbb{E}\left[\left\|\bar{w}_{t}\right\|^2\right] \notag \\
				& \leq 3 \mathbb{E}\left[\left\|\frac{1}{K}{W}_{t}\mathbf{1} - \frac{1}{K}\delta^{\hat{\mathcal{G}}_{h}}({X}_{t}, {Y}_{t},  {Z}_t)\mathbf{1}\right\|^2\right] + 9 \ell_{g_y}^2 \mathbb{E}\left[\left\|\bar{z}_t - z^*(\bar{x}_t) \right\|^2\right]  \notag \\
				& \quad + 9 \Big(\frac{\ell_{g_{yy}}^2c_{f_y}^2}{\mu^2}+ \ell_{f_y}^2\Big)\mathbb{E}\left[\left\|\bar{y}_t  - y^*(\bar{x}_t)\right\|^2\right]+ 9\ell_{g_y}^2\frac{1}{K}\mathbb{E}\left[\left\|{Z}_t- \bar{Z}_{t}\right\|_F^2\right] \notag \\
				& \quad    + 9\Big(\frac{c_{f_y}^2\ell_{g_{yy}}^2}{\mu^2} + \ell_{f_y}^2\Big)\frac{1}{K}\mathbb{E}\left[\left\|\bar{X}_{t} - {X}_{t}\right\|_F^2\right] +  9\Big(\frac{c_{f_y}^2\ell_{g_{yy}}^2}{\mu^2}+ \ell_{f_y}^2\Big)\frac{1}{K}\mathbb{E}\left[\left\| \bar{Y}_t - {Y}_{t}\right\|_F^2\right]  \ . 
		\end{align}
\end{lemma}

\begin{proof}
	\begin{align}
			& \quad \mathbb{E}\left[\left\|\bar{r}_{t}\right\|^2\right] = \mathbb{E}\left[\left\|\bar{w}_{t}\right\|^2\right] \notag \\
			& = \mathbb{E}\Bigg[\Bigg\|\frac{1}{K}{W}_{t}\mathbf{1} - \frac{1}{K}\delta^{\hat{\mathcal{G}}_{h}}({X}_{t}, {Y}_{t},  {Z}_t)\mathbf{1}+  \frac{1}{K}\delta^{\hat{\mathcal{G}}_{h}}({X}_{t}, {Y}_{t},  {Z}_t)\mathbf{1}  \notag \\
			& \quad -  \frac{1}{K}\delta^{\hat{\mathcal{G}}_{h}}(\bar{X}_{t}, \bar{Y}_{t},  \bar{Z}_t)\mathbf{1} + \frac{1}{K}\delta^{\hat{\mathcal{G}}_{h}}(\bar{X}_{t}, \bar{Y}_{t},  \bar{Z}_t)\mathbf{1}\Bigg\|^2\Bigg] \notag \\
			& \leq 3 \mathbb{E}\left[\left\|\frac{1}{K}{W}_{t}\mathbf{1} - \frac{1}{K}\delta^{\hat{\mathcal{G}}_{h}}({X}_{t}, {Y}_{t},  {Z}_t)\mathbf{1}\right\|^2\right] \notag \\
			& \quad + 3 \mathbb{E}\left[\left\| \frac{1}{K}\delta^{\hat{\mathcal{G}}_{h}}({X}_{t}, {Y}_{t},  {Z}_t)\mathbf{1} -  \frac{1}{K}\delta^{\hat{\mathcal{G}}_{h}}(\bar{X}_{t}, \bar{Y}_{t},  \bar{Z}_t)\mathbf{1}\right\|^2\right] \notag \\
			& \quad + 3 \mathbb{E}\left[\left\|\frac{1}{K}\delta^{\hat{\mathcal{G}}_{h}}(\bar{X}_{t}, \bar{Y}_{t},  \bar{Z}_t)\mathbf{1}\right\|^2\right] \notag \\
			& \leq 3 \mathbb{E}\left[\left\|\frac{1}{K}{W}_{t}\mathbf{1} - \frac{1}{K}\delta^{\hat{\mathcal{G}}_{h}}({X}_{t}, {Y}_{t},  {Z}_t)\mathbf{1}\right\|^2\right]  + 9 \ell_{g_y}^2 \mathbb{E}\left[\left\|\bar{z}_t - z^*(\bar{x}_t) \right\|^2\right]   \notag \\
			& \quad + 9 \Big(\frac{\ell_{g_{yy}}^2c_{f_y}^2}{\mu^2}+ \ell_{f_y}^2\Big)\mathbb{E}\left[\left\|\bar{y}_t  - y^*(\bar{x}_t)\right\|^2\right]  + 9\ell_{g_y}^2\frac{1}{K}\mathbb{E}\left[\left\|{Z}_t- \bar{Z}_{t}\right\|_F^2\right]  \notag \\
			& \quad + 9\Big(\frac{c_{f_y}^2\ell_{g_{yy}}^2}{\mu^2} + \ell_{f_y}^2\Big)\frac{1}{K}\mathbb{E}\left[\left\|\bar{X}_{t} - {X}_{t}\right\|_F^2\right] +  9\Big(\frac{c_{f_y}^2\ell_{g_{yy}}^2}{\mu^2}+ \ell_{f_y}^2\Big)\frac{1}{K}\mathbb{E}\left[\left\| \bar{Y}_t - {Y}_{t}\right\|_F^2\right] \ , 
	\end{align}
	where the last step follows from Eq.~(\ref{eq:h-grad-consensus}) and the following inequality:
	\begin{align}
			& \quad \mathbb{E}\left[\left\|\frac{1}{K}\delta^{\hat{\mathcal{G}}_{h}}(\bar{X}_{t}, \bar{Y}_{t},  \bar{Z}_t)\mathbf{1}\right\|^2\right] = \mathbb{E}\left[\left\|\hat{\mathcal{G}}_{h}(\bar{x}_t, \bar{y}_t, \bar{z}_t)\right\|^2\right] \notag \\
			& = \mathbb{E}\left[\left\|\nabla_{22}^2g(\bar{x}_t, \bar{y}_t) \bar{z}_t -  \nabla_2 f(\bar{x}_t, \bar{y}_t)  \right\|^2\right] \notag \\
			& = \mathbb{E}\left[\Big\|\nabla_{22}^2g(\bar{x}_t, \bar{y}_t) \bar{z}_t -  \nabla_2 f(\bar{x}_t, \bar{y}_t) -\nabla_{22}^2g(\bar{x}_t, y^*(\bar{x}_t)) z^*(\bar{x}_t) +   \nabla_2 f(\bar{x}_t, y^*(\bar{x}_t))  \Big\|^2\right] \notag \\
			& \leq 3\mathbb{E}\left[\left\|\nabla_{22}^2g(\bar{x}_t, \bar{y}_t) \bar{z}_t - \nabla_{22}^2g(\bar{x}_t, \bar{y}_t) z^*(\bar{x}_t) \right\|^2\right] \notag \\
			& \quad + 3\mathbb{E}\left[\left\|\nabla_{22}^2g(\bar{x}_t, \bar{y}_t) z^*(\bar{x}_t) -\nabla_{22}^2g(\bar{x}_t, y^*(\bar{x}_t)) z^*(\bar{x}_t)\right\|^2\right] \notag \\
			& \quad + 3\mathbb{E}\left[\left\|-  \nabla_2 f(\bar{x}_t, \bar{y}_t)  +  \nabla_2 f(\bar{x}_t, y^*(\bar{x}_t))  \right\|^2\right] \notag \\
			& \leq 3 \ell_{g_y}^2 \mathbb{E}\left[\left\|\bar{z}_t - z^*(\bar{x}_t) \right\|^2\right] + 3 \Big(\frac{\ell_{g_{yy}}^2c_{f_y}^2}{\mu^2}+ \ell_{f_y}^2\Big)\mathbb{E}\left[\left\|\bar{y}_t  - y^*(\bar{x}_t)\right\|^2\right]   \ .  
	\end{align}
\end{proof}


Given aforementioned lemmas, we are ready to prove Theorem~\ref{theorem-sim}. 
\begin{proof}
For the potential function in Eq.~(\ref{eq:potential-function}), based on Lemmas~\ref{lemma_F_iter},~\ref{lemma_y_opt},~\ref{lemma_z_opt},~\ref{lemma_consensus_x},~\ref{lemma_consensus_p},~\ref{lemma_consensus_q},~\ref{lemma_consensus_r},~\ref{lemma_hyper_storm_var_mean},~\ref{lemma_g_storm_var_mean},~\ref{lemma_h_storm_var_mean}, we can obtain
\begin{align}
		&   \mathcal{L}_{t+1} -  \mathcal{L}_{t}  \leq - \frac{\beta_1\eta}{2}\mathbb{E} \left[ \left\| \nabla F(\bar{x}_{t})\right\|^2 \right]\notag \\
		&  + \Bigg(-\frac{\beta_2\eta\mu}{4}c_0 +  12\beta_1\eta \Big(\ell_{f_x}^2 +\frac{c_{f_y}^2\ell_{g_{xy}}^2}{\mu^2} \Big) + \frac{9\eta\beta_3}{\mu}\Big(\frac{c_{f_y}^2\ell_{g_{yy}}^2}{\mu^2}+\ell_{f_y}^2\Big)c_1 \notag \\
		& \qquad  + 9 \Big(\frac{\ell_{g_{yy}}^2c_{f_y}^2}{\mu^2}+ \ell_{f_y}^2\Big)C_4\Bigg) \mathbb{E}\left[\left\|\bar{   {y}}_{t}   -    {y}^{*}(\bar{   {x}}_t)\right\|^2\right]  \notag \\
		&  + \Big(12\beta_1\eta c_{g_{xy}}^2 -\frac{\eta\beta_3\mu}{8}c_1 + 9 \ell_{g_y}^2 C_4\Big) \mathbb{E}\left[\left\|z^*(\bar{x}_t) -  \bar{z}_t \right\|^2\right]\notag \\
  &  + \Big(4\eta^2\beta_2^2 KC_2- \frac{3\eta\beta_2^2}{4}c_0 \Big) \mathbb{E}\left[\left\|\bar{v}_{t}  \right\|^2\right]\notag \\
  & + \Big(\frac{25\eta\beta_1^2L_{y}^2 }{6\beta_2\mu}c_0 -  \frac{\beta_1\eta}{4} + \frac{9\eta\beta_1^2L_z^2}{\beta_3\mu}c_1 + 4\eta^2\beta_1^2 K C_1\Big)\mathbb{E}\left[\left\|\bar{u}_{t}\right\|^2 \right] \notag \\
  & +  \Bigg( 12\beta_1\eta(\ell_{f_x}^2+\frac{c_{f_y}^2\ell_{g_{xy}}^2}{\mu^2} )\frac{1}{K} + \frac{25\beta_2 \eta \ell_{g_y}^2}{3\mu}  \frac{1}{K}c_0 + \frac{70\eta\beta_3}{\mu}\Big(\frac{c_{f_y}^2\ell_{g_{yy}}^2}{\mu^2} + \ell_{f_y}^2\Big)\frac{1}{K}c_1 -\frac{\eta(1-\lambda^2)}{2}\frac{1}{K}c_2 \notag \\
  & \quad + 8\eta^2 C_1 + 9\Big(\frac{c_{f_y}^2\ell_{g_{yy}}^2}{\mu^2} + \ell_{f_y}^2\Big)\frac{1}{K}C_4\Bigg)\mathbb{E}\left[\left\| \bar{X}_t - {X}_t\right\|_F^2\right] \notag \\
  &  +  \Bigg(12\beta_1\eta(\ell_{f_x}^2+ \frac{c_{f_y}^2\ell_{g_{xy}}^2}{\mu^2})\frac{1}{K} + \frac{25\beta_2 \eta \ell_{g_y}^2}{3\mu}  \frac{1}{K}c_0 + \frac{70\eta\beta_3}{\mu}\Big(\frac{c_{f_y}^2\ell_{g_{yy}}^2}{\mu^2} + \ell_{f_y}^2\Big)\frac{1}{K}c_1 -\frac{\eta(1-\lambda^2)}{2}\frac{1}{K}c_3 \notag \\
  & \quad + 8\eta^2 C_2 +  9\Big(\frac{c_{f_y}^2\ell_{g_{yy}}^2}{\mu^2}+ \ell_{f_y}^2\Big)\frac{1}{K}C_4\Bigg)\mathbb{E}\left[\left\|\bar{Y}_t -  {Y}_t\right\|_F^2\right] \notag \\
  & + \Bigg( 12\beta_1\eta c_{g_{xy}}^2\frac{1}{K} +\Big(\frac{70\eta\beta_3}{\mu}\ell_{g_y}^2+  \frac{9}{4}\Big)\frac{1}{K}c_1 -\frac{\eta(1-\lambda^2)}{2}\frac{1}{K}c_4 + 8\eta^2 C_3 + 9\ell_{g_y}^2\frac{1}{K}C_4\Bigg)\mathbb{E}\left[\left\|{Z}_t- \bar{Z}_{t}\right\|_F^2\right]\notag \\
  & + \Big(\frac{2\eta \beta_1^2}{1-\lambda^2}\frac{1}{K}c_2 + (\lambda-1)\frac{1}{K}c_5 + 4\eta^2\beta_1^2 C_1\Big)\mathbb{E}\left[\left\|P_t- \bar{ P}_t\right\|_F^2\right] \notag \\
  & + \Big(\frac{2\eta \beta_2^2}{1-\lambda^2}\frac{1}{K}c_3 + (\lambda-1)\frac{1}{K}c_6 + 4\eta^2\beta_2^2 C_2\Big)\mathbb{E}\left[\left\|Q_t- \bar{ Q}_t\right\|_F^2\right] \notag \\ 
  & + \Big(\frac{2\eta \beta_3^2}{1-\lambda^2}\frac{1}{K}c_4 + (\lambda-1)\frac{1}{K}c_7 + \frac{9\eta\beta_3^2}{4}\frac{1}{K}c_1 + 4\eta^2\beta_3^2 C_3\Big)\mathbb{E}\left[\left\|R_t- \bar{R}_t\right\|_F^2\right] \notag \\ 
  & + \Big(2\beta_1\eta -\alpha_1 \eta^2 c_8\Big) \mathbb{E} \left[ \left\| \frac{1}{K}\delta^{\hat{\mathcal{G}}_{F}}(X_{t}, Y_{t}, Z_{t}) \mathbf{1} -  \frac{1}{K}U_{t} \mathbf{1}  \right\|^2 \right]\notag \\
  & +  \Big(\frac{25\beta_2 \eta}{3\mu}c_0 -\alpha_2 \eta^2 c_{10} \Big) \mathbb{E}\left[\left\|\frac{1}{K} \delta^{g}  ({X}_t, {Y}_t) \mathbf{1} - \frac{1}{K}{V}_t\mathbf{1} \right\|^2\right] \notag \\
  & +  \Big(\frac{25\eta\beta_3}{\mu}c_1  -\alpha_3 \eta^2 c_{12} + 3 C_4\Big) \mathbb{E}\left[\left\|\frac{1}{K}\delta^{\hat{\mathcal{G}}_{h}}({X}_{t}, {Y}_{t},  {Z}_t)\mathbf{1} - \frac{1}{K}{W}_{t}\mathbf{1}\right\|^2\right]   \notag \\
  & +  \Big(\frac{3\alpha_1^2\eta^4}{1-\lambda}\frac{1}{K}c_5 -\alpha_1 \eta^2 \frac{1}{K}c_9\Big)\mathbb{E}\left[\left\| U_{t}- \delta^{\hat{\mathcal{G}}_{F}}(X_{t}, Y_{t}, Z_{t})   \right\|_F^2\right]\notag \\
  & +  \Big(\frac{3\alpha_2^2\eta^4}{1-\lambda}\frac{1}{K}c_6 -\alpha_2 \eta^2 \frac{1}{K}c_{11}\Big)\mathbb{E}\left[\left\| V_{t}- \delta^{g}(X_{t}, Y_{t})   \right\|_F^2\right]\notag \\
  & +  \Big(\frac{3\alpha_3^2\eta^4}{1-\lambda}\frac{1}{K}c_7 -\alpha_3 \eta^2 \frac{1}{K}c_{13}\Big)\mathbb{E}\left[\left\| W_{t}- \delta^{\hat{\mathcal{G}}_{h}}(X_{t}, Y_{t}, Z_{t})   \right\|_F^2\right]\notag \\
  & +  \frac{6\alpha_1^2\eta^4}{1-\lambda}(1 + \frac{c_{f_y}^2}{\mu^2})\sigma^2 c_5 +  \frac{6\alpha_2^2\eta^4}{1-\lambda}\sigma^2 c_6 +  \frac{6\alpha_3^2\eta^4}{1-\lambda}(1 + \frac{c_{f_y}^2}{\mu^2})\sigma^2 c_7 + 4\alpha_1^2\eta^4 (1 + \frac{c_{f_y}^2}{\mu^2})\sigma^2\frac{1}{K} c_8 \notag \\
  & + 4\alpha_1^2\eta^4 (1 + \frac{c_{f_y}^2}{\mu^2})\sigma^2 c_9 + 2\alpha_2^2\eta^4 \sigma^2\frac{1}{K} c_{10} + 2\alpha_2^2\eta^4 \sigma^2 c_{11} \notag \\
  & + 4\alpha_3^2\eta^4(1 + \frac{c_{f_y}^2}{\mu^2})\sigma^2\frac{1}{K} c_{12} + 4\alpha_3^2\eta^4(1 + \frac{c_{f_y}^2}{\mu^2})\sigma^2 c_{13}  \ , 
\end{align}
where
\begin{align}
        & C_1 =C_2= (\ell_{f_x}^2+\frac{c_{f_y}^2\ell_{g_{xy}}^2}{\mu^2})\frac{9}{1-\lambda}\frac{1}{K}c_5 +\frac{9\ell_{g_y}^2}{1-\lambda}\frac{1}{K}c_6 + (\ell_{f_y}^2+\frac{c_{f_y}^2\ell_{g_{yy}}^2}{\mu^2})\frac{9}{1-\lambda}\frac{1}{K}c_7  \notag \\
  & \quad + 6(\ell_{f_x}^2+\frac{c_{f_y}^2\ell_{g_{xy}}^2}{\mu^2})\frac{1}{K^2}c_8 + 6(\ell_{f_x}^2+\frac{c_{f_y}^2\ell_{g_{xy}}^2}{\mu^2})\frac{1}{K}c_9 + 2\ell_{g_y}^2\frac{1}{K^2}c_{10} + 2\ell_{g_y}^2\frac{1}{K}c_{11} \notag \\
  &\quad  +6(\frac{c_{f_y}^2\ell_{g_{yy}}^2}{\mu^2}+\ell_{f_y}^2)\frac{1}{K^2}c_{12} +6(\frac{c_{f_y}^2\ell_{g_{yy}}^2}{\mu^2}+\ell_{f_y}^2)\frac{1}{K}c_{13}  \ , \notag \\
    & C_3 =  c_{g_{xy}}^2\frac{9}{1-\lambda}\frac{1}{K}c_5 + \ell_{g_y}^2\frac{9}{1-\lambda}\frac{1}{K}c_7 + 6c_{g_{xy}}^2\frac{1}{K^2}c_8 + 6c_{g_{xy}}^2\frac{1}{K}c_9 + 6\ell_{g_y}^2 \frac{1}{K^2}c_{12} + 6\ell_{g_y}^2 \frac{1}{K}c_{13} \ ,  \notag \\
        & C_4 = 4\eta^2\beta_3^2 K C_3 \ . 
\end{align}

By setting 
\begin{align}
        &  c_8 = \frac{2\beta_1}{\alpha_1 \eta} \ , \   c_{10} = \frac{25\beta_2}{3\alpha_2 \eta\mu}c_0\ ,  \ c_5 = c_6 = c_7 = \beta_1(1-\lambda) \ , \ c_9 =c_{11} = c_{13}  = 3\beta_1 \ , 
\end{align}
we can eliminate $\mathbb{E} [ \| \frac{1}{K}\delta^{\hat{\mathcal{G}}_{F}}(X_{t}, Y_{t}, Z_{t}) \mathbf{1} -  \frac{1}{K}U_{t} \mathbf{1}  \|^2 ]$, $\mathbb{E}[\|\frac{1}{K} \delta^{g}  ({X}_t, {Y}_t) \mathbf{1} - \frac{1}{K}{V}_t\mathbf{1} \|^2] $, $\mathbb{E}[\| U_{t}- \delta^{\hat{\mathcal{G}}_{F}}(X_{t}, Y_{t}, Z_{t})   \|_F^2]$, $\mathbb{E}[\| V_{t}- \delta^{g}(X_{t}, Y_{t})   \|_F^2]$, and $\mathbb{E}[\| W_{t}- \delta^{\hat{\mathcal{G}}_{h}}(X_{t}, Y_{t}, Z_{t})   \|_F^2]$. 

To eliminate $\mathbb{E} [ \|\frac{1}{K}\delta^{\hat{\mathcal{G}}_{h}}({X}_{t}, {Y}_{t},  {Z}_t)\mathbf{1} - \frac{1}{K}{W}_{t}\mathbf{1}\|^2]$, we enforce
\begin{align}
 & \frac{25\eta\beta_3}{\mu}c_1  -\alpha_3 \eta^2 c_{12} + 12\eta^2\beta_3^2 K ( c_{g_{xy}}^2\frac{27}{K}\beta_1 + \ell_{g_y}^2\frac{27}{K}\beta_1 + c_{g_{xy}}^2\frac{12}{K^2}\frac{\beta_1}{\alpha_1 \eta} +6\ell_{g_y}^2 \frac{1}{K^2}c_{12} )  \leq 0  \ . 
\end{align}
This can be done by setting
\begin{align}
        & 12\eta^2\beta_3^2 K6\ell_{g_y}^2 \frac{1}{K^2}c_{12} \leq \frac{1}{4}\alpha_3 \eta^2 c_{12}  \ ,  \notag \\
        &  \frac{25\eta\beta_3}{\mu}c_1  \leq \frac{1}{4} \alpha_3 \eta^2 c_{12}  \ ,  \notag \\
        &  12\eta^2\beta_3^2 K ( c_{g_{xy}}^2\frac{27}{K}\beta_1 + \ell_{g_y}^2\frac{27}{K}\beta_1 + c_{g_{xy}}^2\frac{12}{K^2}\frac{\beta_1}{\alpha_1 \eta} )  \leq \frac{1}{2}\alpha_3 \eta^2 c_{12} \ . 
\end{align}
Then, we can set
\begin{align}\label{eq:c-1-1}
    & c_{12} = \frac{100\beta_3}{\alpha_3 \eta\mu}c_1  \ ,     \notag \\
        & \beta_3  \leq \frac{\sqrt{K\alpha_3}}{24 \ell_{g_y}} \ ,  \notag \\
        &  c_1\geq   \frac{ 12\beta_3\beta_1\eta\mu}{50}( 27c_{g_{xy}}^2 + 27\ell_{g_y}^2 + \frac{12c_{g_{xy}}^2}{K\alpha_1}\frac{1}{ \eta} )\ . 
\end{align}

To eliminate $\mathbb{E} [\|z^*(\bar{x}_t) -  \bar{z}_t \|^2]$, we enforce
\begin{align}
    & 12\beta_1\eta c_{g_{xy}}^2 -\frac{\eta\beta_3\mu}{8}c_1 \notag \\
    & \quad + 9 \ell_{g_y}^2 4\eta^2\beta_3^2 K ( c_{g_{xy}}^2\frac{27}{K}\beta_1 + \ell_{g_y}^2\frac{27}{K}\beta_1 + c_{g_{xy}}^2\frac{12}{K^2}\frac{\beta_1}{\alpha_1 \eta}  + 6\ell_{g_y}^2 \frac{1}{K^2}\frac{100\beta_3}{\alpha_3 \eta\mu}c_1)  \leq 0\ . 
\end{align}
This can be done by setting
\begin{align}
        & 9 \ell_{g_y}^2 4\eta^2\beta_3^2 K 6\ell_{g_y}^2 \frac{1}{K^2}\frac{100\beta_3}{\alpha_3 \eta\mu}c_1 \leq \frac{\eta\beta_3\mu}{16}c_1  \ , \notag \\
        &      12\beta_1\eta c_{g_{xy}}^2  + 9 \ell_{g_y}^2 4\eta^2\beta_3^2 K ( c_{g_{xy}}^2\frac{27}{K}\beta_1 + \ell_{g_y}^2\frac{27}{K}\beta_1 + c_{g_{xy}}^2\frac{12}{K^2}\frac{\beta_1}{\alpha_1 \eta}  )  \leq \frac{\eta\beta_3\mu}{16}c_1\ . 
\end{align}
Then, we can set
\begin{align} \label{eq:c-1-2}
        &  \beta_3    \leq \frac{\sqrt{K\alpha_3} \mu}{600 \ell_{g_y}^2}  \ , \notag \\
         & c_1  \geq \frac{192\beta_1 c_{g_{xy}}^2}{\beta_3\mu}  + \frac{576 \ell_{g_y}^2 \eta\beta_1\beta_3}{\mu} (27c_{g_{xy}}^2 + 27\ell_{g_y}^2 + \frac{12c_{g_{xy}}^2}{K\alpha_1}\frac{1}{ \eta}  )   \ . 
\end{align}
Due to $\ell_{g_y}>\mu$, it is easy to know that $ \frac{ 12\beta_3\beta_1\eta\mu}{50} \leq  \frac{ 12\beta_3\beta_1\eta\ell_{g_y}}{50}\leq  \frac{ 12\beta_3\beta_1\eta\ell_{g_y}^2}{50\mu} \leq  \frac{576 \ell_{g_y}^2 \eta\beta_1\beta_3}{\mu}$. Furthermore, due to $\eta\leq 1$ and $\beta_3\leq 1$, we can directly set
\begin{align}
    c_1  = \frac{192\beta_1 c_{g_{xy}}^2}{\beta_3\mu}  + \frac{576 \ell_{g_y}^2 \beta_1}{\beta_3\mu} (27c_{g_{xy}}^2 + 27\ell_{g_y}^2 + \frac{12c_{g_{xy}}^2}{K\alpha_1} )   \ , 
\end{align}
which satisfies both Eq.~(\ref{eq:c-1-1}) and Eq.~(\ref{eq:c-1-2}).

To eliminate $\mathbb{E}[\|\bar{   {y}}_{t}   -    {y}^{*}(\bar{   {x}}_t)\| ^2]$, we enforce
\begin{align}
        &    12\beta_1\eta \Big(\ell_{f_x}^2 +\frac{c_{f_y}^2\ell_{g_{xy}}^2}{\mu^2} \Big) + \Big(\frac{9\eta\beta_3}{\mu}+ \frac{5400\ell_{g_y}^2\eta\beta_3^3}{\alpha_3 K \mu}\Big)\Big(\frac{c_{f_y}^2\ell_{g_{yy}}^2}{\mu^2}+\ell_{f_y}^2\Big)c_1 -\frac{\beta_2\eta\mu}{4}c_0 \notag \\
        & + 36\eta^2\beta_3^2  \Big(\frac{\ell_{g_{yy}}^2c_{f_y}^2}{\mu^2}+ \ell_{f_y}^2\Big) \Big( 27c_{g_{xy}}^2\beta_1 + 27\ell_{g_y}^2\beta_1 + \frac{12c_{g_{xy}}^2\beta_1}{K\alpha_1}\frac{1}{ \eta} \Big)  \leq   0  \ . 
\end{align}
Then, due to $\eta\leq 1$ and $\beta_3\leq 1$, we can set
\begin{align}
        &  c_0  =    \frac{48\beta_1}{\beta_2\mu}\Big(\ell_{f_x}^2 +\frac{c_{f_y}^2\ell_{g_{xy}}^2}{\mu^2} \Big)  +   \frac{144\beta_1}{\beta_2\mu}\Big(\frac{\ell_{g_{yy}}^2c_{f_y}^2}{\mu^2}+ \ell_{f_y}^2\Big) \Big( 27c_{g_{xy}}^2 + 27\ell_{g_y}^2 + \frac{12c_{g_{xy}}^2}{K\alpha_1}\Big)       \notag \\
        & + \frac{768\beta_1 c_{g_{xy}}^2}{\beta_2\mu^2}\Big(\frac{9}{\mu}+ \frac{5400\ell_{g_y}^2}{\alpha_3 K \mu}\Big)\Big(\frac{c_{f_y}^2\ell_{g_{yy}}^2}{\mu^2}+\ell_{f_y}^2\Big)  \notag \\
        & + \frac{2304 \ell_{g_y}^2 \beta_1}{\beta_2\mu^2}\Big(\frac{9}{\mu}+ \frac{5400\ell_{g_y}^2}{\alpha_3 K \mu}\Big)\Big(\frac{c_{f_y}^2\ell_{g_{yy}}^2}{\mu^2}+\ell_{f_y}^2\Big) \Big(27c_{g_{xy}}^2 + 27\ell_{g_y}^2 + \frac{12c_{g_{xy}}^2}{K\alpha_1}\Big )   \ . 
\end{align}

In terms of the following coefficients
\begin{align}
    &  c_0  =   \frac{\beta_1}{\beta_2} \Bigg[\frac{48}{\mu}\Big(\ell_{f_x}^2 +\frac{c_{f_y}^2\ell_{g_{xy}}^2}{\mu^2} \Big)  +   \frac{144}{\mu}\Big(\frac{\ell_{g_{yy}}^2c_{f_y}^2}{\mu^2}+ \ell_{f_y}^2\Big) \Big( 27c_{g_{xy}}^2 + 27\ell_{g_y}^2 + \frac{12c_{g_{xy}}^2}{K\alpha_1}\Big)       \notag \\
        & \quad+ \frac{768 c_{g_{xy}}^2}{\mu^2}\Big(\frac{9}{\mu}+ \frac{5400\ell_{g_y}^2}{\alpha_3 K \mu}\Big)\Big(\frac{c_{f_y}^2\ell_{g_{yy}}^2}{\mu^2}+\ell_{f_y}^2\Big)  \notag \\
        & \quad+ \frac{2304 \ell_{g_y}^2 }{\mu^2}\Big(\frac{9}{\mu}+ \frac{5400\ell_{g_y}^2}{\alpha_3 K \mu}\Big)\Big(\frac{c_{f_y}^2\ell_{g_{yy}}^2}{\mu^2}+\ell_{f_y}^2\Big) \Big(27c_{g_{xy}}^2 + 27\ell_{g_y}^2 + \frac{12c_{g_{xy}}^2}{K\alpha_1} \Big)\Bigg]  \notag \\
        & \triangleq \frac{\beta_1}{\beta_2} \Tilde{c}_0\ , \notag \\
        & c_1  =\frac{\beta_1}{\beta_3}\Bigg[ \frac{192 c_{g_{xy}}^2}{\mu}  + \frac{576 \ell_{g_y}^2}{\mu} \Big(27c_{g_{xy}}^2 + 27\ell_{g_y}^2 + \frac{12c_{g_{xy}}^2}{K\alpha_1} \Big)\Bigg] \triangleq \frac{\beta_1}{\beta_3} \Tilde{c}_1  \ ,\notag \\
       & c_5 = c_6 = c_7 = \beta_1(1-\lambda) \ , \notag \\
        &  c_8 = \frac{2\beta_1}{\alpha_1 \eta} \ , \notag \\
        &   c_9 =c_{11} = c_{13}  = 3\beta_1 \ , \notag \\
        & c_{10} = \frac{25\beta_2}{3\alpha_2 \eta\mu}c_0 = \frac{25\beta_1}{3\alpha_2 \eta\mu} \Tilde{c}_0\ , \notag \\
         & c_{12} = \frac{100\beta_3}{\alpha_3 \eta\mu}c_1 = \frac{100\beta_1}{\alpha_3 \eta\mu}\Tilde{c}_1   \ ,    
\end{align}
we can obtain 
\begin{align}
     & C_3 =  c_{g_{xy}}^2\frac{9}{1-\lambda}\frac{1}{K}c_5 + \ell_{g_y}^2\frac{9}{1-\lambda}\frac{1}{K}c_7 + 6c_{g_{xy}}^2\frac{1}{K^2}c_8 + 6c_{g_{xy}}^2\frac{1}{K}c_9 + 6\ell_{g_y}^2 \frac{1}{K^2}c_{12} + 6\ell_{g_y}^2 \frac{1}{K}c_{13}  \notag \\
    &  =  c_{g_{xy}}^2\frac{9}{1-\lambda}\frac{1}{K}\beta_1(1-\lambda) + \ell_{g_y}^2\frac{9}{1-\lambda}\frac{1}{K}\beta_1(1-\lambda) + 6c_{g_{xy}}^2\frac{1}{K^2}\frac{2\beta_1}{\alpha_1 \eta} \notag \\
    & \quad + 6c_{g_{xy}}^2\frac{1}{K}3\beta_1 + 6\ell_{g_y}^2 \frac{1}{K^2}c_{12} + 6\ell_{g_y}^2 \frac{1}{K}3\beta_1 \ , \notag \\
    & =  27c_{g_{xy}}^2\frac{1}{K}\beta_1 + 27\ell_{g_y}^2\frac{1}{K}\beta_1 + c_{g_{xy}}^2\frac{1}{K^2}\frac{12\beta_1}{\alpha_1 \eta}  + \ell_{g_y}^2 \frac{1}{K^2}\frac{600}{\alpha_3 \eta\mu}\beta_1\Tilde{c}_1   \notag \\
    & = \frac{\beta_1}{\eta} \frac{1}{K}\Big(27c_{g_{xy}}^2\eta + 27\ell_{g_y}^2\eta + \frac{12c_{g_{xy}}^2}{\alpha_1 K }  + \frac{600 \ell_{g_y}^2}{\alpha_3 K\mu}\Tilde{c}_1\Big)   \notag \\
    & \leq  \frac{\beta_1}{\eta} \frac{1}{K}\Big(27c_{g_{xy}}^2 + 27\ell_{g_y}^2 + \frac{12c_{g_{xy}}^2}{\alpha_1 K }  + \frac{600 \ell_{g_y}^2}{\alpha_3 K\mu}\Tilde{c}_1\Big)   \notag \\
    & \triangleq \frac{\beta_1}{\eta K}\Tilde{C}_3 \ ,  
\end{align}  
and
\begin{align}
    & C_4 = 4\eta^2\beta_3^2 K C_3 \leq 4\eta\beta_1\beta_3^2 \Tilde{C}_3 \ , 
\end{align}
and
\begin{align}
     & C_1 =C_2= (\ell_{f_x}^2+\frac{c_{f_y}^2\ell_{g_{xy}}^2}{\mu^2})\frac{9}{1-\lambda}\frac{1}{K}c_5 +\frac{9\ell_{g_y}^2}{1-\lambda}\frac{1}{K}c_6 + (\ell_{f_y}^2+\frac{c_{f_y}^2\ell_{g_{yy}}^2}{\mu^2})\frac{9}{1-\lambda}\frac{1}{K}c_7  \notag \\
  & \quad + 6(\ell_{f_x}^2+\frac{c_{f_y}^2\ell_{g_{xy}}^2}{\mu^2})\frac{1}{K^2}c_8 + 6(\ell_{f_x}^2+\frac{c_{f_y}^2\ell_{g_{xy}}^2}{\mu^2})\frac{1}{K}c_9 + 2\ell_{g_y}^2\frac{1}{K^2}c_{10} + 2\ell_{g_y}^2\frac{1}{K}c_{11} \notag \\
  & \quad +6(\frac{c_{f_y}^2\ell_{g_{yy}}^2}{\mu^2}+\ell_{f_y}^2)\frac{1}{K^2}c_{12} +6(\frac{c_{f_y}^2\ell_{g_{yy}}^2}{\mu^2}+\ell_{f_y}^2)\frac{1}{K}c_{13}   \notag \\
    &  = (\ell_{f_x}^2+\frac{c_{f_y}^2\ell_{g_{xy}}^2}{\mu^2})\frac{9}{1-\lambda}\frac{1}{K}\beta_1(1-\lambda) +\frac{9\ell_{g_y}^2}{1-\lambda}\frac{1}{K}\beta_1(1-\lambda) + (\ell_{f_y}^2+\frac{c_{f_y}^2\ell_{g_{yy}}^2}{\mu^2})\frac{9}{1-\lambda}\frac{1}{K}\beta_1(1-\lambda)  \notag \\
  & \quad + 6(\ell_{f_x}^2+\frac{c_{f_y}^2\ell_{g_{xy}}^2}{\mu^2})\frac{1}{K^2}\frac{2\beta_1}{\alpha_1 \eta} + 6(\ell_{f_x}^2+\frac{c_{f_y}^2\ell_{g_{xy}}^2}{\mu^2})\frac{1}{K}3\beta_1 + 2\ell_{g_y}^2\frac{1}{K^2}\frac{25\beta_1}{3\alpha_2 \eta\mu} \Tilde{c}_0 + 2\ell_{g_y}^2\frac{1}{K}3\beta_1 \notag \\
  & \quad +6(\frac{c_{f_y}^2\ell_{g_{yy}}^2}{\mu^2}+\ell_{f_y}^2)\frac{1}{K^2}\frac{100\beta_1}{\alpha_3 \eta\mu}\Tilde{c}_1   +6(\frac{c_{f_y}^2\ell_{g_{yy}}^2}{\mu^2}+\ell_{f_y}^2)\frac{1}{K}3\beta_1  \ , \notag \\
  &  = \frac{\beta_1}{\eta K}\bigg( \eta27(\ell_{f_x}^2+\frac{c_{f_y}^2\ell_{g_{xy}}^2}{\mu^2}) +\eta15\ell_{g_y}^2 + \eta27(\ell_{f_y}^2+\frac{c_{f_y}^2\ell_{g_{yy}}^2}{\mu^2}) + 6(\ell_{f_x}^2+\frac{c_{f_y}^2\ell_{g_{xy}}^2}{\mu^2})\frac{2}{\alpha_1 K} \notag \\
  &  \quad + \ell_{g_y}^2\frac{50}{3K\alpha_2 \mu} \Tilde{c}_0  +(\frac{c_{f_y}^2\ell_{g_{yy}}^2}{\mu^2}+\ell_{f_y}^2)\frac{600}{K\alpha_3 \mu} \Tilde{c}_1   \bigg) \notag \\
  &  \leq  \frac{\beta_1}{\eta K}\bigg( 27(\ell_{f_x}^2+\frac{c_{f_y}^2\ell_{g_{xy}}^2}{\mu^2}) +15\ell_{g_y}^2 + 27(\ell_{f_y}^2+\frac{c_{f_y}^2\ell_{g_{yy}}^2}{\mu^2}) + (\ell_{f_x}^2+\frac{c_{f_y}^2\ell_{g_{xy}}^2}{\mu^2})\frac{12}{\alpha_1 K} + \frac{50\ell_{g_y}^2}{3K\alpha_2 \mu} \Tilde{c}_0  \notag \\
  & \quad +(\frac{c_{f_y}^2\ell_{g_{yy}}^2}{\mu^2}+\ell_{f_y}^2)\frac{600}{K\alpha_3 \mu} \Tilde{c}_1   \bigg) \notag \\
  & \triangleq \frac{\beta_1}{\eta K}\Tilde{C}_1 \ . 
\end{align}

To eliminate $\mathbb{E}[\| \bar{X}_t - {X}_t\|_F^2]$, we enforce
\begin{align}
        & 12\beta_1\eta(\ell_{f_x}^2+\frac{c_{f_y}^2\ell_{g_{xy}}^2}{\mu^2} )\frac{1}{K} + \frac{25\beta_2 \eta \ell_{g_y}^2}{3\mu}  \frac{1}{K}c_0 + \frac{70\eta\beta_3}{\mu}\Big(\frac{c_{f_y}^2\ell_{g_{yy}}^2}{\mu^2} + \ell_{f_y}^2\Big)\frac{1}{K}c_1 -\frac{\eta(1-\lambda^2)}{2}\frac{1}{K}c_2 \notag \\
  & \quad + 8\eta^2 C_1 + 9\Big(\frac{c_{f_y}^2\ell_{g_{yy}}^2}{\mu^2} + \ell_{f_y}^2\Big)\frac{1}{K}C_4 \leq  0 \ . 
\end{align}
We can obtain
\begin{align}
        & c_2  \geq \frac{2}{(1-\lambda^2)}\Bigg(12\beta_1(\ell_{f_x}^2+\frac{c_{f_y}^2\ell_{g_{xy}}^2}{\mu^2} ) + \frac{25\beta_2  \ell_{g_y}^2}{3\mu}  c_0 + \frac{70\beta_3}{\mu}\Big(\frac{c_{f_y}^2\ell_{g_{yy}}^2}{\mu^2} + \ell_{f_y}^2\Big)c_1  \notag \\
        & \quad \quad + 8\eta K C_1 + \frac{9}{\eta}\Big(\frac{c_{f_y}^2\ell_{g_{yy}}^2}{\mu^2} + \ell_{f_y}^2\Big)C_4\Bigg) \  .  
\end{align}
Then, due to $\beta_3\leq 1$, we can set
\begin{align}
        & c_2  = \frac{2\beta_1}{(1-\lambda^2)}\Bigg(12(\ell_{f_x}^2+\frac{c_{f_y}^2\ell_{g_{xy}}^2}{\mu^2} ) + \frac{25  \ell_{g_y}^2}{3\mu}   \Tilde{c}_0 + \frac{70}{\mu}\Big(\frac{c_{f_y}^2\ell_{g_{yy}}^2}{\mu^2} + \ell_{f_y}^2\Big) \Tilde{c}_1 \notag \\
        & \qquad + 8\Tilde{C}_1+ 36\Big(\frac{c_{f_y}^2\ell_{g_{yy}}^2}{\mu^2} + \ell_{f_y}^2\Big) \Tilde{C}_3\Bigg)  \notag \\
        &\quad  \triangleq  \frac{2\beta_1}{(1-\lambda^2)}\tilde{c}_2\ . 
\end{align}

To eliminate $\mathbb{E}[\|\bar{Y}_t -  {Y}_t\|_F^2] $, we enforce
\begin{align}
        & 12\beta_1\eta(\ell_{f_x}^2+ \frac{c_{f_y}^2\ell_{g_{xy}}^2}{\mu^2})\frac{1}{K} + \frac{25\beta_2 \eta \ell_{g_y}^2}{3\mu}  \frac{1}{K}c_0 + \frac{70\eta\beta_3}{\mu}\Big(\frac{c_{f_y}^2\ell_{g_{yy}}^2}{\mu^2} + \ell_{f_y}^2\Big)\frac{1}{K}c_1 -\frac{\eta(1-\lambda^2)}{2}\frac{1}{K}c_3 \notag \\
  & \quad + 8\eta^2 C_2 +  9\Big(\frac{c_{f_y}^2\ell_{g_{yy}}^2}{\mu^2}+ \ell_{f_y}^2\Big)\frac{1}{K}C_4 \leq  0 \ . 
\end{align}
We can obtain
\begin{align}
        &  \frac{\eta(1-\lambda^2)}{2}\frac{1}{K}c_3 \geq 12\beta_1\eta(\ell_{f_x}^2+ \frac{c_{f_y}^2\ell_{g_{xy}}^2}{\mu^2})\frac{1}{K} + \frac{25\beta_2 \eta \ell_{g_y}^2}{3\mu}  \frac{1}{K}c_0 + \frac{70\eta\beta_3}{\mu}\Big(\frac{c_{f_y}^2\ell_{g_{yy}}^2}{\mu^2} + \ell_{f_y}^2\Big)\frac{1}{K}c_1  \notag \\
  & \quad + 8\eta^2 C_2 +  9\Big(\frac{c_{f_y}^2\ell_{g_{yy}}^2}{\mu^2}+ \ell_{f_y}^2\Big)\frac{1}{K}C_4  \ . 
\end{align}

Then, due to $\beta_3\leq 1$, we can set
\begin{align}
  & c_3 = \frac{2\beta_1}{(1-\lambda^2)}\Bigg(12(\ell_{f_x}^2+ \frac{c_{f_y}^2\ell_{g_{xy}}^2}{\mu^2}) + \frac{25\ell_{g_y}^2}{3\mu}   \Tilde{c}_0 + \frac{70}{\mu}\Big(\frac{c_{f_y}^2\ell_{g_{yy}}^2}{\mu^2} + \ell_{f_y}^2\Big) \Tilde{c}_1  \notag \\
  & \qquad + 8 \Tilde{C}_1 +  36\Big(\frac{c_{f_y}^2\ell_{g_{yy}}^2}{\mu^2}+ \ell_{f_y}^2\Big) \Tilde{C}_3\Bigg)  \ . 
\end{align}
Obviously, $c_3=c_2$. 

To eliminate $\mathbb{E}[\|{Z}_t- \bar{Z}_{t}\|_F^2]$, we enforce
\begin{align}
        &12\beta_1\eta c_{g_{xy}}^2\frac{1}{K} +\Big(\frac{70\eta\beta_3}{\mu}\ell_{g_y}^2+  \frac{9}{4}\Big)\frac{1}{K}c_1 -\frac{\eta(1-\lambda^2)}{2}\frac{1}{K}c_4 + 8\eta^2 C_3 + 9\ell_{g_y}^2\frac{1}{K}C_4 \leq  0 \ . 
\end{align}
We can obtain
\begin{align}
    & c_4 \geq   \frac{2}{\eta(1-\lambda^2)}\Bigg(12\beta_1\eta c_{g_{xy}}^2 +\Big(\frac{70\eta\beta_3}{\mu}\ell_{g_y}^2+  \frac{9}{4}\Big)c_1 + 8\eta^2 K C_3 + 9\ell_{g_y}^2C_4\Bigg) \ . 
\end{align}

Then, due to $\eta\leq 1$ and $\beta_3\leq 1$, we can set 
\begin{align} \label{eq:c-4}
         & c_4 =   \frac{2\beta_1}{(1-\lambda^2)}\Big(12 c_{g_{xy}}^2 +\frac{70}{\mu}\ell_{g_y}^2 \Tilde{c}_1  + 8 \Tilde{C}_3 + 36\ell_{g_y}^2 \Tilde{C}_3\Big) +  \frac{5\beta_1}{\beta_3\eta(1-\lambda^2)}   \Tilde{c}_1  \notag \\
         & \triangleq    \frac{2\beta_1}{(1-\lambda^2)}\tilde{c}_4+  \frac{5\beta_1}{\beta_3\eta(1-\lambda^2)}   \Tilde{c}_1  \ . 
\end{align}

To eliminate $\mathbb{E}[\|P_t- \bar{ P}_t\|_F^2]$, we enforce
\begin{align}
		&\quad \frac{2\eta \beta_1^2}{1-\lambda^2}\frac{1}{K}c_2 + (\lambda-1)\frac{1}{K}c_5 + 4\eta^2\beta_1^2 C_1 \notag \\
		& \leq \frac{2\eta \beta_1^2}{1-\lambda^2}\frac{1}{K}\frac{2\beta_1}{(1-\lambda^2)}\tilde{c}_2  + 4\eta^2\beta_1^2  \frac{\beta_1}{\eta K}\tilde{C}_1- \frac{1}{K}\beta_1(1-\lambda)^2 \notag \\
		& \leq \frac{2 \beta_1^2}{1-\lambda^2}\frac{1}{K}\frac{2\beta_1}{(1-\lambda^2)}\tilde{c}_2  + 4\beta_1^2  \frac{\beta_1}{ K}\tilde{C}_1- \frac{1}{K}\beta_1(1-\lambda)^2\notag \\
		&  \leq  0 \ , 
\end{align}
where the second to last step holds due to $\eta\leq 1$.

Then, we can set
\begin{align}
		& \frac{2 \beta_1^2}{1-\lambda^2}\frac{1}{K}\frac{2\beta_1}{(1-\lambda^2)}\tilde{c}_2 \leq \frac{1}{2}   \frac{1}{K}\beta_1(1-\lambda)^2 \ ,  \notag \\
		&  4\beta_1^2  \frac{\beta_1}{ K}\tilde{C}_1\leq \frac{1}{2}   \frac{1}{K}\beta_1(1-\lambda)^2\ . 
\end{align}
Then, we obtain
\begin{align}
    & \beta_1 \leq \min\left\{\frac{(1-\lambda)^2}{3 \sqrt{\tilde{c}_2} } ,  \frac{1-\lambda}{3\sqrt{\tilde{C}_1}}\right\} \ .
\end{align}

To eliminate $\mathbb{E}[\|Q_t- \bar{ Q}_t\|_F^2]$, we enforce
\begin{align}
		& \quad \frac{2\eta \beta_2^2}{1-\lambda^2}\frac{1}{K}c_3 + (\lambda-1)\frac{1}{K}c_6 + 4\eta^2\beta_2^2 C_2 \notag \\
		&= \frac{2\eta \beta_2^2}{1-\lambda^2}\frac{1}{K}c_2 + (\lambda-1)\frac{1}{K}c_6 + 4\eta^2\beta_2^2 C_1\notag \\
		& \leq \frac{2\eta \beta_2^2}{1-\lambda^2}\frac{1}{K}\frac{2\beta_1}{(1-\lambda^2)}\tilde{c}_2 + 4\eta^2\beta_2^2 \frac{\beta_1}{\eta K}\tilde{C}_1 - \frac{1}{K}\beta_1(1-\lambda)^2  \notag \\
		& \leq \frac{2 \beta_2^2}{1-\lambda^2}\frac{1}{K}\frac{2\beta_1}{(1-\lambda^2)}\tilde{c}_2 + 4\beta_2^2 \frac{\beta_1}{ K}\tilde{C}_1 - \frac{1}{K}\beta_1(1-\lambda)^2\notag \\
		&  \leq 0 \ ,  
\end{align}
where the second to last step holds due to $\eta\leq 1$. 

Then, we can set
\begin{align}
		& \frac{2 \beta_2^2}{1-\lambda^2}\frac{1}{K}\frac{2\beta_1}{(1-\lambda^2)}\tilde{c}_2   \leq \frac{1}{2}\frac{1}{K}\beta_1(1-\lambda)^2  \ , \notag \\
		& 4\beta_2^2 \frac{\beta_1}{ K}\tilde{C}_1\leq \frac{1}{2} \frac{1}{K}\beta_1(1-\lambda)^2  \ . 
\end{align}

We can obtain
\begin{align}
		& \beta_2\leq \min\Bigg\{\frac{(1-\lambda)^2}{3\sqrt{\tilde{c}_2}} ,  \frac{1-\lambda}{3\sqrt{\tilde{C}_1}} \Bigg\}  \ . 
\end{align}

To eliminate $\mathbb{E}[\|R_t- \bar{R}_t\|_F^2] $, we enforce
\begin{align}
		&\quad \frac{2\eta \beta_3^2}{1-\lambda^2}\frac{1}{K}c_4 + (\lambda-1)\frac{1}{K}c_7 + \frac{9\eta\beta_3^2}{4}\frac{1}{K}c_1 + 4\eta^2\beta_3^2 C_3 \notag \\
		& \leq \frac{2\eta \beta_3^2}{1-\lambda^2}\frac{1}{K} \frac{2\beta_1}{(1-\lambda^2)}\tilde{c}_4+   \frac{2\eta \beta_3^2}{1-\lambda^2}\frac{1}{K}\frac{5\beta_1}{\beta_3\eta(1-\lambda^2)}   \Tilde{c}_1 - \frac{1}{K}\beta_1(1-\lambda)^2\notag \\
		& \quad + \frac{9\eta\beta_3^2}{4}\frac{1}{K}\frac{\beta_1}{\beta_3} \Tilde{c}_1 + 4\eta^2\beta_3^2  \frac{\beta_1}{\eta K}\Tilde{C}_3 \notag \\
		& \leq \frac{2 \beta_3^2}{1-\lambda^2}\frac{1}{K} \frac{2\beta_1}{(1-\lambda^2)}\tilde{c}_4+   \frac{2 \beta_3^2}{1-\lambda^2}\frac{1}{K}\frac{5\beta_1}{\beta_3(1-\lambda^2)}   \Tilde{c}_1 - \frac{1}{K}\beta_1(1-\lambda)^2\notag \\
		& \quad + \frac{9\beta_3^2}{4}\frac{1}{K}\frac{\beta_1}{\beta_3} \Tilde{c}_1 + 4\beta_3^2  \frac{\beta_1}{ K}\Tilde{C}_3 \leq 0 \ ,  
\end{align}
where the second to last step holds due to $\eta\leq 1$. 

Then, we can set
\begin{align} \label{eq:beta-3-inequalities}
		& \frac{2 \beta_3^2}{1-\lambda^2}\frac{1}{K} \frac{2\beta_1}{(1-\lambda^2)}\tilde{c}_4 \leq \frac{1}{4} \frac{1}{K}\beta_1(1-\lambda)^2 \ , \notag \\
		& \frac{2 \beta_3^2}{1-\lambda^2}\frac{1}{K}\frac{5\beta_1}{\beta_3(1-\lambda^2)}   \Tilde{c}_1 \leq \frac{1}{4} \frac{1}{K}\beta_1(1-\lambda)^2 \ , \notag \\
		& \frac{9\beta_3^2}{4}\frac{1}{K}\frac{\beta_1}{\beta_3} \Tilde{c}_1 \leq \frac{1}{4} \frac{1}{K}\beta_1(1-\lambda)^2 \ , \notag \\
		&  4\beta_3^2  \frac{\beta_1}{ K}\Tilde{C}_3 \leq \frac{1}{4} \frac{1}{K}\beta_1(1-\lambda)^2 \ . 
\end{align}

We obtain 
\begin{align}
	\label{eq:beta3-upper-1}
	& \beta_3^2\tilde{c}_4 \leq \frac{1}{16} (1-\lambda)^4 \ , \\
	\label{eq:beta3-upper-2}
	& \beta_3  \Tilde{c}_1 \leq \frac{1}{32} (1-\lambda)^4 \ , \\
	\label{eq:beta3-upper-3}
	& \beta_3\Tilde{c}_1 \leq \frac{1}{9} (1-\lambda)^2 \ , \\
	\label{eq:beta3-upper-4}
	&  \beta_3^2  \Tilde{C}_3 \leq \frac{1}{16} (1-\lambda)^2 \ .
\end{align}
As a result, we can set
\begin{align}
		& \beta_3 \leq \min\left\{ \frac{(1-\lambda)^2 }{4\sqrt{\tilde{c}_4}},  \frac{(1-\lambda)^4 }{32{\tilde{c}_1}},   \frac{1-\lambda }{4\sqrt{\tilde{C}_3}}\right\} \ . 
\end{align}

To eliminate $\mathbb{E}[\|\bar{v}_{t}  \|^2]$, we enforce
\begin{align}
        & \quad 4\eta^2\beta_2^2 KC_2 - \frac{3\eta\beta_2^2}{4}c_0 \notag \\
        & \leq 4\eta^2\beta_2^2 K \frac{\beta_1}{\eta K}\Tilde{C}_1 - \frac{3\eta\beta_2^2}{4}\frac{\beta_1}{\beta_2} \Tilde{c}_0 \notag \\
        & = 4\eta\beta_1\beta_2^2  \Tilde{C}_1 - \frac{3\eta\beta_1\beta_2}{4} \Tilde{c}_0 \notag \\
        & \leq  0  \ . 
\end{align}
We can  obtain
\begin{align}
        & \beta_2   \leq \frac{3\Tilde{c}_0}{16 \Tilde{C}_1}\ . 
\end{align}

To eliminate $\mathbb{E}[\|\bar{u}_{t}\| ^2 ] $, we enforce
\begin{align}
        & \quad \frac{25\eta\beta_1^2L_{y}^2 }{6\beta_2\mu}c_0 -  \frac{\beta_1\eta}{4} + \frac{9\eta\beta_1^2L_z^2}{\beta_3\mu}c_1 + 4\eta^2\beta_1^2 K C_1 \notag \\
        & \leq \frac{25\eta\beta_1^2L_{y}^2 }{6\beta_2\mu}\frac{\beta_1}{\beta_2} \Tilde{c}_0 -  \frac{\beta_1\eta}{4} + \frac{9\eta\beta_1^2L_z^2}{\beta_3\mu}\frac{\beta_1}{\beta_3} \Tilde{c}_1  + 4\eta^2\beta_1^2 K \frac{\beta_1}{\eta K}\Tilde{C}_1  \leq 0 \ . 
\end{align}
We can set
\begin{align}
        & \frac{25\eta\beta_1^2L_{y}^2 }{6\beta_2\mu}\frac{\beta_1}{\beta_2} \Tilde{c}_0 \leq  \frac{\beta_1\eta}{8}  \ , \notag \\
        &  \frac{9\eta\beta_1^2L_z^2}{\beta_3\mu}\frac{\beta_1}{\beta_3} \Tilde{c}_1   \leq  \frac{\beta_1\eta}{16}  \ , \notag \\
        &   4\eta^2\beta_1^2 K \frac{\beta_1}{\eta K}\Tilde{C}_1  \leq  \frac{\beta_1\eta}{16}  \ . 
\end{align}
Then, we  obtain
\begin{align}
    & \beta_1\leq \min\left\{\sqrt{\frac{6\mu}{200L_{y}^2 \Tilde{c}_0}} \beta_2, \sqrt{\frac{\mu}{144 L_z^2\Tilde{c}_1}} \beta_3 ,  \frac{1}{8\sqrt{\Tilde{C}_1}} \right\}  \ . 
\end{align}

In summary, by setting
\begin{align} \label{eq:hyper}
		& \beta_1 \leq \min\left\{\sqrt{\frac{6\mu}{200L_{y}^2 \Tilde{c}_0}} \beta_2, \sqrt{\frac{\mu}{144 L_z^2\Tilde{c}_1}} \beta_3 ,  \frac{(1-\lambda)^2}{3 \sqrt{\tilde{c}_2} } ,  \frac{1-\lambda}{8\sqrt{\tilde{C}_1}} \right\} \  ,   \notag \\
		&   \beta_2\leq \min\left\{\frac{1}{6\ell_{g_y}}, \frac{3\Tilde{c}_0}{16 \Tilde{C}_1}, \frac{(1-\lambda)^2}{3 \sqrt{\tilde{c}_2} } ,  \frac{1-\lambda}{3\sqrt{\tilde{C}_1}}\right\}  \  ,  \notag \\
			& \beta_3 \leq \min\left\{ \frac{(1-\lambda)^2 }{4\sqrt{\tilde{c}_4}},  \frac{(1-\lambda)^4 }{32{\tilde{c}_1}},   \frac{1-\lambda }{4\sqrt{\tilde{C}_3}},  \frac{\sqrt{K\alpha_3}}{24 \ell_{g_y}}, \frac{\sqrt{K\alpha_3} \mu}{600 \ell_{g_y}^2}, 1\right\} \ , \notag \\
		& \eta\leq \min\left\{\frac{1}{2\beta_1 L_F}, \frac{1}{\beta_3\mu}, \frac{1}{\sqrt{\alpha_1}}, \frac{1}{\sqrt{\alpha_2}}, \frac{1}{\sqrt{\alpha_3}}, 1\right\} \ , 
\end{align}
and 
\begin{align} \label{eq:coeff}
		&  c_0  = \frac{\beta_1}{\beta_2} \Tilde{c}_0 \ ,  \notag \\
		&   \Tilde{c}_0 = \frac{48}{\mu}\Big(\ell_{f_x}^2 +\frac{c_{f_y}^2\ell_{g_{xy}}^2}{\mu^2} \Big)  +   \frac{144}{\mu}\Big(\frac{\ell_{g_{yy}}^2c_{f_y}^2}{\mu^2}+ \ell_{f_y}^2\Big) \Big( 27c_{g_{xy}}^2 + 27\ell_{g_y}^2 + \frac{12c_{g_{xy}}^2}{K\alpha_1}\Big)       \notag \\
		& \quad+ \frac{768 c_{g_{xy}}^2}{\mu^2}\Big(\frac{9}{\mu}+ \frac{5400\ell_{g_y}^2}{\alpha_3 K \mu}\Big)\Big(\frac{c_{f_y}^2\ell_{g_{yy}}^2}{\mu^2}+\ell_{f_y}^2\Big)  \notag \\
		& \quad+ \frac{2304 \ell_{g_y}^2 }{\mu^2}\Big(\frac{9}{\mu}+ \frac{5400\ell_{g_y}^2}{\alpha_3 K \mu}\Big)\Big(\frac{c_{f_y}^2\ell_{g_{yy}}^2}{\mu^2}+\ell_{f_y}^2\Big) (27c_{g_{xy}}^2 + 27\ell_{g_y}^2 + \frac{12c_{g_{xy}}^2}{K\alpha_1} ) \ ,  \notag \\
		& c_1  =\frac{\beta_1}{\beta_3}\Tilde{c}_1  \ , \quad   \Tilde{c}_1  =  \frac{192 c_{g_{xy}}^2}{\mu}  + \frac{576 \ell_{g_y}^2 }{\mu} \Big(27c_{g_{xy}}^2 + 27\ell_{g_y}^2 + \frac{12c_{g_{xy}}^2}{K\alpha_1} \Big)   \ ,\notag \\
		& c_2  =c_3 = \frac{2\beta_1}{(1-\lambda^2)}\tilde{c}_2 \ , \notag \\
		& \tilde{c}_2=12\Big(\ell_{f_x}^2+\frac{c_{f_y}^2\ell_{g_{xy}}^2}{\mu^2} \Big) + \frac{25  \ell_{g_y}^2}{3\mu}   \Tilde{c}_0 + \frac{70}{\mu}\Big(\frac{c_{f_y}^2\ell_{g_{yy}}^2}{\mu^2} + \ell_{f_y}^2\Big) \Tilde{c}_1  + 8\Tilde{C}_1+ 36\Big(\frac{c_{f_y}^2\ell_{g_{yy}}^2}{\mu^2} + \ell_{f_y}^2\Big) \Tilde{C}_3  \ ,   \notag \\
		& c_4 =   \frac{2\beta_1}{(1-\lambda^2)}\tilde{c}_4+  \frac{4\beta_1}{\beta_3\eta(1-\lambda^2)}   \Tilde{c}_1   \ ,\notag \\
		& \tilde{c}_4 = 12 c_{g_{xy}}^2 +\frac{70}{\mu}\ell_{g_y}^2 \Tilde{c}_1  + 8 \Tilde{C}_3 + 36\ell_{g_y}^2 \Tilde{C}_3 \ , \notag \\
		& c_5 = c_6 = c_7 = \beta_1(1-\lambda) \ ,  \quad  c_8 = \frac{2\beta_1}{\alpha_1 \eta} \ ,  \quad c_9 =c_{11} = c_{13}  = 3\beta_1 \ , \notag \\
		& c_{10} = \frac{25\beta_2}{3\alpha_2 \eta\mu}c_0 = \frac{25\beta_1}{3\alpha_2 \eta\mu} \Tilde{c}_0\ , \quad  c_{12} = \frac{100\beta_3}{\alpha_3 \eta\mu}c_1 = \frac{100\beta_1}{\alpha_3 \eta\mu}\Tilde{c}_1   \ ,     \notag \\
		& C_3 \leq    \frac{\beta_1}{\eta K}\Tilde{C}_3 \ ,\quad \tilde{C}_3 = 27c_{g_{xy}}^2 + 27\ell_{g_y}^2 + \frac{12c_{g_{xy}}^2}{\alpha_1 K }  + \frac{600 \ell_{g_y}^2}{\alpha_3 K\mu}\Tilde{c}_1  \ , \notag \\
		& C_4 \leq   4\eta\beta_1\beta_3^2 \Tilde{C}_3 \ , \notag \\
		& C_1 =C_2 \leq  \frac{\beta_1}{\eta K}\Tilde{C}_1  \ , \notag \\
		& \tilde{C}_1 = 27(\ell_{f_x}^2+\frac{c_{f_y}^2\ell_{g_{xy}}^2}{\mu^2}) +15\ell_{g_y}^2 + 27(\ell_{f_y}^2+\frac{c_{f_y}^2\ell_{g_{yy}}^2}{\mu^2}) + (\ell_{f_x}^2+\frac{c_{f_y}^2\ell_{g_{xy}}^2}{\mu^2})\frac{12}{\alpha_1 K} + \frac{50\ell_{g_y}^2}{3K\alpha_2 \mu} \Tilde{c}_0 \notag \\
		& \quad +(\frac{c_{f_y}^2\ell_{g_{yy}}^2}{\mu^2}+\ell_{f_y}^2)\frac{600}{K\alpha_3 \mu} \Tilde{c}_1    \ ,  
\end{align}
we can obtain
\begin{align}
		&   \mathcal{L}_{t+1} -  \mathcal{L}_{t} \leq - \frac{\beta_1\eta}{2}\mathbb{E} [ \| \nabla F(\bar{x}_{t})\|^2 ] +  6\beta_1\alpha_1^2\eta^4(1 + \frac{c_{f_y}^2}{\mu^2})\sigma^2 +  6\beta_1\alpha_2^2\eta^4\sigma^2 +  6\beta_1\alpha_3^2\eta^4(1 + \frac{c_{f_y}^2}{\mu^2})\sigma^2  \notag \\
  &  + 4\alpha_1^2\eta^3 \frac{2\beta_1}{\alpha_1 K}(1 + \frac{c_{f_y}^2}{\mu^2})\sigma^2 + 12\beta_1\alpha_1^2\eta^4 (1 + \frac{c_{f_y}^2}{\mu^2})\sigma^2  + 12\beta_1\alpha_3^2\eta^4(1 + \frac{c_{f_y}^2}{\mu^2})\sigma^2 + 6\beta_1\alpha_2^2\eta^4 \sigma^2   \notag \\
  & + 4\alpha_3^2\eta^3(1 + \frac{c_{f_y}^2}{\mu^2})\sigma^2 \frac{100\beta_3}{\alpha_3 K \mu}c_1  + 2\alpha_2^2\eta^3 \sigma^2\frac{25\beta_2}{3\alpha_2 K\mu}c_0  \ . 
\end{align}


By summing over $t$ from $0$ to $T-1$, we can obtain
\begin{align} \label{eq:grad_norm_with_L}
		&  \frac{1}{T}\sum_{t=0}^{T-1}\mathbb{E} [ \| \nabla F(\bar{x}_{t})\|^2 ] \leq  \frac{2(\mathcal{L}_{0} -  \mathcal{L}_{T})}{\beta_1\eta} \notag \\
  & +  12\alpha_1^2\eta^3(1 + \frac{c_{f_y}^2}{\mu^2})\sigma^2 +  12\alpha_2^2\eta^3\sigma^2+  12\alpha_3^2\eta^3(1 + \frac{c_{f_y}^2}{\mu^2})\sigma^2  + 16\alpha_1\eta^2 \frac{1}{ K}(1 + \frac{c_{f_y}^2}{\mu^2})\sigma^2 \notag \\
  & + 24\alpha_1^2\eta^3 (1 + \frac{c_{f_y}^2}{\mu^2})\sigma^2   + 12\alpha_2^2\eta^3 \sigma^2    + 24\alpha_3^2\eta^3(1 + \frac{c_{f_y}^2}{\mu^2})\sigma^2 \notag \\
  & + \frac{100\alpha_2\eta^2 }{3 K\mu} \Tilde{c}_0\sigma^2 + \Tilde{c}_1\frac{800\alpha_3\eta^2}{ K \mu} (1 + \frac{c_{f_y}^2}{\mu^2})\sigma^2  \ . 
\end{align}


For the initialization step, due to $x_{0}^{(k)}=x_{0}$, $y_{0}^{(k)}=y_{0}$, $z_{0}^{(k)}=z_{0}$,  we have

\begin{align}
		& \mathcal{L}_{0} =   {\mathbb{E}}[F(\bar{x}_{0})] +  c_0\mathbb{E}[\|\bar{   {y}}_{0} -    {y}^{*}(\bar{   {x}}_{0})\| ^2 ]  +  c_1\mathbb{E}[\|\bar{ z}_{0} -    {z}^{*}(\bar{   {x}}_{0})\| ^2 ]   \notag \\
		& \quad + c_5\frac{1}{K} \mathbb{E}[\|P_{0} - \bar{P}_{0}\|_F^2 ] + c_6 \frac{1}{K}\mathbb{E}[\|Q_{0} - \bar{Q}_{0}\|_F^2 ]  +c_7\frac{1}{K}\mathbb{E}[\| R_{0}-\bar{R}_{0} \|_F^2] \notag \\
		& \quad + c_8 \mathbb{E} [ \|\frac{1}{K}\delta^{\hat{\mathcal{G}}_{F}}(X_{0}, Y_{0}, Z_{0}) \mathbf{1}  -  \frac{1}{K} U_{0} \mathbf{1} \|^2 ]+ c_9 \frac{1}{K}\mathbb{E} [ \|\delta^{\hat{\mathcal{G}}_{F}}(X_{0}, Y_{0}, Z_{0}) -  U_{0} \|_F^2 ] \notag \\
		& \quad  + c_{10}\mathbb{E} [ \|\frac{1}{K}\delta^{g}(X_{0}, Y_{0}) \mathbf{1} -  \frac{1}{K} V_{0} \mathbf{1} \|^2 ] + c_{11} \frac{1}{K}\mathbb{E} [ \|\delta^{g}(X_{0}, Y_{0}) - V_{0}  \|_F^2 ]\notag \\
		& \quad  + c_{12} \mathbb{E} [ \|\frac{1}{K}\delta^{\hat{\mathcal{G}}_{h}}(X_{0}, Y_{0}, Z_{0})\mathbf{1}  -  \frac{1}{K} W_{0} \mathbf{1} \|^2 ] +   c_{13} \frac{1}{K}\mathbb{E} [ \|\delta^{\hat{\mathcal{G}}_{h}}(X_{0}, Y_{0}, Z_{0}) -   W_{0}  \|_F^2 ] \ .
\end{align}

As for $\mathbb{E}[\|P_{0} - \bar{P}_{0}\|_F^2 ]$, we have
\begin{align}
		& \quad\frac{1}{K} \mathbb{E}[\|P_{0} - \bar{P}_{0}\|_F^2 ]  \notag \\
		& = \frac{1}{K} \sum_{k=1}^{K}\mathbb{E}[\|	\hat{\mathcal{G}}_{F}^{(k)}(x_{0}, y_{0}, z_{0}; \hat{\xi}_{0}^{(k)})- 	\frac{1}{K} \sum_{k'=1}^{K}\hat{\mathcal{G}}_{F}^{(k')}(x_{0}, y_{0}, z_{0}; \hat{\xi}_{0}^{(k')})\|^2 ]  \notag \\
		& \leq 2\frac{1}{K} \sum_{k=1}^{K}\mathbb{E}[\|\nabla_{1} { f^{(k)}(x_{0}, y_{0}; \xi_{0}^{(k)})}   - 	\frac{1}{K} \sum_{k'=1}^{K}\nabla_{1} { f^{(k')}(x_{0}, y_{0}; \xi_{0}^{(k')})}  \|^2 ]  \notag \\
		& \quad + 2\frac{1}{K} \sum_{k=1}^{K}\mathbb{E}[\|\nabla_{12}^2 g^{(k)}(x_{0}, y_{0}; \zeta_{0}^{(k)})z_{0}  - \frac{1}{K} \sum_{k'=1}^{K}\nabla_{12}^2 g^{(k')}(x_{0}, y_{0}; \zeta_{0}^{(k')})z_{0} \|^2 ]  \notag \\
		& \leq 6\frac{1}{K} \sum_{k=1}^{K}\mathbb{E}[\|\nabla_{1} { f^{(k)}(x_{0}, y_{0}; \xi_{0}^{(k)})} - \nabla_{1} { f^{(k)}(x_{0}, y_{0})} \|^2 ]\notag \\
		& \quad  + 6\frac{1}{K} \sum_{k=1}^{K}\mathbb{E}[\|\nabla_{1} { f^{(k)}(x_{0}, y_{0})}  - \frac{1}{K} \sum_{k'=1}^{K}\nabla_{1} { f^{(k')}(x_{0}, y_{0})}\|^2 ] \notag \\
		& \quad + 6\frac{1}{K} \sum_{k=1}^{K}\mathbb{E}[\|\frac{1}{K} \sum_{k'=1}^{K}\nabla_{1} { f^{(k')}(x_{0}, y_{0})}   - 	\frac{1}{K} \sum_{k'=1}^{K}\nabla_{1} { f^{(k')}(x_{0}, y_{0}; \xi_{0}^{(k')})}  \|^2 ]  \notag \\
		& \quad + 6\frac{1}{K} \sum_{k=1}^{K}\mathbb{E}[\|\nabla_{12}^2 g^{(k)}(x_{0}, y_{0}; \zeta_{0}^{(k)})z_{0}  - \nabla_{12}^2 g^{(k)}(x_{0}, y_{0})z_{0} \|^2 ]\notag \\
		& \quad  + 6\frac{1}{K} \sum_{k=1}^{K}\mathbb{E}[\|\nabla_{12}^2 g^{(k)}(x_{0}, y_{0})z_{0}    - \frac{1}{K} \sum_{k'=1}^{K}\nabla_{12}^2 g^{(k')}(x_{0}, y_{0})z_{0} \|^2 ] \notag \\
		& \quad + 6\frac{1}{K} \sum_{k=1}^{K}\mathbb{E}[\|\frac{1}{K} \sum_{k'=1}^{K}\nabla_{12}^2 g^{(k')}(x_{0}, y_{0})z_{0}- \frac{1}{K} \sum_{k'=1}^{K}\nabla_{12}^2 g^{(k')}(x_{0}, y_{0}; \zeta_{0}^{(k')})z_{0} \|^2 ]  \notag \\
		& \leq 12(1+  \frac{c_{f_y}^2}{\mu^2})\frac{\sigma^2}{B_0} + \frac{24c_{g_{xy}}^2c_{f_y}^2}{\mu^2} +  24\frac{1}{K} \sum_{k=1}^{K}\mathbb{E}[\|\nabla_{1} { f^{(k)}(x_{0}, y_{0})} \|^2]   \ . 
\end{align}

As for $\mathbb{E}[\|Q_{0} - \bar{Q}_{0}\|_F^2 ]$, we have
\begin{align}
		& \quad \frac{1}{K} \mathbb{E}[\|Q_{0} - \bar{Q}_{0}\|_F^2 ]  \notag \\
		& = \frac{1}{K} \sum_{k=1}^{K}\mathbb{E}[\|\nabla_{2} g^{(k)}(x_0, y_0; {\zeta}_0^{(k)})  - \frac{1}{K} \sum_{k=1}^{K}\nabla_{2} g^{(k')}(x_0, y_0; {\zeta}_0^{(k')})\|^2 ]  \notag \\
		& \leq 3\frac{1}{K} \sum_{k=1}^{K}\mathbb{E}[\|\nabla_{2} g^{(k)}(x_0, y_0; {\zeta}_0^{(k)}) - \nabla_{2} g^{(k)}(x_0, y_0) \|^2 ] \notag \\
		& \quad + 3\frac{1}{K} \sum_{k=1}^{K}\mathbb{E}[\|\nabla_{2} g^{(k)}(x_0, y_0)  - \frac{1}{K} \sum_{k=1}^{K}\nabla_{2} g^{(k')}(x_0, y_0) \|^2 ] \notag \\
		& \quad   +3\frac{1}{K} \sum_{k=1}^{K}\mathbb{E}[\| \frac{1}{K} \sum_{k=1}^{K}\nabla_{2} g^{(k')}(x_0, y_0)- \frac{1}{K} \sum_{k=1}^{K}\nabla_{2} g^{(k')}(x_0, y_0; {\zeta}_0^{(k')})\|^2 ]  \notag \\
		& \leq \frac{6\sigma^2}{B_0} + 12\frac{1}{K} \sum_{k=1}^{K}\mathbb{E}[\|\nabla_{2} g^{(k)}(x_0, y_0)\|^2]  \ . 
\end{align}

As for $\mathbb{E}[\| R_{0}-\bar{R}_{0} \|_F^2]$, we have
\begin{align}
		& \quad \frac{1}{K}\mathbb{E}[\| R_{0}-\bar{R}_{0} \|_F^2] \notag \\
		& =  \frac{1}{K} \sum_{k=1}^{K}\mathbb{E}[\|	\hat{\mathcal{G}}_{h}^{(k)}(x_{0}, y_{0}, z_{0}; \hat{\xi}_{0}^{(k)})- 	\frac{1}{K} \sum_{k'=1}^{K}\hat{\mathcal{G}}_{h}^{(k')}(x_{0}, y_{0}, z_{0}; \hat{\xi}_{0}^{(k')})\|^2 ]  \notag \\
		& \leq 2\frac{1}{K} \sum_{k=1}^{K}\mathbb{E}[\| \nabla_{22}^2g^{(k)}(x_0, y_0; \zeta_{0}^{(k)}) z_0 - \frac{1}{K} \sum_{k'=1}^{K}\nabla_{22}^2g^{(k')}(x_0, y_0; \zeta_{0}^{(k')}) z_0  \|^2 ]  \notag \\
		& \quad  + 2\frac{1}{K} \sum_{k=1}^{K}\mathbb{E}[\|  \nabla_{2}{ f^{(k)}(x_0, y_0; \xi_{0}^{(k)})}  -  \frac{1}{K} \sum_{k'=1}^{K} \nabla_{2}{ f^{(k')}(x_0, y_0; \xi_{0}^{(k')})} \|^2 ]  \notag \\
		& \leq  6\frac{1}{K} \sum_{k=1}^{K}\mathbb{E}[\| \nabla_{22}^2g^{(k)}(x_0, y_0; \zeta_{0}^{(k)}) z_0 - \nabla_{22}^2g^{(k)}(x_0, y_0) z_0  \|^2 ]  \notag \\
		&\quad + 6\frac{1}{K} \sum_{k=1}^{K}\mathbb{E}[\| \nabla_{22}^2g^{(k)}(x_0, y_0) z_0  - \frac{1}{K} \sum_{k'=1}^{K}\nabla_{22}^2g^{(k')}(x_0, y_0) z_0  \|^2 ]  \notag \\
		& \quad+ 6\frac{1}{K} \sum_{k=1}^{K}\mathbb{E}[\| \frac{1}{K} \sum_{k'=1}^{K}\nabla_{22}^2g^{(k')}(x_0, y_0) z_0 - \frac{1}{K} \sum_{k'=1}^{K}\nabla_{22}^2g^{(k')}(x_0, y_0; \zeta_{0}^{(k')}) z_0  \|^2 ]  \notag \\
		& \quad  + 6\frac{1}{K} \sum_{k=1}^{K}\mathbb{E}[\|  \nabla_{2}{ f^{(k)}(x_0, y_0; \xi_{0}^{(k)})} - \nabla_{2}{ f^{(k)}(x_0, y_0)} \|^2 ]\notag \\
		& \quad + 6\frac{1}{K} \sum_{k=1}^{K}\mathbb{E}[\| \nabla_{2}{ f^{(k)}(x_0, y_0)}  - \frac{1}{K} \sum_{k'=1}^{K} \nabla_{2}{ f^{(k')}(x_0, y_0)} \|^2 ]\notag \\
		& \quad + 6\frac{1}{K} \sum_{k=1}^{K}\mathbb{E}[\| \frac{1}{K} \sum_{k'=1}^{K} \nabla_{2}{ f^{(k')}(x_0, y_0)} -  \frac{1}{K} \sum_{k'=1}^{K} \nabla_{2}{ f^{(k')}(x_0, y_0; \xi_{0}^{(k')})} \|^2 ]  \notag \\
		& \leq 12(1+\frac{c_{f_y}^2}{\mu^2})\frac{\sigma^2}{B_0}+ \frac{24\ell_{g_y}^2c_{f_y}^2}{\mu^2} + 24 c_{f_y}^2\ . 
\end{align}

On the other hand, we have
\begin{align}
		&\quad  \mathbb{E} [ \|\frac{1}{K}\delta^{\hat{\mathcal{G}}_{F}}(X_0, Y_0, Z_0) \mathbf{1}-  \frac{1}{K} U_0 \mathbf{1} \|^2 ] \notag \\
		& = \mathbb{E} [ \|\frac{1}{K} \sum_{k=1}^{K}\hat{\mathcal{G}}_{F}^{(k)}(x_{0}, y_{0}, z_{0})-  \frac{1}{K} \sum_{k=1}^{K}\hat{\mathcal{G}}_{F}^{(k)}(x_{0}, y_{0}, z_{0}; \hat{\xi}_{0}^{(k)}) \|^2 ] \notag \\
		& =  \mathbb{E} [ \|\frac{1}{K} \sum_{k=1}^{K}\nabla_{1} { f^{(k)}(x_{0}, y_{0})}   -\frac{1}{K} \sum_{k=1}^{K}\nabla_{12}^2 g^{(k)}(x_{0}, y_{0})z_{0}\notag \\
		& \quad  - \frac{1}{K} \sum_{k=1}^{K}\nabla_{1} { f^{(k)}(x_{0}, y_{0}; \xi_{0}^{(k)})}   + \frac{1}{K} \sum_{k=1}^{K}\nabla_{12}^2 g^{(k)}(x_{0}, y_{0}; \zeta_{0}^{(k)})z_{0}\|^2 ] \notag \\
		& \leq 2\mathbb{E} [ \|\frac{1}{K} \sum_{k=1}^{K}\nabla_{1} { f^{(k)}(x_{0}, y_{0})} - \frac{1}{K} \sum_{k=1}^{K}\nabla_{1} { f^{(k)}(x_{0}, y_{0}; \xi_{0}^{(k)})} \|^2 ] \notag \\
		& \quad  + 2\mathbb{E} [ \|  \frac{1}{K} \sum_{k=1}^{K}\nabla_{12}^2 g^{(k)}(x_{0}, y_{0})z_{0}  - \frac{1}{K} \sum_{k=1}^{K}\nabla_{12}^2 g^{(k)}(x_{0}, y_{0}; \zeta_{0}^{(k)})z_{0}\|^2 ] \notag \\
		& \leq 2(1+\frac{c_{f_y^2}}{\mu^2})\frac{\sigma^2}{KB_0}  \ , 
\end{align}
and $\frac{1}{K}\mathbb{E} [ \|\delta^{\hat{\mathcal{G}}_{F}}(X_0, Y_0, Z_0) -  U_0 \|_F^2 ]\leq 2(1+\frac{c_{f_y^2}}{\mu^2})\frac{\sigma^2}{B_0}$. 

Moreover, we have
\begin{align}
		&\quad  \mathbb{E} [ \|\frac{1}{K}\delta^{\hat{\mathcal{G}}_{h}}(X_0, Y_0, Z_0) \mathbf{1}-  \frac{1}{K} W_0 \mathbf{1} \|^2 ] \notag \\
		& = \mathbb{E} [ \|\frac{1}{K} \sum_{k=1}^{K}\hat{\mathcal{G}}_{h}^{(k)}(x_{0}, y_{0}, z_{0})-  \frac{1}{K} \sum_{k=1}^{K}\hat{\mathcal{G}}_{h}^{(k)}(x_{0}, y_{0}, z_{0}; \hat{\xi}_{0}^{(k)}) \|^2 ] \notag \\
		& =  \mathbb{E} [ \|\frac{1}{K} \sum_{k=1}^{K}\nabla_{22}^2g^{(k)}(x_{0}, y_{0}) z_{0} -  \frac{1}{K} \sum_{k=1}^{K}\nabla_{2}{ f^{(k)}(x_{0}, y_{0})}   \notag \\
		& \quad  - \frac{1}{K} \sum_{k=1}^{K}\nabla_{22}^2g^{(k)}(x_{0}, y_{0}; \zeta_{0}^{(k)}) z_{0} +  \frac{1}{K} \sum_{k=1}^{K}\nabla_{2}{ f^{(k)}(x_{0}, y_{0};  \xi_{0}^{(k)})}  \|^2 ] \notag \\
		& \leq 2\mathbb{E} [ \|\frac{1}{K} \sum_{k=1}^{K}\nabla_{2} { f^{(k)}(x_{0}, y_{0})} - \frac{1}{K} \sum_{k=1}^{K}\nabla_{2} { f^{(k)}(x_{0}, y_{0}; \xi_{0}^{(k)})} \|^2 ] \notag \\
		& \quad  + 2\mathbb{E} [ \|  \frac{1}{K} \sum_{k=1}^{K}\nabla_{22}^2 g^{(k)}(x_{0}, y_{0})z_{0}  - \frac{1}{K} \sum_{k=1}^{K}\nabla_{22}^2 g^{(k)}(x_{0}, y_{0}; \zeta_{0}^{(k)})z_{0}\|^2 ] \notag \\
		& \leq 2(1+\frac{c_{f_y^2}}{\mu^2})\frac{\sigma^2}{KB_0}  \ , 
\end{align}
and $\frac{1}{K}\mathbb{E} [ \|\delta^{\hat{\mathcal{G}}_{h}}(X_0, Y_0, Z_0) -  W_0 \|_F^2 ]\leq 2(1+\frac{c_{f_y^2}}{\mu^2})\frac{\sigma^2}{B_0}$. 

Similarly, we have
\begin{align}
		& \mathbb{E} [ \|\frac{1}{K}\delta^{g}(X_0, Y_0) \mathbf{1} -  \frac{1}{K} V_0 \mathbf{1} \|^2 ] \leq \frac{\sigma^2}{KB_0}  \ , \quad  \frac{1}{K}\mathbb{E} [ \|\delta^{g}(X_0, Y_0) - V_0  \|_F^2 ] \leq \frac{\sigma^2}{B_0} \ . 
\end{align}

By combining them together, we can obtain
\begin{align}\label{eq:L_0}
		& \mathcal{L}_{0} =   {\mathbb{E}}[F(\bar{x}_{0})] +  c_0\mathbb{E}[\|\bar{   {y}}_{0} -    {y}^{*}(\bar{   {x}}_{0})\| ^2 ]  +  c_1\mathbb{E}[\|\bar{ z}_{0} -    {z}^{*}(\bar{   {x}}_{0})\| ^2 ]   \notag \\
		& \quad + c_5\frac{1}{K} \mathbb{E}[\|P_{0} - \bar{P}_{0}\|_F^2 ] + c_6 \frac{1}{K}\mathbb{E}[\|Q_{0} - \bar{Q}_{0}\|_F^2 ]  +c_7\frac{1}{K}\mathbb{E}[\| R_{0}-\bar{R}_{0} \|_F^2] \notag \\
		& \quad + c_8 \mathbb{E} [ \|\frac{1}{K}\delta^{\hat{\mathcal{G}}_{F}}(X_{0}, Y_{0}, Z_{0}) \mathbf{1}  -  \frac{1}{K} U_{0} \mathbf{1} \|^2 ]+ c_9 \frac{1}{K}\mathbb{E} [ \|\delta^{\hat{\mathcal{G}}_{F}}(X_{0}, Y_{0}, Z_{0}) -  U_{0} \|_F^2 ] \notag \\
		& \quad  + c_{10}\mathbb{E} [ \|\frac{1}{K}\delta^{g}(X_{0}, Y_{0}) \mathbf{1} -  \frac{1}{K} V_{0} \mathbf{1} \|^2 ] + c_{11} \frac{1}{K}\mathbb{E} [ \|\delta^{g}(X_{0}, Y_{0}) - V_{0}  \|_F^2 ]\notag \\
		& \quad  + c_{12} \mathbb{E} [ \|\frac{1}{K}\delta^{\hat{\mathcal{G}}_{h}}(X_{0}, Y_{0}, Z_{0})\mathbf{1}  -  \frac{1}{K} W_{0} \mathbf{1} \|^2 ] +   c_{13} \frac{1}{K}\mathbb{E} [ \|\delta^{\hat{\mathcal{G}}_{h}}(X_{0}, Y_{0}, Z_{0}) -   W_{0}  \|_F^2 ] \notag \\
  & \leq {\mathbb{E}}[F(\bar{x}_{0})] +  c_0\mathbb{E}[\|\bar{   {y}}_{0} -    {y}^{*}(\bar{   {x}}_{0})\| ^2 ]  +  c_1\mathbb{E}[\|\bar{ z}_{0} -    {z}^{*}(\bar{   {x}}_{0})\| ^2 ]   \notag \\
		& \quad + \beta_1\Big(12(1+  \frac{c_{f_y}^2}{\mu^2})\frac{\sigma^2}{B_0} + \frac{24c_{g_{xy}}^2c_{f_y}^2}{\mu^2} +  24\frac{1}{K} \sum_{k=1}^{K}\mathbb{E}[\|\nabla_{1} { f^{(k)}(x_{0}, y_{0})} \|^2]\Big) \notag \\
  & \quad + \beta_1 \Big(\frac{6\sigma^2}{B_0} + 12\frac{1}{K} \sum_{k=1}^{K}\mathbb{E}[\|\nabla_{2} g^{(k)}(x_0, y_0)\|^2] \Big)  \notag \\
  & \quad +\beta_1\Big(12(1+\frac{c_{f_y}^2}{\mu^2})\frac{\sigma^2}{B_0}+ \frac{24\ell_{g_y}^2c_{f_y}^2}{\mu^2} + 24 c_{f_y}^2\Big)+ \frac{4\beta_1}{\alpha_1 \eta} (1+\frac{c_{f_y^2}}{\mu^2})\frac{\sigma^2}{KB_0} + 6\beta_1 (1+\frac{c_{f_y^2}}{\mu^2})\frac{\sigma^2}{B_0}\notag \\
		& \quad  + c_0 \frac{25\beta_2}{3\alpha_2 \eta\mu}\frac{\sigma^2}{KB_0}  + 3\beta_1 \frac{\sigma^2}{B_0}  + c_1\frac{100\beta_3}{\alpha_3 \eta\mu} 2(1+\frac{c_{f_y^2}}{\mu^2})\frac{\sigma^2}{KB_0}  +   6\beta_1 (1+\frac{c_{f_y^2}}{\mu^2})\frac{\sigma^2}{B_0} \ . 
\end{align}

Plugging Eq.~(\ref{eq:L_0}) into Eq.~(\ref{eq:grad_norm_with_L}), we can obtain
\begin{align} 
		& \quad  \frac{1}{T}\sum_{t=0}^{T-1}\mathbb{E} [ \| \nabla F(\bar{x}_{t})\|^2 ]\notag \\
		&  \leq  \frac{2(F(x_0) -  F(x_*))}{\beta_1\eta T}  +   \Tilde{c}_0\frac{2}{\beta_2\eta T}\mathbb{E}[\|\bar{   {y}}_{0} -    {y}^{*}(\bar{   {x}}_{0})\| ^2 ]  +  \Tilde{c}_1\frac{2}{\beta_3\eta T}\mathbb{E}[\|\bar{ z}_{0} -    {z}^{*}(\bar{   {x}}_{0})\| ^2 ]   \notag \\
		& \quad +  \frac{48}{\eta T}\frac{1}{K} \sum_{k=1}^{K}\mathbb{E}[\|\nabla_{1} { f^{(k)}(x_{0}, y_{0})} \|^2]+ \frac{24}{\eta T}\frac{1}{K} \sum_{k=1}^{K}\mathbb{E}[\|\nabla_{2} g^{(k)}(x_0, y_0)\|^2]  \notag \\
  & \quad  + \frac{48(c_{g_{xy}}^2+2\ell_{g_y}^2)c_{f_y}^2}{\eta T\mu^2} + \frac{8}{\alpha_1 \eta^2 T KB_0} (1+\frac{c_{f_y^2}}{\mu^2})\sigma^2 + \frac{100}{\eta TB_0} (1+\frac{c_{f_y^2}}{\mu^2})\sigma^2 \notag \\
   & \quad +  36\alpha_1^2\eta^3(1 + \frac{c_{f_y}^2}{\mu^2})\sigma^2 +  24\alpha_2^2\eta^3\sigma^2+  36\alpha_3^2\eta^3(1 + \frac{c_{f_y}^2}{\mu^2})\sigma^2  +  \frac{16\alpha_1\eta^2}{ K}(1 + \frac{c_{f_y}^2}{\mu^2})\sigma^2 \notag \\
  & \quad  + \Tilde{c}_0 \frac{100\alpha_2\eta^2 }{3 K\mu} \sigma^2 + \Tilde{c}_1\frac{800\alpha_3\eta^2}{ K \mu} (1 + \frac{c_{f_y}^2}{\mu^2})\sigma^2  + \Tilde{c}_0 \frac{50}{3\mu\alpha_2 \eta^2 TKB_0}\sigma^2 \notag \\
  & \quad  +  \Tilde{c}_1\frac{400}{\alpha_3 \mu \eta^2 TKB_0} (1+\frac{c_{f_y^2}}{\mu^2})\sigma^2  \ . 
\end{align}


 According to Eq.~(\ref{eq:hyper}) and Eq.~(\ref{eq:coeff}), by setting $\alpha_1=O(\frac{1}{K})$, $\alpha_2=O(\frac{1}{K})$, and $\alpha_3=O(\frac{1}{K})$,  it is easy to know that the constants $\{\tilde{c}_0, \tilde{c}_1, \tilde{c}_2, \tilde{c}_4, \tilde{C}_1, \tilde{C}_3\}$ only depend on the constant in Assumptions~\ref{assumption_bi_strong}-\ref{assumption_lower_smooth_vr}. As a result, we can obtain 
 \begin{align} 
 			&  \frac{1}{T}\sum_{t=0}^{T-1}\mathbb{E} [ \| \nabla F(\bar{x}_{t})\|^2 ] \leq O\left( \frac{1}{\beta_1\eta T}\right)+  O\left( \frac{1}{\beta_2\eta T}\right)  + O\left( \frac{1}{\beta_3\eta T}\right)   \notag \\
 			&   + O\left(\frac{1}{\eta T}\right)+ O\left(\frac{1}{\eta TB_0} \right) + O\left(\frac{1}{\alpha_1 \eta^2 T KB_0}\right)  + O\left( \frac{1}{\alpha_2 \eta^2 TKB_0}\right) +  O\left(\frac{1}{\alpha_3  \eta^2 TKB_0} \right)  \notag \\
 			&    +O\left( \frac{\alpha_1\eta^2}{ K}\right)  + O\left( \frac{\alpha_2\eta^2 }{ K} \right) +O\left(\frac{\alpha_3\eta^2}{ K } \right)   + O\left(\alpha_1^2\eta^3\right)  +  O\left(\alpha_2^2\eta^3\right)+ O\left(\alpha_3^2\eta^3\right)   \  .  
 \end{align}

Then, according to Eq.~(\ref{eq:hyper}),   by setting   $\alpha_1=O(\frac{1}{K})$, $\alpha_2=O(\frac{1}{K})$,  $\alpha_3=O(\frac{1}{K})$, $\beta_1 = O((1-\lambda)^4)$, $\beta_2 = O((1-\lambda)^2)$, $\beta_3 = O((1-\lambda)^4)$, 
$\eta=O(K\epsilon^{1/2})$, the batch size in the first iteration as $B_0=O(1/\epsilon^{1/2})$,  the batch size in other iterations as $B_1=O(1)$,   and  $T=O\left(\frac{1}{K(1-\lambda)^4\epsilon^{3/2}}\right)$,  we can obtain
\begin{align} \label{eq:convergence-upper-bound-sim}
		& O\left(\frac{1}{\beta_1\eta T}\right) = O\left(\epsilon\right) \ , \quad 
		 O\left(\frac{1}{\beta_2\eta T}\right) = O\left((1-\lambda)^2\epsilon\right) \ , \quad 
		 O\left(\frac{1}{\beta_3\eta T}\right) = O\left(\epsilon\right) \ , \notag \\
		& O\left(\frac{1}{\eta T}\right) = O\left((1-\lambda)^4\epsilon\right) \ , \quad  O\left(\frac{1}{\eta TB_0} \right)    =  O\left((1-\lambda)^4\epsilon^{3/2}\right)  \ , \notag \\
		& O\left(\frac{1}{\alpha_1 \eta^2 T KB_0}\right)  = O\left(\frac{(1-\lambda)^4\epsilon^{3/2}}{K}\right)  \ ,  \quad O\left(\frac{1}{\alpha_2 \eta^2 T KB_0}\right)  = O\left(\frac{(1-\lambda)^4\epsilon^{3/2}}{K}\right)  \ ,  \notag \\
		& O\left(\frac{1}{\alpha_3 \eta^2 T KB_0}\right)  = O\left(\frac{(1-\lambda)^4\epsilon^{3/2}}{K}\right)  \ , \notag \\
		& O\left( \frac{\alpha_1\eta^2}{ K}\right)  = O(\epsilon) \ , \quad  O\left( \frac{\alpha_2\eta^2}{ K}\right)  = O(\epsilon) \ , \quad  O\left( \frac{\alpha_3\eta^2}{ K}\right)  = O(\epsilon) \ , \notag \\ 
		& O\left(\alpha_1^2\eta^3\right) = O\left(K\epsilon^{3/2}\right) \ , \quad   O\left(\alpha_2^2\eta^3\right) = O\left(K\epsilon^{3/2}\right) \ , O\left(\alpha_3^2\eta^3\right)   = O\left(K\epsilon^{3/2}\right) \ . 
\end{align}
Therefore, we have $\frac{1}{T}\sum_{t=0}^{T-1}\mathbb{E} [ \| \nabla F(\bar{x}_{t})\|^2 ]\leq \epsilon$.

\end{proof}

\section{Proof of Theorem~\ref{theorem-alt}} \label{sec:proof-alt}

\begin{lemma} \label{lemma_hypergrad_bias-alt}
		Under Assumptions~\ref{assumption_bi_strong}-\ref{assumption_graph}, we have
	\begin{align} 
		& \quad \mathbb{E}[\|\nabla F(\bar{x}_{t}) -    \frac{1}{K}\delta^{\hat{\mathcal{G}}_{F}} (X_{t}, Y_{t+1}, Z_{t+1}) \mathbf{1}  \|^2 ] \notag \\
		&  \leq  9(\ell_{f_x}^2+\frac{c_{f_y}^2\ell_{g_{xy}}^2}{\mu^2})  \mathbb{E}[\| y^*(\bar{x}_{t}) - \bar{y}_{t}\|^2] + 9 c_{g_{xy}}^2  \mathbb{E}[\|z^*(\bar{x}_{t}) - \bar{z}_{t} \|^2] \notag \\
		& \quad   + 9\beta_2^2\eta^2(\ell_{f_x}^2+\frac{c_{f_y}^2\ell_{g_{xy}}^2}{\mu^2} )  \mathbb{E}[ \| \bar{v}_{t}\|^2] + 9 \beta_3^2\eta^2c_{g_{xy}}^2   \mathbb{E}[ \|\bar{w}_{t}\|^2]  \notag \\
		& \quad +  9(\ell_{f_x}^2+\frac{c_{f_y}^2\ell_{g_{xy}}^2}{\mu^2})\frac{1}{K}\sum_{k=1}^{K} \mathbb{E}[\|{x}_{t}^{(k)} - \bar{x}_{t}\|^2] \notag \\
		& \quad+ 9(\ell_{f_x}^2+\frac{c_{f_y}^2\ell_{g_{xy}}^2}{\mu^2}) \frac{1}{K}\sum_{k=1}^{K} \mathbb{E}[\|{y}_{t+1}^{(k)} - \bar{y}_{t+1}\|^2]  +  9c_{g_{xy}}^2\frac{1}{K}\sum_{k=1}^{K} \mathbb{E}[\|{z}_{t+1}^{(k)} - \bar{z}_{t+1}\|^2]  \ .  
\end{align}
\end{lemma}

\begin{proof}
		\begin{align} 
			& \quad \mathbb{E}[\|\nabla F(\bar{x}_{t}) -    \frac{1}{K}\delta^{\hat{\mathcal{G}}_{F}} (X_{t}, Y_{t+1}, Z_{t+1}) \mathbf{1}  \|^2]] \notag \\
			& \leq  9\mathbb{E}[\|\frac{1}{K}\sum_{k=1}^{K}\nabla_{1} f^{(k)}(\bar{x}_{t}, y^*(\bar{x}_{t})) - \frac{1}{K}\sum_{k=1}^{K}\nabla_{1} f^{(k)}(\bar{x}_{t}, \bar{y}_{t}) \|^2] \notag \\
			& \quad + 9\mathbb{E}[\|\frac{1}{K}\sum_{k=1}^{K}\nabla_{1} f^{(k)}(\bar{x}_{t}, \bar{y}_{t}) - \frac{1}{K}\sum_{k=1}^{K}\nabla_{1} f^{(k)}(\bar{x}_{t}, \bar{y}_{t+1}) \|^2] \notag \\
			& \quad  +9\mathbb{E}[\| \frac{1}{K}\sum_{k=1}^{K}\nabla_{1} f^{(k)}(\bar{x}_{t}, \bar{y}_{t+1})-  \frac{1}{K}\sum_{k=1}^{K}\nabla_{1} { f^{(k)}({x}_{t}^{(k)}, {y}_{t+1}^{(k)})} \|^2]  \notag \\
			& \quad+ 9\mathbb{E}[\| - \frac{1}{K}\sum_{k=1}^{K} \nabla_{12}^2 g^{(k)}(\bar{x}_{t}, y^*(\bar{x}_{t}))   z^*(\bar{x}_{t}) + \frac{1}{K}\sum_{k=1}^{K} \nabla_{12}^2 g^{(k)}(\bar{x}_{t}, y^*(\bar{x}_{t})) \bar{z}_{t}  \|^2]\notag \\
			& \quad + 9\mathbb{E}[\|- \frac{1}{K}\sum_{k=1}^{K} \nabla_{12}^2 g^{(k)}(\bar{x}_{t}, y^*(\bar{x}_{t}))   \bar{z}_{t}  + \frac{1}{K}\sum_{k=1}^{K} \nabla_{12}^2 g^{(k)}(\bar{x}_{t}, \bar{y}_{t})   \bar{z}_{t} \|^2]\notag \\
			& \quad + 9\mathbb{E}[\|- \frac{1}{K}\sum_{k=1}^{K} \nabla_{12}^2 g^{(k)}(\bar{x}_{t}, \bar{y}_{t})   \bar{z}_{t} + \frac{1}{K}\sum_{k=1}^{K} \nabla_{12}^2 g^{(k)}(\bar{x}_{t}, \bar{y}_{t+1})   \bar{z}_{t} \|^2]\notag \\
			& \quad + 9\mathbb{E}[\|-  \frac{1}{K}\sum_{k=1}^{K} \nabla_{12}^2 g^{(k)}(\bar{x}_{t}, \bar{y}_{t+1})   \bar{z}_{t} + \frac{1}{K}\sum_{k=1}^{K} \nabla_{12}^2 g^{(k)}(\bar{x}_{t}, \bar{y}_{t+1})   \bar{z}_{t+1} \|^2]\notag \\
			& \quad + 9\mathbb{E}[\| - \frac{1}{K}\sum_{k=1}^{K} \nabla_{12}^2 g^{(k)}(\bar{x}_{t}, \bar{y}_{t+1})   \bar{z}_{t+1}+ \frac{1}{K}\sum_{k=1}^{K} \nabla_{12}^2 g^{(k)}(\bar{x}_{t}, \bar{y}_{t+1})   z_{t+1}^{(k)}  \|^2]\notag \\
			& \quad + 9\mathbb{E}[\|- \frac{1}{K}\sum_{k=1}^{K} \nabla_{12}^2 g^{(k)}(\bar{x}_{t}, \bar{y}_{t+1})  z_{t+1}^{(k)}  +\frac{1}{K}\sum_{k=1}^{K}\nabla_{12}^2 g^{(k)}({x}_{t}^{(k)}, {y}_{t+1}^{(k)})z_{t+1}^{(k)} \|^2] \notag \\
			& \leq  9(\ell_{f_x}^2+\frac{c_{f_y}^2\ell_{g_{xy}}^2}{\mu^2})  \mathbb{E}[\| y^*(\bar{x}_{t}) - \bar{y}_{t}\|^2] + 9 c_{g_{xy}}^2  \mathbb{E}[\|z^*(\bar{x}_{t}) - \bar{z}_{t} \|^2] \notag \\
			& \quad   + 9\beta_2^2\eta^2(\ell_{f_x}^2+\frac{c_{f_y}^2\ell_{g_{xy}}^2}{\mu^2} )  \mathbb{E}[ \| \bar{v}_{t}\|^2] + 9 \beta_3^2\eta^2c_{g_{xy}}^2   \mathbb{E}[ \|\bar{w}_{t}\|^2] \notag \\
			& \quad  +  9(\ell_{f_x}^2+\frac{c_{f_y}^2\ell_{g_{xy}}^2}{\mu^2})\frac{1}{K}\sum_{k=1}^{K} \mathbb{E}[\|{x}_{t}^{(k)} - \bar{x}_{t}\|^2]+ 9(\ell_{f_x}^2+\frac{c_{f_y}^2\ell_{g_{xy}}^2}{\mu^2}) \frac{1}{K}\sum_{k=1}^{K} \mathbb{E}[\|{y}_{t+1}^{(k)} - \bar{y}_{t+1}\|^2]  \notag \\
			& \quad +  9c_{g_{xy}}^2\frac{1}{K}\sum_{k=1}^{K} \mathbb{E}[\|{z}_{t+1}^{(k)} - \bar{z}_{t+1}\|^2]  \ .   
	\end{align}

\end{proof}

\begin{lemma} \label{lemma_F_iter-alt}
	Under Assumptions~\ref{assumption_bi_strong}-\ref{assumption_graph}, if $\eta\leq\frac{1}{2\beta_1 L_F}$, we have
		\begin{align}
			& \mathbb{E}[F(\bar{x}_{t+1})] \leq \mathbb{E}[F(\bar{x}_{t})] - \frac{\beta_1\eta}{2}\mathbb{E}[\| \nabla F(\bar{x}_{t})\|^2]  -  \frac{\beta_1\eta}{4}\mathbb{E}[\|\bar{u} _{t}\|^2] \notag \\
			& \quad + 9\beta_1\eta(\ell_{f_x}^2+\frac{c_{f_y}^2\ell_{g_{xy}}^2}{\mu^2}) \mathbb{E}[\| y^*(\bar{x}_{t}) - \bar{y}_{t}\|^2] + 9 \beta_1\eta c_{g_{xy}}^2 \mathbb{E}[\|z^*(\bar{x}_{t}) - \bar{z}_{t} \|^2]    \notag \\
			& \quad + 9\beta_1\beta_2^2\eta^3(\ell_{f_x}^2+\frac{c_{f_y}^2\ell_{g_{xy}}^2}{\mu^2} )  \mathbb{E}[\|\bar{v}_{t}\|^2] + 9\beta_1\beta_3^2\eta^3 c_{g_{xy}}^2  \mathbb{E}[ \| \bar{w}_{t}\|^2] \notag \\
			& \quad  +  9\beta_1\eta(\ell_{f_x}^2+\frac{c_{f_y}^2\ell_{g_{xy}}^2}{\mu^2})\frac{1}{K}\mathbb{E}[\|{X}_{t} - \bar{X}_{t}\|_F^2]\notag \\
			& \quad + 9\beta_1\eta(\ell_{f_x}^2+\frac{c_{f_y}^2\ell_{g_{xy}}^2}{\mu^2}) \frac{1}{K}\mathbb{E}[\|{Y}_{t+1} - \bar{Y}_{t+1}\|_F^2]  \notag \\
			& \quad +  9\beta_1\eta c_{g_{xy}}^2\frac{1}{K}\mathbb{E}[\|{Z}_{t+1} - \bar{Z}_{t+1}\|_F^2]+\beta_1\eta\mathbb{E}[\|\frac{1}{K}\delta^{\hat{\mathcal{G}}_{F}} (X_{t}, Y_{t+1}, Z_{t+1}) \mathbf{1} - \frac{1}{K}{U}_{t} \mathbf{1}\|^2]   \ .   
	\end{align}
\end{lemma}

\begin{proof}
	Using the similar approach as Lemma~\ref{lemma_F_iter}, by setting $\eta\leq\frac{1}{2\beta_1 L_F}$, we can obtain 
	\begin{align}
			& \mathbb{E}[F(\bar{x}_{t+1})] \leq \mathbb{E}[F(\bar{x}_{t})] - \frac{\beta_1\eta}{2}\mathbb{E}[\| \nabla F(\bar{x}_{t})\|^2]  -  \frac{\beta_1\eta}{4}\mathbb{E}[\|\bar{u} _{t}\|^2 ]+  \frac{\beta_1\eta}{2}\mathbb{E}[\|\nabla F(\bar{x}_{t}) - \bar{u} _{t}\|^2] \notag \\
			& \leq \mathbb{E}[F(\bar{x}_{t})] - \frac{\beta_1\eta}{2}\mathbb{E}[\| \nabla F(\bar{x}_{t})\|^2]  -  \frac{\beta_1\eta}{4}\mathbb{E}[\|\bar{u} _{t}\|^2] \notag \\
			& \quad +\beta_1\eta\mathbb{E}[\|\frac{1}{K}\delta^{\hat{\mathcal{G}}_{F}} (X_{t}, Y_{t+1}, Z_{t+1}) \mathbf{1} - \frac{1}{K}{U}_{t} \mathbf{1}\|^2] \notag \\
			& \quad +  \beta_1\eta\mathbb{E}[\|\nabla F(\bar{x}_{t}) -  \frac{1}{K}\delta^{\hat{\mathcal{G}}_{F}} (X_{t}, Y_{t+1}, Z_{t+1}) \mathbf{1}  \|^2] \ .  
	\end{align}
  Then, due to Lemma~\ref{lemma_hypergrad_bias-alt}, we complete the proof. 
  

\end{proof}

\begin{lemma} \label{lemma_z_opt-alt}

 	Under Assumptions~\ref{assumption_bi_strong}-\ref{assumption_graph}, if  $\eta \leq \frac{1}{\beta_3\mu}$, we have
  \begin{align}
     & \quad  \mathbb{E} [ \|\bar{z}_{t+1} - z^{*}(\bar{{x}}_{t+1})\| ^2] \notag \\
     &  \leq  (1-\frac{\eta\beta_3\mu}{8}) \mathbb{E} [ \|\bar{z}_{t} - z^*(\bar{x}_{t})\|^2 ]+ \frac{18\eta\beta_3}{\mu}\Big(\frac{c_{f_y}^2\ell_{g_{yy}}^2}{\mu^2}+\ell_{f_y}^2\Big) \mathbb{E} [ \| \bar{y}_{t}  - y^*(\bar{x}_{t})\|^2] \notag \\
     & \quad +  \frac{9\eta\beta_1^2L_z^2}{\beta_3\mu} \mathbb{E} [ \|\bar{u}_{t}\|^2]  +  \frac{18\eta^3\beta_2^2\beta_3}{\mu}\Big(\frac{c_{f_y}^2\ell_{g_{yy}}^2}{\mu^2}+\ell_{f_y}^2\Big)  \mathbb{E}[\|  \bar{v}_{t}\|^2]\notag \\
     & \quad+ \frac{9\eta\beta_3^2}{4}\frac{1}{K}\mathbb{E} [ \|   R_{t} -   \bar{R}_{t} \|_F^2] +\Big(\frac{70\eta\beta_3}{\mu}\ell_{g_y}^2+  \frac{9}{4}\Big)\frac{1}{K}\mathbb{E} [ \|{Z}_{t}- \bar{Z}_{t}\|_F^2]  \notag \\
     & \quad + \frac{70\eta\beta_3}{\mu}\Big(\frac{c_{f_y}^2\ell_{g_{yy}}^2}{\mu^2} + \ell_{f_y}^2\Big)\frac{1}{K}\mathbb{E} [ \|\bar{X}_{t} - {X}_{t}\|_F^2]\notag \\
     & \quad + \frac{70\eta\beta_3}{\mu}\Big(\frac{c_{f_y}^2\ell_{g_{yy}}^2}{\mu^2}+ \ell_{f_y}^2\Big)\frac{1}{K}\mathbb{E} [ \| \bar{Y}_{t+1} - {Y}_{t+1}\|_F^2] \notag \\
     & \quad +\frac{25\eta\beta_3}{\mu} \mathbb{E} [ \| \frac{1}{K}\delta^{\hat{\mathcal{G}}_{h}}({X}_{t}, {Y}_{t+1},  {Z}_{t})\mathbf{1}-   \frac{1}{K}{W}_{t}\mathbf{1}\|^2] \ . 
 \end{align}
	
\end{lemma}

\begin{proof}

	At first, we have
		\begin{align}\label{eq:z-bar-z-star-2-alt}
			& \quad \mathbb{E}[\| \bar{z}_{t}  -\eta\beta_3 \hat{\mathcal{G}}_{h}(\bar{x}_{t}, \bar{y}_{t+1}, \bar{z}_{t}) -  z^*(\bar{x}_{t}) + \eta\beta_3 \nabla_z h(\bar{x}_{t}, z^*(\bar{x}_{t}))\|^2]  \notag \\
			& =  \mathbb{E}[\| \bar{z}_{t}  -\eta\beta_3 (\nabla_{22}^2g(\bar{x}_{t}, \bar{y}_{t+1}) \bar{z}_{t}   -  \nabla_2 f(\bar{x}_{t},  \bar{y}_{t+1}))  \notag \\
			& \quad -  z^*(\bar{x}_{t}) + \eta\beta_3 (\nabla_{22}^2g(\bar{x}_{t}, y^*(\bar{x}_{t})) z^*(\bar{x}_{t})   -  \nabla_2{ f(\bar{x}_{t}, y^*(\bar{x}_{t}))})\|^2]  \notag \\
			& \leq (1+a)\mathbb{E}[\|(I-\eta\beta_3\nabla_{22}^2g(\bar{x}_{t}, \bar{y}_{t+1}) )\bar{z}_{t} - (I-\eta\beta_3\nabla_{22}^2g(\bar{x}_{t}, \bar{y}_{t+1}) )z^*(\bar{x}_{t})\|^2]\notag \\
			& \quad + 4(1+1/a)\mathbb{E}[\|(I-\eta\beta_3\nabla_{22}^2g(\bar{x}_{t}, \bar{y}_{t+1}) )z^*(\bar{x}_{t})  - (I-\eta\beta_3\nabla_{22}^2g(\bar{x}_{t}, \bar{y}_{t}) )z^*(\bar{x}_{t}) \|^2]\notag \\
			& \quad + 4(1+1/a)\mathbb{E}[\|(I-\eta\beta_3\nabla_{22}^2g(\bar{x}_{t}, \bar{y}_{t}) )z^*(\bar{x}_{t})  - (I-\eta\beta_3\nabla_{22}^2g(\bar{x}_{t}, y^*(\bar{x}_{t})) )z^*(\bar{x}_{t})\|^2]\notag \\
			& \quad +  4(1+1/a)\mathbb{E}[\|\eta\beta_3(\nabla_2 f(\bar{x}_{t},  \bar{y}_{t+1})-\nabla_2 f(\bar{x}_{t},  \bar{y}_{t}))\|^2] \notag \\
			& \quad +  4(1+1/a)\mathbb{E}[\|\eta\beta_3(\nabla_2 f(\bar{x}_{t},  \bar{y}_{t})-\nabla_2{ f(\bar{x}_{t}, y^*(\bar{x}_{t}))})\|^2] \notag \\
			& \leq (1+a) (1-\eta\beta_3\mu)^2 \mathbb{E}[\|\bar{z}_{t} - z^*(\bar{x}_{t})\|^2]  \notag \\
			& \quad + 4 \eta^2\beta_3^2\frac{c_{f_y}^2\ell_{g_{yy}}^2}{\mu^2}(1+1/a) \mathbb{E}[\| \bar{y}_{t+1} - \bar{y}_{t}  \|^2] + 4 \eta^2\beta_3^2\frac{c_{f_y}^2\ell_{g_{yy}}^2}{\mu^2}(1+1/a) \mathbb{E}[\| \bar{y}_{t}  - y^*(\bar{x}_{t})\|^2] \notag \\
			& \quad + 4\eta^2\beta_3^2\ell_{f_y}^2(1+1/a) \mathbb{E}[\| \bar{y}_{t+1} - \bar{y}_{t}\|^2] + 4\eta^2\beta_3^2\ell_{f_y}^2(1+1/a)\mathbb{E}[\| \bar{y}_{t}- y^*(\bar{x}_{t})\|^2] \notag \\
			& \leq   (1-\frac{\eta\beta_3\mu}{2}) \mathbb{E}[\|\bar{z}_{t} - z^*(\bar{x}_{t})\|^2] + \frac{12\eta\beta_3}{\mu}\Big(\frac{c_{f_y}^2\ell_{g_{yy}}^2}{\mu^2}+\ell_{f_y}^2\Big) \mathbb{E}[\| \bar{y}_{t}  - y^*(\bar{x}_{t})\|^2] \notag \\
			& \quad + \frac{12\beta_3\beta_2^2\eta^3}{\mu}\Big(\frac{c_{f_y}^2\ell_{g_{yy}}^2}{\mu^2}+\ell_{f_y}^2\Big)\mathbb{E}[\| \bar{v}_{t}\|^2]  \ ,  
	\end{align}
	where the third step follows from Assumptions~\ref{assumption_bi_strong}-\ref{assumption_lower_smooth_vr}, the last step follows from $a=\frac{\eta\beta_3\mu}{2}$ and $\eta \leq \frac{1}{\beta_3\mu}$. 
	
	Then, we have
		\begin{align} \label{eq:z-bar-z-star-3-alt}
			& \quad \mathbb{E}[\| \bar{z}_{t}  - \eta\beta_3  \bar{r}_{t} -  z^*(\bar{x}_{t})\|^2]  = \mathbb{E}[\| \bar{z}_{t}  - \eta\beta_3  \bar{w}_{t} -  z^*(\bar{x}_{t})\|^2] \notag \\
			& = \mathbb{E}[\| \bar{z}_{t}  -\eta\beta_3 \hat{\mathcal{G}}_{h}(\bar{x}_{t}, \bar{y}_{t+1}, \bar{z}_{t})+ \eta\beta_3 \hat{\mathcal{G}}_{h}(\bar{x}_{t}, \bar{y}_{t+1}, \bar{z}_{t}) - \eta\beta_3  \bar{w}_{t} -  z^*(\bar{x}_{t})\|^2] \notag \\
			& \leq (1+a)\mathbb{E}[\| \bar{z}_{t}  -\eta\beta_3 \hat{\mathcal{G}}_{h}(\bar{x}_{t}, \bar{y}_{t+1}, \bar{z}_{t})-  z^*(\bar{x}_{t})\|^2]  \notag \\
			& \quad +(1+1/a)\eta^2\beta_3^2 \mathbb{E}[\|  \hat{\mathcal{G}}_{h}(\bar{x}_{t}, \bar{y}_{t+1}, \bar{z}_{t}) -  \frac{1}{K}{W}_{t}\mathbf{1}\|^2]\notag \\
			& \leq  (1+a)\mathbb{E}[\| \bar{z}_{t}  -\eta\beta_3 \hat{\mathcal{G}}_{h}(\bar{x}_{t}, \bar{y}_{t+1}, \bar{z}_{t}) -  z^*(\bar{x}_{t}) + \eta\beta_3 \nabla_z h(\bar{x}_{t}, z^*(\bar{x}_{t}))\|^2]  \notag \\
			& \quad +(1+1/a)\eta^2\beta_3^2 \mathbb{E}[\|  \hat{\mathcal{G}}_{h}(\bar{x}_{t}, \bar{y}_{t+1}, \bar{z}_{t}) -  \frac{1}{K}{W}_{t}\mathbf{1}\|^2]\notag \\
			& \leq (1+a) (1-\frac{\eta\beta_3\mu}{2}) \mathbb{E}[\|\bar{z}_{t} - z^*(\bar{x}_{t})\|^2] \notag \\
			& \quad + (1+a)\frac{12\eta\beta_3}{\mu}\Big(\frac{c_{f_y}^2\ell_{g_{yy}}^2}{\mu^2}+\ell_{f_y}^2\Big) (\mathbb{E}[\| \bar{y}_{t}  - y^*(\bar{x}_{t})\|^2]+  \beta_2^2\eta^2\mathbb{E}[\|  \bar{v}_{t}\|^2])  \notag \\
			& \quad +2(1+1/a)\eta^2\beta_3^2 \mathbb{E}[\|  \hat{\mathcal{G}}_{h}(\bar{x}_{t}, \bar{y}_{t+1}, \bar{z}_{t}) -  \frac{1}{K}\delta^{\hat{\mathcal{G}}_{h}}({X}_{t}, {Y}_{t+1},  {Z}_{t})\mathbf{1}\|^2]\notag \\
			& \quad +2(1+1/a)\eta^2\beta_3^2 \mathbb{E}[\| \frac{1}{K}\delta^{\hat{\mathcal{G}}_{h}}({X}_{t}, {Y}_{t+1},  {Z}_{t})\mathbf{1} -  \frac{1}{K}{W}_{t}\mathbf{1}\|^2]\notag \\
			& \leq  (1-\frac{\eta\beta_3\mu}{4}) \mathbb{E}[\|\bar{z}_{t} - z^*(\bar{x}_{t})\|^2] + \frac{15\eta\beta_3}{\mu}\Big(\frac{c_{f_y}^2\ell_{g_{yy}}^2}{\mu^2}+\ell_{f_y}^2\Big)(\mathbb{E}[\| \bar{y}_{t}  - y^*(\bar{x}_{t})\|^2]+   \beta_2^2\eta^2\mathbb{E}[\|  \bar{v}_{t}\|^2]) \notag \\
			& \quad +\frac{20\eta\beta_3}{\mu}\mathbb{E}[\| \hat{\mathcal{G}}_{h}(\bar{x}_{t}, \bar{y}_{t+1}, \bar{z}_{t}) - \frac{1}{K}\delta^{\hat{\mathcal{G}}_{h}}({X}_{t}, {Y}_{t+1},  {Z}_{t})\mathbf{1}\|^2] \notag \\
			& \quad +\frac{20\eta\beta_3}{\mu} \mathbb{E}[\| \frac{1}{K}\delta^{\hat{\mathcal{G}}_{h}}({X}_{t}, {Y}_{t+1},  {Z}_{t})\mathbf{1}-   \frac{1}{K}{W}_{t}\mathbf{1}\|^2]\notag \\
			& \leq  (1-\frac{\eta\beta_3\mu}{4}) \mathbb{E}[\|\bar{z}_{t} - z^*(\bar{x}_{t})\|^2] + \frac{15\eta\beta_3}{\mu}\Big(\frac{c_{f_y}^2\ell_{g_{yy}}^2}{\mu^2}+\ell_{f_y}^2\Big) (\mathbb{E}[\| \bar{y}_{t}  - y^*(\bar{x}_{t})\|^2]+  \beta_2^2\eta^2\mathbb{E}[\|  \bar{v}_{t}\|^2]) \notag \\
			& \quad +\frac{60\eta\beta_3}{\mu}\ell_{g_y}^2\frac{1}{K}\mathbb{E}[\|{Z}_{t}- \bar{Z}_{t}\mathbb{E}[\|_F^2  + \frac{60\eta\beta_3}{\mu}\Big(\frac{c_{f_y}^2\ell_{g_{yy}}^2}{\mu^2} + \ell_{f_y}^2\Big)\frac{1}{K}\mathbb{E}[\|\bar{X}_{t} - {X}_{t}\mathbb{E}[\|_F^2 \notag \\
			& \quad +  \frac{60\eta\beta_3}{\mu}\Big(\frac{c_{f_y}^2\ell_{g_{yy}}^2}{\mu^2}+ \ell_{f_y}^2\Big)\frac{1}{K}\mathbb{E}[\| \bar{Y}_{t+1} - {Y}_{t+1}\mathbb{E}[\|_F^2 \notag \\
			& \quad +\frac{20\eta\beta_3}{\mu} \mathbb{E}[\| \frac{1}{K}\delta^{\hat{\mathcal{G}}_{h}}({X}_{t}, {Y}_{t+1},  {Z}_{t})\mathbf{1} -   \frac{1}{K}{W}_{t}\mathbf{1}\|^2] \ ,
	\end{align}
	where the third step follows from $\nabla_z h(\bar{x}_{t}, z^*(\bar{x}_{t})) = 0$, the fourth step follows from Eq.~(\ref{eq:z-bar-z-star-2-alt}), the fifth step follows from  $a=\frac{\eta\beta_3\mu}{4}$ and $\eta\leq \frac{1}{\beta_3\mu}$, and the last step follows from the following inequality.
		\begin{align} \label{eq:h-grad-consensus-alt}
			&\quad  \mathbb{E}[\|  \hat{\mathcal{G}}_{h}(\bar{x}_{t}, \bar{y}_{t+1}, \bar{z}_{t}) -  \frac{1}{K}\delta^{\hat{\mathcal{G}}_{h}}({X}_{t}, {Y}_{t+1},  {Z}_{t})\mathbf{1}\|^2]\notag \\
			&  \leq  \frac{1}{K}\sum_{k=1}^{K}\mathbb{E}[\|\nabla_{22}^2g^{(k)}(\bar{x}_{t}, \bar{y}_{t+1}) \bar{z}_{t}-  \nabla_{2}{ f^{(k)}(\bar{x}_{t}, \bar{y}_{t+1})} \notag \\
			& \quad - \nabla_{22}^2g^{(k)}({x}_{t}^{(k)}, {y}_{t+1}^{(k)}) {z}_{t}^{(k)} +  \nabla_{2}{ f^{(k)}({x}_{t}^{(k)}, {y}_{t+1}^{(k)})}\|^2]  \notag \\
			& \leq  3\frac{1}{K}\sum_{k=1}^{K}\mathbb{E}[\|\nabla_{22}^2g^{(k)}(\bar{x}_{t}, \bar{y}_{t+1}) \bar{z}_{t} - \nabla_{22}^2g^{(k)}(\bar{x}_{t}, \bar{y}_{t+1}) {z}_{t}^{(k)}  \|^2] \notag \\
   & \quad + 3\frac{1}{K}\sum_{k=1}^{K}\mathbb{E}[\|\nabla_{22}^2g^{(k)}(\bar{x}_{t}, \bar{y}_{t+1}) {z}_{t}^{(k)}  - \nabla_{22}^2g^{(k)}({x}_{t}^{(k)}, {y}_{t+1}^{(k)}) {z}_{t}^{(k)} \|^2]  \notag \\
			& \quad + 3\frac{1}{K}\sum_{k=1}^{K}\mathbb{E}[\|- \nabla_{2}{ f^{(k)}(\bar{x}_{t}, \bar{y}_{t+1})} +  \nabla_{2}{ f^{(k)}({x}_{t}^{(k)}, {y}_{t+1}^{(k)})}\|^2]  \notag \\
			& \leq  3\ell_{g_y}^2\frac{1}{K}\mathbb{E}[\|{Z}_{t}- \bar{Z}_{t}\|_F^2]  + 3\Big(\frac{c_{f_y}^2\ell_{g_{yy}}^2}{\mu^2} + \ell_{f_y}^2\Big)\frac{1}{K}\mathbb{E}[\|\bar{X}_{t} - {X}_{t}\|_F^2] \notag \\
			& \quad +  3\Big(\frac{c_{f_y}^2\ell_{g_{yy}}^2}{\mu^2}+ \ell_{f_y}^2\Big)\frac{1}{K}\mathbb{E}[\| \bar{Y}_{t+1} - {Y}_{t+1}\|_F^2] \ , 
	\end{align}
	where the last step follows from Assumptions~\ref{assumption_bi_strong}-\ref{assumption_lower_smooth_vr}. 
Then, we have
\begin{align}
			&\quad   \mathbb{E}[\|\bar{z}_{ t+1} - z^*(\bar{x}_{t+1})\|^2] \notag \\
			&   \leq (1+a)  \mathbb{E}[\|\bar{z}_{ t+1} - z^*(\bar{x}_{t})\|^2] + (1+1/a)  \mathbb{E}[\|z^*(\bar{x}_{t+1})- z^*(\bar{x}_{t})\|^2] \notag \\
			& \leq (1+a)  \frac{2}{K}\mathbb{E}[\|  Z_{t}  - \bar{Z}_{t}  \|_F^2]+ (1+a)2\eta\beta_3^2\frac{1}{K}\mathbb{E}[\|   R_{t} -   \bar{R}_{t} \|_F^2]\notag \\
			& \quad  +  (1+a)\mathbb{E}[\| \bar{z}_{t}  - \eta\beta_3  \bar{r}_{t} -  z^*(\bar{x}_{t})\|^2]  + (1+1/a)  \mathbb{E}[\|z^*(\bar{x}_{t+1})- z^*(\bar{x}_{t})\|^2] \notag \\
			& \leq (1+a)  \frac{2}{K}\mathbb{E}[\|  Z_{t}  - \bar{Z}_{t}  \|_F^2]+ (1+a)2\eta\beta_3^2\frac{1}{K}\mathbb{E}[\|   R_{t} -   \bar{R}_{t} \|_F^2] + (1+1/a)  L_z^2\mathbb{E}[\|\bar{x}_{t+1}- \bar{x}_{t}\|^2] \notag \\
			& \quad   +  (1+a)(1-\frac{\eta\beta_3\mu}{4}) \mathbb{E}[\|\bar{z}_{t} - z^*(\bar{x}_{t})\|^2] \notag \\
			& \quad +(1+a) \frac{15\eta\beta_3}{\mu}\Big(\frac{c_{f_y}^2\ell_{g_{yy}}^2}{\mu^2}+\ell_{f_y}^2\Big) ) (\mathbb{E}[\| \bar{y}_{t}  - y^*(\bar{x}_{t})\|^2]+  \beta_2^2\eta^2\mathbb{E}[\|  \bar{v}_{t}\|^2]) \notag \\
			& \quad +(1+a)\frac{60\eta\beta_3}{\mu}\ell_{g_y}^2\frac{1}{K}\mathbb{E}[\|{Z}_{t}- \bar{Z}_{t}\|_F^2]  +(1+a) \frac{60\eta\beta_3}{\mu}\Big(\frac{c_{f_y}^2\ell_{g_{yy}}^2}{\mu^2} + \ell_{f_y}^2\Big)\frac{1}{K}\mathbb{E}[\|\bar{X}_{t} - {X}_{t}\|_F^2] \notag \\
			& \quad +  (1+a)\frac{60\eta\beta_3}{\mu}\Big(\frac{c_{f_y}^2\ell_{g_{yy}}^2}{\mu^2}+ \ell_{f_y}^2\Big)\frac{1}{K}\mathbb{E}[\| \bar{Y}_{t+1} - {Y}_{t+1}\|_F^2] \notag \\
			& \quad  +(1+a)\frac{20\eta\beta_3}{\mu} \mathbb{E}[\| \frac{1}{K}\sum_{k=1}^{K}\hat{\mathcal{G}}_{h}^{(k)}({x}_{t}^{(k)}, {y}_{t+1}^{(k)}, {z}_{t}^{(k)})-   \frac{1}{K}{W}_{t}\mathbf{1}\|^2]\notag \\
			& \leq  (1-\frac{\eta\beta_3\mu}{8}) \mathbb{E}[\|\bar{z}_{t} - z^*(\bar{x}_{t})\|^2]  +  \frac{18\eta\beta_3}{\mu}\Big(\frac{c_{f_y}^2\ell_{g_{yy}}^2}{\mu^2}+\ell_{f_y}^2\Big)  \mathbb{E}[\| \bar{y}_{t}  - y^*(\bar{x}_{t})\|^2]  \notag \\
			&  \quad + \frac{9\eta\beta_1^2L_z^2}{\beta_3\mu} \mathbb{E}[\|\bar{u}_{t}\|^2] + \frac{18\eta^3\beta_2^2\beta_3}{\mu}\Big(\frac{c_{f_y}^2\ell_{g_{yy}}^2}{\mu^2}+\ell_{f_y}^2\Big)  \mathbb{E}[\|  \bar{v}_{t}\|^2] + \frac{9\eta\beta_3^2}{4}\frac{1}{K}\mathbb{E}[\|   R_{t} -   \bar{R}_{t} \|_F^2]\notag \\
			& \quad +\Big(\frac{70\eta\beta_3}{\mu}\ell_{g_y}^2+  \frac{9}{4}\Big)\frac{1}{K}\mathbb{E}[\|{Z}_{t}- \bar{Z}_{t}\|_F^2]  + \frac{70\eta\beta_3}{\mu}\Big(\frac{c_{f_y}^2\ell_{g_{yy}}^2}{\mu^2} + \ell_{f_y}^2\Big)\frac{1}{K}\mathbb{E}[\|\bar{X}_{t} - {X}_{t}\|_F^2] \notag \\
			& \quad + \frac{70\eta\beta_3}{\mu}\Big(\frac{c_{f_y}^2\ell_{g_{yy}}^2}{\mu^2}+ \ell_{f_y}^2\Big)\frac{1}{K}\mathbb{E}[\| \bar{Y}_{t+1} - {Y}_{t+1}\|_F^2] \notag \\
			& \quad +\frac{25\eta\beta_3}{\mu} \mathbb{E}[\| \frac{1}{K}\sum_{k=1}^{K}\hat{\mathcal{G}}_{h}^{(k)}({x}_{t}^{(k)}, {y}_{t+1}^{(k)}, {z}_{t}^{(k)})-   \frac{1}{K}{W}_{t}\mathbf{1}\|^2] \ ,
	\end{align}
	where the second step follows from Eq.~(\ref{eq:z-bar-z-star-0}),  the third step follows from Eq.~(\ref{eq:z-bar-z-star-3-alt}), the last step follows from $a = \frac{\eta\beta_3\mu}{8}$ and $\eta\leq\frac{1}{\beta_3\mu}$.

\end{proof}

\begin{lemma} \label{lemma_y_opt-alt}
		Under Assumptions~\ref{assumption_bi_strong}-\ref{assumption_graph}, if  $\beta_2\leq \frac{1}{6\ell_{g_y}}$, we have
\begin{align}
		& \quad  \mathbb{E}[\|\bar{   {y}}_{t+1} -    {y}^{*}(\bar{   {x}}_{t+1})\| ^2 ] \notag \\
		& \leq  (1-\frac{\beta_2\eta\mu}{4}) \mathbb{E}[\|\bar{   {y}}_{t}   -    {y}^{*}(\bar{   {x}}_{t})\| ^2] - \frac{3\eta\beta_2^2}{4} \mathbb{E}[\|\bar{v}_{t}  \|^2]   +\frac{25\eta\beta_1^2L_{y}^2 }{6\beta_2\mu} \mathbb{E}[\|\bar{u}_{t}\| ^2 ] \notag \\
		& \quad +  \frac{25\beta_2 \eta \ell_{g_y}^2}{3\mu}  \frac{1}{K}\mathbb{E}[\| \bar{X}_{t} - {X}_{t}\|_F^2] +  \frac{25\beta_2 \eta \ell_{g_y}^2}{3\mu}  \frac{1}{K}\mathbb{E}[\|\bar{Y}_{t} -  {Y}_{t}\|_F^2]\notag \\
		& \quad  +  \frac{25\beta_2 \eta}{3\mu}  \mathbb{E}[\|\frac{1}{K} \delta^{g}  ({X}_{t}, {Y}_{t}) \mathbf{1} - \frac{1}{K}{V}_{t}\mathbf{1} \|^2]  \ . 
\end{align}
\end{lemma}
This lemma is the same as Lemma~\ref{lemma_y_opt}.

\begin{lemma} \label{lemma_hyper_storm_var_mean-alt}
	Under Assumptions~\ref{assumption_bi_strong}-\ref{assumption_graph}, we have the following two inequalities:
		\begin{align} \label{eq:hyper_storm_var_mean-alt}
			& \quad\mathbb{E} [ \|\frac{1}{K}\delta^{\hat{\mathcal{G}}_{F}}(X_{t}, Y_{t+1}, Z_{t+1}) \mathbf{1} -  \frac{1}{K} U_{t} \mathbf{1} \|^2 ] \notag \\
			& \leq  (1-\alpha_1 \eta^2) \mathbb{E} [ \| \frac{1}{K}\delta^{\hat{\mathcal{G}}_{F}}(X_{t-1}, Y_{t}, Z_{t}) \mathbf{1} -  \frac{1}{K}U_{t-1} \mathbf{1}  \|^2 ]\notag \\
			& \quad + 6(\ell_{f_x}^2+\frac{c_{f_y}^2\ell_{g_{xy}}^2}{\mu^2})\frac{1}{K^2}\mathbb{E}[\|X_{t} - X_{t-1}\|_F^2] \notag \\
			& \quad  + 6(\ell_{f_x}^2+\frac{c_{f_y}^2\ell_{g_{xy}}^2}{\mu^2})\frac{1}{K^2}\mathbb{E}[\|Y_{t+1} - Y_{t}\|_F^2]\notag \\
			& \quad + 6c_{g_{xy}}^2\frac{1}{K^2}\mathbb{E}[\|Z_{t+1} - Z_{t}\|_F^2] + 4\alpha_1^2\eta^4 (1 + \frac{c_{f_y}^2}{\mu^2})\sigma^2\frac{1}{K} \ ,  
	\end{align}
	and 
\begin{align}\label{eq:hyper_storm_var_individual-alt}
		& \quad\frac{1}{K}\mathbb{E} [ \|\delta^{\hat{\mathcal{G}}_{F}}(X_{t}, Y_{t+1}, Z_{t+1}) -   U_{t}  \|_F^2 ] \notag \\
		& \leq  (1-\alpha_1 \eta^2) \frac{1}{K}\mathbb{E} [ \| \delta^{\hat{\mathcal{G}}_{F}}(X_{t-1}, Y_{t}, Z_{t}) - U_{t-1}  \|_F^2 ]\notag \\
		& \quad + 6(\ell_{f_x}^2+\frac{c_{f_y}^2\ell_{g_{xy}}^2}{\mu^2})\frac{1}{K}\mathbb{E}[\|X_{t} - X_{t-1}\|_F^2] \notag \\
		& \quad   + 6(\ell_{f_x}^2+\frac{c_{f_y}^2\ell_{g_{xy}}^2}{\mu^2})\frac{1}{K}\mathbb{E}[\|Y_{t+1} - Y_{t}\|_F^2]\notag \\
		& \quad + 6c_{g_{xy}}^2\frac{1}{K}\mathbb{E}[\|Z_{t+1} - Z_{t}\|_F^2] + 4\alpha_1^2\eta^4 (1 + \frac{c_{f_y}^2}{\mu^2})\sigma^2 \ . 
\end{align}
	
\end{lemma}

\begin{proof}
	Eq.~(\ref{eq:hyper_storm_var_mean-alt}) can be proved as follows:
	\begin{align}
			& \quad\mathbb{E} [ \|(\delta^{\hat{\mathcal{G}}_{F}}(X_{t}, Y_{t+1}, Z_{t+1}) -   U_{t}) \frac{1}{K} \mathbf{1} \|^2 ]\notag \\
			& = \mathbb{E} [ \|(\delta^{\hat{\mathcal{G}}_{F}}(X_{t}, Y_{t+1}, Z_{t+1}) \notag \\
			& \quad -  (1-\alpha_1 \eta^2)(U_{t-1} - \delta^{\hat{\mathcal{G}}_{F}}(X_{t-1}, Y_{t}, Z_{t}; \hat{\xi}_{t}) ) - \delta^{\hat{\mathcal{G}}_{F}}(X_{t}, Y_{t+1}, Z_{t+1}; \hat{\xi}_{t}) ) \frac{1}{K} \mathbf{1} \|^2 ]\notag \\
			& = \mathbb{E} [ \| \Big( (1-\alpha_1 \eta^2)(\delta^{\hat{\mathcal{G}}_{F}}(X_{t-1}, Y_{t}, Z_{t}) -  U_{t-1}) \notag \\
			& \quad -(1-\alpha_1 \eta^2)(\delta^{\hat{\mathcal{G}}_{F}}(X_{t}, Y_{t+1}, Z_{t+1}; \hat{\xi}_{t})  - \delta^{\hat{\mathcal{G}}_{F}}(X_{t-1}, Y_{t}, Z_{t}; \hat{\xi}_{t})   \notag \\
			& \quad - \delta^{\hat{\mathcal{G}}_{F}}(X_{t}, Y_{t+1}, Z_{t+1})  + \delta^{\hat{\mathcal{G}}_{F}}(X_{t-1}, Y_{t}, Z_{t}) ) \notag \\
			& \quad - \alpha_1\eta^2(\delta^{\hat{\mathcal{G}}_{F}}(X_{t}, Y_{t+1}, Z_{t+1}; \hat{\xi}_{t})- \delta^{\hat{\mathcal{G}}_{F}}(X_{t}, Y_{t+1}, Z_{t+1}))\Big )\frac{1}{K} \mathbf{1}  \|^2 ]\notag \\
			& = \mathbb{E} [ \| \Big( (1-\alpha_1 \eta^2)(\delta^{\hat{\mathcal{G}}_{F}}(X_{t-1}, Y_{t}, Z_{t}) -  U_{t-1}) \Big )\frac{1}{K} \mathbf{1}  \|^2 ]\notag \\
			& \quad + \mathbb{E} [ \|\Big( (1-\alpha_1 \eta^2)(\delta^{\hat{\mathcal{G}}_{F}}(X_{t}, Y_{t+1}, Z_{t+1}; \hat{\xi}_{t})  - \delta^{\hat{\mathcal{G}}_{F}}(X_{t-1}, Y_{t}, Z_{t}; \hat{\xi}_{t})   \notag \\
			& \quad - \delta^{\hat{\mathcal{G}}_{F}}(X_{t}, Y_{t+1}, Z_{t+1})  + \delta^{\hat{\mathcal{G}}_{F}}(X_{t-1}, Y_{t}, Z_{t}) ) \notag \\
			& \quad + \alpha_1\eta^2(\delta^{\hat{\mathcal{G}}_{F}}(X_{t}, Y_{t+1}, Z_{t+1}; \hat{\xi}_{t})- \delta^{\hat{\mathcal{G}}_{F}}(X_{t}, Y_{t+1}, Z_{t+1}))\Big )\frac{1}{K} \mathbf{1}  \|^2 ]\notag \\
			& \leq   (1-\alpha_1 \eta^2)^2 \mathbb{E} [ \| (\delta^{\hat{\mathcal{G}}_{F}}(X_{t-1}, Y_{t}, Z_{t}) -  U_{t-1}) \frac{1}{K} \mathbf{1}  \|^2 ]\notag \\
			& \quad +2(1-\alpha_1 \eta^2)^2\frac{1}{K^2}\mathbb{E} [ \|  \delta^{\hat{\mathcal{G}}_{F}}(X_{t}, Y_{t+1}, Z_{t+1}; \hat{\xi}_{t})  - \delta^{\hat{\mathcal{G}}_{F}}(X_{t-1}, Y_{t}, Z_{t}; \hat{\xi}_{t}) \notag \\
			& \quad   - \delta^{\hat{\mathcal{G}}_{F}}(X_{t}, Y_{t+1}, Z_{t+1})  + \delta^{\hat{\mathcal{G}}_{F}}(X_{t-1}, Y_{t}, Z_{t})    \|_F^2 ]\notag \\
			& \quad + 2\alpha_1^2\eta^4\frac{1}{K^2}\mathbb{E} [ \|\delta^{\hat{\mathcal{G}}_{F}}(X_{t}, Y_{t+1}, Z_{t+1}; \hat{\xi}_{t})- \delta^{\hat{\mathcal{G}}_{F}}(X_{t}, Y_{t+1}, Z_{t+1}) \|_F^2 ]\notag \\
			& \leq   (1-\alpha_1 \eta^2) \mathbb{E} [ \| (\delta^{\hat{\mathcal{G}}_{F}}(X_{t-1}, Y_{t}, Z_{t}) -  U_{t-1}) \frac{1}{K} \mathbf{1}  \|^2 ]\notag \\
			& \quad +2\frac{1}{K^2}\mathbb{E} [ \|  \delta^{\hat{\mathcal{G}}_{F}}(X_{t}, Y_{t+1}, Z_{t+1}; \hat{\xi}_{t})  - \delta^{\hat{\mathcal{G}}_{F}}(X_{t-1}, Y_{t}, Z_{t}; \hat{\xi}_{t})   \|_F^2 ]\notag \\
			& \quad + 2\alpha_1^2\eta^4\frac{1}{K^2}\mathbb{E} [ \|\delta^{\hat{\mathcal{G}}_{F}}(X_{t}, Y_{t+1}, Z_{t+1}; \hat{\xi}_{t})- \delta^{\hat{\mathcal{G}}_{F}}(X_{t}, Y_{t+1}, Z_{t+1}) \|_F^2 ]\notag \\
			& \leq  (1-\alpha_1 \eta^2) \mathbb{E} [ \| (\delta^{\hat{\mathcal{G}}_{F}}(X_{t-1}, Y_{t}, Z_{t}) -  U_{t-1}) \frac{1}{K} \mathbf{1}  \|^2 ]\notag \\
			& \quad + 6(\ell_{f_x}^2+\frac{c_{f_y}^2\ell_{g_{xy}}^2}{\mu^2})\frac{1}{K^2}\mathbb{E}[\|X_{t} - X_{t-1}\|_F^2] + 6(\ell_{f_x}^2+\frac{c_{f_y}^2\ell_{g_{xy}}^2}{\mu^2})\frac{1}{K^2}\mathbb{E}[\|Y_{t+1} - Y_{t}\|_F^2]\notag \\
			& \quad + 6c_{g_{xy}}^2\frac{1}{K^2}\mathbb{E}[\|Z_{t+1} - Z_{t}\|_F^2] + 4\alpha_1^2\eta^4 (1 + \frac{c_{f_y}^2}{\mu^2})\sigma^2\frac{1}{K} \ ,
	\end{align}
where the last step follows from the following two inequalities:
	\begin{align}
			& \quad \mathbb{E} [ \|\delta^{\hat{\mathcal{G}}_{F}}(X_{t}, Y_{t+1}, Z_{t+1}; \hat{\xi}_{t})- \delta^{\hat{\mathcal{G}}_{F}}(X_{t}, Y_{t+1}, Z_{t+1})\|_F^2 ]\notag \\
			& = \sum_{k=1}^{K}\mathbb{E} [ \|\hat{\mathcal{G}}_{F}^{(k)}(x_{t}^{(k)}, y_{t+1}^{(k)}, z_{t+1}^{(k)};  \hat{\xi}_{t}^{(k)}) - \hat{\mathcal{G}}_{F}^{(k)}(x_{t}^{(k)}, y_{t+1}^{(k)}, z_{t+1}^{(k)}) \|^2 ]\notag \\
			& =\sum_{k=1}^{K} \mathbb{E} [ \|\nabla_{1} { f^{(k)}(x_{t}^{(k)}, y_{t+1}^{(k)}; {\xi}_{t}^{(k)})} - \nabla_{12}^2 g^{(k)}(x_{t}^{(k)}, y_{t+1}^{(k)}; {\zeta}_{t}^{(k)})z_{t+1}^{(k)} \notag \\
			& \quad - \nabla_{1} { f^{(k)}(x_{t}^{(k)}, y_{t+1}^{(k)})} + \nabla_{12}^2 g^{(k)}(x_{t}^{(k)}, y_{t+1}^{(k)})z_{t+1}^{(k)}  \|^2 ]\notag \\
			& \leq 2(1 + \frac{c_{f_y}^2}{\mu^2})\sigma^2K \ , 
	\end{align}
and 
\begin{align}
		&\quad  \mathbb{E} [ \|  \delta^{\hat{\mathcal{G}}_{F}}(X_{t}, Y_{t+1}, Z_{t+1}; \hat{\xi}_{t})  - \delta^{\hat{\mathcal{G}}_{F}}(X_{t-1}, Y_{t}, Z_{t}; \hat{\xi}_{t}) \|_F^2 ]\notag \\
		&   = \sum_{k=1}^{K} \mathbb{E} [ \|\nabla_{1} { f^{(k)}(x_{t}^{(k)}, y_{t+1}^{(k)}; {\xi}_{t}^{(k)})}  -  \nabla_{1} { f^{(k)}(x_{t-1}^{(k)}, y_{t}^{(k)}; {\xi}_{t}^{(k)})}  \notag \\
		& \quad - \nabla_{12}^2 g^{(k)}(x_{t}^{(k)}, y_{t+1}^{(k)}; {\zeta}_{t}^{(k)})z_{t+1}^{(k)}  + \nabla_{12}^2 g^{(k)}(x_{t}^{(k)}, y_{t+1}^{(k)}; {\zeta}_{t}^{(k)})z_{t}^{(k)} \notag \\
		& \quad - \nabla_{12}^2 g^{(k)}(x_{t}^{(k)}, y_{t+1}^{(k)}; {\zeta}_{t}^{(k)})z_{t}^{(k)} + \nabla_{12}^2 g^{(k)}(x_{t-1}^{(k)}, y_{t}^{(k)}; {\zeta}_{t}^{(k)})z_{t}^{(k)} \|^2 ]\notag \\
		& \leq 3(\ell_{f_x}^2+\frac{c_{f_y}^2\ell_{g_{xy}}^2}{\mu^2})\mathbb{E}[\|X_{t} - X_{t-1}\|_F^2]+ 3(\ell_{f_x}^2+\frac{c_{f_y}^2\ell_{g_{xy}}^2}{\mu^2})\mathbb{E}[\|Y_{t+1} - Y_{t}\|_F^2]\notag \\
		& \quad + 3c_{g_{xy}}^2\mathbb{E}[\|Z_{t+1} - Z_{t}\|_F^2] \ . 
\end{align}
Eq.~(\ref{eq:hyper_storm_var_individual-alt}) can be proved by following the above proof so that we omit the detailed steps. 

\end{proof}

\begin{lemma}  \label{lemma_g_storm_var_mean-alt}
		Under Assumptions~\ref{assumption_bi_strong}-\ref{assumption_graph}, we have the following two inequalities:
	\begin{align} \label{eq:g_storm_var_mean-alt}
			& \mathbb{E} [ \|(\delta^{g}(X_{t}, Y_{t} ) -   V_{t}) \frac{1}{K} \mathbf{1} \|^2 ]\leq  (1-\alpha_2 \eta^2) \mathbb{E} [ \| (\delta^{g}(X_{t-1}, Y_{t-1} ) -  V_{t-1}) \frac{1}{K} \mathbf{1}  \|^2 ]\notag \\
			& \quad + 2\ell_{g_y}^2\frac{1}{K^2}\mathbb{E}[\|X_{t} - X_{t-1}\|_F^2] + 2\ell_{g_y}^2\frac{1}{K^2}\mathbb{E}[\|Y_{t} - Y_{t-1}\|_F^2] + 2\alpha_2^2\eta^4 \sigma^2\frac{1}{K} \ , 
	\end{align}
	and
\begin{align} \label{eq:g_storm_var_individual-alt}
		& \frac{1}{K} \mathbb{E} [ \|\delta^{g}(X_{t}, Y_{t} ) -   V_{t}\|_F^2 ] \leq  (1-\alpha_2 \eta^2) \frac{1}{K} \mathbb{E} [ \| \delta^{g}(X_{t-1}, Y_{t-1} ) -  V_{t-1} \|_F^2 ]\notag \\
		& \quad + 2\ell_{g_y}^2\frac{1}{K}\mathbb{E}[\|X_{t} - X_{t-1}\|_F^2] + 2\ell_{g_y}^2\frac{1}{K}\mathbb{E}[\|Y_{t} - Y_{t-1}\|_F^2] + 2\alpha_2^2\eta^4 \sigma^2 \ .
\end{align}
\end{lemma}

\begin{proof}
	Eq.~(\ref{eq:g_storm_var_mean-alt}) can be proved as follows:
		\begin{align}
			& \quad\mathbb{E} [ \|(\delta^{g}(X_{t}, Y_{t} ) -   V_{t}) \frac{1}{K} \mathbf{1} \|^2 ]\notag \\
			& = \mathbb{E} [ \| \Big( (1-\alpha_2 \eta^2)(\delta^{g}(X_{t-1}, Y_{t-1} ) -  V_{t-1}) \notag \\
			& \quad -(1-\alpha_2 \eta^2)(\delta^{g}(X_{t}, Y_{t} ; {\zeta}_{t})  - \delta^{g}(X_{t-1}, Y_{t-1} ; {\zeta}_{t})  - \delta^{g}(X_{t}, Y_{t} )  + \delta^{g}(X_{t-1}, Y_{t-1} ) ) \notag \\
			& \quad  - \alpha_2\eta^2(\delta^{g}(X_{t}, Y_{t} ; {\zeta}_{t})- \delta^{g}(X_{t}, Y_{t} ))\Big )\frac{1}{K} \mathbf{1}  \|^2 ]\notag \\
			& \leq   (1-\alpha_2 \eta^2)^2 \mathbb{E} [ \| (\delta^{g}(X_{t-1}, Y_{t-1} ) -  V_{t-1}) \frac{1}{K} \mathbf{1}  \|^2 ]\notag \\
			& \quad +2(1-\alpha_2 \eta^2)^2\frac{1}{K^2}\mathbb{E} [ \|  \delta^{g}(X_{t}, Y_{t} ; {\zeta}_{t})  - \delta^{g}(X_{t-1}, Y_{t-1} ; {\zeta}_{t})  - \delta^{g}(X_{t}, Y_{t} )  + \delta^{g}(X_{t-1}, Y_{t-1} )    \|_F^2 ]\notag \\
			& \quad + 2\alpha_2^2\eta^4\frac{1}{K^2}\mathbb{E} [ \|\delta^{g}(X_{t}, Y_{t} ; {\zeta}_{t})- \delta^{g}(X_{t}, Y_{t} ) \|_F^2 ]\notag \\
			& \leq   (1-\alpha_2 \eta^2) \mathbb{E} [ \| (\delta^{g}(X_{t-1}, Y_{t-1} ) -  V_{t-1}) \frac{1}{K} \mathbf{1}  \|^2 ]\notag \\
			& \quad +2\frac{1}{K^2}\mathbb{E} [ \|  \delta^{g}(X_{t}, Y_{t} ; {\zeta}_{t})  - \delta^{g}(X_{t-1}, Y_{t-1} ; {\zeta}_{t})   \|_F^2 ] \notag \\
			& \quad + 2\alpha_2^2\eta^4\frac{1}{K^2}\mathbb{E} [ \|\delta^{g}(X_{t}, Y_{t} ; {\zeta}_{t})- \delta^{g}(X_{t}, Y_{t} ) \|_F^2 ]\notag \\
			& \leq  (1-\alpha_2 \eta^2) \mathbb{E} [ \| (\delta^{g}(X_{t-1}, Y_{t-1} ) -  V_{t-1}) \frac{1}{K} \mathbf{1}  \|^2 ]\notag \\
			& \quad + 2\ell_{g_y}^2\frac{1}{K^2}\mathbb{E}[\|X_{t} - X_{t-1}\|_F^2] + 2\ell_{g_y}^2\frac{1}{K^2}\mathbb{E}[\|Y_{t} - Y_{t-1}\|_F^2] + 2\alpha_2^2\eta^4 \sigma^2\frac{1}{K} \ .
	\end{align}
Eq.~(\ref{eq:g_storm_var_individual-alt}) can be proved by following the above proof so that we omit the detailed steps. 
\end{proof}

\begin{lemma} \label{lemma_h_storm_var_mean-alt}
			Under Assumptions~\ref{assumption_bi_strong}-\ref{assumption_graph}, we have the following two inequalities:
		\begin{align} \label{eq:h_storm_var_mean-alt}
			& \quad \mathbb{E} [ \|(\delta^{\hat{\mathcal{G}}_{h}}(X_{t}, Y_{t+1}, Z_{t}) -   W_{t}) \frac{1}{K} \mathbf{1} \|^2 ] \notag \\
			&\leq   (1-\alpha_3 \eta^2) \mathbb{E} [ \| (\delta^{\hat{\mathcal{G}}_{h}}(X_{t-1}, Y_{t}, Z_{t-1}) -  W_{t-1}) \frac{1}{K} \mathbf{1}  \|^2 ]\notag \\
			& \quad +6(\frac{c_{f_y}^2\ell_{g_{yy}}^2}{\mu^2}+\ell_{f_y}^2)\frac{1}{K^2}\mathbb{E}[\|X_{t} - X_{t-1}\|_F^2]  + 6(\frac{c_{f_y}^2\ell_{g_{yy}}^2}{\mu^2}+\ell_{f_y}^2)\frac{1}{K^2}\mathbb{E}[\|Y_{t+1} - Y_{t}\|_F^2] \notag \\
			& \quad + 6\ell_{g_y}^2 \frac{1}{K^2}\mathbb{E}[\|Z_{t} - Z_{t-1}\|_F^2] + 4\alpha_3^2\eta^4(1 + \frac{c_{f_y}^2}{\mu^2})\sigma^2\frac{1}{K} \ , 
	\end{align}
	and
	\begin{align} \label{eq:h_storm_var_individual-alt}
		& \quad \frac{1}{K}\mathbb{E} [ \|\delta^{\hat{\mathcal{G}}_{h}}(X_{t}, Y_{t+1}, Z_{t}) -   W_{t}\|^2 ] \notag \\
		&\leq   (1-\alpha_3 \eta^2) \frac{1}{K}\mathbb{E} [ \| \delta^{\hat{\mathcal{G}}_{h}}(X_{t-1}, Y_{t}, Z_{t-1}) -  W_{t-1}  \|^2 ]\notag \\
		& \quad +6(\frac{c_{f_y}^2\ell_{g_{yy}}^2}{\mu^2}+\ell_{f_y}^2)\frac{1}{K}\mathbb{E}[\|X_{t} - X_{t-1}\|_F^2]  + 6(\frac{c_{f_y}^2\ell_{g_{yy}}^2}{\mu^2}+\ell_{f_y}^2)\frac{1}{K}\mathbb{E}[\|Y_{t+1} - Y_{t}\|_F^2] \notag \\
		& \quad + 6\ell_{g_y}^2 \frac{1}{K}\mathbb{E}[\|Z_{t} - Z_{t-1}\|_F^2] + 4\alpha_3^2\eta^4(1 + \frac{c_{f_y}^2}{\mu^2})\sigma^2 \ . 
\end{align}
\end{lemma}

\begin{proof}
		Eq.~(\ref{eq:h_storm_var_mean-alt}) can be proved as follows:
		\begin{align}
			& \quad\mathbb{E} [ \|(\delta^{\hat{\mathcal{G}}_{h}}(X_{t}, Y_{t+1}, Z_{t}) -   W_{t}) \frac{1}{K} \mathbf{1} \|^2 ]\notag \\
			& = \mathbb{E} [ \| \Big( (1-\alpha_3 \eta^2)(\delta^{\hat{\mathcal{G}}_{h}}(X_{t-1}, Y_{t}, Z_{t-1}) -  W_{t-1}) \notag \\
			& \quad -(1-\alpha_3 \eta^2)(\delta^{\hat{\mathcal{G}}_{h}}(X_{t}, Y_{t+1}, Z_{t}; \hat{\xi}_{t})  - \delta^{\hat{\mathcal{G}}_{h}}(X_{t-1}, Y_{t}, Z_{t-1}; \hat{\xi}_{t})  \notag\\
			& \qquad - \delta^{\hat{\mathcal{G}}_{h}}(X_{t}, Y_{t}, Z_{t})  + \delta^{\hat{\mathcal{G}}_{h}}(X_{t-1}, Y_{t}, Z_{t-1}) ) \notag \\
			& \quad - \alpha_3\eta^2(\delta^{\hat{\mathcal{G}}_{h}}(X_{t}, Y_{t+1}, Z_{t}; \hat{\xi}_{t})- \delta^{\hat{\mathcal{G}}_{h}}(X_{t}, Y_{t+1}, Z_{t}))\Big )\frac{1}{K} \mathbf{1}  \|^2 ]\notag \\
			& \leq   (1-\alpha_3 \eta^2)^2 \mathbb{E} [ \| (\delta^{\hat{\mathcal{G}}_{h}}(X_{t-1}, Y_{t}, Z_{t-1}) -  W_{t-1}) \frac{1}{K} \mathbf{1}  \|^2 ]\notag \\
			& \quad +2(1-\alpha_3 \eta^2)^2\frac{1}{K^2}\mathbb{E} [ \|  \delta^{\hat{\mathcal{G}}_{h}}(X_{t}, Y_{t+1}, Z_{t}; \hat{\xi}_{t})  - \delta^{\hat{\mathcal{G}}_{h}}(X_{t-1}, Y_{t}, Z_{t-1}; \hat{\xi}_{t})  \notag \\
			& \quad - \delta^{\hat{\mathcal{G}}_{h}}(X_{t}, Y_{t+1}, Z_{t})  + \delta^{\hat{\mathcal{G}}_{h}}(X_{t-1}, Y_{t}, Z_{t-1})    \|_F^2 ]\notag \\
			& \quad + 2\alpha_3^2\eta^4\frac{1}{K^2}\mathbb{E} [ \|\delta^{\hat{\mathcal{G}}_{h}}(X_{t}, Y_{t+1}, Z_{t}; \hat{\xi}_{t})- \delta^{\hat{\mathcal{G}}_{h}}(X_{t}, Y_{t+1}, Z_{t}) \|_F^2 ]\notag \\
			& \leq   (1-\alpha_3 \eta^2) \mathbb{E} [ \| (\delta^{\hat{\mathcal{G}}_{h}}(X_{t-1}, Y_{t}, Z_{t-1}) -  W_{t-1}) \frac{1}{K} \mathbf{1}  \|^2 ]\notag \\
			& \quad +2\frac{1}{K^2} \mathbb{E} [ \|  \delta^{\hat{\mathcal{G}}_{h}}(X_{t}, Y_{t+1}, Z_{t}; \hat{\xi}_{t})  - \delta^{\hat{\mathcal{G}}_{h}}(X_{t-1}, Y_{t}, Z_{t-1}; \hat{\xi}_{t})   \|_F^2 ] \notag \\
			& \quad  + 2\alpha_3^2\eta^4\frac{1}{K^2}\mathbb{E} [ \|\delta^{\hat{\mathcal{G}}_{h}}(X_{t}, Y_{t+1}, Z_{t}; \hat{\xi}_{t})- \delta^{\hat{\mathcal{G}}_{h}}(X_{t}, Y_{t+1}, Z_{t}) \|_F^2 ]\notag \\
			& \leq   (1-\alpha_3 \eta^2) \mathbb{E} [ \| (\delta^{\hat{\mathcal{G}}_{h}}(X_{t-1}, Y_{t}, Z_{t-1}) -  W_{t-1}) \frac{1}{K} \mathbf{1}  \|^2 ]\notag \\
			& \quad +6(\frac{c_{f_y}^2\ell_{g_{yy}}^2}{\mu^2}+\ell_{f_y}^2)\frac{1}{K^2}\mathbb{E}[\|X_{t} - X_{t-1}\|_F^2]  + 6(\frac{c_{f_y}^2\ell_{g_{yy}}^2}{\mu^2}+\ell_{f_y}^2)\frac{1}{K^2}\mathbb{E}[\|Y_{t+1} - Y_{t}\|_F^2] \notag \\
			& \quad + 6\ell_{g_y}^2 \frac{1}{K^2}\mathbb{E}[\|Z_{t} - Z_{t-1}\|_F^2] + 4\alpha_3^2\eta^4(1 + \frac{c_{f_y}^2}{\mu^2})\sigma^2\frac{1}{K} \ , 
	\end{align}
where the last step follows from the following two inequalities:
\begin{align}
		& \quad \mathbb{E} [ \|  \delta^{\hat{\mathcal{G}}_{h}}(X_{t}, Y_{t+1}, Z_{t}; \hat{\xi}_{t})  - \delta^{\hat{\mathcal{G}}_{h}}(X_{t-1}, Y_{t}, Z_{t-1}; \hat{\xi}_{t})   \|_F^2 ]\notag \\
		& = \sum_{k=1}^{K}\mathbb{E} [ \| \nabla_{22}^2g^{(k)}(x_{t}^{(k)}, y_{t+1}^{(k)}; \zeta_{t}^{(k)}) z_{t}^{(k)}-  \nabla_{2}{ f^{(k)}(x_{t}^{(k)}, y_{t+1}^{(k)}; \xi_{t}^{(k)})} \notag \\
		& \quad  - \nabla_{22}^2g^{(k)}(x_{t-1}^{(k)}, y_{t}^{(k)}; \zeta_{t}^{(k)}) z_{t-1}^{(k)} + \nabla_{2}{ f^{(k)}(x_{t-1}^{(k)}, y_{t}^{(k)}; \xi_{t}^{(k)})} \|^2 ]\notag \\
		& = \sum_{k=1}^{K}\mathbb{E} [ \| \nabla_{22}^2g^{(k)}(x_{t}^{(k)}, y_{t+1}^{(k)}; \zeta_{t}^{(k)}) z_{t}^{(k)} - \nabla_{22}^2g^{(k)}(x_{t-1}^{(k)}, y_{t}^{(k)}; \zeta_{t}^{(k)}) z_{t}^{(k)} \notag \\
		& \quad + \nabla_{22}^2g^{(k)}(x_{t-1}^{(k)}, y_{t}^{(k)}; \zeta_{t}^{(k)}) z_{t}^{(k)} - \nabla_{22}^2g^{(k)}(x_{t-1}^{(k)}, y_{t}^{(k)}; \zeta_{t}^{(k)}) z_{t-1}^{(k)} \notag \\
		& \quad -  \nabla_{2}{ f^{(k)}(x_{t}^{(k)}, y_{t+1}^{(k)}; \xi_{t}^{(k)})} + \nabla_{2}{ f^{(k)}(x_{t-1}^{(k)}, y_{t}^{(k)}; \xi_{t}^{(k)})} \|^2 ]\notag \\
		& \leq 3(\frac{c_{f_y}^2\ell_{g_{yy}}^2}{\mu^2}+\ell_{f_y}^2)\mathbb{E}[\|X_{t} - X_{t-1}\|_F^2]  + 3(\frac{c_{f_y}^2\ell_{g_{yy}}^2}{\mu^2}+\ell_{f_y}^2)\mathbb{E}[\|Y_{t+1} - Y_{t}\|_F^2] \notag \\
		& \quad + 3\ell_{g_y}^2 \mathbb{E}[\|Z_{t} - Z_{t-1}\|_F^2]  \ , 
\end{align}
and 
\begin{align}
		& \mathbb{E} [ \|\delta^{\hat{\mathcal{G}}_{h}}(X_{t}, Y_{t+1}, Z_{t}; \hat{\xi}_{t})- \delta^{\hat{\mathcal{G}}_{h}}(X_{t}, Y_{t+1}, Z_{t}) \|_F^2 ]\notag \\
		& =\sum_{k=1}^{K} \mathbb{E} [ \|\nabla_{22}^2g^{(k)}(x_{t}^{(k)}, y_{t+1}^{(k)}; \zeta_{t}^{(k)}) z_{t}^{(k)}-  \nabla_2{ f^{(k)}(x_{t}^{(k)}, y_{t+1}^{(k)}; \xi_{t}^{(k)})}  \notag \\
		& \quad - \nabla_{22}^2g^{(k)}(x_{t}^{(k)}, y_{t+1}^{(k)}) z_{t}^{(k)} +  \nabla_2{ f^{(k)}(x_{t}^{(k)}, y_{t+1}^{(k)})}  \|^2 ]\notag \\
		& \leq 2(1 + \frac{c_{f_y}^2}{\mu^2})\sigma^2K \ . 
\end{align}
Eq.~(\ref{eq:h_storm_var_individual-alt}) can be proved by following the above proof so that we omit the detailed steps. 

\end{proof}

\begin{lemma} \label{lemma_consensus_p-alt}
			Under Assumptions~\ref{assumption_bi_strong}-\ref{assumption_graph}, we have 
		\begin{align}
			&  \quad \frac{1}{K}\mathbb{E}[\|P_{t} - \bar{P}_{t}\|_F^2] \notag\\
			&  \leq  \lambda\frac{1}{K}\mathbb{E}[\| P_{t-1} -  \bar{P}_{t-1} \|_F^2] +  \frac{3\alpha_1^2\eta^4}{1-\lambda}\frac{1}{K}\mathbb{E}[\| U_{t-1}- \delta^{\hat{\mathcal{G}}_{F}}(X_{t-1}, Y_{t}, Z_{t})   \|_F^2]\notag \\
			& \quad  + (\ell_{f_x}^2+\frac{c_{f_y}^2\ell_{g_{xy}}^2}{\mu^2})\frac{9}{1-\lambda}\frac{1}{K} \mathbb{E}[\|X_{t} - X_{t-1}\|_F^2] \notag\\
			& \quad+ (\ell_{f_x}^2+\frac{c_{f_y}^2\ell_{g_{xy}}^2}{\mu^2})\frac{9}{1-\lambda}\frac{1}{K} \mathbb{E}[\|Y_{t+1} - Y_{t}\|_F^2]\notag \\
			& \quad + c_{g_{xy}}^2\frac{9}{1-\lambda}\frac{1}{K} \mathbb{E}[\|Z_{t+1} - Z_{t}\|_F^2] +  \frac{6\alpha_1^2\eta^4}{1-\lambda}(1 + \frac{c_{f_y}^2}{\mu^2})\sigma^2 \ .  
	\end{align}
\end{lemma}

\begin{proof}
	\begin{align}
			& \quad \frac{1}{K}\mathbb{E}[\|P_{t} - \bar{P}_{t}\|_F^2] \notag \\
			& = \frac{1}{K}\mathbb{E}[\| P_{t-1}E +U_{t} - U_{t-1} -  \bar{P}_{t-1}- \bar{U}_{t} +\bar{U}_{t-1}\|_F^2] \notag \\
			&\leq \frac{1}{K}\lambda\mathbb{E}[\| P_{t-1} -  \bar{P}_{t-1} \|_F^2]  + \frac{1}{1-\lambda}\frac{1}{K}\mathbb{E}[\|U_{t} - U_{t-1} - \bar{U}_{t} +\bar{U}_{t-1}\|_F^2]\notag \\
			& \leq \lambda\frac{1}{K}\mathbb{E}[\| P_{t-1} -  \bar{P}_{t-1} \|_F^2]  + \frac{1}{1-\lambda}\frac{1}{K}\mathbb{E}[\|U_{t} - U_{t-1} \|_F^2]\notag \\
			& = \lambda\frac{1}{K}\mathbb{E}[\| P_{t-1} -  \bar{P}_{t-1} \|_F^2] \notag \\
			& \quad  + \frac{1}{1-\lambda}\frac{1}{K}\mathbb{E}[\|(1-\alpha_1\eta^2)(U_{t-1} - \delta^{\hat{\mathcal{G}}_{F}}(X_{t-1}, Y_{t}, Z_{t}; \hat{\xi}_{t}))  \notag \\
			& \qquad+ \delta^{\hat{\mathcal{G}}_{F}}(X_{t}, Y_{t+1}, Z_{t+1}; \hat{\xi}_{t})- U_{t-1} \|_F^2]\notag \\
			& = \lambda\frac{1}{K}\mathbb{E}[\| P_{t-1} -  \bar{P}_{t-1} \|_F^2] \notag \\
			& \quad  + \frac{1}{1-\lambda}\frac{1}{K}\mathbb{E}[\|\delta^{\hat{\mathcal{G}}_{F}}(X_{t}, Y_{t+1}, Z_{t+1}; \hat{\xi}_{t}) - \delta^{\hat{\mathcal{G}}_{F}}(X_{t-1}, Y_{t}, Z_{t}; \hat{\xi}_{t})  \notag \\
			& \qquad -\alpha_1\eta^2 (U_{t-1}- \delta^{\hat{\mathcal{G}}_{F}}(X_{t-1}, Y_{t}, Z_{t}))  \notag \\
			& \qquad  -\alpha_1\eta^2(\delta^{\hat{\mathcal{G}}_{F}}(X_{t-1}, Y_{t}, Z_{t})  - \delta^{\hat{\mathcal{G}}_{F}}(X_{t-1}, Y_{t}, Z_{t}; \hat{\xi}_{t}))     \|_F^2]\notag \\
			& \leq  \lambda\frac{1}{K}\mathbb{E}[\| P_{t-1} -  \bar{P}_{t-1} \|_F^2]  + \frac{3}{1-\lambda}\frac{1}{K}\mathbb{E}[\|\delta^{\hat{\mathcal{G}}_{F}}(X_{t}, Y_{t+1}, Z_{t+1}; \hat{\xi}_{t}) - \delta^{\hat{\mathcal{G}}_{F}}(X_{t-1}, Y_{t}, Z_{t}; \hat{\xi}_{t})    \|_F^2]\notag \\
			& \quad +  \frac{3\alpha_1^2\eta^4}{1-\lambda}\frac{1}{K}\mathbb{E}[\| U_{t-1}- \delta^{\hat{\mathcal{G}}_{F}}(X_{t-1}, Y_{t}, Z_{t})   \|_F^2]\notag \\
			& \quad  +  \frac{3\alpha_1^2\eta^4}{1-\lambda}\frac{1}{K}\mathbb{E}[\|\delta^{\hat{\mathcal{G}}_{F}}(X_{t-1}, Y_{t}, Z_{t})  - \delta^{\hat{\mathcal{G}}_{F}}(X_{t-1}, Y_{t}, Z_{t}; \hat{\xi}_{t})   \|_F^2]\notag \\
			& \leq  \lambda\frac{1}{K}\mathbb{E}[\| P_{t-1} -  \bar{P}_{t-1} \|_F^2] +  \frac{3\alpha_1^2\eta^4}{1-\lambda}\frac{1}{K}\mathbb{E}[\| U_{t-1}- \delta^{\hat{\mathcal{G}}_{F}}(X_{t-1}, Y_{t}, Z_{t})   \|_F^2]\notag \\
			& \quad  + (\ell_{f_x}^2+\frac{c_{f_y}^2\ell_{g_{xy}}^2}{\mu^2})\frac{9}{1-\lambda}\frac{1}{K} \mathbb{E}[\|X_{t} - X_{t-1}\|_F^2]+ (\ell_{f_x}^2+\frac{c_{f_y}^2\ell_{g_{xy}}^2}{\mu^2})\frac{9}{1-\lambda}\frac{1}{K} \mathbb{E}[\|Y_{t+1} - Y_{t}\|_F^2]\notag \\
			& \quad + c_{g_{xy}}^2\frac{9}{1-\lambda}\frac{1}{K} \mathbb{E}[\|Z_{t+1} - Z_{t}\|_F^2] +  \frac{6\alpha_1^2\eta^4}{1-\lambda}(1 + \frac{c_{f_y}^2}{\mu^2})\sigma^2 \ . 
	\end{align}

\end{proof}

\begin{lemma} \label{lemma_consensus_r-alt}
				Under Assumptions~\ref{assumption_bi_strong}-\ref{assumption_graph}, we have 
\begin{align}
		& \quad \frac{1}{K}\mathbb{E}[\|R_{t} - \bar{R}_{t}\|_F^2] \notag\\
		&  \leq   \lambda\frac{1}{K}\mathbb{E}[\| R_{t-1} -  \bar{R}_{t-1} \|_F^2] +  \frac{3\alpha_3^2\eta^4}{1-\lambda}\frac{1}{K}\mathbb{E}[\| W_{t-1}- \delta^{\hat{\mathcal{G}}_{h}}(X_{t-1}, Y_{t}, Z_{t-1})   \|_F^2]\notag \\
		& \quad  + (\ell_{f_y}^2+\frac{c_{f_y}^2\ell_{g_{yy}}^2}{\mu^2})\frac{9}{1-\lambda}\frac{1}{K} \mathbb{E}[\|X_{t} - X_{t-1}\|_F^2] \notag \\
		& \quad + (\ell_{f_y}^2+\frac{c_{f_y}^2\ell_{g_{yy}}^2}{\mu^2})\frac{9}{1-\lambda}\frac{1}{K} \mathbb{E}[\|Y_{t+1} - Y_{t}\|_F^2]\notag \\
		& \quad + \ell_{g_y}^2\frac{9}{1-\lambda}\frac{1}{K} \mathbb{E}[\|Z_{t} - Z_{t-1}\|_F^2] +  \frac{6\alpha_3^2\eta^4}{1-\lambda}(1 + \frac{c_{f_y}^2}{\mu^2})\sigma^2  \ . 
\end{align}
\end{lemma}

\begin{proof}
		\begin{align}
			& \quad \frac{1}{K}\mathbb{E}[\|R_{t} - \bar{R}_{t}\|_F^2]  = \frac{1}{K}\mathbb{E}[\| R_{t-1}E +W_{t} - W_{t-1} -  \bar{R}_{t-1}- \bar{W}_{t} +\bar{W}_{t-1}\|_F^2] \notag \\
			&\leq \frac{1}{K}\lambda\mathbb{E}[\| R_{t-1} -  \bar{R}_{t-1} \|_F^2]  + \frac{1}{1-\lambda}\frac{1}{K}\mathbb{E}[\|W_{t} - W_{t-1} - \bar{W}_{t} +\bar{W}_{t-1}\|_F^2]\notag \\
			& \leq \lambda\frac{1}{K}\mathbb{E}[\| R_{t-1} -  \bar{R}_{t-1} \|_F^2]  + \frac{1}{1-\lambda}\frac{1}{K}\mathbb{E}[\|W_{t} - W_{t-1} \|_F^2]\notag \\
			& = \lambda\frac{1}{K}\mathbb{E}[\| R_{t-1} -  \bar{R}_{t-1} \|_F^2]   + \frac{1}{1-\lambda}\frac{1}{K}\mathbb{E}[\|(1-\alpha_3\eta^2)(W_{t-1} - \delta^{\hat{\mathcal{G}}_{h}}(X_{t-1}, Y_{t}, Z_{t-1}; \hat{\xi}_{t}))  \notag \\
			& \quad \quad  + \delta^{\hat{\mathcal{G}}_{h}}(X_{t}, Y_{t+1}, Z_{t}; \hat{\xi}_{t})- W_{t-1} \|_F^2]\notag \\
			& = \lambda\frac{1}{K}\mathbb{E}[\| R_{t-1} -  \bar{R}_{t-1} \|_F^2]  + \frac{1}{1-\lambda}\frac{1}{K}\mathbb{E}[\|\delta^{\hat{\mathcal{G}}_{h}}(X_{t}, Y_{t+1}, Z_{t}; \hat{\xi}_{t}) - \delta^{\hat{\mathcal{G}}_{h}}(X_{t-1}, Y_{t}, Z_{t-1}; \hat{\xi}_{t}) \notag \\
			& \quad  -\alpha_3\eta^2 (W_{t-1}- \delta^{\hat{\mathcal{G}}_{h}}(X_{t-1}, Y_{t}, Z_{t-1}))   -\alpha_3\eta^2(\delta^{\hat{\mathcal{G}}_{h}}(X_{t-1}, Y_{t}, Z_{t-1})  - \delta^{\hat{\mathcal{G}}_{h}}(X_{t-1}, Y_{t}, Z_{t-1}; \hat{\xi}_{t}))     \|_F^2]\notag \\
			& \leq  \lambda\frac{1}{K}\mathbb{E}[\| R_{t-1} -  \bar{R}_{t-1} \|_F^2] + \frac{3}{1-\lambda}\frac{1}{K}\mathbb{E}[\|\delta^{\hat{\mathcal{G}}_{h}}(X_{t}, Y_{t+1}, Z_{t}; \hat{\xi}_{t}) - \delta^{\hat{\mathcal{G}}_{h}}(X_{t-1}, Y_{t}, Z_{t-1}; \hat{\xi}_{t})    \|_F^2]\notag \\
			& \quad +  \frac{3\alpha_3^2\eta^4}{1-\lambda}\frac{1}{K}\mathbb{E}[\| W_{t-1}- \delta^{\hat{\mathcal{G}}_{h}}(X_{t-1}, Y_{t}, Z_{t-1})   \|_F^2] \notag \\
			& \quad  +  \frac{3\alpha_3^2\eta^4}{1-\lambda}\frac{1}{K}\mathbb{E}[\|\delta^{\hat{\mathcal{G}}_{h}}(X_{t-1}, Y_{t}, Z_{t-1})  - \delta^{\hat{\mathcal{G}}_{h}}(X_{t-1}, Y_{t}, Z_{t-1}; \hat{\xi}_{t})   \|_F^2]\notag \\
			& \leq  \lambda\frac{1}{K}\mathbb{E}[\| R_{t-1} -  \bar{R}_{t-1} \|_F^2] +  \frac{3\alpha_3^2\eta^4}{1-\lambda}\frac{1}{K}\mathbb{E}[\| W_{t-1}- \delta^{\hat{\mathcal{G}}_{h}}(X_{t-1}, Y_{t}, Z_{t-1})   \|_F^2]\notag \\
			& \quad  + (\ell_{f_y}^2+\frac{c_{f_y}^2\ell_{g_{yy}}^2}{\mu^2})\frac{9}{1-\lambda}\frac{1}{K} \mathbb{E}[\|X_{t} - X_{t-1}\|_F^2]+ (\ell_{f_y}^2+\frac{c_{f_y}^2\ell_{g_{yy}}^2}{\mu^2})\frac{9}{1-\lambda}\frac{1}{K} \mathbb{E}[\|Y_{t+1} - Y_{t}\|_F^2]\notag \\
			& \quad + \ell_{g_y}^2\frac{9}{1-\lambda}\frac{1}{K} \mathbb{E}[\|Z_{t} - Z_{t-1}\|_F^2] +  \frac{6\alpha_3^2\eta^4}{1-\lambda}(1 + \frac{c_{f_y}^2}{\mu^2})\sigma^2  \ . 
	\end{align}

\end{proof}

\begin{lemma} \label{lemma_consensus_q-alt}
				Under Assumptions~\ref{assumption_bi_strong}-\ref{assumption_graph}, we have 
\begin{align}
		& \quad \frac{1}{K}\mathbb{E}[\|Q_{t} - \bar{Q}_{t}\|_F^2] \notag \\
		&  \leq  \lambda\frac{1}{K}\mathbb{E}[\| Q_{t-1} -  \bar{Q}_{t-1} \|_F^2] +  \frac{3\alpha_2^2\eta^4}{1-\lambda}\frac{1}{K}\mathbb{E}[\| V_{t-1}- \delta^{g}(X_{t-1}, Y_{t-1})   \|_F^2]\notag \\
		& \quad  +\frac{9\ell_{g_y}^2}{1-\lambda}\frac{1}{K} \mathbb{E}[\|X_{t} - X_{t-1}\|_F^2]+ \frac{9\ell_{g_y}^2}{1-\lambda}\frac{1}{K} \mathbb{E}[\|Y_{t} - Y_{t-1}\|_F^2] +  \frac{6\alpha_2^2\eta^4}{1-\lambda}\sigma^2  \ . 
\end{align}
\end{lemma}
This lemma is the same as Lemma~\ref{lemma_consensus_q}.

\begin{lemma} \label{lemma_consensus_x-alt}
				Under Assumptions~\ref{assumption_bi_strong}-\ref{assumption_graph}, we have 
	\begin{align}
			& \quad\mathbb{E}[\|X_{t+1} - \bar{X}_{t+1}\|_F^2] \leq  \Big(1-\frac{\eta(1-\lambda^2)}{2}\Big)\mathbb{E}[\|X_{t} -\bar{X}_{t}\|_F^2]+ \frac{2\eta \beta_1^2}{1-\lambda^2}\mathbb{E}[\|P_{t}- \bar{ P}_{t}\|_F^2]  \ ,\notag \\ 
			& \quad\mathbb{E}[\|Y_{t+1} - \bar{Y}_{t+1}\|_F^2] \leq  \Big(1-\frac{\eta(1-\lambda^2)}{2}\Big)\mathbb{E}[\|Y_{t} -\bar{Y}_{t}\|_F^2]+ \frac{2\eta \beta_2^2}{1-\lambda^2}\mathbb{E}[\|Q_{t}- \bar{ Q}_{t}\|_F^2]  \ ,\notag \\ 
			& \quad\mathbb{E}[\|Z_{t+1} - \bar{Z}_{t+1}\|_F^2] \leq  \Big(1-\frac{\eta(1-\lambda^2)}{2}\Big)\mathbb{E}[\|Z_{t} -\bar{Z}_{t}\|_F^2]+ \frac{2\eta \beta_3^2}{1-\lambda^2}\mathbb{E}[\|R_{t}- \bar{R}_{t}\|_F^2]  \ . 
	\end{align}
\end{lemma}
This lemma is the same as Lemma~\ref{lemma_consensus_x}.

\begin{lemma} \label{lemma_incremental_x-alt}
					Under Assumptions~\ref{assumption_bi_strong}-\ref{assumption_graph}, we have 
	\begin{align}
			& \quad \mathbb{E}[\|X_{t+1}-X_{t}\|_F^2]\leq 8\eta^2\mathbb{E}[\|X_{t}-\bar{X}_{t}\|_F^2] + 4\eta^2\beta_1^2\mathbb{E}[\|P_{t}-\bar{P}_{t}\|_F^2]+4\eta^2\beta_1^2\mathbb{E}[\|\bar{P}_{t}\|_F^2]  \ , \notag \\
			& \quad \mathbb{E}[\|Y_{t+1}-Y_{t}\|_F^2]\leq 8\eta^2\mathbb{E}[\|Y_{t}-\bar{Y}_{t}\|_F^2] + 4\eta^2\beta_2^2\mathbb{E}[\|Q_{t}-\bar{Q}_{t}\|_F^2]+4\eta^2\beta_2^2\mathbb{E}[\|\bar{Q}_{t}\|_F^2]  \ ,\notag \\
			& \quad \mathbb{E}[\|Z_{t+1}-Z_{t}\|_F^2]\leq 8\eta^2\mathbb{E}[\|Z_{t}-\bar{Z}_{t}\|_F^2] + 4\eta^2\beta_3^2\mathbb{E}[\|R_{t}-\bar{R}_{t}\|_F^2]+4\eta^2\beta_3^2\mathbb{E}[\|\bar{R}_{t}\|_F^2]  \ . 
	\end{align}
\end{lemma}
This lemma is the same as Lemma~\ref{lemma_incremental_x}. 

%

\begin{lemma}
		Under Assumptions~\ref{assumption_bi_strong}-\ref{assumption_graph}, we have 
		\begin{align}
				& \quad \mathbb{E}[\|\bar{r}_{t}\|^2] = \mathbb{E}[\|\bar{w}_{t}\|^2]   \leq 3 \mathbb{E}[\|\frac{1}{K}{W}_{t}\mathbf{1} - \frac{1}{K}\delta^{\hat{\mathcal{G}}_{h}}({X}_{t}, {Y}_{t+1},  {Z}_{t})\mathbf{1}\|^2]   \notag \\
				& \quad + 9\ell_{g_y}^2\frac{1}{K}\mathbb{E}[\|{Z}_{t}- \bar{Z}_{t}\|_F^2]  + 9\Big(\frac{c_{f_y}^2\ell_{g_{yy}}^2}{\mu^2} + \ell_{f_y}^2\Big)\frac{1}{K}\mathbb{E}[\|\bar{X}_{t} - {X}_{t}\|_F^2]  \notag \\
				& \quad +  15 \ell_{g_y}^2 \mathbb{E}[\|\bar{z}_{t} - z^*(\bar{x}_{t}) \|^2] + 15\Big(\frac{\ell_{g_{yy}}^2c_{f_y}^2}{\mu^2}+ \ell_{f_y}^2\Big)\mathbb{E}[\|\bar{y}_{t}  - y^*(\bar{x}_{t})\|^2]  \notag \\
				& \quad + 15\beta_2^2\eta^2\Big(\frac{\ell_{g_{yy}}^2c_{f_y}^2}{\mu^2}+ \ell_{f_y}^2\Big)\mathbb{E}[\|\bar{v}_{t} \|^2]   \notag \\
				& \quad +  9\Big(\frac{c_{f_y}^2\ell_{g_{yy}}^2}{\mu^2}+ \ell_{f_y}^2\Big)\frac{1}{K}\mathbb{E}[\|Y_{t} -\bar{Y}_{t}\|_F^2]+ \frac{18\eta \beta_2^2}{1-\lambda^2} \Big(\frac{c_{f_y}^2\ell_{g_{yy}}^2}{\mu^2}+ \ell_{f_y}^2\Big)\frac{1}{K}\mathbb{E}[\|Q_{t}- \bar{ Q}_{t}\|_F^2] \ . 
		\end{align}
\end{lemma}

\begin{proof}
	\begin{align}
			& \quad \mathbb{E}[\|\bar{r}_{t}\|^2] = \mathbb{E}[\|\bar{w}_{t}\|^2] \notag \\
			& = \mathbb{E}[\|\frac{1}{K}{W}_{t}\mathbf{1} - \frac{1}{K}\delta^{\hat{\mathcal{G}}_{h}}({X}_{t}, {Y}_{t+1},  {Z}_{t})\mathbf{1}+  \frac{1}{K}\delta^{\hat{\mathcal{G}}_{h}}({X}_{t}, {Y}_{t+1},  {Z}_{t})\mathbf{1}  \notag \\
			& \quad -  \frac{1}{K}\delta^{\hat{\mathcal{G}}_{h}}(\bar{X}_{t}, \bar{Y}_{t+1},  \bar{Z}_{t})\mathbf{1} + \frac{1}{K}\delta^{\hat{\mathcal{G}}_{h}}(\bar{X}_{t}, \bar{Y}_{t+1},  \bar{Z}_{t})\mathbf{1}\|^2] \notag \\
			& \leq 3 \mathbb{E}[\|\frac{1}{K}{W}_{t}\mathbf{1} - \frac{1}{K}\delta^{\hat{\mathcal{G}}_{h}}({X}_{t}, {Y}_{t+1},  {Z}_{t})\mathbf{1}\|^2]  \notag \\
			& \quad+ 3 \mathbb{E}[\| \frac{1}{K}\delta^{\hat{\mathcal{G}}_{h}}({X}_{t}, {Y}_{t+1},  {Z}_{t})\mathbf{1} -  \frac{1}{K}\delta^{\hat{\mathcal{G}}_{h}}(\bar{X}_{t}, \bar{Y}_{t+1},  \bar{Z}_{t})\mathbf{1}\|^2] \notag \\
			& \quad + 3 \mathbb{E}[\|\frac{1}{K}\delta^{\hat{\mathcal{G}}_{h}}(\bar{X}_{t}, \bar{Y}_{t+1},  \bar{Z}_{t})\mathbf{1}\|^2] \notag \\
			& \leq 3 \mathbb{E}[\|\frac{1}{K}{W}_{t}\mathbf{1} - \frac{1}{K}\delta^{\hat{\mathcal{G}}_{h}}({X}_{t}, {Y}_{t+1},  {Z}_{t})\mathbf{1}\|^2]   \notag \\
			& \quad + 9\ell_{g_y}^2\frac{1}{K}\mathbb{E}[\|{Z}_{t}- \bar{Z}_{t}\|_F^2]  + 9\Big(\frac{c_{f_y}^2\ell_{g_{yy}}^2}{\mu^2} + \ell_{f_y}^2\Big)\frac{1}{K}\mathbb{E}[\|\bar{X}_{t} - {X}_{t}\|_F^2]  \notag \\
			& \quad+  9\Big(\frac{c_{f_y}^2\ell_{g_{yy}}^2}{\mu^2}+ \ell_{f_y}^2\Big)\frac{1}{K}\mathbb{E}[\| \bar{Y}_{t+1} - {Y}_{t+1}\|_F^2]  \notag \\
			& \quad +  15 \ell_{g_y}^2 \mathbb{E}[\|\bar{z}_{t} - z^*(\bar{x}_{t}) \|^2] + 15\Big(\frac{\ell_{g_{yy}}^2c_{f_y}^2}{\mu^2}+ \ell_{f_y}^2\Big)\mathbb{E}[\|\bar{y}_{t}  - y^*(\bar{x}_{t})\|^2]   \notag \\
			& \quad+ 15\beta_2^2\eta^2\Big(\frac{\ell_{g_{yy}}^2c_{f_y}^2}{\mu^2}+ \ell_{f_y}^2\Big)\mathbb{E}[\|\bar{v}_{t} \|^2]  \notag \\
			& \leq 3 \mathbb{E}[\|\frac{1}{K}{W}_{t}\mathbf{1} - \frac{1}{K}\delta^{\hat{\mathcal{G}}_{h}}({X}_{t}, {Y}_{t+1},  {Z}_{t})\mathbf{1}\|^2]   \notag \\
			& \quad + 9\ell_{g_y}^2\frac{1}{K}\mathbb{E}[\|{Z}_{t}- \bar{Z}_{t}\|_F^2]  + 9\Big(\frac{c_{f_y}^2\ell_{g_{yy}}^2}{\mu^2} + \ell_{f_y}^2\Big)\frac{1}{K}\mathbb{E}[\|\bar{X}_{t} - {X}_{t}\|_F^2]  \notag \\
			& \quad +  15 \ell_{g_y}^2 \mathbb{E}[\|\bar{z}_{t} - z^*(\bar{x}_{t}) \|^2] + 15\Big(\frac{\ell_{g_{yy}}^2c_{f_y}^2}{\mu^2}+ \ell_{f_y}^2\Big)\mathbb{E}[\|\bar{y}_{t}  - y^*(\bar{x}_{t})\|^2] \notag \\
			& \quad  + 15\beta_2^2\eta^2\Big(\frac{\ell_{g_{yy}}^2c_{f_y}^2}{\mu^2}+ \ell_{f_y}^2\Big)\mathbb{E}[\|\bar{v}_{t} \|^2]   \notag \\
			& \quad +  9\Big(\frac{c_{f_y}^2\ell_{g_{yy}}^2}{\mu^2}+ \ell_{f_y}^2\Big)\frac{1}{K}\mathbb{E}[\|Y_{t} -\bar{Y}_{t}\|_F^2]+ \frac{18\eta \beta_2^2}{1-\lambda^2} \Big(\frac{c_{f_y}^2\ell_{g_{yy}}^2}{\mu^2}+ \ell_{f_y}^2\Big)\frac{1}{K}\mathbb{E}[\|Q_{t}- \bar{ Q}_{t}\|_F^2] \ , 
	\end{align}
	where the second to last step follows from Eq.~(\ref{eq:h-grad-consensus-alt}) and the following inequality, the last step follows from Lemma~\ref{lemma_incremental_x-alt}.
	\begin{align}
			& \quad \mathbb{E}[\|\frac{1}{K}\delta^{\hat{\mathcal{G}}_{h}}(\bar{X}_{t}, \bar{Y}_{t+1},  \bar{Z}_{t})\mathbf{1}\|^2] = \mathbb{E}[\|\hat{\mathcal{G}}_{h}(\bar{x}_{t}, \bar{y}_{t+1}, \bar{z}_{t})\|^2] \notag \\
			&  = \mathbb{E}[\|\nabla_{22}^2g(\bar{x}_{t}, \bar{y}_{t+1}) \bar{z}_{t} -  \nabla_2 f(\bar{x}_{t}, \bar{y}_{t+1})  \|^2] \notag \\
			& = \mathbb{E}[\|\nabla_{22}^2g(\bar{x}_{t}, \bar{y}_{t+1}) \bar{z}_{t} -  \nabla_2 f(\bar{x}_{t}, \bar{y}_{t+1}) -\nabla_{22}^2g(\bar{x}_{t}, y^*(\bar{x}_{t})) z^*(\bar{x}_{t}) +   \nabla_2 f(\bar{x}_{t}, y^*(\bar{x}_{t}))  \|^2] \notag \\
			& \leq 5\mathbb{E}[\|\nabla_{22}^2g(\bar{x}_{t}, \bar{y}_{t+1}) \bar{z}_{t} - \nabla_{22}^2g(\bar{x}_{t}, \bar{y}_{t+1}) z^*(\bar{x}_{t}) \|^2]\notag \\
			& \quad + 5\mathbb{E}[\|\nabla_{22}^2g(\bar{x}_{t}, \bar{y}_{t+1}) z^*(\bar{x}_{t}) -\nabla_{22}^2g(\bar{x}_{t},\bar{y}_{t}) z^*(\bar{x}_{t})\|^2] \notag \\
			& \quad + 5\mathbb{E}[\|\nabla_{22}^2g(\bar{x}_{t}, \bar{y}_{t}) z^*(\bar{x}_{t}) -\nabla_{22}^2g(\bar{x}_{t}, y^*(\bar{x}_{t})) z^*(\bar{x}_{t})\|^2] \notag \\
			& \quad + 5\mathbb{E}[\|-  \nabla_2 f(\bar{x}_{t}, \bar{y}_{t+1})  +  \nabla_2 f(\bar{x}_{t}, \bar{y}_{t})  \|^2]  + 5\mathbb{E}[\|-  \nabla_2 f(\bar{x}_{t}, \bar{y}_{t})  +  \nabla_2 f(\bar{x}_{t}, y^*(\bar{x}_{t}))  \|^2] \notag \\
			& \leq 5 \ell_{g_y}^2 \mathbb{E}[\|\bar{z}_{t} - z^*(\bar{x}_{t}) \|^2] + 5\Big(\frac{\ell_{g_{yy}}^2c_{f_y}^2}{\mu^2}+ \ell_{f_y}^2\Big)\mathbb{E}[\|\bar{y}_{t}  - y^*(\bar{x}_{t})\|^2]   \notag \\
			& \quad+ 5\beta_2^2\eta^2\Big(\frac{\ell_{g_{yy}}^2c_{f_y}^2}{\mu^2}+ \ell_{f_y}^2\Big)\mathbb{E}[\|\bar{v}_{t} \|^2]  \ . 
	\end{align}
	
\end{proof}


\begin{lemma} \label{lemma:y-incremental-2-alt}
			Under Assumptions~\ref{assumption_bi_strong}-\ref{assumption_graph}, we have 
	\begin{align}
			& \quad \frac{1}{K}\mathbb{E}[\|Y_{t+2}-Y_{t+1}\|_F^2]  \leq 	 8\eta^2\frac{1}{K}\mathbb{E}[\|Y_{t} -\bar{Y}_{t}\|_F^2]+\frac{20\eta^2 \beta_2^2}{1-\lambda^2}\frac{1}{K}\mathbb{E}[\|Q_{t}- \bar{ Q}_{t}\|_F^2]     \notag \\
			& \quad + \frac{36\alpha^2_2\beta_2^2\eta^6}{1-\lambda}\frac{1}{K}\mathbb{E}[\| V_{t}- \delta^{g}(X_{t}, Y_{t})   \|_F^2] +\frac{60 \beta_2^2\eta^2\ell_{g_y}^2}{1-\lambda}\frac{1}{K} \mathbb{E}[\|X_{t+1} - X_{t}\|_F^2] \notag \\
			& \quad + \frac{60 \beta_2^2\eta^2\ell_{g_y}^2}{1-\lambda}\frac{1}{K} \mathbb{E}[\|Y_{t+1} - Y_{t}\|_F^2] + 8\eta^2\beta_2^2\frac{1}{K}\mathbb{E}[\|\bar{V}_{t}\|_F^2] +  \frac{48\alpha^2_2\beta_2^2\eta^6}{1-\lambda}\sigma^2 \ . 
	\end{align}
\end{lemma}

\begin{proof}
	
	In terms of our algorithm, we can obtain
	\begin{align}
			&   \quad \mathbb{E}[\|\bar{v}_{t+1} - \bar{v}_{t}\|^2]   \notag \\
			& = \mathbb{E}[\|((1-\alpha_2\eta^2)(V_{t} - \delta_{t}^{g}(X_{t}, Y_{t}; {\zeta}_{t+1}))+\delta_{t}^{g}(X_{t+1}, Y_{t+1}; {\zeta}_{t+1})- V_{t})\frac{1}{K} \mathbf{1}	\|^2]   \notag \\
			& = \mathbb{E}[\|(-\alpha_2\eta^2V_{t} + \alpha_2\eta^2 \delta_{t}^{g}(X_{t}, Y_{t}; {\zeta}_{t+1}) -  \delta_{t}^{g}(X_{t}, Y_{t}; {\zeta}_{t+1}) + \delta_{t}^{g}(X_{t+1}, Y_{t+1}; {\zeta}_{t+1}))\frac{1}{K} \mathbf{1}	\|^2]   \notag \\
			& = \mathbb{E}[\|(-\alpha_2\eta^2(V_{t} -  \delta_{t}^{g}(X_{t}, Y_{t}) ) -  \alpha_2\eta^2 \delta_{t}^{g}(X_{t}, Y_{t}) + \alpha_2\eta^2 \delta_{t}^{g}(X_{t}, Y_{t}; {\zeta}_{t+1}) \notag \\
			& \quad -  \delta_{t}^{g}(X_{t}, Y_{t}; {\zeta}_{t+1}) + \delta_{t}^{g}(X_{t+1}, Y_{t+1}; {\zeta}_{t+1}))\frac{1}{K} \mathbf{1}	\|^2]   \notag \\
			& \leq 3\alpha^2_2\eta^4\frac{1}{K} \mathbb{E}[\|V_{t}- \delta_{t}^{g}(X_{t}, Y_{t})	\|_F^2]  \notag \\
			& \quad+3\ell_{g_y}^2 \frac{1}{K} \mathbb{E}[\| X_{t+1}-  X_{t}\|_F^2] +3\ell_{g_y}^2 \frac{1}{K} \mathbb{E}[\| Y_{t+1}-   Y_{t}\|_F^2]   + 3\alpha_2^2\eta^4\sigma^2 \ . 
	\end{align}
	
	
	Consequently, we have
	\begin{align}
			& \quad \frac{1}{K}\mathbb{E}[\|Y_{t+2}-Y_{t+1}\|_F^2]\notag \\
			& \leq 8\eta^2\frac{1}{K}\mathbb{E}[\|Y_{t+1}-\bar{Y}_{t+1}\|_F^2] + 4\eta^2\beta_2^2\frac{1}{K}\mathbb{E}[\|Q_{t+1}-\bar{Q}_{t+1}\|_F^2]+4\eta^2\beta_2^2\frac{1}{K}\mathbb{E}[\|\bar{Q}_{t+1}\|_F^2]  \notag \\
			& \leq 8\eta^2\frac{1}{K}\mathbb{E}[\|Y_{t+1}-\bar{Y}_{t+1}\|_F^2] + 4\eta^2\beta_2^2\frac{1}{K}\mathbb{E}[\|Q_{t+1}-\bar{Q}_{t+1}\|_F^2] \notag \\
			& \quad +8\eta^2\beta_2^2\frac{1}{K}\mathbb{E}[\|\bar{V}_{t+1} - \bar{V}_{t}\|_F^2]   + 8\eta^2\beta_2^2\frac{1}{K}\mathbb{E}[\|\bar{V}_{t}\|_F^2]  \notag \\
			& \leq 8\eta^2\frac{1}{K}\mathbb{E}[\|Y_{t} -\bar{Y}_{t}\|_F^2]+ \frac{16\eta^3 \beta_2^2}{1-\lambda^2}\frac{1}{K}\mathbb{E}[\|Q_{t}- \bar{ Q}_{t}\|_F^2] \notag \\
			& \quad + 4\lambda\eta^2\beta_2^2\frac{1}{K}\mathbb{E}[\| Q_{t} -  \bar{Q}_{t} \|_F^2] +  4\eta^2\beta_2^2\frac{3\alpha_2^2\eta^4}{1-\lambda}\frac{1}{K}\mathbb{E}[\| V_{t}- \delta^{g}(X_{t}, Y_{t})   \|_F^2]\notag \\
			& \quad  +4\eta^2\beta_2^2\frac{9\ell_{g_y}^2}{1-\lambda}\frac{1}{K} \mathbb{E}[\|X_{t+1} - X_{t}\|_F^2]+ 4\eta^2\beta_2^2\frac{9\ell_{g_y}^2}{1-\lambda}\frac{1}{K} \mathbb{E}[\|Y_{t+1} - Y_{t}\|_F^2] +  4\eta^2\beta_2^2\frac{6\alpha_2^2\eta^4}{1-\lambda}\sigma^2  \notag \\
			& \quad +24\alpha^2_2\beta_2^2\eta^6\frac{1}{K} \mathbb{E}[\|V_{t}- \delta_{t}^{g}(X_{t}, Y_{t})	\|_F^2] +24 \beta_2^2\eta^2\ell_{g_y}^2\frac{1}{K}\mathbb{E}[\| X_{t+1}-  X_{t}\|_F^2]  \notag \\
			& \quad+24 \beta_2^2\eta^2\ell_{g_y}^2\frac{1}{K}\mathbb{E}[\| Y_{t+1}-   Y_{t}\|_F^2]    + 8\eta^2\beta_2^2\frac{1}{K}\mathbb{E}[\|\bar{V}_{t}\|_F^2]  +  24\alpha_2^2\beta_2^2\eta^6\sigma^2  \notag \\
			& \leq 8\eta^2\frac{1}{K}\mathbb{E}[\|Y_{t} -\bar{Y}_{t}\|_F^2]+\frac{20\eta^2 \beta_2^2}{1-\lambda^2}\frac{1}{K}\mathbb{E}[\|Q_{t}- \bar{ Q}_{t}\|_F^2] \notag \\
			& \quad + \frac{36\alpha^2_2\beta_2^2\eta^6}{1-\lambda}\frac{1}{K}\mathbb{E}[\| V_{t}- \delta^{g}(X_{t}, Y_{t})   \|_F^2]\notag \\
			& \quad  +\frac{60 \beta_2^2\eta^2\ell_{g_y}^2}{1-\lambda}\frac{1}{K} \mathbb{E}[\|X_{t+1} - X_{t}\|_F^2]+ \frac{60 \beta_2^2\eta^2\ell_{g_y}^2}{1-\lambda}\frac{1}{K} \mathbb{E}[\|Y_{t+1} - Y_{t}\|_F^2] \notag \\
			& \quad    + 8\eta^2\beta_2^2\frac{1}{K}\mathbb{E}[\|\bar{V}_{t}\|_F^2] +  \frac{48\alpha^2_2\beta_2^2\eta^6}{1-\lambda}\sigma^2   \ , 
	\end{align}
	where the third step follows from Lemma~\ref{lemma_consensus_x-alt} and Lemma~\ref{lemma_consensus_q-alt}, the last step follows from $\eta<1$. 
\end{proof}

\begin{lemma} \label{lemma:z-incremental-2-alt}
				Under Assumptions~\ref{assumption_bi_strong}-\ref{assumption_graph}, we have 
	\begin{align}
			& \quad \frac{1}{K}\mathbb{E}[\|Z_{t+2}-Z_{t+1}\|_F^2] \notag \\
			& \leq  8\eta^2\frac{1}{K}\mathbb{E}[\|Z_{t} -\bar{Z}_{t}\|_F^2]+  \frac{20\eta^3 \beta_3^2}{1-\lambda^2}\frac{1}{K}\mathbb{E}[\|R_{t}- \bar{R}_{t}\|_F^2]  +8\eta^2\beta_3^2\frac{1}{K}\mathbb{E}[\| \bar{R}_{t}\|_F^2]  \notag \\
			& \quad+  \frac{72\alpha^2_3\beta_3^2\eta^6}{1-\lambda}(1 + \frac{c_{f_y}^2}{\mu^2})\sigma^2 +  \frac{36\beta_3^2\alpha^2_3\eta^6}{1-\lambda} \frac{1}{K}\mathbb{E}[\| W_{t}- \delta^{\hat{\mathcal{G}}_{h}}(X_{t}, Y_{t+1}, Z_{t})   \|_F^2]  \notag \\
			& \quad+ \frac{108\eta^2\beta_3^2}{1-\lambda}(\ell_{f_y}^2+\frac{c_{f_y}^2\ell_{g_{yy}}^2}{\mu^2})\frac{1}{K} \mathbb{E}[\|X_{t+1} - X_{t}\|_F^2] \notag \\
			& \quad+  \frac{108\eta^2\beta_3^2}{1-\lambda}(\ell_{f_y}^2+\frac{c_{f_y}^2\ell_{g_{yy}}^2}{\mu^2})\frac{1}{K} \mathbb{E}[\|Y_{t+2} - Y_{t+1}\|_F^2]+\frac{ 108\eta^2\beta_3^2\ell_{g_y}^2 }{1-\lambda} \frac{1}{K} \mathbb{E}[\|Z_{t+1} - Z_{t}\|_F^2]    \ . 
	\end{align}
\end{lemma}

\begin{proof}
	Similarly, we can obtain
	\begin{align}
			&   \quad \mathbb{E}[\|\bar{w}_{t+1} - \bar{w}_{t}\|^2]   
			 = \mathbb{E}[\|(-\alpha_3\eta^2(W_{t} - \delta_{t}^{\hat{\mathcal{G}}_{h}}(X_{t}, Y_{t+1}, Z_{t})) -  \alpha_3\eta^2\delta_{t}^{\hat{\mathcal{G}}_{h}}(X_{t}, Y_{t+1}, Z_{t})\notag \\
			& \quad + \alpha_3\eta^2\delta_{t}^{\hat{\mathcal{G}}_{h}}(X_{t}, Y_{t+1}, Z_{t}; \hat{\xi}_{t+1})  - \delta_{t}^{\hat{\mathcal{G}}_{h}}(X_{t}, Y_{t+1}, Z_{t}; \hat{\xi}_{t+1}) + \delta_{t}^{\hat{\mathcal{G}}_{h}}(X_{t+1}, Y_{t+2}, Z_{t+1}; \hat{\xi}_{t+1}))\frac{1}{K} \mathbf{1}	\|^2]   \notag \\
			& \leq 3\alpha^2_3\eta^4\frac{1}{K} \mathbb{E}[\|W_{t}- \delta_{t}^{\hat{\mathcal{G}}_{h}}(X_{t}, Y_{t+1}, Z_{t})	\|_F^2] + 6\alpha^2_3\eta^4(1 + \frac{c_{f_y}^2}{\mu^2})\sigma^2K+ 9\ell_{g_y}^2 \frac{1}{K}\mathbb{E}[\|Z_{t+1} - Z_{t}\|_F^2]  \notag \\
			& \quad +9(\frac{c_{f_y}^2\ell_{g_{yy}}^2}{\mu^2}+\ell_{f_y}^2)\frac{1}{K}\mathbb{E}[\|X_{t+1} - X_{t}\|_F^2]  + 9(\frac{c_{f_y}^2\ell_{g_{yy}}^2}{\mu^2}+\ell_{f_y}^2)\frac{1}{K}\mathbb{E}[\|Y_{t+2} - Y_{t+1}\|_F^2]  \ . 
	\end{align}
	Then, we have
	\begin{align}
 \small
			& \quad \frac{1}{K}\mathbb{E}[\|Z_{t+2}-Z_{t+1}\|_F^2]  \notag \\
			& \leq 8\eta^2\frac{1}{K}\mathbb{E}[\|Z_{t+1}-\bar{Z}_{t+1}\|_F^2] + 4\eta^2\beta_3^2\frac{1}{K}\mathbb{E}[\|R_{t+1}-\bar{R}_{t+1}\|_F^2]+4\eta^2\beta_3^2\frac{1}{K}\mathbb{E}[\|\bar{R}_{t+1}\|_F^2]   \notag \\
			& \leq 8\eta^2\frac{1}{K}\mathbb{E}[\|Z_{t+1}-\bar{Z}_{t+1}\|_F^2] + 4\eta^2\beta_3^2\frac{1}{K}\mathbb{E}[\|R_{t+1}-\bar{R}_{t+1}\|_F^2] \notag \\
			& \quad+8\eta^2\beta_3^2\frac{1}{K}\mathbb{E}[\|\bar{R}_{t+1}- \bar{R}_{t}\|_F^2] +8\eta^2\beta_3^2\frac{1}{K}\mathbb{E}[\| \bar{R}_{t}\|_F^2]   \notag \\
			& \leq 8\eta^2\frac{1}{K}\mathbb{E}[\|Z_{t} -\bar{Z}_{t}\|_F^2]+ \frac{16\eta^3 \beta_3^2}{1-\lambda^2}\frac{1}{K}\mathbb{E}[\|R_{t}- \bar{R}_{t}\|_F^2]  \notag \\
			& \quad + 4 \lambda \eta^2\beta_3^2\frac{1}{K}\mathbb{E}[\| R_{t} -  \bar{R}_{t} \|_F^2] +  4\eta^2\beta_3^2\frac{3\alpha_3^2\eta^4}{1-\lambda}\frac{1}{K}\mathbb{E}[\| W_{t}- \delta^{\hat{\mathcal{G}}_{h}}(X_{t}, Y_{t+1}, Z_{t})   \|_F^2]\notag \\
			& \quad  + 4\eta^2\beta_3^2(\ell_{f_y}^2+\frac{c_{f_y}^2\ell_{g_{yy}}^2}{\mu^2})\frac{9}{1-\lambda}\frac{1}{K} \mathbb{E}[\|X_{t+1} - X_{t}\|_F^2] \notag \\
			& \quad+ 4\eta^2\beta_3^2(\ell_{f_y}^2+\frac{c_{f_y}^2\ell_{g_{yy}}^2}{\mu^2})\frac{9}{1-\lambda}\frac{1}{K} \mathbb{E}[\|Y_{t+2} - Y_{t+1}\|_F^2]\notag \\
			& \quad +4\eta^2\beta_3^2 \ell_{g_y}^2\frac{9}{1-\lambda}\frac{1}{K} \mathbb{E}[\|Z_{t+1} - Z_{t}\|_F^2] +  4\eta^2\beta_3^2\frac{6\alpha_3^2\eta^4}{1-\lambda}(1 + \frac{c_{f_y}^2}{\mu^2})\sigma^2   \notag \\
			& \quad + 24\beta_3^2\alpha^2_3\eta^6\frac{1}{K} \mathbb{E}[\|W_{t}- \delta_{t}^{\hat{\mathcal{G}}_{h}}(X_{t}, Y_{t+1}, Z_{t})	\|_F^2] + 48\alpha^2_3\beta_3^2\eta^6(1 + \frac{c_{f_y}^2}{\mu^2})\sigma^2  \notag \\
			& \quad +72\eta^2\beta_3^2(\frac{c_{f_y}^2\ell_{g_{yy}}^2}{\mu^2}+\ell_{f_y}^2)\frac{1}{K}\mathbb{E}[\|X_{t+1} - X_{t}\|_F^2]   \notag \\
			& \quad+ 72\eta^2\beta_3^2(\frac{c_{f_y}^2\ell_{g_{yy}}^2}{\mu^2}+\ell_{f_y}^2)\frac{1}{K}\mathbb{E}[\|Y_{t+2} - Y_{t+1}\|_F^2] \notag \\
			& \quad + 72\eta^2\beta_3^2\ell_{g_y}^2 \frac{1}{K}\mathbb{E}[\|Z_{t+1} - Z_{t}\|_F^2]  +8\eta^2\beta_3^2\frac{1}{K}\mathbb{E}[\| \bar{R}_{t}\|_F^2]   \notag \\
			& \leq 8\eta^2\frac{1}{K}\mathbb{E}[\|Z_{t} -\bar{Z}_{t}\|_F^2]+  \frac{20\eta^3 \beta_3^2}{1-\lambda^2}\frac{1}{K}\mathbb{E}[\|R_{t}- \bar{R}_{t}\|_F^2]  +8\eta^2\beta_3^2\frac{1}{K}\mathbb{E}[\| \bar{R}_{t}\|_F^2] \notag \\
			& \quad+  \frac{72\alpha^2_3\beta_3^2\eta^6}{1-\lambda}(1 + \frac{c_{f_y}^2}{\mu^2})\sigma^2 +  \frac{36\beta_3^2\alpha^2_3\eta^6}{1-\lambda} \frac{1}{K}\mathbb{E}[\| W_{t}- \delta^{\hat{\mathcal{G}}_{h}}(X_{t}, Y_{t+1}, Z_{t})   \|_F^2]   \notag \\
			& \quad+ \frac{108\eta^2\beta_3^2}{1-\lambda}(\ell_{f_y}^2+\frac{c_{f_y}^2\ell_{g_{yy}}^2}{\mu^2})\frac{1}{K} \mathbb{E}[\|X_{t+1} - X_{t}\|_F^2] \notag \\
			& \quad+  \frac{108\eta^2\beta_3^2}{1-\lambda}(\ell_{f_y}^2+\frac{c_{f_y}^2\ell_{g_{yy}}^2}{\mu^2})\frac{1}{K} \mathbb{E}[\|Y_{t+2} - Y_{t+1}\|_F^2] +\frac{ 108\eta^2\beta_3^2\ell_{g_y}^2 }{1-\lambda} \frac{1}{K} \mathbb{E}[\|Z_{t+1} - Z_{t}\|_F^2]  \ . 
	\end{align}

\end{proof}
Based on these lemmas, we are ready to prove Theorem~\ref{theorem-alt}.
\begin{proof}
	Based on aforementioned lemmas, we have
	\begin{align}
		\small
			&   \mathcal{L}_{t+1} -  \mathcal{L}_{t}  \leq  - \frac{\beta_1\eta}{2}\mathbb{E}[\| \nabla F(\bar{x}_{t})\|^2]  +\Bigg(c_{0}\frac{25\eta\beta_1^2L_{y}^2 }{6\beta_2\mu}+  c_{1}\frac{9\eta\beta_1^2L_z^2}{\beta_3\mu} +4\eta^2\beta_1^2D_1-  \frac{\beta_1\eta}{4}\Bigg) \mathbb{E}[\|\bar{u}_{t}\| ^2 ] \notag \\
			& \quad    +\Bigg(9\beta_1\beta_2^2\eta^3(\ell_{f_x}^2+\frac{c_{f_y}^2\ell_{g_{xy}}^2}{\mu^2} ) + c_{1} \frac{18\eta^3\beta_2^2\beta_3}{\mu}\Big(\frac{c_{f_y}^2\ell_{g_{yy}}^2}{\mu^2}+\ell_{f_y}^2\Big) \notag \\
			& \qquad + 8\eta^2\beta_2^2\left(D_y+ D_z \frac{108\eta^2\beta_3^2}{1-\lambda}(\ell_{f_y}^2+\frac{c_{f_y}^2\ell_{g_{yy}}^2}{\mu^2})\right)\notag \\
			& \quad \quad +4\eta^2\beta_2^2D_2+ 15D_4 \beta_2^2\eta^2\Big(\frac{\ell_{g_{yy}}^2c_{f_y}^2}{\mu^2}+ \ell_{f_y}^2\Big)-c_{0} \frac{3\eta\beta_2^2}{4}\Bigg) \mathbb{E}[\|\bar{v}_{t}  \|^2]  \notag \\
			& \quad + \Bigg(9\beta_1\eta(\ell_{f_x}^2+\frac{c_{f_y}^2\ell_{g_{xy}}^2}{\mu^2})+c_{1} \frac{18\eta\beta_3}{\mu}\Big(\frac{c_{f_y}^2\ell_{g_{yy}}^2}{\mu^2}+\ell_{f_y}^2\Big) \notag \\
			& \qquad+ 15D_4 \Big(\frac{\ell_{g_{yy}}^2c_{f_y}^2}{\mu^2}+ \ell_{f_y}^2\Big)-c_{0}\frac{\beta_2\eta\mu}{4} \Bigg)\mathbb{E}[\| y^*(\bar{x}_{t}) - \bar{y}_{t}\|^2] \notag \\
			& \quad + \Big(9 \beta_1\eta c_{g_{xy}}^2 +  15 D_4 \ell_{g_y}^2-c_{1}\frac{\eta\beta_3\mu}{8}\Big) \mathbb{E}[\|z^*(\bar{x}_{t}) - \bar{z}_{t} \|^2]    \notag \\
			& \quad  +  \Bigg(9\beta_1\eta(\ell_{f_x}^2+\frac{c_{f_y}^2\ell_{g_{xy}}^2}{\mu^2})+  c_{0}\frac{25\beta_2 \eta \ell_{g_y}^2}{3\mu}+c_{1}\frac{70\eta\beta_3}{\mu}\Big(\frac{c_{f_y}^2\ell_{g_{yy}}^2}{\mu^2} + \ell_{f_y}^2\Big)+  8\eta^2D_1 \notag \\
			& \quad \quad   + 9D_4 \Big(\frac{c_{f_y}^2\ell_{g_{yy}}^2}{\mu^2} + \ell_{f_y}^2\Big) -c_{2}\frac{\eta(1-\lambda^2)}{2} \Bigg)\frac{1}{K}\mathbb{E}[\|{X}_{t} - \bar{X}_{t}\|_F^2] \notag \\
			& \quad + \Bigg(c_{0} \frac{25\beta_2 \eta \ell_{g_y}^2}{3\mu}+ 8\eta^2D_2+8\eta^2\left(D_y+ D_z \frac{108\eta^2\beta_3^2}{1-\lambda}(\ell_{f_y}^2+\frac{c_{f_y}^2\ell_{g_{yy}}^2}{\mu^2})\right)+D_5 \notag \\
			& \quad \quad +  9D_4 \Big(\frac{c_{f_y}^2\ell_{g_{yy}}^2}{\mu^2}+ \ell_{f_y}^2\Big)-c_{3}\frac{\eta(1-\lambda^2)}{2}\Bigg)  \frac{1}{K}\mathbb{E}[\|\bar{Y}_{t} -  {Y}_{t}\|_F^2]\notag \\
			& \quad +\Bigg(c_{1}\Big(\frac{70\eta\beta_3}{\mu}\ell_{g_y}^2+  \frac{9}{4}\Big)+8\eta^2D_3+ 8D_z\eta^2+9\beta_1\eta c_{g_{xy}}^2+ 9D_4 \ell_{g_y}^2-c_{4}\frac{\eta(1-\lambda^2)}{2}\Bigg)\frac{1}{K}\mathbb{E} [ \|{Z}_{t}- \bar{Z}_{t}\|_F^2] \notag \\
			& \quad  +\Bigg( c_{3}\frac{2\eta \beta_2^2}{1-\lambda^2}+ 4\eta^2\beta_2^2D_2+\frac{20\eta^2 \beta_2^2}{1-\lambda^2}\left(D_y+ D_z \frac{108\eta^2\beta_3^2}{1-\lambda}(\ell_{f_y}^2+\frac{c_{f_y}^2\ell_{g_{yy}}^2}{\mu^2})\right) \notag \\
			& \quad \quad + D_5\frac{2\eta \beta_2^2}{1-\lambda^2} +D_4  \frac{18\eta \beta_2^2}{1-\lambda^2} \Big(\frac{c_{f_y}^2\ell_{g_{yy}}^2}{\mu^2}+ \ell_{f_y}^2\Big)- c_{6}(1-\lambda)\Bigg)\frac{1}{K}\mathbb{E}[\|Q_{t}- \bar{ Q}_{t}\|_F^2]  \ ,\notag \\ 
			& \quad + \Bigg(c_{4}\frac{2\eta \beta_3^2}{1-\lambda^2}+c_{1}\frac{9\eta\beta_3^2}{4}+ 4\eta^2\beta_3^2D_3+ D_z \frac{20\eta^3 \beta_3^2}{1-\lambda^2}+ 9\beta_1\eta c_{g_{xy}}^2\frac{2\eta \beta_3^2}{1-\lambda^2}- c_{7}(1-\lambda)\Bigg)\frac{1}{K}\mathbb{E}[\|R_{t}- \bar{R}_{t}\|_F^2]  \notag \\
			& \quad  + \Bigg(c_{2}\frac{2\eta \beta_1^2}{1-\lambda^2} + 4\eta^2\beta_1^2D_1-c_{5}(1-\lambda)\Bigg)\frac{1}{K}\mathbb{E}[\| P_{t} -  \bar{P}_{t} \|_F^2] \notag \\
			& \quad +\Bigg(\beta_1\eta-c_{8}\alpha_1 \eta^2\Bigg)\mathbb{E}[\|\frac{1}{K}\delta^{\hat{\mathcal{G}}_{F}} (X_{t}, Y_{t+1}, Z_{t+1}) \mathbf{1} - \frac{1}{K}{U}_{t} \mathbf{1}\|^2]  \notag \\
			& \quad + \Big(c_{0} \frac{25\beta_2 \eta}{3\mu} - c_{10}\alpha_2 \eta^2 \Big)\mathbb{E}[\|\frac{1}{K} \delta^{g}  ({X}_{t}, {Y}_{t}) \mathbf{1} - \frac{1}{K}{V}_{t}\mathbf{1} \|^2]   \notag \\
			& \quad +\Bigg(c_{1}\frac{25\eta\beta_3}{\mu}+3 D_4-c_{12}\alpha_3 \eta^2\Bigg) \mathbb{E} [ \| \frac{1}{K}\delta^{\hat{\mathcal{G}}_{h}}({X}_{t}, {Y}_{t+1},  {Z}_{t})\mathbf{1}-   \frac{1}{K}{W}_{t}\mathbf{1}\|^2] \notag \\
			& \quad +  \Bigg(c_{5}\frac{3\alpha_1^2\eta^4}{1-\lambda}-c_{9}\alpha_1 \eta^2\Bigg)\frac{1}{K}\mathbb{E}[\| U_{t}- \delta^{\hat{\mathcal{G}}_{F}}(X_{t}, Y_{t+1}, Z_{t+1})   \|_F^2]\notag \\
			&\quad 	+ \Bigg(c_{6} \frac{3\alpha_2^2\eta^4}{1-\lambda}+ \frac{36\alpha^2_2\beta_2^2\eta^6}{1-\lambda}\left(D_y+ D_z \frac{108\eta^2\beta_3^2}{1-\lambda}(\ell_{f_y}^2+\frac{c_{f_y}^2\ell_{g_{yy}}^2}{\mu^2})\right) \notag\\
			& \qquad -c_{11}\alpha_2 \eta^2\Bigg)\frac{1}{K}\mathbb{E}[\| V_{t}- \delta^{g}(X_{t}, Y_{t})   \|_F^2]\notag \\
			& \quad +  \Bigg(c_{7}\frac{3\alpha_3^2\eta^4}{1-\lambda}+  D_z\frac{36\beta_3^2\alpha^2_3\eta^6}{1-\lambda} -c_{13}\alpha_3 \eta^2 \Bigg)\frac{1}{K}\mathbb{E}[\| W_{t}- \delta^{\hat{\mathcal{G}}_{h}}(X_{t}, Y_{t+1}, Z_{t})   \|_F^2]\notag \\
			& \quad + \left(D_y+ D_z \frac{108\eta^2\beta_3^2}{1-\lambda}(\ell_{f_y}^2+\frac{c_{f_y}^2\ell_{g_{yy}}^2}{\mu^2})\right) \frac{48\alpha^2_2\beta_2^2\eta^6}{1-\lambda}\sigma^2 + c_{7} \frac{6\alpha_3^2\eta^4}{1-\lambda}(1 + \frac{c_{f_y}^2}{\mu^2})\sigma^2 \notag \\
			& \quad +  D_z\frac{72\alpha^2_3\beta_3^2\eta^6}{1-\lambda}(1 + \frac{c_{f_y}^2}{\mu^2})\sigma^2 \notag \\
			& \quad + 4c_{8}\alpha_1^2\eta^4 (1 + \frac{c_{f_y}^2}{\mu^2})\sigma^2\frac{1}{K}  + 4c_{9}\alpha_1^2\eta^4 (1 + \frac{c_{f_y}^2}{\mu^2})\sigma^2  + 2c_{10}\alpha_2^2\eta^4 \sigma^2\frac{1}{K}  + 2c_{11}\alpha_2^2\eta^4 \sigma^2 \ .\notag \\
			& \quad  + 4c_{12}\alpha_3^2\eta^4(1 + \frac{c_{f_y}^2}{\mu^2})\sigma^2\frac{1}{K} + 4c_{13}\alpha_3^2\eta^4(1 + \frac{c_{f_y}^2}{\mu^2})\sigma^2 + c_{6} \frac{6\alpha_2^2\eta^4}{1-\lambda}\sigma^2+  c_{5}\frac{6\alpha_1^2\eta^4}{1-\lambda}(1 + \frac{c_{f_y}^2}{\mu^2})\sigma^2 , 
	\end{align}
	where 
	\begin{align}
		\small
				& D_y = c_{5} (\ell_{f_x}^2+\frac{c_{f_y}^2\ell_{g_{xy}}^2}{\mu^2})\frac{9}{1-\lambda}+ c_{7}(\ell_{f_y}^2+\frac{c_{f_y}^2\ell_{g_{yy}}^2}{\mu^2})\frac{9}{1-\lambda} + 6c_{8}(\ell_{f_x}^2+\frac{c_{f_y}^2\ell_{g_{xy}}^2}{\mu^2})\frac{1}{K}\notag \\
				& \quad \quad + 6c_{9}(\ell_{f_x}^2+\frac{c_{f_y}^2\ell_{g_{xy}}^2}{\mu^2}) + 6c_{12}(\frac{c_{f_y}^2\ell_{g_{yy}}^2}{\mu^2}+\ell_{f_y}^2)\frac{1}{K}+ 6c_{13}(\frac{c_{f_y}^2\ell_{g_{yy}}^2}{\mu^2}+\ell_{f_y}^2) \ ,  \notag \\
				& D_z = c_{5}c_{g_{xy}}^2\frac{9}{1-\lambda}+ 6c_{8}c_{g_{xy}}^2\frac{1}{K}+ 6c_{9}c_{g_{xy}}^2  \ , \notag \\
				& \hat{D}_y = \left(D_y+D_z \left(4\eta^2\beta_3^2(\ell_{f_y}^2+\frac{c_{f_y}^2\ell_{g_{yy}}^2}{\mu^2})\frac{9}{1-\lambda}+ 72\eta^2\beta_3^2(\frac{c_{f_y}^2\ell_{g_{yy}}^2}{\mu^2}+\ell_{f_y}^2)\right)\right) \ , 
	\end{align}
	and 
	\begin{align}
		\small
				& D_1 = \Bigg(c_{5}(\ell_{f_x}^2+\frac{c_{f_y}^2\ell_{g_{xy}}^2}{\mu^2})\frac{9}{1-\lambda}+c_{6}\frac{9\ell_{g_y}^2}{1-\lambda}+ c_{7}(\ell_{f_y}^2+\frac{c_{f_y}^2\ell_{g_{yy}}^2}{\mu^2})\frac{9}{1-\lambda}+ 6c_{8}(\ell_{f_x}^2+\frac{c_{f_y}^2\ell_{g_{xy}}^2}{\mu^2})\frac{1}{K}\notag \\
				& \quad \quad + 6c_{9}(\ell_{f_x}^2+\frac{c_{f_y}^2\ell_{g_{xy}}^2}{\mu^2})+ 2c_{10}\ell_{g_y}^2\frac{1}{K}+ 2c_{11}\ell_{g_y}^2+6c_{12}(\frac{c_{f_y}^2\ell_{g_{yy}}^2}{\mu^2}+\ell_{f_y}^2)\frac{1}{K}\notag \\
				& \quad \quad + 6c_{13}(\frac{c_{f_y}^2\ell_{g_{yy}}^2}{\mu^2}+\ell_{f_y}^2)+ \frac{60 \beta_2^2\eta^2\ell_{g_y}^2}{1-\lambda}\left(D_y+ D_z \frac{108\eta^2\beta_3^2}{1-\lambda}(\ell_{f_y}^2+\frac{c_{f_y}^2\ell_{g_{yy}}^2}{\mu^2})\right)\notag \\
				& \quad \quad  + D_z\frac{108\eta^2\beta_3^2}{1-\lambda}(\ell_{f_y}^2+\frac{c_{f_y}^2\ell_{g_{yy}}^2}{\mu^2})\Bigg)  \ , \notag \\
				& D_2 = \Bigg(c_{6} \frac{9\ell_{g_y}^2}{1-\lambda}+ 2c_{10}\ell_{g_y}^2\frac{1}{K}+ 2c_{11}\ell_{g_y}^2+ \frac{60 \beta_2^2\eta^2\ell_{g_y}^2}{1-\lambda}\left(D_y+ D_z \frac{108\eta^2\beta_3^2}{1-\lambda}(\ell_{f_y}^2+\frac{c_{f_y}^2\ell_{g_{yy}}^2}{\mu^2})\right)\Bigg)  \ , \notag \\
				& D_3 = \Bigg(c_{7} \ell_{g_y}^2\frac{9}{1-\lambda}+ 6c_{12}\ell_{g_y}^2 \frac{1}{K}+ 6c_{13}\ell_{g_y}^2+D_z\frac{ 108\eta^2\beta_3^2\ell_{g_y}^2 }{1-\lambda} \Bigg) \  , \notag \\
								& D_4 = 9\beta_1\beta_3^2\eta^3 c_{g_{xy}}^2+4\eta^2\beta_3^2D_3+8D_z\eta^2\beta_3^2 \notag \\
				& D_5 = 9\beta_1\eta(\ell_{f_x}^2+\frac{c_{f_y}^2\ell_{g_{xy}}^2}{\mu^2}) + c_{1}\frac{70\eta\beta_3}{\mu}\Big(\frac{c_{f_y}^2\ell_{g_{yy}}^2}{\mu^2}+ \ell_{f_y}^2\Big) \ . 
	\end{align}
	
	To establish the convergence rate of our algorithm, in the following, we choose proper hyperparameters to eliminate all the terms in the potential function Eq.~(\ref{eq:potential-function-alt}). 
	
	At first,  by setting
	\begin{align}
			& c_8 = \frac{\beta_1}{\alpha_1\eta}  \ , \notag \\
			&c_{10} =   \frac{25\beta_2 }{3\alpha_2 \eta\mu}   c_{0}  \ , \notag \\
			& c_5 = \beta_1(1-\lambda)  \ , \notag \\
			& c_9 = 3\beta_1  \ ,  
	\end{align}
	we can eliminate the terms: $\mathbb{E}[\|\frac{1}{K}\delta^{\hat{\mathcal{G}}_{F}} (X_{t}, Y_{t+1}, Z_{t+1}) \mathbf{1} - \frac{1}{K}{U}_{t} \mathbf{1}\|^2] $, $\mathbb{E}[\|\frac{1}{K} \delta^{g}  ({X}_{t}, {Y}_{t}) \mathbf{1} - \frac{1}{K}{V}_{t}\mathbf{1} \|^2] $,  $\mathbb{E}[\| U_{t}- \delta^{\hat{\mathcal{G}}_{F}}(X_{t}, Y_{t+1}, Z_{t+1})   \|_F^2]$.

	Based on these values, we can obtain 
		\begin{align}
						& D_z = c_{5}c_{g_{xy}}^2\frac{9}{1-\lambda}+ 6c_{8}c_{g_{xy}}^2\frac{1}{K}+ 6c_{9}c_{g_{xy}}^2 \notag \\
						& = 9\beta_1c_{g_{xy}}^2 + \frac{6\beta_1}{\alpha_1\eta K} c_{g_{xy}}^2+ 18\beta_1c_{g_{xy}}^2 \notag \\
						& = 27\beta_1c_{g_{xy}}^2 + \frac{6\beta_1}{\alpha_1\eta K} c_{g_{xy}}^2  \ ,  
	\end{align}
	and 
\begin{align}
		& D_3 =c_{7} \ell_{g_y}^2\frac{9}{1-\lambda}+ 6c_{12}\ell_{g_y}^2 \frac{1}{K}+ 6c_{13}\ell_{g_y}^2+D_z\frac{ 108\eta^2\beta_3^2\ell_{g_y}^2 }{1-\lambda} \notag \\
		& = \beta_1(1-\lambda)  \ell_{g_y}^2\frac{9}{1-\lambda}+ 6c_{12}\ell_{g_y}^2 \frac{1}{K}+ 18\ell_{g_y}^2\beta_1 +  6\ell_{g_y}^2\frac{972\beta_1\beta_3^2c_{g_{xy}}^2}{1-\lambda} \left(1+\frac{1}{\alpha_1 K} \right)\notag \\
		& \quad  +27\beta_1c_{g_{xy}}^2\frac{ 108\eta^2\beta_3^2\ell_{g_y}^2 }{1-\lambda}  + \frac{6\beta_1}{\alpha_1\eta K} c_{g_{xy}}^2 \frac{ 108\eta^2\beta_3^2\ell_{g_y}^2 }{1-\lambda}  \notag \\
		& \leq  27\beta_1 \ell_{g_y}^2 + 6c_{12}\ell_{g_y}^2 \frac{1}{K} +  \frac{8748\beta_1\beta_3^2c_{g_{xy}}^2\ell_{g_y}^2}{1-\lambda} \left(1+\frac{1}{\alpha_1 K} \right)\notag \\
		& \leq  27\beta_1 \ell_{g_y}^2 + 6c_{12}\ell_{g_y}^2 \frac{1}{K} +8748\beta_1c_{g_{xy}}^2\ell_{g_y}^2 \left(1+\frac{1}{\alpha_1 K} \right) \ , 
\end{align}
where the last step follows from $\beta_3\leq 1-\lambda$.

	To eliminate $\mathbb{E}[\| W_{t}- \delta^{\hat{\mathcal{G}}_{h}}(X_{t}, Y_{t+1}, Z_{t})   \|_F^2]$,  we set 
		\begin{align}
			& c_7 = \beta_1(1-\lambda) \ . 
	\end{align}
	Then, due to $\alpha_3\eta^2\leq 1$,  $\eta\leq 1$, and $\beta_3\leq 1-\lambda$, we can obtain
	\begin{align}
			& \quad c_{7}\frac{3\alpha_3^2\eta^4}{1-\lambda}+  D_z\frac{36\beta_3^2\alpha^2_3\eta^6}{1-\lambda} -c_{13}\alpha_3 \eta^2 \notag \\
			& \leq 3\beta_1\alpha_3\eta^2+  9\beta_1c_{g_{xy}}^2\frac{36\beta_3^2\alpha^2_3\eta^6}{1-\lambda} + \frac{36\beta_3^2\alpha^2_3\eta^6}{1-\lambda}\frac{6\beta_1}{\alpha_1\eta K} c_{g_{xy}}^2+ 18\beta_1c_{g_{xy}}^2\frac{36\beta_3^2\alpha^2_3\eta^6}{1-\lambda} -c_{13}\alpha_3 \eta^2 \notag \\
			& \leq  3\beta_1\alpha_3\eta^2+  9\beta_1c_{g_{xy}}^2\frac{36\beta_3^2\alpha_3\eta^4}{1-\lambda} + \frac{36\beta_3^2\alpha_3\eta^3}{1-\lambda}\frac{6\beta_1}{\alpha_1 K} c_{g_{xy}}^2+ 18\beta_1c_{g_{xy}}^2\frac{36\beta_3^2\alpha_3\eta^4}{1-\lambda} -c_{13}\alpha_3 \eta^2 \notag \\
			& \leq  3\beta_1\alpha_3\eta^2+  324\beta_1c_{g_{xy}}^2\beta_3\alpha_3\eta^4 +216\beta_3\alpha_3\eta^3\frac{\beta_1}{\alpha_1 K} c_{g_{xy}}^2+ 648\beta_1c_{g_{xy}}^2\beta_3 \alpha_3\eta^4-c_{13}\alpha_3 \eta^2 \ ,  
	\end{align}
By enforcing this upper bound to be non-positive, we can obtain
	\begin{align}
		& c_{13} \geq 3\beta_1 +  324\beta_1\beta_3c_{g_{xy}}^2 + 216\beta_1\beta_3c_{g_{xy}}^2\frac{1}{\alpha_1 K} +648\beta_1\beta_3 c_{g_{xy}}^2 \ . 
\end{align}
Due to $\beta_3\leq 1-\lambda \leq 1$, we can set
	\begin{align}
		& c_{13} = 3\beta_1 + 972\beta_1c_{g_{xy}}^2 \left(1+\frac{1}{\alpha_1 K} \right) \ . 
\end{align}
	
	To eliminate $ \mathbb{E} [ \| \frac{1}{K}\delta^{\hat{\mathcal{G}}_{h}}({X}_{t}, {Y}_{t+1},  {Z}_{t})\mathbf{1}-   \frac{1}{K}{W}_{t}\mathbf{1}\|^2]$, we enforce
	\begin{align}
			&\quad  c_{1}\frac{25\eta\beta_3}{\mu}+3 D_4-c_{12}\alpha_3 \eta^2 \notag \\
			& \leq c_{1}\frac{25\eta\beta_3}{\mu}+681\eta^2\beta_1\beta_3^2 c_{g_{xy}}^2 + \frac{144\eta\beta_1 \beta_3^2}{\alpha_1 K} c_{g_{xy}}^2+324\eta^2 \beta_1\beta_3^2 \ell_{g_y}^2 + 72\eta^2\beta_3^2\ell_{g_y}^2 \frac{1}{K} c_{12} \notag \\
			& \quad + 12\eta^2\beta_3^2 \frac{8748\beta_1\beta_3^2c_{g_{xy}}^2\ell_{g_y}^2}{1-\lambda} \left(1+\frac{1}{\alpha_1 K} \right) -c_{12}\alpha_3 \eta^2  \notag \\
			& \leq c_{1}\frac{25\eta\beta_3}{\mu}+681\eta^2\beta_1\beta_3^2 c_{g_{xy}}^2 + \frac{144\eta\beta_1 \beta_3^2}{\alpha_1 K} c_{g_{xy}}^2+324\eta^2 \beta_1\beta_3^2 \ell_{g_y}^2 + 72\eta^2\beta_3^2\ell_{g_y}^2 \frac{1}{K} c_{12} \notag \\
			& \quad + 104976\eta^2\beta_3^3\beta_1c_{g_{xy}}^2\ell_{g_y}^2 \left(1+\frac{1}{\alpha_1 K} \right) -c_{12}\alpha_3 \eta^2\notag \\
			&  \leq  0  \ , 
	\end{align}
	where the second step follows from $\beta_3\leq 1-\lambda$. 
	
	To this end, we enforce
		\begin{align}
			&  72\eta^2\beta_3^2\ell_{g_y}^2 \frac{1}{K} c_{12} \leq c_{12}\alpha_3 \eta^2 \frac{1}{4}  \ , \notag \\
			&  c_{1}\frac{25\eta\beta_3}{\mu}\leq c_{12}\alpha_3 \eta^2 \frac{1}{4}  \ , \notag \\ 
			& 681\eta^2\beta_1\beta_3^2 c_{g_{xy}}^2 + \frac{144\eta\beta_1 \beta_3^2}{\alpha_1 K} c_{g_{xy}}^2+324\eta^2 \beta_1\beta_3^2 \ell_{g_y}^2  \notag \\
			& \quad +104976\eta^2\beta_3^3\beta_1c_{g_{xy}}^2\ell_{g_y}^2 \left(1+\frac{1}{\alpha_1 K} \right) \leq c_{12}\alpha_3 \eta^2 \frac{1}{2} \ . 
	\end{align}
	We can obtain
		\begin{align}\label{eq:c-1-1-alt}
			& \beta_3 \leq \frac{\sqrt{\alpha_3K}}{20\ell_{g_y}}  \ , \notag \\
			& c_{12} =   c_{1}\frac{100\beta_3}{\alpha_3 \eta \mu} \ , \notag \\
			& c_1 \geq 15\mu\beta_1\beta_3 c_{g_{xy}}^2\left(1+ \frac{1}{\alpha_1 K}\right)+7\mu\beta_1\beta_3 \ell_{g_y}^2  +  2100\mu\beta_1\beta_3^2c_{g_{xy}}^2\ell_{g_y}^2 \left(1+\frac{1}{\alpha_1 K} \right)  \ . 
	\end{align}
	
	Based on these values, we can obtain
		\begin{align}
						& D_4 = 9\beta_1\beta_3^2\eta^3 c_{g_{xy}}^2+4\eta^2\beta_3^2D_3+8D_z\eta^2\beta_3^2 \notag \\
						& = 9\beta_1\beta_3^2\eta^3 c_{g_{xy}}^2+4\eta^2\beta_3^2D_3+216\eta^2\beta_3^2\beta_1c_{g_{xy}}^2 + \frac{48\eta\beta_3^2\beta_1}{\alpha_1 K} c_{g_{xy}}^2 \notag \\
						& \leq 227\eta^2\beta_1\beta_3^2 c_{g_{xy}}^2 + \frac{48\eta\beta_1 \beta_3^2}{\alpha_1 K} c_{g_{xy}}^2+4\eta^2\beta_3^2D_3 \notag \\
						& \leq 227\eta^2\beta_1\beta_3^2 c_{g_{xy}}^2 + \frac{48\eta\beta_1 \beta_3^2}{\alpha_1 K} c_{g_{xy}}^2+108\eta^2 \beta_1\beta_3^2 \ell_{g_y}^2 + 24\eta^2\beta_3^2\ell_{g_y}^2 \frac{1}{K} c_{12}  \notag \\
						& \quad + 34992\eta^2\beta_3^3 \beta_1 c_{g_{xy}}^2\ell_{g_y}^2 \left(1+\frac{1}{\alpha_1 K} \right)\notag \\
						& \leq  227\eta \beta_1\beta_3^2 c_{g_{xy}}^2 \left(1 + \frac{1}{\alpha_1 K}\right) + \frac{2400\eta\beta_3^3 \ell_{g_y}^2}{\alpha_3 K \mu } c_{1}+108\eta^2 \beta_1\beta_3^2 \ell_{g_y}^2 \notag \\
						& \quad  + 34992\eta^2\beta_3^3 \beta_1 c_{g_{xy}}^2\ell_{g_y}^2 \left(1+\frac{1}{\alpha_1 K} \right) \ . 
			\end{align}
			where the second step follows from $\eta\leq 1$, the last step follows from $c_{12} =   c_{1}\frac{100\beta_3}{\alpha_3 \eta \mu} $. 
	
	To eliminate $ \mathbb{E}[\|z^*(\bar{x}_{t}) - \bar{z}_{t} \|^2]$,  we can enforce
	\begin{align}
			& \quad 9 \beta_1\eta c_{g_{xy}}^2 +  15 D_4 \ell_{g_y}^2-c_{1}\frac{\eta\beta_3\mu}{8} \notag \\
& \leq 9 \beta_1\eta c_{g_{xy}}^2 +  15  \ell_{g_y}^2\Bigg(227\eta \beta_1\beta_3^2 c_{g_{xy}}^2 \left(1 + \frac{1}{\alpha_1 K}\right) + \frac{2400\eta\beta_3^3 \ell_{g_y}^2}{\alpha_3 K \mu } c_{1}+108\eta^2 \beta_1\beta_3^2 \ell_{g_y}^2  \notag \\
& \quad + 34992\eta^2\beta_3^3 \beta_1 c_{g_{xy}}^2\ell_{g_y}^2 \left(1+\frac{1}{\alpha_1 K} \right) \Bigg) -c_{1}\frac{\eta\beta_3\mu}{8} \notag \\
& \leq 9 \beta_1\eta c_{g_{xy}}^2 +  15  \ell_{g_y}^2\Bigg(227\eta \beta_1\beta_3^2 c_{g_{xy}}^2 \left(1 + \frac{1}{\alpha_1 K}\right) + \frac{2400\eta\beta_3^3 \ell_{g_y}^2}{\alpha_3 K \mu } c_{1}+108\eta \beta_1\beta_3^2 \ell_{g_y}^2  \notag \\
& \quad + 34992\eta \beta_3^3 \beta_1 c_{g_{xy}}^2\ell_{g_y}^2 \left(1+\frac{1}{\alpha_1 K} \right) \Bigg) -c_{1}\frac{\eta\beta_3\mu}{8} \notag \\
			& \leq  0  \  . 
	\end{align}
	where the last step follows from $\eta\leq 1$.

	This can be done by enforcing
	\begin{align}
			& 15  \ell_{g_y}^2\frac{2400\eta\beta_3^3 \ell_{g_y}^2}{\alpha_3 K \mu } c_{1}  \leq c_{1}\frac{\eta\beta_3\mu}{16}  \ , \notag \\
			& 9 \beta_1\eta c_{g_{xy}}^2 +  15  \ell_{g_y}^2\Bigg(227\eta \beta_1\beta_3^2 c_{g_{xy}}^2 \left(1 + \frac{1}{\alpha_1 K}\right) + 108\eta \beta_1\beta_3^2 \ell_{g_y}^2  \notag \\
			& \quad + 34992\eta \beta_3^3 \beta_1 c_{g_{xy}}^2\ell_{g_y}^2 \left(1+\frac{1}{\alpha_1 K} \right) \Bigg) \leq c_{1}\frac{\eta\beta_3\mu}{16} \ . 
	\end{align}
	As a result, we can obtain
	\begin{align}\label{eq:c-1-2-alt}
			& \beta_3  \leq \frac{\mu\sqrt{\alpha_3 K} }{800  \ell_{g_y}^2}  \ , \notag \\
			& c_{1} \geq \frac{144 \beta_1 c_{g_{xy}}^2}{\beta_3\mu}+   \frac{240\beta_1\ell_{g_y}^2}{\mu}\Bigg(227 \beta_3 c_{g_{xy}}^2 \left(1 + \frac{1}{\alpha_1 K}\right) + 108 \beta_3 \ell_{g_y}^2 \notag \\
			& \qquad + 34992\beta_3^2  c_{g_{xy}}^2\ell_{g_y}^2 \left(1+\frac{1}{\alpha_1 K} \right) \Bigg)   \ . 
	\end{align}
	
	Due to $ \mu \leq \ell_{g_y}$ and $\beta_3<1-\lambda$, we can set
		\begin{align}
			& c_{1} = \frac{\beta_1}{\beta_3}  \Bigg[\frac{144  c_{g_{xy}}^2}{\mu}+  \frac{ 54480 c_{g_{xy}}^2 \ell_{g_y}^2}{\mu }   \left(1 + \frac{1}{\alpha_1K}\right) + \frac{25920    \ell_{g_y}^4}{\mu}  \notag \\
			& \qquad +   \frac{8748\times 960 c_{g_{xy}}^2\ell_{g_y}^4}{\mu} \left(1+\frac{1}{\alpha_1 K} \right) \Bigg]  \notag \\
			& \triangleq  \frac{\beta_1}{\beta_3}  \tilde{c}_1\ , 
	\end{align}
	which satisfies both Eq.~(\ref{eq:c-1-1-alt}) and Eq.~(\ref{eq:c-1-2-alt}). 
	
	To eliminate $ \mathbb{E}[\|y^*(\bar{x}_{t}) - \bar{y}_{t} \|^2]$,  we can enforce
	\begin{align}
			& \quad 9\beta_1\eta\Big(\ell_{f_x}^2+\frac{c_{f_y}^2\ell_{g_{xy}}^2}{\mu^2}\Big)+c_{1} \frac{18\eta\beta_3}{\mu}\Big(\frac{c_{f_y}^2\ell_{g_{yy}}^2}{\mu^2}+\ell_{f_y}^2\Big) + 15D_4 \Big(\frac{\ell_{g_{yy}}^2c_{f_y}^2}{\mu^2}+ \ell_{f_y}^2\Big)-c_{0}\frac{\beta_2\eta\mu}{4}  \notag \\
			& \leq 9\beta_1\eta\Big(\ell_{f_x}^2+\frac{c_{f_y}^2\ell_{g_{xy}}^2}{\mu^2}\Big) +c_{1} \frac{18\eta\beta_3}{\mu}\Big(\frac{c_{f_y}^2\ell_{g_{yy}}^2}{\mu^2}+\ell_{f_y}^2\Big) + 15 \Big(\frac{\ell_{g_{yy}}^2c_{f_y}^2}{\mu^2}+ \ell_{f_y}^2\Big)  \frac{2400\eta\beta_3^3 \ell_{g_y}^2}{\alpha_3 K \mu } c_{1}  \notag \\
			& \quad -c_{0}\frac{\beta_2\eta\mu}{4} + 15 \Big(\frac{\ell_{g_{yy}}^2c_{f_y}^2}{\mu^2}+ \ell_{f_y}^2\Big) \Bigg( 227\eta \beta_1\beta_3^2 c_{g_{xy}}^2 \left(1 + \frac{1}{\alpha_1 K}\right) +108\eta^2 \beta_1\beta_3^2 \ell_{g_y}^2  \notag \\
			& \qquad + 34992\eta^2\beta_3^3 \beta_1 c_{g_{xy}}^2\ell_{g_y}^2 \left(1+\frac{1}{\alpha_1 K} \right)\Bigg)\notag \\
			& \leq 9\eta\beta_1\Big(\ell_{f_x}^2+\frac{c_{f_y}^2\ell_{g_{xy}}^2}{\mu^2}\Big) +c_{1} \eta\beta_3\Big(\frac{18}{\mu}+ 15\frac{2400 \ell_{g_y}^2}{\alpha_3 K \mu } \Big)\Big(\frac{c_{f_y}^2\ell_{g_{yy}}^2}{\mu^2}+\ell_{f_y}^2\Big) -c_{0}\frac{\beta_2\eta\mu}{4}  \notag \\
			& \quad + 15\eta \beta_1\Big(\frac{\ell_{g_{yy}}^2c_{f_y}^2}{\mu^2}+ \ell_{f_y}^2\Big) \Bigg( 227   \left(1 + \frac{1}{\alpha_1 K}\right)c_{g_{xy}}^2 +108 \ell_{g_y}^2  + 34992 c_{g_{xy}}^2\ell_{g_y}^2 \left(1+\frac{1}{\alpha_1 K} \right)\Bigg)\notag \\
			&\leq  0  \ , 
	\end{align}
	where the second to last step follows from $\eta\leq 1$ and $\beta_3\leq 1-\lambda$. Then, we can set
	\begin{align}
			& c_0 =\frac{\beta_1}{\beta_2} \Bigg[ \frac{36}{\mu}\Big(\ell_{f_x}^2+\frac{c_{f_y}^2\ell_{g_{xy}}^2}{\mu^2}\Big)  \notag \\
			& \quad+ \frac{60}{\mu}\Big(\frac{\ell_{g_{yy}}^2c_{f_y}^2}{\mu^2}+ \ell_{f_y}^2\Big) \Bigg( \left(227c_{g_{xy}}^2 + 34992c_{g_{xy}}^2\ell_{g_y}^2 \right)   \left(1 + \frac{1}{\alpha_1 K}\right) +108 \ell_{g_y}^2 \Bigg)\notag \\
			&\quad +\frac{4}{\mu} \Big(\frac{18}{\mu}+ \frac{36000 \ell_{g_y}^2}{\alpha_3 K \mu } \Big)\Big(\frac{c_{f_y}^2\ell_{g_{yy}}^2}{\mu^2}+\ell_{f_y}^2\Big)\Bigg(  \frac{ (54480  +8748\times 960 \ell_{g_y}^2)\ell_{g_y}^2c_{g_{xy}}^2}{\mu }   \left(1 + \frac{1}{\alpha_1K}\right) \notag \\ 
			& \quad \quad +  \frac{144  c_{g_{xy}}^2+25920    \ell_{g_y}^4}{\mu}   \Bigg)  \Bigg] \notag \\
			& \triangleq \frac{\beta_1}{\beta_2} \tilde{c}_0  \ . 
	\end{align}
	
	Based on these values,  $\beta_3\leq 1-\lambda$, and $\eta\leq 1$, we can obtain
	\begin{align}
			& D_z  = 27\beta_1c_{g_{xy}}^2 + \frac{6\beta_1}{\alpha_1\eta K} c_{g_{xy}}^2 \leq  \frac{27\beta_1c_{g_{xy}}^2}{\eta}\left(1+ \frac{1}{\alpha_1 K}\right)   \ , \notag \\
			& D_y = c_{5} (\ell_{f_x}^2+\frac{c_{f_y}^2\ell_{g_{xy}}^2}{\mu^2})\frac{9}{1-\lambda}+ c_{7}(\ell_{f_y}^2+\frac{c_{f_y}^2\ell_{g_{yy}}^2}{\mu^2})\frac{9}{1-\lambda} + 6c_{8}(\ell_{f_x}^2+\frac{c_{f_y}^2\ell_{g_{xy}}^2}{\mu^2})\frac{1}{K}\notag \\
			& \quad \quad + 6c_{9}(\ell_{f_x}^2+\frac{c_{f_y}^2\ell_{g_{xy}}^2}{\mu^2}) + 6c_{12}(\frac{c_{f_y}^2\ell_{g_{yy}}^2}{\mu^2}+\ell_{f_y}^2)\frac{1}{K}+ 6c_{13}(\frac{c_{f_y}^2\ell_{g_{yy}}^2}{\mu^2}+\ell_{f_y}^2) \notag \\
			&\leq 27 \beta_1 (\ell_{f_x}^2+\frac{c_{f_y}^2\ell_{g_{xy}}^2}{\mu^2})  + 9\beta_1(\ell_{f_y}^2+\frac{c_{f_y}^2\ell_{g_{yy}}^2}{\mu^2})+  \frac{6\beta_1}{\alpha_1\eta K} (\ell_{f_x}^2+\frac{c_{f_y}^2\ell_{g_{xy}}^2}{\mu^2})  \notag \\
			& \quad  + \frac{600\beta_3}{\alpha_3 \eta \mu K} (\frac{c_{f_y}^2\ell_{g_{yy}}^2}{\mu^2}+\ell_{f_y}^2)c_{1}+ 18\beta_1(\frac{c_{f_y}^2\ell_{g_{yy}}^2}{\mu^2}+\ell_{f_y}^2) \notag \\
			& \quad  +5832\beta_1\beta_3 c_{g_{xy}}^2 (\frac{c_{f_y}^2\ell_{g_{yy}}^2}{\mu^2}+\ell_{f_y}^2)  \left(1+\frac{1}{\alpha_1 K} \right) \notag \\
			& \leq 27\beta_1 \left(1+\frac{1}{\alpha_1K\eta} \right)\left(\ell_{f_x}^2+\frac{c_{f_y}^2\ell_{g_{xy}}^2}{\mu^2}\right)  + 9\beta_1\left(\ell_{f_y}^2+\frac{c_{f_y}^2\ell_{g_{yy}}^2}{\mu^2}\right)   \notag \\
			& \quad  + 18\beta_1\left(1+324 \beta_3 c_{g_{xy}}^2\left(1+\frac{1}{\alpha_1 K} \right)  \right)\left(\ell_{f_y}^2+\frac{c_{f_y}^2\ell_{g_{yy}}^2}{\mu^2}\right)  + \frac{600\beta_1}{\alpha_3 K \eta \mu } \left(\frac{c_{f_y}^2\ell_{g_{yy}}^2}{\mu^2}+\ell_{f_y}^2\right)\tilde{c}_1 \notag \\
			& \leq \frac{\beta_1}{\eta}\Bigg[27 \left(1+\frac{1}{\alpha_1K} \right)\left(\ell_{f_x}^2+\frac{c_{f_y}^2\ell_{g_{xy}}^2}{\mu^2})\right)   \notag \\
			& \quad + 18\left(2+324  c_{g_{xy}}^2\left(1+\frac{1}{\alpha_1 K} \right)  \right)\left(\ell_{f_y}^2+\frac{c_{f_y}^2\ell_{g_{yy}}^2}{\mu^2}\right)   + \frac{600}{\alpha_3 K  \mu } \left(\frac{c_{f_y}^2\ell_{g_{yy}}^2}{\mu^2}+\ell_{f_y}^2\right)\tilde{c}_1 \Bigg] \notag \\
			& \triangleq  \frac{\beta_1}{\eta} \tilde{D}_y  \ , \notag \\
			& \hat{D}_y = D_y+D_z \left(4\eta^2\beta_3^2(\ell_{f_y}^2+\frac{c_{f_y}^2\ell_{g_{yy}}^2}{\mu^2})\frac{9}{1-\lambda}+ 72\eta^2\beta_3^2(\frac{c_{f_y}^2\ell_{g_{yy}}^2}{\mu^2}+\ell_{f_y}^2)\right)  \notag \\
			& \leq D_y+D_z\frac{108\eta^2\beta_3^2}{1-\lambda} \left(\ell_{f_y}^2+\frac{c_{f_y}^2\ell_{g_{yy}}^2}{\mu^2}\right) \notag \\
			& \leq2916\eta\beta_1 c_{g_{xy}}^2\left(1+ \frac{1}{\alpha_1 K}\right)  \left(\ell_{f_y}^2+\frac{c_{f_y}^2\ell_{g_{yy}}^2}{\mu^2}\right)  + \frac{\beta_1}{\eta} \tilde{D}_y \ . 
	\end{align}
	
	Then, to eliminate $\mathbb{E}[\| V_{t}- \delta^{g}(X_{t}, Y_{t})   \|_F^2]$, we can enforce
	\begin{align}
			&  c_{6} = \beta_1(1-\lambda) 
	\end{align}
	and 
		\begin{align}
			&\quad  c_{6} \frac{3\alpha_2^2\eta^4}{1-\lambda}+ \frac{36\alpha^2_2\beta_2^2\eta^6}{1-\lambda}\left(D_y+ D_z \frac{108\eta^2\beta_3^2}{1-\lambda}(\ell_{f_y}^2+\frac{c_{f_y}^2\ell_{g_{yy}}^2}{\mu^2})\right)-c_{11}\alpha_2 \eta^2  \notag \\
			&\leq 3 \beta_1  \alpha_2\eta^2+ \frac{36\alpha_2\beta_2^2\eta^4}{1-\lambda}\left(\frac{\beta_1}{\eta} \tilde{D}_y+  \frac{27\beta_1c_{g_{xy}}^2}{\eta}\left(1+ \frac{1}{\alpha_1 K}\right)   \frac{108\eta^2\beta_3^2}{1-\lambda}(\ell_{f_y}^2+\frac{c_{f_y}^2\ell_{g_{yy}}^2}{\mu^2})\right)\notag \\
			&\leq 3 \beta_1  \alpha_2\eta^2+ \frac{36\alpha_2\beta_2^2\eta^4}{1-\lambda}\left(\frac{\beta_1}{\eta} \tilde{D}_y+  2916\beta_1\eta\beta_3c_{g_{xy}}^2\left(1+ \frac{1}{\alpha_1 K}\right)   (\ell_{f_y}^2+\frac{c_{f_y}^2\ell_{g_{yy}}^2}{\mu^2})\right)\notag \\
			&\leq 3 \beta_1  \alpha_2\eta^2+ \frac{36\alpha_2\eta^2\beta_2^2}{1-\lambda}\left(\beta_1 \tilde{D}_y+  2916\beta_1c_{g_{xy}}^2\left(1+ \frac{1}{\alpha_1 K}\right)   (\ell_{f_y}^2+\frac{c_{f_y}^2\ell_{g_{yy}}^2}{\mu^2})\right)-c_{11}\alpha_2 \eta^2  \notag \\
			& \leq 0 \ ,  
	\end{align}
	where we use the fact $\beta_3\leq 1-\lambda$,  $\eta\leq 1$, and $\alpha_2\eta<1$ in different steps. Then, we can set
		\begin{align}
			&c_{11}=  3 \beta_1  + \frac{36\beta_1\beta_2^2}{1-\lambda}\left(\tilde{D}_y+  2916 c_{g_{xy}}^2\left(1+ \frac{1}{\alpha_1 K}\right)   \left(\ell_{f_y}^2+\frac{c_{f_y}^2\ell_{g_{yy}}^2}{\mu^2}\right)\right) \ . 
	\end{align}
	
	Consequently, due to $\beta_3\leq 1-\lambda$ , $\beta_2\leq 1-\lambda$ , and $\eta\leq 1$, we have
	\begin{align}
			& D_1 = c_{5}(\ell_{f_x}^2+\frac{c_{f_y}^2\ell_{g_{xy}}^2}{\mu^2})\frac{9}{1-\lambda}+c_{6}\frac{9\ell_{g_y}^2}{1-\lambda}+ c_{7}(\ell_{f_y}^2+\frac{c_{f_y}^2\ell_{g_{yy}}^2}{\mu^2})\frac{9}{1-\lambda}+ 6c_{8}(\ell_{f_x}^2+\frac{c_{f_y}^2\ell_{g_{xy}}^2}{\mu^2})\frac{1}{K}\notag \\
			& \quad \quad + 6c_{9}(\ell_{f_x}^2+\frac{c_{f_y}^2\ell_{g_{xy}}^2}{\mu^2})+ 2c_{10}\ell_{g_y}^2\frac{1}{K}+ 2c_{11}\ell_{g_y}^2+6c_{12}(\frac{c_{f_y}^2\ell_{g_{yy}}^2}{\mu^2}+\ell_{f_y}^2)\frac{1}{K} \notag \\
			& \qquad   + 6c_{13}(\frac{c_{f_y}^2\ell_{g_{yy}}^2}{\mu^2}+\ell_{f_y}^2) + \frac{60 \beta_2^2\eta^2\ell_{g_y}^2}{1-\lambda}\left(D_y+ D_z \frac{108\eta^2\beta_3^2}{1-\lambda}(\ell_{f_y}^2+\frac{c_{f_y}^2\ell_{g_{yy}}^2}{\mu^2})\right)  \notag \\
			& \quad \quad+ D_z\frac{108\eta^2\beta_3^2}{1-\lambda}(\ell_{f_y}^2+\frac{c_{f_y}^2\ell_{g_{yy}}^2}{\mu^2})\notag \\
			& \leq  \beta_1(1-\lambda) (\ell_{f_x}^2+\frac{c_{f_y}^2\ell_{g_{xy}}^2}{\mu^2})\frac{9}{1-\lambda}+\beta_1(1-\lambda) \frac{9\ell_{g_y}^2}{1-\lambda}+ \beta_1(1-\lambda) (\ell_{f_y}^2+\frac{c_{f_y}^2\ell_{g_{yy}}^2}{\mu^2})\frac{9}{1-\lambda}\notag \\
			&\quad + 6 \frac{\beta_1}{\alpha_1\eta}(\ell_{f_x}^2+\frac{c_{f_y}^2\ell_{g_{xy}}^2}{\mu^2})\frac{1}{K}+ 18\beta_1(\ell_{f_x}^2+\frac{c_{f_y}^2\ell_{g_{xy}}^2}{\mu^2})+ 2 \frac{25\beta_2 }{3\alpha_2 \eta\mu}   c_{0}\ell_{g_y}^2\frac{1}{K}+ 2c_{11}\ell_{g_y}^2 \notag \\
			& \quad +6\frac{100\beta_3}{\alpha_3 \eta \mu} c_{1}(\frac{c_{f_y}^2\ell_{g_{yy}}^2}{\mu^2}+\ell_{f_y}^2)\frac{1}{K} + 6\left(3\beta_1 + 972\beta_1 c_{g_{xy}}^2 \left(1+\frac{1}{\alpha_1 K} \right) \right)(\frac{c_{f_y}^2\ell_{g_{yy}}^2}{\mu^2}+\ell_{f_y}^2) \notag \\
			& \quad + \frac{60 \beta_2^2\eta^2\ell_{g_y}^2}{1-\lambda}\left(\frac{\beta_1}{\eta} \tilde{D}_y+   2916\eta\beta_1\beta_3 c_{g_{xy}}^2\left(1+ \frac{1}{\alpha_1 K}\right) \frac{}{}(\ell_{f_y}^2+\frac{c_{f_y}^2\ell_{g_{yy}}^2}{\mu^2})\right)\notag \\
			& \quad   +   2916\eta\beta_1\beta_3 c_{g_{xy}}^2\left(1+ \frac{1}{\alpha_1 K}\right)(\ell_{f_y}^2+\frac{c_{f_y}^2\ell_{g_{yy}}^2}{\mu^2})\notag \\
			& \leq  \frac{27\beta_1}{\eta}\left(1+ \frac{1}{\alpha_1 K}\right)\left(\ell_{f_x}^2+\frac{c_{f_y}^2\ell_{g_{xy}}^2}{\mu^2}\right)+9\ell_{g_y}^2\beta_1+ 9\beta_1 (\ell_{f_y}^2+\frac{c_{f_y}^2\ell_{g_{yy}}^2}{\mu^2}) \notag \\
			& \quad  + \frac{\beta_1}{\eta} \frac{50 \ell_{g_y}^2}{3\alpha_2K \mu }   \tilde{c}_0 + \frac{\beta_1}{\eta}  \frac{600}{\alpha_3 K \mu} \left(\frac{c_{f_y}^2\ell_{g_{yy}}^2}{\mu^2}+\ell_{f_y}^2\right) \tilde{c}_1\notag \\
			& \quad + 2\ell_{g_y}^2\frac{\beta_1}{\eta}\left(3   + \frac{36\beta_2^2}{1-\lambda}\left(\tilde{D}_y+  2916\beta_3c_{g_{xy}}^2\left(1+ \frac{1}{\alpha_1 K}\right)   \left(\ell_{f_y}^2+\frac{c_{f_y}^2\ell_{g_{yy}}^2}{\mu^2}\right)\right) \right) \notag \\
			& \quad  + 6\frac{\beta_1}{\eta}\left(3 + 972\beta_3c_{g_{xy}}^2\left(1+\frac{1}{\alpha_1 K} \right) \right)\left(\frac{c_{f_y}^2\ell_{g_{yy}}^2}{\mu^2}+\ell_{f_y}^2\right) \notag \\
			& \quad + \frac{\beta_1}{\eta}\frac{60 \beta_2^2\ell_{g_y}^2}{1-\lambda}\left(\tilde{D}_y+   2916\beta_3 c_{g_{xy}}^2\left(1+ \frac{1}{\alpha_1 K}\right) \left(\ell_{f_y}^2+\frac{c_{f_y}^2\ell_{g_{yy}}^2}{\mu^2}\right)\right)\notag \\
			& \quad   + \frac{\beta_1}{\eta} 2916\beta_3 c_{g_{xy}}^2\left(1+ \frac{1}{\alpha_1 K}\right)\left(\ell_{f_y}^2+\frac{c_{f_y}^2\ell_{g_{yy}}^2}{\mu^2}\right)\notag \\
& \leq  \frac{\beta_1}{\eta}\Bigg[27\left(1+ \frac{1}{\alpha_1 K}\right)\left(\ell_{f_x}^2+\frac{c_{f_y}^2\ell_{g_{xy}}^2}{\mu^2}\right) \notag \\
& \qquad + 6\left(5 + (1458 + 64152\ell_{g_y}^2) c_{g_{xy}}^2\left(1+\frac{1}{\alpha_1 K} \right) \right)\left(\frac{c_{f_y}^2\ell_{g_{yy}}^2}{\mu^2}+\ell_{f_y}^2\right) \notag \\
& \quad \quad + \frac{50 \ell_{g_y}^2}{3\alpha_2K \mu }   \tilde{c}_0 +  \frac{600}{\alpha_3 K \mu} \left(\frac{c_{f_y}^2\ell_{g_{yy}}^2}{\mu^2}+\ell_{f_y}^2\right) \tilde{c}_1 + 132 \ell_{g_y}^2\tilde{D}_y+ 15\ell_{g_y}^2  \Bigg] \notag \\
			& = \frac{\beta_1}{\eta}\tilde{D}_{1} \ . 
	\end{align}
	Similarly, we can obtain 
	\begin{align}
			& \quad  D_2 = c_{6} \frac{9\ell_{g_y}^2}{1-\lambda}+ 2c_{10}\ell_{g_y}^2\frac{1}{K}+ 2c_{11}\ell_{g_y}^2+ \frac{60 \beta_2^2\eta^2\ell_{g_y}^2}{1-\lambda}\left(D_y+ D_z \frac{108\eta^2\beta_3^2}{1-\lambda}(\ell_{f_y}^2+\frac{c_{f_y}^2\ell_{g_{yy}}^2}{\mu^2})\right) \notag \\
			& \leq \beta_1(1-\lambda)  \frac{9\ell_{g_y}^2}{1-\lambda}+ 2 \frac{25\beta_2 }{3\alpha_2 \eta\mu}    \frac{\beta_1}{\beta_2}\tilde{c}_0\ell_{g_y}^2\frac{1}{K}\notag \\
			& \quad + 2\ell_{g_y}^2\left(3 \beta_1  + \frac{36\beta_1\beta_2^2}{1-\lambda}\left(\tilde{D}_y+  2916\beta_3c_{g_{xy}}^2\left(1+ \frac{1}{\alpha_1 K}\right)   \left(\ell_{f_y}^2+\frac{c_{f_y}^2\ell_{g_{yy}}^2}{\mu^2}\right)\right) \right)\notag \\
			& \quad + \frac{60 \beta_2^2\eta^2\ell_{g_y}^2}{1-\lambda}\left(\frac{\beta_1}{\eta} \tilde{D}_y+  2916\eta\beta_3\beta_1c_{g_{xy}}^2\left(1+ \frac{1}{\alpha_1 K}\right)  (\ell_{f_y}^2+\frac{c_{f_y}^2\ell_{g_{yy}}^2}{\mu^2})\right) \notag \\
			& \leq 9\beta_1 \ell_{g_y}^2+  \frac{50 \beta_1 \ell_{g_y}^2}{3\alpha_2 K\eta\mu}   \tilde{c}_0 \notag \\
			& \quad + 6 \beta_1 \ell_{g_y}^2 + \frac{72\beta_1\beta_2^2\ell_{g_y}^2}{1-\lambda}\left(\tilde{D}_y+  2916\beta_3c_{g_{xy}}^2\left(1+ \frac{1}{\alpha_1 K}\right)   \left(\ell_{f_y}^2+\frac{c_{f_y}^2\ell_{g_{yy}}^2}{\mu^2}\right)\right) \notag \\
			& \quad + \frac{60 \beta_1\beta_2^2\ell_{g_y}^2}{1-\lambda}\left( \tilde{D}_y+  2916\beta_3 c_{g_{xy}}^2\left(1+ \frac{1}{\alpha_1 K}\right)  (\ell_{f_y}^2+\frac{c_{f_y}^2\ell_{g_{yy}}^2}{\mu^2})\right) \notag \\
			& \leq \frac{\beta_1}{\eta} \Bigg[15 \ell_{g_y}^2+  \frac{50  \ell_{g_y}^2}{3\alpha_2 K\mu}   \tilde{c}_0  + 132\ell_{g_y}^2\left(\tilde{D}_y+  2916c_{g_{xy}}^2\left(1+ \frac{1}{\alpha_1 K}\right)   \left(\ell_{f_y}^2+\frac{c_{f_y}^2\ell_{g_{yy}}^2}{\mu^2}\right)\right)\Bigg] \notag \\
			& \triangleq\frac{\beta_1}{\eta}\tilde{D}_2 \ , 
	\end{align}
where we use the fact $\beta_2\leq 1-\lambda$ in the fourth step. 

	Moreover, due to $\beta_3\leq 1-\lambda$ , $\beta_2\leq 1-\lambda$ , and $\eta\leq 1$, we have
	\begin{align}
			& D_3  \leq  27\beta_1 \ell_{g_y}^2 + 6c_{12}\ell_{g_y}^2 \frac{1}{K} +  \frac{8748\beta_1\beta_3^2c_{g_{xy}}^2\ell_{g_y}^2}{1-\lambda} \left(1+\frac{1}{\alpha_1 K} \right)\notag \\
			& \leq  27\beta_1 \ell_{g_y}^2 + 6\ell_{g_y}^2 \frac{1}{K}\frac{100\beta_3}{\alpha_3 \eta \mu} \frac{\beta_1}{\beta_3}\tilde{c}_1 +  \frac{8748\beta_1\beta_3^2c_{g_{xy}}^2\ell_{g_y}^2}{1-\lambda} \left(1+\frac{1}{\alpha_1 K} \right)\notag \\
			& \leq \frac{\beta_1}{\eta}\left[27 \ell_{g_y}^2 + \frac{600\ell_{g_y}^2 }{\alpha_3  K\mu} \frac{}{}\tilde{c}_1 +  8748c_{g_{xy}}^2\ell_{g_y}^2 \left(1+\frac{1}{\alpha_1 K} \right) \right] \notag \\
			& \triangleq \frac{\beta_1}{\eta}\tilde{D}_3  \ , 
	\end{align}
	and 
	\begin{align}
			& D_4  \leq   227\eta \beta_1\beta_3^2 c_{g_{xy}}^2 \left(1 + \frac{1}{\alpha_1 K}\right) + \frac{2400\eta\beta_3^3 \ell_{g_y}^2}{\alpha_3 K \mu } c_{1}+108\eta^2 \beta_1\beta_3^2 \ell_{g_y}^2  \notag \\
			& \quad + 34992\eta^2\beta_3^3 \beta_1 c_{g_{xy}}^2\ell_{g_y}^2 \left(1+\frac{1}{\alpha_1 K} \right)\notag \\
			& \leq \eta \beta_1\beta_3^2 \left[ 227 c_{g_{xy}}^2 \left(1 + \frac{1}{\alpha_1 K}\right) + \frac{2400 \ell_{g_y}^2}{\alpha_3 K \mu } \tilde{c}_1+108 \ell_{g_y}^2  +34992 c_{g_{xy}}^2\ell_{g_y}^2 \left(1+\frac{1}{\alpha_1 K} \right)\right]\notag \\
			& \triangleq \eta \beta_1\beta_3^2 \tilde{D}_4 \ , 
	\end{align}
	and 
	\begin{align}
			& D_5 = 9\beta_1\eta(\ell_{f_x}^2+\frac{c_{f_y}^2\ell_{g_{xy}}^2}{\mu^2}) + c_{1}\frac{70\eta\beta_3}{\mu}\Big(\frac{c_{f_y}^2\ell_{g_{yy}}^2}{\mu^2}+ \ell_{f_y}^2\Big) \notag \\
			&  \leq 9\beta_1\eta(\ell_{f_x}^2+\frac{c_{f_y}^2\ell_{g_{xy}}^2}{\mu^2}) + \frac{\beta_1}{\beta_3}\tilde{c}_1 \frac{70\eta\beta_3}{\mu}\Big(\frac{c_{f_y}^2\ell_{g_{yy}}^2}{\mu^2}+ \ell_{f_y}^2\Big) \notag \\
			&  \leq \beta_1\eta \left[9(\ell_{f_x}^2+\frac{c_{f_y}^2\ell_{g_{xy}}^2}{\mu^2}) + \tilde{c}_1 \frac{70}{\mu}\Big(\frac{c_{f_y}^2\ell_{g_{yy}}^2}{\mu^2}+ \ell_{f_y}^2\Big)\right] \notag \\
			& \triangleq \beta_1\eta\tilde{D}_5 \ . 
	\end{align}
	
	Then, to eliminate $\mathbb{E}[\|{X}_{t} - \bar{X}_{t}\|_F^2] $, we enforce
	\begin{align}
			& 9\beta_1\eta(\ell_{f_x}^2+\frac{c_{f_y}^2\ell_{g_{xy}}^2}{\mu^2})+  c_{0}\frac{25\beta_2 \eta \ell_{g_y}^2}{3\mu}+c_{1}\frac{70\eta\beta_3}{\mu}\Big(\frac{c_{f_y}^2\ell_{g_{yy}}^2}{\mu^2} + \ell_{f_y}^2\Big)+  8\eta^2D_1 \notag \\
			& \quad \quad   + 9D_4 \Big(\frac{c_{f_y}^2\ell_{g_{yy}}^2}{\mu^2} + \ell_{f_y}^2\Big) -c_{2}\frac{\eta(1-\lambda^2)}{2}  \notag \\
			& \leq 9\beta_1\eta(\ell_{f_x}^2+\frac{c_{f_y}^2\ell_{g_{xy}}^2}{\mu^2})+  \frac{\beta_1}{\beta_2}\tilde{c}_0\frac{25\beta_2 \eta \ell_{g_y}^2}{3\mu}+\frac{\beta_1}{\beta_3}\tilde{c}_1\frac{70\eta\beta_3}{\mu}\Big(\frac{c_{f_y}^2\ell_{g_{yy}}^2}{\mu^2} + \ell_{f_y}^2\Big)+  8\eta^2\frac{\beta_1}{\eta}\tilde{D}_1  \notag \\
			& \quad \quad   + 9\eta \beta_1\beta_3^2 \tilde{D}_4 \Big(\frac{c_{f_y}^2\ell_{g_{yy}}^2}{\mu^2} + \ell_{f_y}^2\Big) -c_{2}\frac{\eta(1-\lambda^2)}{2} \notag \\
			& \leq 9\beta_1\eta(\ell_{f_x}^2+\frac{c_{f_y}^2\ell_{g_{xy}}^2}{\mu^2})+  \tilde{c}_0\frac{25\beta_1 \eta \ell_{g_y}^2}{3\mu}+\tilde{c}_1\frac{70\eta\beta_1}{\mu}\Big(\frac{c_{f_y}^2\ell_{g_{yy}}^2}{\mu^2} + \ell_{f_y}^2\Big)+  8\eta\beta_1\tilde{D}_1  \notag \\
			& \quad \quad   + 9\eta \beta_1\beta_3^2 \tilde{D}_4 \Big(\frac{c_{f_y}^2\ell_{g_{yy}}^2}{\mu^2} + \ell_{f_y}^2\Big) -c_{2}\frac{\eta(1-\lambda^2)}{2}  \notag \\
			&  \leq  0 \ . 
	\end{align}
	Therefore, we can set
		\begin{align}
			& c_{2} =  \frac{2\beta_1}{(1-\lambda^2)} \left(9\left(\ell_{f_x}^2+\frac{c_{f_y}^2\ell_{g_{xy}}^2}{\mu^2}\right)+\left(\frac{70\tilde{c}_1}{\mu}+9  \tilde{D}_4 \right)\Big(\frac{c_{f_y}^2\ell_{g_{yy}}^2}{\mu^2} + \ell_{f_y}^2\Big)+  \tilde{c}_0\frac{25  \ell_{g_y}^2}{3\mu}+  8\tilde{D}_1  \right) \ . 
	\end{align}
	
		To eliminate $\mathbb{E}[\|{Y}_{t} - \bar{Y}_{t}\|_F^2] $, we enforce
	\begin{align}
			& c_{0} \frac{25\beta_2 \eta \ell_{g_y}^2}{3\mu}+ 8\eta^2D_2+8\eta^2\left(D_y+ D_z \frac{108\eta^2\beta_3^2}{1-\lambda}(\ell_{f_y}^2+\frac{c_{f_y}^2\ell_{g_{yy}}^2}{\mu^2})\right) \notag \\
			& \quad +D_5+  9D_4 \Big(\frac{c_{f_y}^2\ell_{g_{yy}}^2}{\mu^2}+ \ell_{f_y}^2\Big)-c_{3}\frac{\eta(1-\lambda^2)}{2} \notag \\
			& \leq  \frac{\beta_1}{\beta_2}\tilde{c}_0 \frac{25\beta_2 \eta \ell_{g_y}^2}{3\mu}+ 8\eta^2\frac{\beta_1}{\eta}\tilde{D}_2 +8\eta^2\left( \frac{\beta_1}{\eta} \tilde{D}_y+ 2916\beta_1\eta\beta_3c_{g_{xy}}^2 \left(1+ \frac{1}{\alpha_1 K}\right)  (\ell_{f_y}^2+\frac{c_{f_y}^2\ell_{g_{yy}}^2}{\mu^2})\right)\notag \\
			& \quad +\beta_1\eta\tilde{D}_5+  9\eta \beta_1\beta_3^2 \tilde{D}_4 \Big(\frac{c_{f_y}^2\ell_{g_{yy}}^2}{\mu^2}+ \ell_{f_y}^2\Big)-c_{3}\frac{\eta(1-\lambda^2)}{2} \leq  0 \ . 
	\end{align}
	Therefore, due to $\beta_3\leq 1-\lambda$ and $\eta\leq 1$, we can set
		\begin{align}
			& c_{3} = \frac{2\beta_1}{(1-\lambda^2)}\Bigg(\tilde{c}_0 \frac{25  \ell_{g_y}^2}{3\mu}+ 8\tilde{D}_2 +8\left( \tilde{D}_y+ 2916c_{g_{xy}}^2\left(1+ \frac{1}{\alpha_1 K}\right)  \left(\ell_{f_y}^2+\frac{c_{f_y}^2\ell_{g_{yy}}^2}{\mu^2}\right)\right)\notag \\
			& \quad +\tilde{D}_5+  9  \tilde{D}_4 \Big(\frac{c_{f_y}^2\ell_{g_{yy}}^2}{\mu^2}+ \ell_{f_y}^2\Big)\Bigg) \ . 
	\end{align}
	
	 To eliminate $\mathbb{E}[\|{Z}_{t} - \bar{Z}_{t}\|_F^2] $, we enforce
	\begin{align}
			& \quad c_{1}\Big(\frac{70\eta\beta_3}{\mu}\ell_{g_y}^2+  \frac{9}{4}\Big)+8\eta^2D_3+ 8D_z\eta^2+9\beta_1\eta c_{g_{xy}}^2+ 9D_4 \ell_{g_y}^2-c_{4}\frac{\eta(1-\lambda^2)}{2} \notag \\
			& \leq \frac{\beta_1}{\beta_3}\tilde{c}_1\Big(\frac{70\eta\beta_3}{\mu}\ell_{g_y}^2+  \frac{9}{4}\Big)+8\eta^2\frac{\beta_1}{\eta}\tilde{D}_3+ 8\frac{27\beta_1c_{g_{xy}}^2}{\eta}\left(1+ \frac{1}{\alpha_1 K}\right) \eta^2+9\beta_1\eta c_{g_{xy}}^2\notag \\
			& \quad + 9\eta \beta_1\beta_3^2 \tilde{D}_4 \ell_{g_y}^2-c_{4}\frac{\eta(1-\lambda^2)}{2} \leq  0\ . 
	\end{align}
	Therefore, due to $\beta_3\leq 1-\lambda$, we can set
	\begin{align} \label{eq:c-4-alt}
			& c_{4}  = \frac{2\beta_1}{(1-\lambda^2)}\Bigg(\frac{70\ell_{g_y}^2}{\mu}\tilde{c}_1+8\tilde{D}_3+216c_{g_{xy}}^2\left(1+ \frac{1}{\alpha_1 K}\right) +9 c_{g_{xy}}^2 + 9   \ell_{g_y}^2 \tilde{D}_4\Bigg) + \frac{9\beta_1}{2\eta\beta_3(1-\lambda^2)}\tilde{c}_1 \ .
	\end{align}
	
	In summary, by setting
	\begin{align}  \label{eq:coefficient-alt}
			& c_0 = \frac{\beta_1}{\beta_2} \tilde{c}_0 \ , \notag \\
			&  \tilde{c}_0 \triangleq\frac{36}{\mu}\Big(\ell_{f_x}^2+\frac{c_{f_y}^2\ell_{g_{xy}}^2}{\mu^2}\Big) + \frac{60}{\mu}\Big(\frac{\ell_{g_{yy}}^2c_{f_y}^2}{\mu^2}+ \ell_{f_y}^2\Big) \Bigg( \left(227c_{g_{xy}}^2 + 34992c_{g_{xy}}^2\ell_{g_y}^2 \right)   \left(1 + \frac{1}{\alpha_1 K}\right) +108 \ell_{g_y}^2 \Bigg)\notag \\
			&\quad \quad +\frac{4}{\mu} \Big(\frac{18}{\mu}+ \frac{36000 \ell_{g_y}^2}{\alpha_3 K \mu } \Big)\Big(\frac{c_{f_y}^2\ell_{g_{yy}}^2}{\mu^2}+\ell_{f_y}^2\Big)\Bigg(  \frac{ (54480  +8748\times 960 \ell_{g_y}^2)\ell_{g_y}^2c_{g_{xy}}^2}{\mu }   \left(1 + \frac{1}{\alpha_1K}\right) \notag \\ 
			& \quad \quad \quad +  \frac{144  c_{g_{xy}}^2+25920    \ell_{g_y}^4}{\mu}   \Bigg)  \ ,  \notag \\
			& c_{1}  =  \frac{\beta_1}{\beta_3}  \tilde{c}_1  \ , \notag \\
			&   \tilde{c}_1  \triangleq  \frac{144  c_{g_{xy}}^2}{\mu}+  \frac{ 54480 c_{g_{xy}}^2 \ell_{g_y}^2}{\mu }   \left(1 + \frac{1}{\alpha_1K}\right) + \frac{25920    \ell_{g_y}^4}{\mu}   +   \frac{8748\times 960 c_{g_{xy}}^2\ell_{g_y}^4}{\mu} \left(1+\frac{1}{\alpha_1 K} \right)  \ ,   \notag \\
			& c_{2} =  \frac{2\beta_1}{(1-\lambda^2)}\tilde{c}_2 \ ,  \notag \\
			& \quad \tilde{c}_2 \triangleq9\left(\ell_{f_x}^2+\frac{c_{f_y}^2\ell_{g_{xy}}^2}{\mu^2}\right)+\left(\frac{70\tilde{c}_1}{\mu}+9  \tilde{D}_4 \right)\left(\frac{c_{f_y}^2\ell_{g_{yy}}^2}{\mu^2} + \ell_{f_y}^2\right)+  \tilde{c}_0\frac{25  \ell_{g_y}^2}{3\mu}+  8\tilde{D}_1   \ , \notag \\
			& c_{3} = \frac{2\beta_1}{(1-\lambda^2)} \tilde{c}_3 \ , \notag \\
			& \tilde{c}_3 \triangleq 8\left( \tilde{D}_y+ 2916c_{g_{xy}}^2\left(1+ \frac{1}{\alpha_1 K}\right)  \left(\ell_{f_y}^2+\frac{c_{f_y}^2\ell_{g_{yy}}^2}{\mu^2}\right)\right)  \notag \\
			& \quad \quad +\tilde{c}_0 \frac{25  \ell_{g_y}^2}{3\mu}+ 8\tilde{D}_2 +\tilde{D}_5+  9  \tilde{D}_4 \Big(\frac{c_{f_y}^2\ell_{g_{yy}}^2}{\mu^2}+ \ell_{f_y}^2\Big) \ , \notag \\
			& c_{4}  =  \frac{2\beta_1}{(1-\lambda^2)}\tilde{c}_{4} + \frac{9\beta_1}{2\eta\beta_3(1-\lambda^2)}\tilde{c}_1   \ , \notag \\
			&  \tilde{c}_{4} \triangleq \frac{70\ell_{g_y}^2}{\mu}\tilde{c}_1+8\tilde{D}_3+216c_{g_{xy}}^2\left(1+ \frac{1}{\alpha_1 K}\right) +9 c_{g_{xy}}^2 + 9  \ell_{g_y}^2 \tilde{D}_4 \ , \notag \\
			& c_5 = c_{6} = c_7 = \beta_1(1-\lambda) \ ,  \notag \\
			& c_8 = \frac{\beta_1}{\alpha_1\eta} ,\quad  c_9 = 3\beta_1 , \quad  c_{10} =   \frac{25\beta_2 }{3\alpha_2 \eta\mu}   c_{0}  = \frac{25\beta_1 }{3\alpha_2 \eta\mu}  \tilde{c}_0  \ ,   \notag \\
			&c_{11}=  3 \beta_1  + \frac{36\beta_1\beta_2^2}{1-\lambda}\left(\tilde{D}_y+  2916 c_{g_{xy}}^2\left(1+ \frac{1}{\alpha_1 K}\right)   \left(\ell_{f_y}^2+\frac{c_{f_y}^2\ell_{g_{yy}}^2}{\mu^2}\right)\right)  \ , \notag \\
			& c_{12} =   \frac{100\beta_1}{\alpha_3 \eta \mu}  \tilde{c}_1  \ , \notag \\
			& c_{13} = 3\beta_1 +  972\beta_1 c_{g_{xy}}^2\left(1+\frac{1}{\alpha_1 K} \right)  \  , 
	\end{align}
	and
		\begin{align}  \label{eq:coefficient-alt-D}
			\small
			& D_z   \leq  \frac{27\beta_1c_{g_{xy}}^2}{\eta}\left(1+ \frac{1}{\alpha_1 K}\right)   \ , \notag \\
			& D_y  \leq  \frac{\beta_1}{\eta} \tilde{D}_y \ , \notag \\
			& \quad  \tilde{D}_y \triangleq 27 \left(1+\frac{1}{\alpha_1K} \right)\left(\ell_{f_x}^2+\frac{c_{f_y}^2\ell_{g_{xy}}^2}{\mu^2})\right)   + 18\left(2+324  c_{g_{xy}}^2\left(1+\frac{1}{\alpha_1 K} \right)  \right)\left(\ell_{f_y}^2+\frac{c_{f_y}^2\ell_{g_{yy}}^2}{\mu^2}\right)  \notag \\
			& \quad \quad  + \frac{600}{\alpha_3 K  \mu } \left(\frac{c_{f_y}^2\ell_{g_{yy}}^2}{\mu^2}+\ell_{f_y}^2\right)\tilde{c}_1 \ ,  \notag \\
			& D_1  \leq\frac{\beta_1}{\eta}\tilde{D}_{1} \ ,  \notag \\
			&   \tilde{D}_{1}  \triangleq 27\left(1+ \frac{1}{\alpha_1 K}\right)\left(\ell_{f_x}^2+\frac{c_{f_y}^2\ell_{g_{xy}}^2}{\mu^2}\right) \notag \\
			& \quad + 6\left(5 + (1458 + 64152\ell_{g_y}^2) c_{g_{xy}}^2\left(1+\frac{1}{\alpha_1 K} \right) \right)\left(\frac{c_{f_y}^2\ell_{g_{yy}}^2}{\mu^2}+\ell_{f_y}^2\right) \notag \\
			& \quad \quad + \frac{50 \ell_{g_y}^2}{3\alpha_2K \mu }   \tilde{c}_0 +  \frac{600}{\alpha_3 K \mu} \left(\frac{c_{f_y}^2\ell_{g_{yy}}^2}{\mu^2}+\ell_{f_y}^2\right) \tilde{c}_1 + 132 \ell_{g_y}^2\tilde{D}_y+ 15\ell_{g_y}^2   \ ,  \notag \\
			&   D_2 \leq \frac{\beta_1}{\eta}\tilde{D}_2 \ ,   \notag \\
			&  \tilde{D}_2 \triangleq  15 \ell_{g_y}^2+  \frac{50  \ell_{g_y}^2}{3\alpha_2 K\mu}   \tilde{c}_0  + 132\ell_{g_y}^2\left(\tilde{D}_y+  2916c_{g_{xy}}^2\left(1+ \frac{1}{\alpha_1 K}\right)   \left(\ell_{f_y}^2+\frac{c_{f_y}^2\ell_{g_{yy}}^2}{\mu^2}\right)\right) \ ,  \notag \\
			& D_3  \leq \frac{\beta_1}{\eta} \ , \quad  \tilde{D}_3 \triangleq 27 \ell_{g_y}^2 + \frac{600\ell_{g_y}^2 }{\alpha_3  K\mu} \frac{}{}\tilde{c}_1 +  8748c_{g_{xy}}^2\ell_{g_y}^2 \left(1+\frac{1}{\alpha_1 K} \right)  \ , \notag \\
			& D_4  \leq  \eta \beta_1\beta_3^2 \tilde{D}_4  \ ,   \notag \\
			&   \tilde{D}_4 \triangleq 227 c_{g_{xy}}^2 \left(1 + \frac{1}{\alpha_1 K}\right) + \frac{2400 \ell_{g_y}^2}{\alpha_3 K \mu } \tilde{c}_1+108 \ell_{g_y}^2  +34992 c_{g_{xy}}^2\ell_{g_y}^2 \left(1+\frac{1}{\alpha_1 K} \right)  \ , \notag \\
			& D_5   \leq \beta_1\eta\tilde{D}_5 , \quad \tilde{D}_5 \triangleq 9(\ell_{f_x}^2+\frac{c_{f_y}^2\ell_{g_{xy}}^2}{\mu^2}) + \tilde{c}_1 \frac{70}{\mu}\Big(\frac{c_{f_y}^2\ell_{g_{yy}}^2}{\mu^2}+ \ell_{f_y}^2\Big) \ , 
	\end{align}
	we can obtain
	\begin{align}
		\small
			&   \mathcal{L}_{t+1} -  \mathcal{L}_{t}  \leq  - \frac{\beta_1\eta}{2}\mathbb{E}[\| \nabla F(\bar{x}_{t})\|^2]  +\Bigg(c_{0}\frac{25\eta\beta_1^2L_{y}^2 }{6\beta_2\mu}+  c_{1}\frac{9\eta\beta_1^2L_z^2}{\beta_3\mu} +4\eta^2\beta_1^2D_1-  \frac{\beta_1\eta}{4}\Bigg) \mathbb{E}[\|\bar{u}_{t}\| ^2 ] \notag \\
			& \quad    +\Bigg(9\beta_1\beta_2^2\eta^3(\ell_{f_x}^2+\frac{c_{f_y}^2\ell_{g_{xy}}^2}{\mu^2} ) + c_{1} \frac{18\eta^3\beta_2^2\beta_3}{\mu}\Big(\frac{c_{f_y}^2\ell_{g_{yy}}^2}{\mu^2}+\ell_{f_y}^2\Big) \notag \\
			& \qquad + 8\eta^2\beta_2^2\left(D_y+ D_z \frac{108\eta^2\beta_3^2}{1-\lambda}(\ell_{f_y}^2+\frac{c_{f_y}^2\ell_{g_{yy}}^2}{\mu^2})\right)\notag \\
			& \quad \quad +4\eta^2\beta_2^2D_2+ 15D_4 \beta_2^2\eta^2\Big(\frac{\ell_{g_{yy}}^2c_{f_y}^2}{\mu^2}+ \ell_{f_y}^2\Big)-c_{0} \frac{3\eta\beta_2^2}{4}\Bigg) \mathbb{E}[\|\bar{v}_{t}  \|^2]  \notag \\
			& \quad  +\Bigg( c_{3}\frac{2\eta \beta_2^2}{1-\lambda^2}+ 4\eta^2\beta_2^2D_2+\frac{20\eta^2 \beta_2^2}{1-\lambda^2}\left(D_y+ D_z \frac{108\eta^2\beta_3^2}{1-\lambda}(\ell_{f_y}^2+\frac{c_{f_y}^2\ell_{g_{yy}}^2}{\mu^2})\right) \notag \\
			& \quad \quad + D_5\frac{2\eta \beta_2^2}{1-\lambda^2} +D_4  \frac{18\eta \beta_2^2}{1-\lambda^2} \Big(\frac{c_{f_y}^2\ell_{g_{yy}}^2}{\mu^2}+ \ell_{f_y}^2\Big)- c_{6}(1-\lambda)\Bigg)\frac{1}{K}\mathbb{E}[\|Q_{t}- \bar{ Q}_{t}\|_F^2]  \ ,\notag \\ 
			& \quad + \Bigg(c_{4}\frac{2\eta \beta_3^2}{1-\lambda^2}+c_{1}\frac{9\eta\beta_3^2}{4}+ 4\eta^2\beta_3^2D_3+ D_z \frac{20\eta^3 \beta_3^2}{1-\lambda^2}+ 9\beta_1\eta c_{g_{xy}}^2\frac{2\eta \beta_3^2}{1-\lambda^2} \notag \\
			& \qquad - c_{7}(1-\lambda)\Bigg)\frac{1}{K}\mathbb{E}[\|R_{t}- \bar{R}_{t}\|_F^2]  \notag \\
			& \quad  + \Bigg(c_{2}\frac{2\eta \beta_1^2}{1-\lambda^2} + 4\eta^2\beta_1^2D_1-c_{5}(1-\lambda)\Bigg)\frac{1}{K}\mathbb{E}[\| P_{t} -  \bar{P}_{t} \|_F^2] \notag \\
			& \quad + \left(D_y+ D_z \frac{108\eta^2\beta_3^2}{1-\lambda}(\ell_{f_y}^2+\frac{c_{f_y}^2\ell_{g_{yy}}^2}{\mu^2})\right) \frac{48\alpha^2_2\beta_2^2\eta^6}{1-\lambda}\sigma^2 + c_{7} \frac{6\alpha_3^2\eta^4}{1-\lambda}(1 + \frac{c_{f_y}^2}{\mu^2})\sigma^2 \notag \\
			& \quad +  D_z\frac{72\alpha^2_3\beta_3^2\eta^6}{1-\lambda}(1 + \frac{c_{f_y}^2}{\mu^2})\sigma^2 \notag \\
			& \quad + 4c_{8}\alpha_1^2\eta^4 (1 + \frac{c_{f_y}^2}{\mu^2})\sigma^2\frac{1}{K}  + 4c_{9}\alpha_1^2\eta^4 (1 + \frac{c_{f_y}^2}{\mu^2})\sigma^2  + 2c_{10}\alpha_2^2\eta^4 \sigma^2\frac{1}{K}  + 2c_{11}\alpha_2^2\eta^4 \sigma^2 \ .\notag \\
			& \quad  + 4c_{12}\alpha_3^2\eta^4(1 + \frac{c_{f_y}^2}{\mu^2})\sigma^2\frac{1}{K} + 4c_{13}\alpha_3^2\eta^4(1 + \frac{c_{f_y}^2}{\mu^2})\sigma^2 + c_{6} \frac{6\alpha_2^2\eta^4}{1-\lambda}\sigma^2+  c_{5}\frac{6\alpha_1^2\eta^4}{1-\lambda}(1 + \frac{c_{f_y}^2}{\mu^2})\sigma^2  \ .  
	\end{align}
	
	Then, to eliminate $\mathbb{E}[\| P_{t} -  \bar{P}_{t} \|_F^2]$, we enforce
	\begin{align}
			& c_{2}\frac{2\eta \beta_1^2}{1-\lambda^2} + 4\eta^2\beta_1^2D_1-c_{5}(1-\lambda) \notag \\
			& \leq  \frac{2\beta_1}{(1-\lambda^2)}\tilde{c}_{2} \frac{2\eta \beta_1^2}{1-\lambda^2} + 4\eta^2\beta_1^2\frac{\beta_1}{\eta}\tilde{D}_1 - \beta_1(1-\lambda)(1-\lambda) \notag \\
			& \leq  \frac{2\beta_1}{(1-\lambda^2)}\tilde{c}_{2} \frac{2 \beta_1^2}{1-\lambda^2} + 4\beta_1^2\beta_1\tilde{D}_1 - \beta_1(1-\lambda)(1-\lambda) \notag \\
			& \leq  0 \  , 
	\end{align}
	where the second step follows from $\eta\leq 1$.  Then, we enforce
		\begin{align}
			&  \frac{2\beta_1}{(1-\lambda^2)}\tilde{c}_{2} \frac{2 \beta_1^2}{1-\lambda^2} \leq \frac{1}{2}\beta_1(1-\lambda)^2  , \notag \\
			& 4\beta_1^2\beta_1\tilde{D}_1 \leq  \frac{1}{2}\beta_1(1-\lambda)^2  \  . 
	\end{align}
	We can obtain
	\begin{align}
		\beta_1 \leq \min\left\{\frac{(1-\lambda)^2}{3\sqrt{\tilde{c}_{2}}} ,\frac{1-\lambda}{3\sqrt{\tilde{D}_{1}}} \right\} \ . 
	\end{align}
	
	To eliminate $\mathbb{E}[\| Q_{t} -  \bar{Q}_{t} \|_F^2]$, we enforce
	\begin{align}
			& c_{3}\frac{2\eta \beta_2^2}{1-\lambda^2}+ 4\eta^2\beta_2^2D_2+\frac{20\eta^2 \beta_2^2}{1-\lambda^2}\left(D_y+ D_z \frac{108\eta^2\beta_3^2}{1-\lambda}(\ell_{f_y}^2+\frac{c_{f_y}^2\ell_{g_{yy}}^2}{\mu^2})\right)\notag \\
			& \quad \quad + D_5\frac{2\eta \beta_2^2}{1-\lambda^2} +D_4  \frac{18\eta \beta_2^2}{1-\lambda^2} \Big(\frac{c_{f_y}^2\ell_{g_{yy}}^2}{\mu^2}+ \ell_{f_y}^2\Big)- c_{6}(1-\lambda) \notag \\
			& \leq \frac{2\beta_1}{(1-\lambda^2)}\tilde{c}_{3}\frac{2\eta \beta_2^2}{1-\lambda^2}+ 4\eta^2\beta_2^2\frac{\beta_1}{\eta}\tilde{D}_2 \notag \\
			& \quad +\frac{20\eta^2 \beta_2^2}{1-\lambda^2}\left( \frac{\beta_1}{\eta} \tilde{D}_y+  \frac{27\beta_1c_{g_{xy}}^2}{\eta}\left(1+ \frac{1}{\alpha_1 K}\right)  \frac{108\eta^2\beta_3^2}{1-\lambda}(\ell_{f_y}^2+\frac{c_{f_y}^2\ell_{g_{yy}}^2}{\mu^2})\right)\notag \\
			& \quad  + \beta_1\eta\tilde{D}_5\frac{2\eta \beta_2^2}{1-\lambda^2} +\eta \beta_1\beta_3^2 \tilde{D}_4  \frac{18\eta \beta_2^2}{1-\lambda^2} \Big(\frac{c_{f_y}^2\ell_{g_{yy}}^2}{\mu^2}+ \ell_{f_y}^2\Big)- \beta_1(1-\lambda)(1-\lambda) \notag \\
			& \leq \frac{2\beta_1}{(1-\lambda^2)}\tilde{c}_{3}\frac{2 \beta_2^2}{1-\lambda^2}+ 4\beta_2^2\beta_1\tilde{D}_2 \notag \\
			& \quad +\frac{20 \beta_2^2}{1-\lambda^2}\left( \beta_1\tilde{D}_y+ 2916\beta_1{g_{xy}}^2\left(1+ \frac{1}{\alpha_1 K}\right)  (\ell_{f_y}^2+\frac{c_{f_y}^2\ell_{g_{yy}}^2}{\mu^2})\right)\notag \\
			& \quad  + \beta_1\tilde{D}_5\frac{2\eta \beta_2^2}{1-\lambda^2} + \beta_1 \tilde{D}_4  \frac{18 \beta_2^2}{1-\lambda^2} \Big(\frac{c_{f_y}^2\ell_{g_{yy}}^2}{\mu^2}+ \ell_{f_y}^2\Big)- \beta_1(1-\lambda)^2 \notag \\
			& \leq  0  \ ,  
	\end{align}
	where the second step follows from $\eta \leq 1$ and $\beta_3\leq 1-\lambda$. Then, we enforce
	\begin{align}
			& \frac{2\beta_1}{(1-\lambda^2)}\tilde{c}_{3}\frac{2 \beta_2^2}{1-\lambda^2} \leq \frac{1}{5} \beta_1(1-\lambda)^2 \ ,   \notag \\
			&  4\beta_1\beta_2^2\tilde{D}_2 \leq \frac{1}{5} \beta_1(1-\lambda)^2  \ ,  \notag \\
			& \frac{20 \beta_2^2}{1-\lambda^2}\left( \beta_1\tilde{D}_y+ 2916\beta_1{g_{xy}}^2\left(1+ \frac{1}{\alpha_1 K}\right)  (\ell_{f_y}^2+\frac{c_{f_y}^2\ell_{g_{yy}}^2}{\mu^2})\right) \leq \frac{1}{5} \beta_1(1-\lambda)^2  \ , \notag \\
			&  \beta_1\tilde{D}_5\frac{2\eta \beta_2^2}{1-\lambda^2}  \leq \frac{1}{5} \beta_1(1-\lambda)^2  \ ,  \notag \\
			&  \beta_1 \tilde{D}_4  \frac{18 \beta_2^2}{1-\lambda^2} \Big(\frac{c_{f_y}^2\ell_{g_{yy}}^2}{\mu^2}+ \ell_{f_y}^2\Big)\leq \frac{1}{5} \beta_1(1-\lambda)^2  \ . 
	\end{align}
	We can obtain
	\begin{align}
	& 	\beta_2 \leq \min \Bigg\{ \frac{(1-\lambda)^2}{5\sqrt{\tilde{c}_{3}}} , \frac{1-\lambda}{5\sqrt{\tilde{D}_2 }},  \frac{(1-\lambda)^{3/2}}{10\sqrt{\tilde{D}_y+ 2916 c_{g_{xy}}^2 \left(1+ \frac{1}{\alpha_1 K}\right) \left(\ell_{f_y}^2+\frac{c_{f_y}^2\ell_{g_{yy}}^2}{\mu^2}\right)}}, \notag \\
		&  \frac{(1-\lambda)^{3/2}}{4\sqrt{ \tilde{D}_5}},  \frac{(1-\lambda)^{3/2}}{10\sqrt{\tilde{D}_4\Big(\frac{c_{f_y}^2\ell_{g_{yy}}^2}{\mu^2}+ \ell_{f_y}^2\Big)} }    \Bigg\} \ . 
	\end{align}

		To eliminate $\mathbb{E}[\| R_{t} -  \bar{R}_{t} \|_F^2]$, we enforce
	\begin{align}
			& c_{4}\frac{2\eta \beta_3^2}{1-\lambda^2}+c_{1}\frac{9\eta\beta_3^2}{4}+ 4\eta^2\beta_3^2D_3+ D_z \frac{20\eta^3 \beta_3^2}{1-\lambda^2}+ 9\beta_1\eta c_{g_{xy}}^2\frac{2\eta \beta_3^2}{1-\lambda^2}- c_{7}(1-\lambda) \notag \\
			& \leq  \frac{2\beta_1}{(1-\lambda^2)}\tilde{c}_{4} \frac{2\eta \beta_3^2}{1-\lambda^2}+ \frac{9\beta_1}{2\eta\beta_3(1-\lambda^2)}\tilde{c}_1 \frac{2\eta \beta_3^2}{1-\lambda^2}+ \frac{\beta_1}{\beta_3}\tilde{c}_1 \frac{9\eta\beta_3^2}{4}+ 4\eta^2\beta_3^2\frac{\beta_1}{\eta}\tilde{D}_3\notag \\
			& \quad +   \frac{27\beta_1c_{g_{xy}}^2}{\eta}\left(1+ \frac{1}{\alpha_1 K}\right)   \frac{20\eta^3 \beta_3^2}{1-\lambda^2}+ 9\beta_1\eta c_{g_{xy}}^2\frac{2\eta \beta_3^2}{1-\lambda^2}- \beta_1(1-\lambda)(1-\lambda) \notag \\
			& \leq  \frac{2\beta_1}{(1-\lambda^2)}\tilde{c}_{4} \frac{2 \beta_3^2}{1-\lambda^2}+ \frac{9\beta_1}{(1-\lambda^2)}\tilde{c}_1 \frac{ \beta_3}{1-\lambda^2}+ \beta_1\tilde{c}_1 \frac{9\beta_3}{4}+ 4\beta_3^2\beta_1\tilde{D}_3\notag \\
			& \quad +   27\beta_1c_{g_{xy}}^2\left(1+ \frac{1}{\alpha_1 K}\right)   \frac{20\beta_3^2}{1-\lambda^2}+ 9\beta_1 c_{g_{xy}}^2\frac{2 \beta_3^2}{1-\lambda^2}- \beta_1(1-\lambda)(1-\lambda) \notag \\
			& \leq 0 \ , 
	\end{align}
	where the second step follows from $\eta\leq 1$. Then, we enforce
	
	\begin{align}
			& \frac{2\beta_1}{(1-\lambda^2)}\tilde{c}_{4} \frac{2 \beta_3^2}{1-\lambda^2} \leq \frac{1}{6}\beta_1(1-\lambda)^2 \ ,  \notag \\
			& \frac{9\beta_1}{(1-\lambda^2)}\tilde{c}_1 \frac{ \beta_3}{1-\lambda^2} \leq \frac{1}{6}\beta_1(1-\lambda)^2 \ ,  \notag \\
			& \beta_1\tilde{c}_1 \frac{9\beta_3}{4} \leq \frac{1}{6}\beta_1(1-\lambda)^2  \ , \notag \\
			& 4\beta_3^2\beta_1\tilde{D}_3 \leq \frac{1}{6}\beta_1(1-\lambda)^2  \ , \notag \\
			& 27\beta_1c_{g_{xy}}^2\left(1+ \frac{1}{\alpha_1 K}\right)  \frac{20\beta_3^2}{1-\lambda^2}\leq \frac{1}{6}\beta_1(1-\lambda)^2 \ ,  \notag \\
			& 9\beta_1 c_{g_{xy}}^2\frac{2 \beta_3^2}{1-\lambda^2}\leq \frac{1}{6}\beta_1(1-\lambda)^2 \ . 
	\end{align}
	We can obtain 
	\begin{align}
			& \beta_3\leq \min\left\{\frac{(1-\lambda)^2}{5\sqrt{\tilde{c}_{4}}} , \frac{(1-\lambda)^4}{54\tilde{c}_1 }, \frac{2(1-\lambda)^2}{27\tilde{c}_1 }, \frac{1-\lambda}{5\sqrt{\tilde{D}_3}}, \frac{(1-\lambda)^{3/2} }{60c_{g_{xy}}\sqrt{1+ \frac{1}{\alpha_1 K}}}, \frac{(1-\lambda)^{3/2}}{11c_{g_{xy}}}  \right\} \ . 
	\end{align}
	
	To eliminate $ \mathbb{E}[\|\bar{u}_{t}\| ^2 ]$, we enforce 
	\begin{align}
			& \quad c_{0}\frac{25\eta\beta_1^2L_{y}^2 }{6\beta_2\mu}+  c_{1}\frac{9\eta\beta_1^2L_z^2}{\beta_3\mu} +4\eta^2\beta_1^2D_1-  \frac{\beta_1\eta}{4} \notag \\
			& \leq  \frac{\beta_1}{\beta_2} \tilde{c}_0\frac{25\eta\beta_1^2L_{y}^2 }{6\beta_2\mu}+  \frac{\beta_1}{\beta_3}\tilde{c}_1\frac{9\eta\beta_1^2L_z^2}{\beta_3\mu} +4\eta^2\beta_1^2 \frac{\beta_1}{\eta}\tilde{D}_{1}-  \frac{\beta_1\eta}{4} \notag \\
			&  \leq 0  \ . 
	\end{align}
	Then, we enforce
		\begin{align}
			& \frac{\beta_1}{\beta_2} \tilde{c}_0\frac{25\eta\beta_1^2L_{y}^2 }{6\beta_2\mu} \leq \frac{\beta_1\eta}{12}  \ , \notag \\  
			& \frac{\beta_1}{\beta_3}\tilde{c}_1\frac{9\eta\beta_1^2L_z^2}{\beta_3\mu} \leq \frac{\beta_1\eta}{12} \ ,  \notag \\  
			& 4\eta^2\beta_1^2 \frac{\beta_1}{\eta}\tilde{D}_{1}\leq   \frac{\beta_1\eta}{4} \ . 
	\end{align}
	We can obtain
	\begin{align}
		\beta_1 \leq \min \left\{\frac{\beta_2\sqrt{\mu} }{8 L_{y}\sqrt{ \tilde{c}_0}}, \frac{\beta_3\sqrt{\mu}}{11L_z \sqrt{\tilde{c}_1}},  \frac{1}{4\sqrt{\tilde{D}_{1}}} \right\} \ . 
	\end{align}
	
	To eliminate $ \mathbb{E}[\|\bar{v}_{t}\| ^2 ]$, we enforce 
	\begin{align}
			& 9\beta_1\beta_2^2\eta^3(\ell_{f_x}^2+\frac{c_{f_y}^2\ell_{g_{xy}}^2}{\mu^2} ) + c_{1} \frac{18\eta^3\beta_2^2\beta_3}{\mu}\Big(\frac{c_{f_y}^2\ell_{g_{yy}}^2}{\mu^2}+\ell_{f_y}^2\Big) \notag \\
			& \quad+ 8\eta^2\beta_2^2\left(D_y+ D_z \frac{108\eta^2\beta_3^2}{1-\lambda}(\ell_{f_y}^2+\frac{c_{f_y}^2\ell_{g_{yy}}^2}{\mu^2})\right)\notag \\
			& \quad  +4\eta^2\beta_2^2D_2+ 15D_4 \beta_2^2\eta^2\Big(\frac{\ell_{g_{yy}}^2c_{f_y}^2}{\mu^2}+ \ell_{f_y}^2\Big)-c_{0} \frac{3\eta\beta_2^2}{4} \notag \\
			& \leq 9\beta_1\beta_2^2\eta^3(\ell_{f_x}^2+\frac{c_{f_y}^2\ell_{g_{xy}}^2}{\mu^2} ) +  \frac{\beta_1}{\beta_3}\tilde{c}_1 \frac{18\eta^3\beta_2^2\beta_3}{\mu}\Big(\frac{c_{f_y}^2\ell_{g_{yy}}^2}{\mu^2}+\ell_{f_y}^2\Big) \notag \\
			& \quad + 8\eta^2\beta_2^2\left( \frac{\beta_1}{\eta} \tilde{D}_y + \frac{27\beta_1c_{g_{xy}}^2}{\eta}\left(1+ \frac{1}{\alpha_1 K}\right)  \frac{108\eta^2\beta_3^2}{1-\lambda}(\ell_{f_y}^2+\frac{c_{f_y}^2\ell_{g_{yy}}^2}{\mu^2})\right)\notag \\
			& \quad  +4\eta^2\beta_2^2 \frac{\beta_1}{\eta}\tilde{D}_2+ 15\eta \beta_1\beta_3^2 \tilde{D}_4\beta_2^2\eta^2\Big(\frac{\ell_{g_{yy}}^2c_{f_y}^2}{\mu^2}+ \ell_{f_y}^2\Big)-  \tilde{c}_0 \frac{3\eta\beta_1\beta_2}{4} \notag \\
			& \leq 9\beta_1\beta_2^2\eta\left(\ell_{f_x}^2+\frac{c_{f_y}^2\ell_{g_{xy}}^2}{\mu^2} \right) +  \tilde{c}_1 \frac{18\eta\beta_1\beta_2^2}{\mu}\Big(\frac{c_{f_y}^2\ell_{g_{yy}}^2}{\mu^2}+\ell_{f_y}^2\Big) \notag \\
			& \quad + 8\eta\beta_2^2\left( \beta_1\tilde{D}_y +2916\beta_1 c_{g_{xy}}^2\left(1+ \frac{1}{\alpha_1 K}\right)  (\ell_{f_y}^2+\frac{c_{f_y}^2\ell_{g_{yy}}^2}{\mu^2})\right)\notag \\
			& \quad  +4\eta^2\beta_2^2 \frac{\beta_1}{\eta}\tilde{D}_2+ 15\eta \beta_1\beta_2^2 \tilde{D}_4\Big(\frac{\ell_{g_{yy}}^2c_{f_y}^2}{\mu^2}+ \ell_{f_y}^2\Big)-  \tilde{c}_0 \frac{3\eta\beta_1\beta_2}{4} \notag \\
			&  \leq 0 \ . 
	\end{align}
	where the second step follows from $\eta\leq 1$ and $\beta_3\leq 1-\lambda\leq 1$. 
	Then, we enforce
		\begin{align}
			& 9\beta_1\beta_2^2\eta(\ell_{f_x}^2+\frac{c_{f_y}^2\ell_{g_{xy}}^2}{\mu^2} )  \leq  \tilde{c}_0 \frac{3\eta\beta_1\beta_2}{20}  , \notag \\ 
			& \tilde{c}_1 \frac{18\eta\beta_1\beta_2^2}{\mu}\Big(\frac{c_{f_y}^2\ell_{g_{yy}}^2}{\mu^2}+\ell_{f_y}^2\Big) \leq  \tilde{c}_0 \frac{3\eta\beta_1\beta_2}{20}  \ , \notag \\
			& 8\eta\beta_2^2\left( \beta_1\tilde{D}_y +2916\beta_1 c_{g_{xy}}^2\left(1+ \frac{1}{\alpha_1 K}\right)  (\ell_{f_y}^2+\frac{c_{f_y}^2\ell_{g_{yy}}^2}{\mu^2})\right) \leq  \tilde{c}_0 \frac{3\eta\beta_1\beta_2}{20}  \ , \notag \\
			& 4\eta^2\beta_2^2 \frac{\beta_1}{\eta}\tilde{D}_2\leq \tilde{c}_0 \frac{3\eta\beta_1\beta_2}{20} \ ,  \notag \\
& 15\eta \beta_1 \beta_2^2 \tilde{D}_4\Big(\frac{\ell_{g_{yy}}^2c_{f_y}^2}{\mu^2}+ \ell_{f_y}^2\Big) \leq  \tilde{c}_0 \frac{3\eta\beta_1\beta_2}{20} \ . 
	\end{align}
	
	As a result, we can obtain
	\begin{align}
		\small
		& \beta_2 \leq \min\Bigg\{ \frac{\tilde{c}_0}{12\left(\ell_{f_x}^2+\frac{c_{f_y}^2\ell_{g_{xy}}^2}{\mu^2} \right)} , \frac{\tilde{c}_0}{\tilde{c}_1} \frac{\mu}{120\Big(\frac{c_{f_y}^2\ell_{g_{yy}}^2}{\mu^2}+\ell_{f_y}^2\Big)} ,  \frac{3\tilde{c}_0}{80 \tilde{D}_2},  \frac{\tilde{c}_0}{100\tilde{D}_4\Big(\frac{\ell_{g_{yy}}^2c_{f_y}^2}{\mu^2}+ \ell_{f_y}^2\Big) } , \notag \\
		& \quad\quad \quad \quad \quad  \frac{3 \tilde{c}_0 }{160\left(  \tilde{D}_y +2916 c_{g_{xy}}^2\left(1+ \frac{1}{\alpha_1 K}\right)  \left(\ell_{f_y}^2+\frac{c_{f_y}^2\ell_{g_{yy}}^2}{\mu^2}\right) \right) }  \Bigg\} \ . 
	\end{align}
	
	In summary, by setting
	\begin{align} \label{eq:hyper-alt}
		\small
			& \beta_1 \leq \min \left\{\frac{(1-\lambda)^2}{3\sqrt{\tilde{c}_{2}}} ,\frac{1-\lambda}{3\sqrt{\tilde{D}_{1}}}, \frac{\beta_2\sqrt{\mu} }{8 L_{y}\sqrt{ \tilde{c}_0}}, \frac{\beta_3\sqrt{\mu}}{11L_z \sqrt{\tilde{c}_1}},  \frac{1}{4\sqrt{\tilde{D}_{1}}}  \right\} \  , \notag \\
			& \beta_2 \leq \min\Bigg\{ \frac{(1-\lambda)^2}{5\sqrt{\tilde{c}_{3}}} , \frac{1-\lambda}{5\sqrt{\tilde{D}_2 }},  \frac{(1-\lambda)^{3/2}}{10\sqrt{\tilde{D}_y+ 2916 c_{g_{xy}}^2 \left(1+ \frac{1}{\alpha_1 K}\right) \left(\ell_{f_y}^2+\frac{c_{f_y}^2\ell_{g_{yy}}^2}{\mu^2}\right)}}, \frac{(1-\lambda)^{3/2}}{4\sqrt{ \tilde{D}_5}},  \notag \\
			& \quad \frac{(1-\lambda)^{3/2}}{10\sqrt{\tilde{D}_4\Big(\frac{c_{f_y}^2\ell_{g_{yy}}^2}{\mu^2}+ \ell_{f_y}^2\Big)} } ,  \frac{\tilde{c}_0}{12\left(\ell_{f_x}^2+\frac{c_{f_y}^2\ell_{g_{xy}}^2}{\mu^2} \right)} , \frac{\tilde{c}_0}{\tilde{c}_1} \frac{\mu}{120\Big(\frac{c_{f_y}^2\ell_{g_{yy}}^2}{\mu^2}+\ell_{f_y}^2\Big)} , 1-\lambda, \frac{1}{6\ell_{g_y}}  ,  \notag \\
			& \quad \frac{3 \tilde{c}_0 }{160\left(  \tilde{D}_y +2916 c_{g_{xy}}^2\left(1+ \frac{1}{\alpha_1 K}\right)  \left(\ell_{f_y}^2+\frac{c_{f_y}^2\ell_{g_{yy}}^2}{\mu^2}\right) \right) },   \frac{3\tilde{c}_0}{80 \tilde{D}_2},  \frac{\tilde{c}_0}{100\tilde{D}_4\Big(\frac{\ell_{g_{yy}}^2c_{f_y}^2}{\mu^2}+ \ell_{f_y}^2\Big) }  \Bigg\} \  , \notag \\
			& \beta_3\leq \min\Bigg\{\frac{(1-\lambda)^2}{5\sqrt{\tilde{c}_{4}}} , \frac{(1-\lambda)^4}{54\tilde{c}_1 }, \frac{2(1-\lambda)^2}{27\tilde{c}_1 }, \frac{1-\lambda}{5\sqrt{\tilde{D}_3}}, \frac{(1-\lambda)^{3/2} }{60c_{g_{xy}}\sqrt{1+ \frac{1}{\alpha_1 K}}}, \frac{(1-\lambda)^{3/2}}{11c_{g_{xy}}} , \notag \\
			& \qquad \frac{\sqrt{\alpha_3K}}{20\ell_{g_y}}  , \frac{\mu\sqrt{\alpha_3 K} }{800  \ell_{g_y}^2}, 1-\lambda \Bigg\} \ ,  \notag \\
			& \eta \leq \min\left\{\frac{1}{2\beta_1 L_F},  \frac{1}{\beta_3\mu}, \frac{1}{\sqrt{\alpha_1}},  \frac{1}{\sqrt{\alpha_2}},  \frac{1}{\sqrt{\alpha_3}}, 1\right\} \ , 
	\end{align}
	we can obtain
	\begin{align}
			& \quad  \mathcal{L}_{t+1} -  \mathcal{L}_{t} \notag \\
& \leq  - \frac{\beta_1\eta}{2}\mathbb{E}[\| \nabla F(\bar{x}_{t})\|^2]  \notag \\
& \quad + \left( \tilde{D}_y+ \frac{2916c_{g_{xy}}^2\beta_3^2}{1-\lambda}\left(1+ \frac{1}{\alpha_1 K}\right)  \left(\ell_{f_y}^2+\frac{c_{f_y}^2\ell_{g_{yy}}^2}{\mu^2}\right)\right) \frac{48\beta_1\alpha^2_2\beta_2^2\eta^5}{1-\lambda}\sigma^2 \notag \\
& \quad  +  6\beta_1\alpha_3^2\eta^4\left(1 + \frac{c_{f_y}^2}{\mu^2}\right)\sigma^2 +   \frac{1944\beta_1\alpha^2_3\beta_3^2\eta^5c_{g_{xy}}^2}{1-\lambda}\left(1+ \frac{1}{\alpha_1 K}\right) \left(1 + \frac{c_{f_y}^2}{\mu^2}\right)\sigma^2 \notag \\
& \quad + 4 \beta_1\alpha_1\eta^3 \left(1 + \frac{c_{f_y}^2}{\mu^2}\right)\frac{\sigma^2}{K}  + 12\beta_1\alpha_1^2\eta^4 \left(1 + \frac{c_{f_y}^2}{\mu^2}\right)\sigma^2  + \tilde{c}_0\frac{50\beta_1 \alpha_2\eta^3}{3 \mu}   \frac{\sigma^2}{K}  \notag \\
& \quad + 2\beta_1\alpha_2^2\eta^4  \left( 3   + \frac{36\beta_2^2}{1-\lambda}\left(\tilde{D}_y+  2916 c_{g_{xy}}^2\left(1+ \frac{1}{\alpha_1 K}\right)   \left(\ell_{f_y}^2+\frac{c_{f_y}^2\ell_{g_{yy}}^2}{\mu^2}\right)\right)\right)\sigma^2\notag \\
& \quad  + \tilde{c}_1\frac{400\beta_1\alpha_3\eta^3}{  \mu}  \left(1 + \frac{c_{f_y}^2}{\mu^2}\right)\frac{\sigma^2}{K} + 4\beta_1\alpha_3^2\eta^4\left(1 + \frac{c_{f_y}^2}{\mu^2}\right)\left(3 +  972\beta_3 c_{g_{xy}}^2\left(1+\frac{1}{\alpha_1 K} \right) \right)\sigma^2 \notag \\
& \quad + 6\beta_1\alpha_2^2\eta^4\sigma^2 +  6\beta_1\alpha_1^2\eta^4\left(1 + \frac{c_{f_y}^2}{\mu^2}\right)\sigma^2\ . 
	\end{align}
	By summing over $t$ from $0$ to $T-1$, we can obtain
		\begin{align}
			& \quad \frac{1}{T}\sum_{t=0}^{T-1}\mathbb{E}[\| \nabla F(\bar{x}_{t})\|^2] \notag \\
& \leq  \frac{2(\mathcal{L}_{0} -  \mathcal{L}_{T})}{\beta_1\eta T} + \left( \tilde{D}_y+ \frac{2916c_{g_{xy}}^2\beta_3^2}{1-\lambda}\left(1+ \frac{1}{\alpha_1 K}\right)  \left(\ell_{f_y}^2+\frac{c_{f_y}^2\ell_{g_{yy}}^2}{\mu^2}\right)\right) \frac{96\alpha^2_2\beta_2^2\eta^4}{1-\lambda}\sigma^2 \notag \\
& \quad  +  12\alpha_3^2\eta^3\left(1 + \frac{c_{f_y}^2}{\mu^2}\right)\sigma^2 +   \frac{3888\alpha^2_3\beta_3^2\eta^4c_{g_{xy}}^2}{1-\lambda}\left(1+ \frac{1}{\alpha_1 K}\right) \left(1 + \frac{c_{f_y}^2}{\mu^2}\right)\sigma^2 \notag \\
& \quad + 8 \alpha_1\eta^2 \left(1 + \frac{c_{f_y}^2}{\mu^2}\right)\frac{\sigma^2}{K}  + 24\alpha_1^2\eta^3 \left(1 + \frac{c_{f_y}^2}{\mu^2}\right)\sigma^2  + \tilde{c}_0\frac{100 \alpha_2\eta^2}{3 \mu}   \frac{\sigma^2}{K}  \notag \\
& \quad +4\alpha_2^2\eta^3  \left( 3   + \frac{36\beta_2^2}{1-\lambda}\left(\tilde{D}_y+  2916 c_{g_{xy}}^2\left(1+ \frac{1}{\alpha_1 K}\right)   \left(\ell_{f_y}^2+\frac{c_{f_y}^2\ell_{g_{yy}}^2}{\mu^2}\right)\right)\right)\sigma^2\notag \\
& \quad  +\tilde{c}_1\frac{800\alpha_3\eta^2}{  \mu}  \left(1 + \frac{c_{f_y}^2}{\mu^2}\right)\frac{\sigma^2}{K} + 8\alpha_3^2\eta^3\left(1 + \frac{c_{f_y}^2}{\mu^2}\right)\left(3 +  972\beta_3 c_{g_{xy}}^2\left(1+\frac{1}{\alpha_1 K} \right) \right)\sigma^2 \notag \\
& \quad +12\alpha_2^2\eta^3\sigma^2 + 12\alpha_1^2\eta^3\left(1 + \frac{c_{f_y}^2}{\mu^2}\right)\sigma^2\ . 
		\end{align}

For the initialization step, due to $x_{0}^{(k)}=x_{0}$, $y_{0}^{(k)}=y_{0}$, $z_{0}^{(k)}=z_{0}$,  we have


\begin{align}
		& \mathcal{L}_{0} =   {\mathbb{E}}[F(\bar{x}_{0})] +  c_0\mathbb{E}[\|\bar{   {y}}_{0} -    {y}^{*}(\bar{   {x}}_{0})\| ^2 ]  +  c_1\mathbb{E}[\|\bar{ z}_{0} -    {z}^{*}(\bar{   {x}}_{0})\| ^2 ]   \notag \\
		& \quad + c_5\frac{1}{K} \mathbb{E}[\|P_{0} - \bar{P}_{0}\|_F^2 ] + c_6 \frac{1}{K}\mathbb{E}[\|Q_{0} - \bar{Q}_{0}\|_F^2 ]  +c_7\frac{1}{K}\mathbb{E}[\| R_{0}-\bar{R}_{0} \|_F^2] \notag \\
		& \quad + c_8 \mathbb{E} [ \|\frac{1}{K}\delta^{\hat{\mathcal{G}}_{F}}(X_{0}, Y_{1}, Z_{1}) \mathbf{1}  -  \frac{1}{K} U_{0} \mathbf{1} \|^2 ]+ c_9 \frac{1}{K}\mathbb{E} [ \|\delta^{\hat{\mathcal{G}}_{F}}(X_{0}, Y_{1}, Z_{1}) -  U_{0} \|_F^2 ] \notag \\
		& \quad  + c_{10}\mathbb{E} [ \|\frac{1}{K}\delta^{g}(X_{0}, Y_{0}) \mathbf{1} -  \frac{1}{K} V_{0} \mathbf{1} \|^2 ] + c_{11} \frac{1}{K}\mathbb{E} [ \|\delta^{g}(X_{0}, Y_{0}) - V_{0}  \|_F^2 ]\notag \\
		& \quad  + c_{12} \mathbb{E} [ \|\frac{1}{K}\delta^{\hat{\mathcal{G}}_{h}}(X_{0}, Y_{1}, Z_{0})\mathbf{1}  -  \frac{1}{K} W_{0} \mathbf{1} \|^2 ] +   c_{13} \frac{1}{K}\mathbb{E} [ \|\delta^{\hat{\mathcal{G}}_{h}}(X_{0}, Y_{1}, Z_{0}) -   W_{0}  \|_F^2 ] \ .
\end{align}

As for $\mathbb{E}[\|P_{0} - \bar{P}_{0}\|_F^2 ]$, we have
\begin{align}
		& \quad\frac{1}{K} \mathbb{E}[\|P_{0} - \bar{P}_{0}\|_F^2 ]  \notag \\
		& = \frac{1}{K} \sum_{k=1}^{K}\mathbb{E}[\|	\hat{\mathcal{G}}_{F}^{(k)}(x_{0}, y_{1}^{(k)}, z_{1}^{(k)}; \hat{\xi}_{0}^{(k)})- 	\frac{1}{K} \sum_{k'=1}^{K}\hat{\mathcal{G}}_{F}^{(k')}(x_{0}, y_{1}^{(k')}, z_{1}^{(k')}; \hat{\xi}_{0}^{(k')})\|^2 ]  \notag \\
& \leq 2\frac{1}{K} \sum_{k=1}^{K}\mathbb{E}[\|\nabla_{1} { f^{(k)}(x_{0}, y_{1}^{(k)}; \xi_{0}^{(k)})}   - 	\frac{1}{K} \sum_{k'=1}^{K}\nabla_{1} { f^{(k')}(x_{0}, y_{1}^{(k')}; \xi_{0}^{(k')})}  \|^2 ]  \notag \\
& \quad + 2\frac{1}{K} \sum_{k=1}^{K}\mathbb{E}[\|\nabla_{12}^2 g^{(k)}(x_{0}, y_{1}^{(k)}; \zeta_{0}^{(k)})z_{1}^{(k)}  - \frac{1}{K} \sum_{k'=1}^{K}\nabla_{12}^2 g^{(k')}(x_{0}, y_{1}^{(k')}; \zeta_{0}^{(k')})z_{1}^{(k')} \|^2 ]  \notag \\
& \leq 10\frac{1}{K} \sum_{k=1}^{K}\mathbb{E}[\|\nabla_{1} { f^{(k)}(x_{0}, y_{1}^{(k)}; \xi_{0}^{(k)})}  - \nabla_{1} { f^{(k)}(x_{0}, y_{0}^{(k)}; \xi_{0}^{(k)})}  \|^2]\notag \\
& \quad + 	10\frac{1}{K} \sum_{k=1}^{K}\mathbb{E}[\|\frac{1}{K} \sum_{k'=1}^{K}\nabla_{1} { f^{(k')}(x_{0}, y_{0}^{(k')}; \xi_{0}^{(k')})} - 	\frac{1}{K} \sum_{k'=1}^{K}\nabla_{1} { f^{(k')}(x_{0}, y_{1}^{(k')}; \xi_{0}^{(k')})}  \|^2 ]  \notag \\
& \quad + 10\frac{1}{K} \sum_{k=1}^{K}\mathbb{E}[\|\nabla_{12}^2 g^{(k)}(x_{0}, y_{1}^{(k)}; \zeta_{0}^{(k)})z_{1}^{(k)} - \nabla_{12}^2 g^{(k)}(x_{0}, y_{0}^{(k)}; \zeta_{0}^{(k)})z_{1}^{(k)} \|^2]\notag \\
& \quad + 10\frac{1}{K} \sum_{k=1}^{K}\mathbb{E}[\|\nabla_{12}^2 g^{(k)}(x_{0}, y_{0}^{(k)}; \zeta_{0}^{(k)})z_{1}^{(k)} -\nabla_{12}^2 g^{(k)}(x_{0}, y_{0}^{(k)}; \zeta_{0}^{(k)})z_{0}^{(k)}  \|^2] \notag \\
& \quad  +  10\frac{1}{K} \sum_{k=1}^{K}\mathbb{E}[\|  \frac{1}{K} \sum_{k'=1}^{K}\nabla_{12}^2 g^{(k')}(x_{0}, y_{0}^{(k')}; \zeta_{0}^{(k')})z_{0}^{(k')}   \notag\\
& \qquad-    \frac{1}{K} \sum_{k'=1}^{K}\nabla_{12}^2 g^{(k')}(x_{0}, y_{1}^{(k')}; \zeta_{0}^{(k')})z_{0}^{(k')} \|^2]\notag \\
& \quad   +10\frac{1}{K} \sum_{k=1}^{K}\mathbb{E}[\|  \frac{1}{K} \sum_{k'=1}^{K}\nabla_{12}^2 g^{(k')}(x_{0}, y_{1}^{(k')}; \zeta_{0}^{(k')})z_{0}^{(k')}   \notag\\
& \qquad- \frac{1}{K} \sum_{k'=1}^{K}\nabla_{12}^2 g^{(k')}(x_{0}, y_{1}^{(k')}; \zeta_{0}^{(k')})z_{1}^{(k')} \|^2 ]  \notag \\
& \quad + 10\frac{1}{K} \sum_{k=1}^{K}\mathbb{E}[\|\nabla_{1} { f^{(k)}(x_{0}, y_{0}^{(k)}; \xi_{0}^{(k)})}  - 	\frac{1}{K} \sum_{k'=1}^{K}\nabla_{1} { f^{(k')}(x_{0}, y_{0}^{(k')}; \xi_{0}^{(k')})}  \|^2] \notag \\
& \quad +10\frac{1}{K} \sum_{k=1}^{K}\mathbb{E}[\| \nabla_{12}^2 g^{(k)}(x_{0}, y_{0}^{(k)}; \zeta_{0}^{(k)})z_{0}^{(k)}   - \frac{1}{K} \sum_{k'=1}^{K}\nabla_{12}^2 g^{(k')}(x_{0}, y_{0}^{(k')}; \zeta_{0}^{(k')})z_{0}^{(k')} \|^2 ]\notag \\
& \leq 20\ell_{f_x}^2\frac{1}{K} \sum_{k=1}^{K}\mathbb{E}[\|y_{1}^{(k)} -y_{0}^{(k)}\|^2]  + 20\frac{c_{f_y}^2\ell_{g_{xy}}^2}{\mu^2}\frac{1}{K} \sum_{k=1}^{K}\mathbb{E}[\| y_{1}^{(k)}-y_{0}^{(k)} \|^2]   \notag\\
& \quad+ 20c_{g_{xy}}^2\frac{1}{K} \sum_{k=1}^{K}\mathbb{E}[\|z_{1}^{(k)} -z_{0}^{(k)}  \|^2] \notag \\
& \quad + 10\frac{1}{K} \sum_{k=1}^{K}\mathbb{E}[\|\nabla_{1} { f^{(k)}(x_{0}, y_{0}^{(k)}; \xi_{0}^{(k)})}  - 	\frac{1}{K} \sum_{k'=1}^{K}\nabla_{1} { f^{(k')}(x_{0}, y_{0}^{(k')}; \xi_{0}^{(k')})}  \|^2] \notag \\
& \quad +10\frac{1}{K} \sum_{k=1}^{K}\mathbb{E}[\| \nabla_{12}^2 g^{(k)}(x_{0}, y_{0}^{(k)}; \zeta_{0}^{(k)})z_{0}^{(k)}   - \frac{1}{K} \sum_{k'=1}^{K}\nabla_{12}^2 g^{(k')}(x_{0}, y_{0}^{(k')}; \zeta_{0}^{(k')})z_{0}^{(k')} \|^2 ]\notag \\
		& \leq 20\left(\ell_{f_x}^2+\frac{c_{f_y}^2\ell_{g_{xy}}^2}{\mu^2}\right)\frac{1}{K} \sum_{k=1}^{K}\mathbb{E}[\|y_{1}^{(k)} -y_{0}^{(k)}\|^2]   + 20c_{g_{xy}}^2\frac{1}{K} \sum_{k=1}^{K}\mathbb{E}[\|z_{1}^{(k)} -z_{0}^{(k)}  \|^2] \notag \\
		& \quad + 48(1+  \frac{c_{f_y}^2}{\mu^2})\frac{\sigma^2}{B_0} + \frac{48c_{g_{xy}}^2c_{f_y}^2}{\mu^2} +  48\frac{1}{K} \sum_{k=1}^{K}\mathbb{E}[\|\nabla_{1} { f^{(k)}(x_{0}, y_{0})} \|^2] \ . 
\end{align}

Then, according to the proof of Lemma~\ref{lemma_incremental_x-alt}, we can obtain
	\begin{align}
		& \quad  \mathbb{E}[\|Y_{1}-Y_{0}\|_F^2] \notag \\
		& \leq 8\eta^2\mathbb{E}[\|Y_{0}-\bar{Y}_{0}\|_F^2] + 2\eta^2\beta_2^2\mathbb{E}[\|Q_{0}\|_F^2] \notag \\
		&   =  2\eta^2\beta_2^2\sum_{k=1}^{K}\mathbb{E}[\|\nabla_{2} g^{(k)}(x_0, y_0; \xi_{0}^{(k)})\|^2] \notag \\
		&   =  4\eta^2\beta_2^2\sum_{k=1}^{K}\mathbb{E}[\|\nabla_{2} g^{(k)}(x_0, y_0; \xi_{0}^{(k)}) - \nabla_{2} g^{(k)}(x_0, y_0)\|^2] + 4\eta^2\beta_2^2\sum_{k=1}^{K}\mathbb{E}[\| \nabla_{2} g^{(k)}(x_0, y_0)\|^2] \notag \\
		& \leq 4K\eta^2\beta_2^2\frac{\sigma^2 }{B_0}+ 4\eta^2\beta_2^2\sum_{k=1}^{K}\| \nabla_{2} g^{(k)}(x_0, y_0)\|^2 \ . 
\end{align}
Similarly, we have
\begin{align}
		& \quad  \mathbb{E}[\|Z_{1}-Z_{0}\|_F^2] \notag \\
		& \leq 8\eta^2\mathbb{E}[\|Z_{0}-\bar{Z}_{0}\|_F^2] + 2\eta^2\beta_3^2\mathbb{E}[\|R_{0}\|_F^2]   \notag \\
		& =  2\eta^2\beta_3^2\sum_{k=1}^{K}\mathbb{E}[\|	\hat{\mathcal{G}}_{h}^{(k)}(x_{0}, y_{1}^{(k)}, z_{0}; \hat{\xi}_{0}^{(k)})\|^2]  \notag \\
		& = 2\eta^2\beta_3^2\sum_{k=1}^{K}\mathbb{E}[\|	 \nabla_{22}^2g^{(k)}(x_0, y_{1}^{(k)}; \zeta_{0}^{(k)}) z_0  - \nabla_{2}{ f^{(k)}(x_0, y_{1}^{(k)}; \xi_{0}^{(k)})}   \|^2]  \notag \\
		& \leq  6\eta^2\beta_3^2\sum_{k=1}^{K}\mathbb{E}[\|	 \nabla_{22}^2g^{(k)}(x_0, y_{1}^{(k)}; \zeta_{0}^{(k)}) z_0 - \nabla_{22}^2g^{(k)}(x_0, y_{0}^{(k)}; \zeta_{0}^{(k)}) z_0 \|^2]\notag \\
		& \quad +6\eta^2\beta_3^2\sum_{k=1}^{K}\mathbb{E}[\|	 \nabla_{22}^2g^{(k)}(x_0, y_{0}^{(k)}; \zeta_{0}^{(k)}) z_0 -  \nabla_{2}{ f^{(k)}(x_0, y_{0}^{(k)}; \xi_{0}^{(k)})} \|^2]\notag \\
		& \quad + 6\eta^2\beta_3^2\sum_{k=1}^{K}\mathbb{E}[\|	 \nabla_{2}{ f^{(k)}(x_0, y_{0}^{(k)}; \xi_{0}^{(k)})}  - \nabla_{2}{ f^{(k)}(x_0, y_{1}^{(k)}; \xi_{0}^{(k)})}   \|^2]  \notag \\
		& \leq  6\eta^2\beta_3^2\left(\frac{c_{f_y}^2\ell_{g_{yy}}^2}{\mu^2} + \ell_{f_y}^2\right)\sum_{k=1}^{K}\mathbb{E}[\|	 y_{1}^{(k)}-  y_{0}^{(k)} \|^2] +12K\eta^2\beta_3^2 \left(\frac{c_{f_y}^2\ell_{g_y}^2}{\mu^2} + c_{f_y}^2\right) \notag \\
		& \leq 24K\eta^4\beta_2^2\beta_3^2\left(\frac{c_{f_y}^2\ell_{g_{yy}}^2}{\mu^2} + \ell_{f_y}^2\right)\frac{\sigma^2 }{B_0}+ 24\eta^4\beta_2^2\beta_3^2\left(\frac{c_{f_y}^2\ell_{g_{yy}}^2}{\mu^2} + \ell_{f_y}^2\right)\sum_{k=1}^{K}\| \nabla_{2} g^{(k)}(x_0, y_0)\|^2 \notag \\
		& \quad +12K\eta^2\beta_3^2 \left(\frac{c_{f_y}^2\ell_{g_y}^2}{\mu^2} + c_{f_y}^2\right)  \ . 
\end{align}

As a result, we have
\begin{align}
		& \quad\frac{1}{K} \mathbb{E}[\|P_{0} - \bar{P}_{0}\|_F^2 ]  \notag \\
& \leq  80\eta^2\beta_2^2\left(\ell_{f_x}^2+\frac{c_{f_y}^2\ell_{g_{xy}}^2}{\mu^2}\right)\frac{\sigma^2 }{B_0} +80\eta^2\beta_2^2\left(\ell_{f_x}^2+\frac{c_{f_y}^2\ell_{g_{xy}}^2}{\mu^2}\right)\frac{1}{K} \sum_{k=1}^{K}\| \nabla_{2} g^{(k)}(x_0, y_0)\|^2  \notag \\
& \quad   + 480\eta^4\beta_2^2\beta_3^2c_{g_{xy}}^2\left(\frac{c_{f_y}^2\ell_{g_{yy}}^2}{\mu^2} + \ell_{f_y}^2\right)\frac{\sigma^2 }{B_0}   \notag\\
& \quad+ 480\eta^4\beta_2^2\beta_3^2c_{g_{xy}}^2 \left(\frac{c_{f_y}^2\ell_{g_{yy}}^2}{\mu^2} + \ell_{f_y}^2\right)\frac{1}{K}\sum_{k=1}^{K}\| \nabla_{2} g^{(k)}(x_0, y_0)\|^2 \notag \\
& \quad +240\eta^2\beta_3^2c_{g_{xy}}^2 \left(\frac{c_{f_y}^2\ell_{g_y}^2}{\mu^2} + c_{f_y}^2\right) + 48\left(1+  \frac{c_{f_y}^2}{\mu^2}\right)\frac{\sigma^2}{B_0} + \frac{48c_{g_{xy}}^2c_{f_y}^2}{\mu^2}   \notag\\
& \quad+  48\frac{1}{K} \sum_{k=1}^{K}\mathbb{E}[\|\nabla_{1} { f^{(k)}(x_{0}, y_{0})} \|^2]    \ . 
\end{align}

%

As for $\mathbb{E}[\|Q_{0} - \bar{Q}_{0}\|_F^2 ]$,  it is easy to know that 
\begin{align}
		&  \frac{1}{K} \mathbb{E}[\|Q_{0} - \bar{Q}_{0}\|_F^2 ] \leq \frac{6\sigma^2}{B_0} + 12\frac{1}{K} \sum_{k=1}^{K}\|\nabla_{2} g^{(k)}(x_0, y_0)\|^2  \ . 
\end{align}

As for $\mathbb{E}[\| R_{0}-\bar{R}_{0} \|_F^2]$, we have
\begin{align}
		& \quad \frac{1}{K}\mathbb{E}[\| R_{0}-\bar{R}_{0} \|_F^2] \notag \\
		& =  \frac{1}{K} \sum_{k=1}^{K}\mathbb{E}[\|	\hat{\mathcal{G}}_{h}^{(k)}(x_{0}, y_{1}^{(k)}, z_{0}; \hat{\xi}_{0}^{(k)})- 	\frac{1}{K} \sum_{k'=1}^{K}\hat{\mathcal{G}}_{h}^{(k')}(x_{0}, y_{1}^{(k')}, z_{0}; \hat{\xi}_{0}^{(k')})\|^2 ]  \notag \\
		& \leq 2\frac{1}{K} \sum_{k=1}^{K}\mathbb{E}[\| \nabla_{22}^2g^{(k)}(x_0, y_{1}^{(k)}; \zeta_{0}^{(k)}) z_0 - \frac{1}{K} \sum_{k'=1}^{K}\nabla_{22}^2g^{(k')}(x_0, y_{1}^{(k')}; \zeta_{0}^{(k')}) z_0  \|^2 ]  \notag \\
		& \quad  + 2\frac{1}{K} \sum_{k=1}^{K}\mathbb{E}[\|  \nabla_{2}{ f^{(k)}(x_0, y_{1}^{(k)}; \xi_{0}^{(k)})}  -  \frac{1}{K} \sum_{k'=1}^{K} \nabla_{2}{ f^{(k')}(x_0, y_{1}^{(k')}; \xi_{0}^{(k')})} \|^2 ]  \notag \\
		& \leq 6\frac{1}{K} \sum_{k=1}^{K}\mathbb{E}[\| \nabla_{22}^2g^{(k)}(x_0, y_{1}^{(k)}; \zeta_{0}^{(k)}) z_0 -\nabla_{22}^2g^{(k)}(x_0, y_{0}^{(k)}; \zeta_{0}^{(k)}) z_0  \|^2 ]  \notag \\
		&\quad + 6\frac{1}{K} \sum_{k=1}^{K}\mathbb{E}[\| \nabla_{22}^2g^{(k)}(x_0, y_{0}; \zeta_{0}^{(k)}) z_0 - \frac{1}{K} \sum_{k'=1}^{K}\nabla_{22}^2g^{(k')}(x_0, y_{0}; \zeta_{0}^{(k')}) z_0  \|^2 ]  \notag \\
		&\quad + 6\frac{1}{K} \sum_{k=1}^{K}\mathbb{E}[\| \frac{1}{K} \sum_{k'=1}^{K}\nabla_{22}^2g^{(k')}(x_0, y_{0}^{(k')}; \zeta_{0}^{(k')}) z_0- \frac{1}{K} \sum_{k'=1}^{K}\nabla_{22}^2g^{(k')}(x_0, y_{1}^{(k')}; \zeta_{0}^{(k')}) z_0  \|^2 ]  \notag \\
		& \quad  + 6\frac{1}{K} \sum_{k=1}^{K}\mathbb{E}[\|  \nabla_{2}{ f^{(k)}(x_0, y_{1}^{(k)}; \xi_{0}^{(k)})}  -  \nabla_{2}{ f^{(k)}(x_0, y_{0}^{(k)}; \xi_{0}^{(k)})} \|^2 ]  \notag \\
		& \quad  + 6\frac{1}{K} \sum_{k=1}^{K}\mathbb{E}[\|  \nabla_{2}{ f^{(k)}(x_0, y_{0}; \xi_{0}^{(k)})}  -  \frac{1}{K} \sum_{k'=1}^{K} \nabla_{2}{ f^{(k')}(x_0, y_{0}; \xi_{0}^{(k')})} \|^2 ]  \notag \\
		& \quad  + 6\frac{1}{K} \sum_{k=1}^{K}\mathbb{E}[\|  \frac{1}{K} \sum_{k'=1}^{K} \nabla_{2}{ f^{(k')}(x_0, y_{0}^{(k')}; \xi_{0}^{(k')})} -  \frac{1}{K} \sum_{k'=1}^{K} \nabla_{2}{ f^{(k')}(x_0, y_{1}^{(k')}; \xi_{0}^{(k')})} \|^2 ]  \notag \\
		& \leq 12\left(\frac{c_{f_y}^2\ell_{g_{yy}}^2}{\mu^2} + \ell_{f_y}^2\right)\frac{1}{K} \sum_{k=1}^{K}\mathbb{E}[\|  y_{1}^{(k)} - y_{0}^{(k)} \|^2 ]  \notag \\
		&\quad + 6\frac{1}{K} \sum_{k=1}^{K}\mathbb{E}[\| \nabla_{22}^2g^{(k)}(x_0, y_{0}; \zeta_{0}^{(k)}) z_0 - \frac{1}{K} \sum_{k'=1}^{K}\nabla_{22}^2g^{(k')}(x_0, y_{0}; \zeta_{0}^{(k')}) z_0  \|^2 ]  \notag \\
		& \quad  + 6\frac{1}{K} \sum_{k=1}^{K}\mathbb{E}[\|  \nabla_{2}{ f^{(k)}(x_0, y_{0}; \xi_{0}^{(k)})}  -  \frac{1}{K} \sum_{k'=1}^{K} \nabla_{2}{ f^{(k')}(x_0, y_{0}; \xi_{0}^{(k')})} \|^2 ]  \notag \\
		& \leq 36\left(1+\frac{c_{f_y}^2}{\mu^2}\right)\frac{\sigma^2}{B_0}+ \frac{72\ell_{g_y}^2c_{f_y}^2}{\mu^2} + 72 c_{f_y}^2 \notag \\
		& \quad + 48\eta^2\beta_2^2\left(\frac{c_{f_y}^2\ell_{g_{yy}}^2}{\mu^2} + \ell_{f_y}^2\right)\frac{\sigma^2 }{B_0}+48\eta^2\beta_2^2\left(\frac{c_{f_y}^2\ell_{g_{yy}}^2}{\mu^2} + \ell_{f_y}^2\right)\frac{1}{K}   \sum_{k=1}^{K}\| \nabla_{2} g^{(k)}(x_0, y_0)\|^2 \ . 
\end{align}

On the other hand, we have
\begin{align}
		&\quad  \mathbb{E} [ \|\frac{1}{K}\delta^{\hat{\mathcal{G}}_{F}}(X_0, Y_1, Z_1) \mathbf{1}-  \frac{1}{K} U_0 \mathbf{1} \|^2 ] \notag \\
		& = \mathbb{E} [ \|\frac{1}{K} \sum_{k=1}^{K}\hat{\mathcal{G}}_{F}^{(k)}(x_{0}, y_{1}^{(k)}, z_{1}^{(k)})-  \frac{1}{K} \sum_{k=1}^{K}\hat{\mathcal{G}}_{F}^{(k)}(x_{0}, y_{1}^{(k)}, z_{1}^{(k)}; \hat{\xi}_{0}^{(k)}) \|^2 ] \notag \\
		& =  \mathbb{E} [ \|\frac{1}{K} \sum_{k=1}^{K}\nabla_{1} { f^{(k)}(x_{0}, y_{1}^{(k)})}   -\frac{1}{K} \sum_{k=1}^{K}\nabla_{12}^2 g^{(k)}(x_{0}, y_{1}^{(k)})z_{1}^{(k)}\notag \\
		& \quad  - \frac{1}{K} \sum_{k=1}^{K}\nabla_{1} { f^{(k)}(x_{0}, y_{1}^{(k)}; \xi_{0}^{(k)})}   + \frac{1}{K} \sum_{k=1}^{K}\nabla_{12}^2 g^{(k)}(x_{0}, y_{1}^{(k)}; \zeta_{0}^{(k)})z_{1}^{(k)}\|^2 ] \notag \\
		& \leq 6\mathbb{E} [ \|\frac{1}{K} \sum_{k=1}^{K}\nabla_{1} { f^{(k)}(x_{0}, y_{1}^{(k)})} - \frac{1}{K} \sum_{k=1}^{K}\nabla_{1} { f^{(k)}(x_{0}, y_{1}^{(k)}; \xi_{0}^{(k)})} \|^2 ] \notag \\
		& \quad  + 6\mathbb{E} [ \|  \frac{1}{K} \sum_{k=1}^{K}\nabla_{12}^2 g^{(k)}(x_{0}, y_{1}^{(k)})z_{1}^{(k)} - \frac{1}{K} \sum_{k=1}^{K}\nabla_{12}^2 g^{(k)}(x_{0}, y_{0}^{(k)})z_{1}^{(k)}  \|^2 ] \notag \\
		& \quad + 6\mathbb{E} [ \|\frac{1}{K} \sum_{k=1}^{K}\nabla_{12}^2 g^{(k)}(x_{0}, y_{0}^{(k)})z_{1}^{(k)} -\frac{1}{K} \sum_{k=1}^{K}\nabla_{12}^2 g^{(k)}(x_{0}, y_{0}^{(k)})z_{0}^{(k)}  \|^2 ] \notag \\
		& \quad   + 6\mathbb{E} [ \|\frac{1}{K} \sum_{k=1}^{K}\nabla_{12}^2 g^{(k)}(x_{0}, y_{0}^{(k)})z_{0}^{(k)} -\frac{1}{K} \sum_{k=1}^{K}\nabla_{12}^2 g^{(k)}(x_{0}, y_{0}^{(k)}; \zeta_{0}^{(k)})z_{0}^{(k)} \|^2 ] \notag \\
		& \quad  + 6\mathbb{E} [ \|\frac{1}{K} \sum_{k=1}^{K}\nabla_{12}^2 g^{(k)}(x_{0}, y_{0}^{(k)}; \zeta_{0}^{(k)})z_{0}^{(k)} -  \frac{1}{K} \sum_{k=1}^{K}\nabla_{12}^2 g^{(k)}(x_{0}, y_{0}^{(k)}; \zeta_{0}^{(k)})z_{1}^{(k)}  \|^2 ] \notag \\
		& \quad + 6\mathbb{E} [ \| \frac{1}{K} \sum_{k=1}^{K}\nabla_{12}^2 g^{(k)}(x_{0}, y_{0}^{(k)}; \zeta_{0}^{(k)})z_{1}^{(k)} - \frac{1}{K} \sum_{k=1}^{K}\nabla_{12}^2 g^{(k)}(x_{0}, y_{1}^{(k)}; \zeta_{0}^{(k)})z_{1}^{(k)}\|^2 ] \notag \\
		& \leq 6\mathbb{E} [ \|\frac{1}{K} \sum_{k=1}^{K}\nabla_{1} { f^{(k)}(x_{0}, y_{1}^{(k)})} - \frac{1}{K} \sum_{k=1}^{K}\nabla_{1} { f^{(k)}(x_{0}, y_{1}^{(k)}; \xi_{0}^{(k)})} \|^2 ] \notag \\
				& \quad   + 6\mathbb{E} [ \|\frac{1}{K} \sum_{k=1}^{K}\nabla_{12}^2 g^{(k)}(x_{0}, y_{0}^{(k)})z_{0}^{(k)} -\frac{1}{K} \sum_{k=1}^{K}\nabla_{12}^2 g^{(k)}(x_{0}, y_{0}^{(k)}; \zeta_{0}^{(k)})z_{0}^{(k)} \|^2 ] \notag \\
		& \quad  + 12\frac{c_{f_y}^2\ell_{g_{xy}}^2}{\mu^2} \frac{1}{K} \sum_{k=1}^{K}\mathbb{E} [ \|  y_{1}^{(k)} -y_{0}^{(k)} \|^2 ] + 12c_{g_{xy}}^2\frac{1}{K} \sum_{k=1}^{K}\mathbb{E} [ \|z_{1}^{(k)} -z_{0}^{(k)}  \|^2 ] \notag \\
		& \leq 6(1+\frac{c_{f_y^2}}{\mu^2})\frac{\sigma^2}{KB_0} +  48\eta^2\beta_2^2\frac{c_{f_y}^2\ell_{g_{xy}}^2}{\mu^2} \frac{\sigma^2 }{B_0}+ 48\eta^2\beta_2^2\frac{c_{f_y}^2\ell_{g_{xy}}^2}{\mu^2} \frac{1}{K}\sum_{k=1}^{K}\| \nabla_{2} g^{(k)}(x_0, y_0)\|^2 \notag \\
		& \quad + 288\eta^4\beta_2^2\beta_3^2c_{g_{xy}}^2\left(\frac{c_{f_y}^2\ell_{g_{yy}}^2}{\mu^2} + \ell_{f_y}^2\right)\frac{\sigma^2 }{B_0}  \notag\\
		& \quad+ 288\eta^4\beta_2^2\beta_3^2c_{g_{xy}}^2 \left(\frac{c_{f_y}^2\ell_{g_{yy}}^2}{\mu^2} + \ell_{f_y}^2\right)\frac{1}{K}\sum_{k=1}^{K}\| \nabla_{2} g^{(k)}(x_0, y_0)\|^2 \notag \\
		& \quad +244\eta^2\beta_3^2 c_{g_{xy}}^2\left(\frac{c_{f_y}^2\ell_{g_y}^2}{\mu^2} + c_{f_y}^2\right)  \ , 
\end{align}
and 
\begin{align}
		& \quad \frac{1}{K}\mathbb{E} [ \|\delta^{\hat{\mathcal{G}}_{F}}(X_0, Y_0, Z_0) -  U_0 \|_F^2 ] \notag \\
		& \leq 6(1+\frac{c_{f_y^2}}{\mu^2})\frac{\sigma^2}{B_0} +  48\eta^2\beta_2^2\frac{c_{f_y}^2\ell_{g_{xy}}^2}{\mu^2} \frac{\sigma^2 }{B_0}+ 48\eta^2\beta_2^2\frac{c_{f_y}^2\ell_{g_{xy}}^2}{\mu^2} \frac{1}{K}\sum_{k=1}^{K}\| \nabla_{2} g^{(k)}(x_0, y_0)\|^2 \notag \\
		& \quad + 288\eta^4\beta_2^2\beta_3^2c_{g_{xy}}^2\left(\frac{c_{f_y}^2\ell_{g_{yy}}^2}{\mu^2} + \ell_{f_y}^2\right)\frac{\sigma^2 }{B_0}  \notag\\
		& \quad+ 288\eta^4\beta_2^2\beta_3^2c_{g_{xy}}^2 \left(\frac{c_{f_y}^2\ell_{g_{yy}}^2}{\mu^2} + \ell_{f_y}^2\right)\frac{1}{K}\sum_{k=1}^{K}\| \nabla_{2} g^{(k)}(x_0, y_0)\|^2 \notag \\
		& \quad +244\eta^2\beta_3^2 c_{g_{xy}}^2\left(\frac{c_{f_y}^2\ell_{g_y}^2}{\mu^2} + c_{f_y}^2\right)  \ . 
\end{align}


Moreover, we have
\begin{align}
		&\quad  \mathbb{E} [ \|\frac{1}{K}\delta^{\hat{\mathcal{G}}_{h}}(X_0, Y_1, Z_0) \mathbf{1}-  \frac{1}{K} W_0 \mathbf{1} \|^2 ] \notag \\
		& = \mathbb{E} [ \|\frac{1}{K} \sum_{k=1}^{K}\hat{\mathcal{G}}_{h}^{(k)}(x_{0}, y_{1}^{(k)}, z_{0})-  \frac{1}{K} \sum_{k=1}^{K}\hat{\mathcal{G}}_{h}^{(k)}(x_{0}, y_{1}^{(k)}, z_{0}; \hat{\xi}_{0}^{(k)}) \|^2 ] \notag \\
		& =  \mathbb{E} [ \|\frac{1}{K} \sum_{k=1}^{K}\nabla_{22}^2g^{(k)}(x_{0}, y_{1}^{(k)}) z_{0} -  \frac{1}{K} \sum_{k=1}^{K}\nabla_{2}{ f^{(k)}(x_{0}, y_{1}^{(k)})}   \notag \\
		& \quad  - \frac{1}{K} \sum_{k=1}^{K}\nabla_{22}^2g^{(k)}(x_{0}, y_{1}^{(k)}; \zeta_{0}^{(k)}) z_{0} +  \frac{1}{K} \sum_{k=1}^{K}\nabla_{2}{ f^{(k)}(x_{0}, y_{1}^{(k)};  \xi_{0}^{(k)})}  \|^2 ] \notag \\
		& \leq 2\mathbb{E} [ \|\frac{1}{K} \sum_{k=1}^{K}\nabla_{2} { f^{(k)}(x_{0}, y_{1}^{(k)})} - \frac{1}{K} \sum_{k=1}^{K}\nabla_{2} { f^{(k)}(x_{0}, y_{1}^{(k)}; \xi_{0}^{(k)})} \|^2 ] \notag \\
		& \quad  + 2\mathbb{E} [ \|  \frac{1}{K} \sum_{k=1}^{K}\nabla_{22}^2 g^{(k)}(x_{0}, y_{1}^{(k)})z_{0}  - \frac{1}{K} \sum_{k=1}^{K}\nabla_{22}^2 g^{(k)}(x_{0}, y_{1}^{(k)}; \zeta_{0}^{(k)})z_{0}\|^2 ] \notag \\
		& \leq 2(1+\frac{c_{f_y^2}}{\mu^2})\frac{\sigma^2}{KB_0}  \ , 
\end{align}
and $\frac{1}{K}\mathbb{E} [ \|\delta^{\hat{\mathcal{G}}_{h}}(X_0, Y_1, Z_0) -  W_0 \|_F^2 ]\leq 2(1+\frac{c_{f_y^2}}{\mu^2})\frac{\sigma^2}{B_0}$. 

Similarly, we have
\begin{align}
		& \mathbb{E} [ \|\frac{1}{K}\delta^{g}(X_0, Y_0) \mathbf{1} -  \frac{1}{K} V_0 \mathbf{1} \|^2 ] \leq \frac{\sigma^2}{KB_0}  \ , \notag \\
		& \frac{1}{K}\mathbb{E} [ \|\delta^{g}(X_0, Y_0) - V_0  \|_F^2 ] \leq \frac{\sigma^2}{B_0} \ . 
\end{align}

By combining them together, we can obtain
\begin{align}
& \frac{2}{\beta_1\eta T} \mathcal{L}_{0}  \leq  \frac{2}{\beta_1\eta T} F({x}_{0})+    \frac{2}{\eta \beta_2T}\tilde{c}_0\mathbb{E}[\|\bar{   {y}}_{0} -    {y}^{*}(\bar{   {x}}_{0})\| ^2 ]  +  \frac{2}{\beta_3\eta T}  \tilde{c}_1\mathbb{E}[\|\bar{ z}_{0} -    {z}^{*}(\bar{   {x}}_{0})\| ^2 ]   \notag \\
& \quad +\Bigg[ \frac{160\eta\beta_2^2}{ T} \left(\ell_{f_x}^2+\frac{c_{f_y}^2\ell_{g_{xy}}^2}{\mu^2}\right)+ \frac{960\eta^3\beta_2^2\beta_3^2c_{g_{xy}}^2}{ T}\left(\frac{c_{f_y}^2\ell_{g_{yy}}^2}{\mu^2} + \ell_{f_y}^2\right)+ \frac{24}{\eta T} \notag \\
& \quad \quad + \frac{96\eta\beta_2^2}{ T}\left(\frac{c_{f_y}^2\ell_{g_{yy}}^2}{\mu^2} + \ell_{f_y}^2\right)+\frac{96\beta_2^2}{ \alpha_1T} \frac{c_{f_y}^2\ell_{g_{xy}}^2}{\mu^2}+ \frac{576}{ \alpha_1T} \eta^2\beta_2^2\beta_3^2c_{g_{xy}}^2 \left(\frac{c_{f_y}^2\ell_{g_{yy}}^2}{\mu^2} + \ell_{f_y}^2\right)\notag \\
& \quad \quad + \frac{288\eta\beta_2^2}{ T}\frac{c_{f_y}^2\ell_{g_{xy}}^2}{\mu^2}+ \frac{1728\eta^3\beta_2^2\beta_3^2c_{g_{xy}}^2}{ T} \left(\frac{c_{f_y}^2\ell_{g_{yy}}^2}{\mu^2} + \ell_{f_y}^2\right)\Bigg]\frac{1}{K} \sum_{k=1}^{K}\| \nabla_{2} g^{(k)}(x_0, y_0)\|^2  \notag \\
& \quad +   \frac{96}{\eta T}\frac{1}{K} \sum_{k=1}^{K}\mathbb{E}[\|\nabla_{1} { f^{(k)}(x_{0}, y_{0})} \|^2]   +  \frac{160\eta\beta_2^2}{ T} \left(\ell_{f_x}^2+\frac{c_{f_y}^2\ell_{g_{xy}}^2}{\mu^2}\right)\frac{\sigma^2 }{B_0} \notag \\
& \quad   +  \frac{960\eta^3\beta_2^2\beta_3^2c_{g_{xy}}^2}{ T}\left(\frac{c_{f_y}^2\ell_{g_{yy}}^2}{\mu^2} + \ell_{f_y}^2\right)\frac{\sigma^2 }{B_0} + \frac{480\eta\beta_3^2c_{g_{xy}}^2}{ T} \left(\frac{c_{f_y}^2\ell_{g_y}^2}{\mu^2} + c_{f_y}^2\right) +  \frac{96}{\eta T}\left(1+  \frac{c_{f_y}^2}{\mu^2}\right)\frac{\sigma^2}{B_0}    \notag \\
& \quad +  \frac{96}{\eta T}\frac{c_{g_{xy}}^2c_{f_y}^2}{\mu^2} + \frac{12}{\eta T}  \frac{\sigma^2}{B_0} +  \frac{72}{\eta T}\left(1+\frac{c_{f_y}^2}{\mu^2}\right)\frac{\sigma^2}{B_0}+ \frac{144}{\eta T} \frac{\ell_{g_y}^2c_{f_y}^2}{\mu^2} +  \frac{144}{\eta T}c_{f_y}^2 \notag \\
& \quad + \frac{96\eta\beta_2^2}{ T} \left(\frac{c_{f_y}^2\ell_{g_{yy}}^2}{\mu^2} + \ell_{f_y}^2\right)\frac{\sigma^2 }{B_0}+ \frac{12\sigma^2}{\alpha_1\eta^2 TKB_0}\left(1+\frac{c_{f_y^2}}{\mu^2}\right)+  \frac{96\beta_2^2\sigma^2}{\alpha_1 TB_0}   \frac{c_{f_y}^2\ell_{g_{xy}}^2}{\mu^2}  \notag \\
& \quad +  \frac{576}{ \alpha_1T} \eta^2\beta_2^2\beta_3^2c_{g_{xy}}^2\left(\frac{c_{f_y}^2\ell_{g_{yy}}^2}{\mu^2} + \ell_{f_y}^2\right)\frac{\sigma^2 }{B_0} + \frac{2}{\beta_1\eta T}244 \frac{\beta_1}{\alpha_1\eta} \eta^2\beta_3^2 c_{g_{xy}}^2\left(\frac{c_{f_y}^2\ell_{g_y}^2}{\mu^2} + c_{f_y}^2\right)  \notag \\
& \quad +  \frac{36}{\eta T} \left(1+\frac{c_{f_y^2}}{\mu^2}\right)\frac{\sigma^2}{B_0} +  \frac{288\eta\beta_2^2}{ T} \frac{c_{f_y}^2\ell_{g_{xy}}^2}{\mu^2} \frac{\sigma^2 }{B_0} +  \frac{1728\eta^3\beta_2^2\beta_3^2c_{g_{xy}}^2\sigma^2}{ TB_0} \left(\frac{c_{f_y}^2\ell_{g_{yy}}^2}{\mu^2} + \ell_{f_y}^2\right) \notag \\
& \quad + \frac{1464\eta \beta_3^2 c_{g_{xy}}^2}{ T}\left(\frac{c_{f_y}^2\ell_{g_y}^2}{\mu^2} + c_{f_y}^2\right) +  \tilde{c}_0 \frac{50 }{3\alpha_2 \eta^2T\mu}  \frac{\sigma^2}{KB_0}  \notag \\
& \quad + \frac{2}{\eta T} \left(3   + \frac{36\beta_2^2}{1-\lambda}\left(\tilde{D}_y+  2916 c_{g_{xy}}^2\left(1+ \frac{1}{\alpha_1 K}\right)   \left(\ell_{f_y}^2+\frac{c_{f_y}^2\ell_{g_{yy}}^2}{\mu^2}\right)\right) \right)\frac{\sigma^2}{B_0} \notag \\
& \quad  + \tilde{c}_1 \frac{400}{\alpha_3 \eta^2T \mu}  \left(1+\frac{c_{f_y^2}}{\mu^2}\right)\frac{\sigma^2}{KB_0}  +    \frac{2}{\eta T}\left(3 +  972\beta_3 c_{g_{xy}}^2\left(1+\frac{1}{\alpha_1 K} \right) \right) \frac{\sigma^2}{B_0}  \ . 
\end{align}

Finally, due to $\beta_3\leq 1-\lambda$ and  $\beta_2\leq 1-\lambda$, we can obtain 

	\begin{align}
		& \quad \frac{1}{T}\sum_{t=0}^{T-1}\mathbb{E}[\| \nabla F(\bar{x}_{t})\|^2] \notag \\
		& \leq \frac{2(F({x}_{0}) - F(x_*))}{\beta_1\eta T} +    \frac{2}{\eta \beta_2T}\tilde{c}_0\mathbb{E}[\|\bar{   {y}}_{0} -    {y}^{*}(\bar{   {x}}_{0})\| ^2 ]  +  \frac{2}{\beta_3\eta T}  \tilde{c}_1\mathbb{E}[\|\bar{ z}_{0} -    {z}^{*}(\bar{   {x}}_{0})\| ^2 ]   \notag \\
		& \quad +\Bigg[ \frac{160\eta\beta_2^2}{ T} \left(\ell_{f_x}^2+\frac{c_{f_y}^2\ell_{g_{xy}}^2}{\mu^2}\right)+ \frac{960\eta^3\beta_2^2\beta_3^2c_{g_{xy}}^2}{ T}\left(\frac{c_{f_y}^2\ell_{g_{yy}}^2}{\mu^2} + \ell_{f_y}^2\right)+ \frac{24}{\eta T} \notag \\
		& \quad \quad + \frac{96\eta\beta_2^2}{ T}\left(\frac{c_{f_y}^2\ell_{g_{yy}}^2}{\mu^2} + \ell_{f_y}^2\right)+\frac{96\beta_2^2}{ \alpha_1T} \frac{c_{f_y}^2\ell_{g_{xy}}^2}{\mu^2}+ \frac{576}{ \alpha_1T} \eta^2\beta_2^2\beta_3^2c_{g_{xy}}^2 \left(\frac{c_{f_y}^2\ell_{g_{yy}}^2}{\mu^2} + \ell_{f_y}^2\right)\notag \\
		& \quad \quad + \frac{288\eta\beta_2^2}{ T}\frac{c_{f_y}^2\ell_{g_{xy}}^2}{\mu^2}+ \frac{1728\eta^3\beta_2^2\beta_3^2c_{g_{xy}}^2}{ T} \left(\frac{c_{f_y}^2\ell_{g_{yy}}^2}{\mu^2} + \ell_{f_y}^2\right)\Bigg]\frac{1}{K} \sum_{k=1}^{K}\| \nabla_{2} g^{(k)}(x_0, y_0)\|^2  \notag \\
		& \quad +   \frac{96}{\eta T}\frac{1}{K} \sum_{k=1}^{K}\mathbb{E}[\|\nabla_{1} { f^{(k)}(x_{0}, y_{0})} \|^2]   +  \frac{160\eta\beta_2^2}{ T} \left(\ell_{f_x}^2+\frac{c_{f_y}^2\ell_{g_{xy}}^2}{\mu^2}\right)\frac{\sigma^2 }{B_0} \notag \\
		& \quad   +  \frac{960\eta^3\beta_2^2\beta_3^2c_{g_{xy}}^2}{ T}\left(\frac{c_{f_y}^2\ell_{g_{yy}}^2}{\mu^2} + \ell_{f_y}^2\right)\frac{\sigma^2 }{B_0} + \frac{480\eta\beta_3^2c_{g_{xy}}^2}{ T} \left(\frac{c_{f_y}^2\ell_{g_y}^2}{\mu^2} + c_{f_y}^2\right) +  \frac{96}{\eta T}\left(1+  \frac{c_{f_y}^2}{\mu^2}\right)\frac{\sigma^2}{B_0}    \notag \\
		& \quad +  \frac{96}{\eta T}\frac{c_{g_{xy}}^2c_{f_y}^2}{\mu^2} + \frac{12}{\eta T}  \frac{\sigma^2}{B_0} +  \frac{72}{\eta T}\left(1+\frac{c_{f_y}^2}{\mu^2}\right)\frac{\sigma^2}{B_0}+ \frac{144}{\eta T} \frac{\ell_{g_y}^2c_{f_y}^2}{\mu^2} +  \frac{144}{\eta T}c_{f_y}^2 \notag \\
		& \quad + \frac{96\eta\beta_2^2}{ T} \left(\frac{c_{f_y}^2\ell_{g_{yy}}^2}{\mu^2} + \ell_{f_y}^2\right)\frac{\sigma^2 }{B_0}+ \frac{12\sigma^2}{\alpha_1\eta^2 TKB_0}\left(1+\frac{c_{f_y^2}}{\mu^2}\right)+  \frac{96\beta_2^2\sigma^2}{\alpha_1 TB_0}   \frac{c_{f_y}^2\ell_{g_{xy}}^2}{\mu^2}  \notag \\
		& \quad +  \frac{576}{ \alpha_1T} \eta^2\beta_2^2\beta_3^2c_{g_{xy}}^2\left(\frac{c_{f_y}^2\ell_{g_{yy}}^2}{\mu^2} + \ell_{f_y}^2\right)\frac{\sigma^2 }{B_0} + \frac{2}{\beta_1\eta T}244 \frac{\beta_1}{\alpha_1\eta} \eta^2\beta_3^2 c_{g_{xy}}^2\left(\frac{c_{f_y}^2\ell_{g_y}^2}{\mu^2} + c_{f_y}^2\right)  \notag \\
		& \quad +  \frac{36}{\eta T} \left(1+\frac{c_{f_y^2}}{\mu^2}\right)\frac{\sigma^2}{B_0} +  \frac{288\eta\beta_2^2}{ T} \frac{c_{f_y}^2\ell_{g_{xy}}^2}{\mu^2} \frac{\sigma^2 }{B_0} +  \frac{1728\eta^3\beta_2^2\beta_3^2c_{g_{xy}}^2\sigma^2}{ TB_0} \left(\frac{c_{f_y}^2\ell_{g_{yy}}^2}{\mu^2} + \ell_{f_y}^2\right) \notag \\
		& \quad + \frac{1464\eta \beta_3^2 c_{g_{xy}}^2}{ T}\left(\frac{c_{f_y}^2\ell_{g_y}^2}{\mu^2} + c_{f_y}^2\right) +  \tilde{c}_0 \frac{50 }{3\alpha_2 \eta^2T\mu}  \frac{\sigma^2}{KB_0}  \notag \\
		& \quad + \frac{2}{\eta T} \left(3   + 36\left(\tilde{D}_y+  2916 c_{g_{xy}}^2\left(1+ \frac{1}{\alpha_1 K}\right)   \left(\ell_{f_y}^2+\frac{c_{f_y}^2\ell_{g_{yy}}^2}{\mu^2}\right)\right) \right)\frac{\sigma^2}{B_0} \notag \\
		& \quad  + \tilde{c}_1 \frac{400}{\alpha_3 \eta^2T \mu}  \left(1+\frac{c_{f_y^2}}{\mu^2}\right)\frac{\sigma^2}{KB_0}  +    \frac{2}{\eta T}\left(3 +  972 c_{g_{xy}}^2\left(1+\frac{1}{\alpha_1 K} \right) \right) \frac{\sigma^2}{B_0}  \notag \\
		& \quad + \left( \tilde{D}_y+2916c_{g_{xy}}^2\left(1+ \frac{1}{\alpha_1 K}\right)  \left(\ell_{f_y}^2+\frac{c_{f_y}^2\ell_{g_{yy}}^2}{\mu^2}\right)\right) 96\alpha^2_2\eta^4\sigma^2 \notag \\
		& \quad  +  12\alpha_3^2\eta^3\left(1 + \frac{c_{f_y}^2}{\mu^2}\right)\sigma^2 +  3888\alpha^2_3\eta^4c_{g_{xy}}^2\left(1+ \frac{1}{\alpha_1 K}\right) \left(1 + \frac{c_{f_y}^2}{\mu^2}\right)\sigma^2 \notag \\
		& \quad + 8 \alpha_1\eta^2 \left(1 + \frac{c_{f_y}^2}{\mu^2}\right)\frac{\sigma^2}{K}  + 24\alpha_1^2\eta^3 \left(1 + \frac{c_{f_y}^2}{\mu^2}\right)\sigma^2  + \tilde{c}_0\frac{100 \alpha_2\eta^2}{3 \mu}   \frac{\sigma^2}{K}  \notag \\
		& \quad +4\alpha_2^2\eta^3  \left( 3   + 36\left(\tilde{D}_y+  2916 c_{g_{xy}}^2\left(1+ \frac{1}{\alpha_1 K}\right)   \left(\ell_{f_y}^2+\frac{c_{f_y}^2\ell_{g_{yy}}^2}{\mu^2}\right)\right)\right)\sigma^2\notag \\
		& \quad  +\tilde{c}_1\frac{800\alpha_3\eta^2}{  \mu}  \left(1 + \frac{c_{f_y}^2}{\mu^2}\right)\frac{\sigma^2}{K} + 8\alpha_3^2\eta^3\left(1 + \frac{c_{f_y}^2}{\mu^2}\right)\left(3 +  972\beta_3 c_{g_{xy}}^2\left(1+\frac{1}{\alpha_1 K} \right) \right)\sigma^2 \notag \\
		& \quad +12\alpha_2^2\eta^3\sigma^2 + 12\alpha_1^2\eta^3\left(1 + \frac{c_{f_y}^2}{\mu^2}\right)\sigma^2\ . 
\end{align}

	According to Eq.~(\ref{eq:hyper-alt}),  Eq.~(\ref{eq:coefficient-alt}), and Eq.~(\ref{eq:coefficient-alt-D}), by setting $\alpha_1=O(\frac{1}{K})$, $\alpha_2=O(\frac{1}{K})$, and $\alpha_3=O(\frac{1}{K})$,  it is easy to know that  $\{\tilde{c}_0, \tilde{c}_1, \tilde{c}_2,  \tilde{c}_3, \tilde{c}_4, \tilde{D}_1, \tilde{D}_2, \tilde{D}_3,  \tilde{D}_4, \tilde{D}_5, \tilde{D}_y \}$ are constant values, which only depend on the constant in Assumptions~\ref{assumption_bi_strong}-\ref{assumption_lower_smooth_vr}. Therefore, we can obtain
\begin{align}
	\small
		&  \frac{1}{T}\sum_{t=0}^{T-1}\mathbb{E}[\| \nabla F(\bar{x}_{t})\|^2] \leq O\left(\frac{1}{\beta_1\eta T}\right)+    O\left(\frac{1}{\beta_2\eta T}\right) +  O\left(\frac{1}{\beta_3\eta T}\right) + O\left(\frac{1}{\eta T} \right)  + O\left(\frac{1}{\eta TB_0} \right)   \notag \\
		&  + O\left(\frac{1}{\alpha_1\eta^2 TKB_0}\right) + O\left(\frac{1}{\alpha_2\eta^2 TKB_0}\right)  + O\left(\frac{1}{\alpha_3\eta^2 TKB_0}\right) +  O\left(\frac{\alpha_1\eta^2}{K}\right)   + O\left(\frac{\alpha_2\eta^2}{K}\right)    \notag \\
		&  + O\left(\frac{\alpha_3\eta^2}{K}\right)   + O\left(\alpha_3^2\eta^3\right) + O\left(\alpha_1^2\eta^3\right)   + O\left(\alpha_2^2\eta^3\right) + O\left(\alpha^2_2\eta^4\right)     \notag \\
		&  + O\left(\frac{\eta\beta_2^2}{T}\right) + O\left(\frac{\eta\beta_3^2}{T}\right)  +   O\left(\frac{\eta^3\beta_2^2\beta_3^2}{T}\right)   + O\left(\frac{\eta\beta_2^2}{TB_0}\right)  + O\left( \frac{\eta^3\beta_2^2\beta_3^2}{ TB_0}\right) \notag \\
		&    + O\left( \frac{\beta_2^2}{ \alpha_1T} \right)   + O\left(\frac{\beta_3^2}{\alpha_1 T}\right)+  O\left( \frac{ \eta^2\beta_2^2\beta_3^2}{ \alpha_1T} \right)  +  O\left( \frac{ \beta_2^2}{ \alpha_1TB_0} \right)   +  O\left( \frac{ \eta^2\beta_2^2\beta_3^2}{ \alpha_1TB_0} \right)  \ . 
\end{align}
Then, according to Eq.~(\ref{eq:hyper-alt}),  Eq.~(\ref{eq:coefficient-alt}), and Eq.~(\ref{eq:coefficient-alt-D}), by setting $\alpha_1=O(\frac{1}{K})$, $\alpha_2=O(\frac{1}{K})$, and $\alpha_3=O(\frac{1}{K})$,  $\beta_1 = O((1-\lambda)^4)$, $\beta_2 = O((1-\lambda)^2)$, $\beta_3 = O((1-\lambda)^4)$, 
$\eta=O(K\epsilon^{1/2})$, the batch size in the first iteration as $B_0=O(1/\epsilon^{1/2})$,  the batch size in other iterations as $B_1=O(1)$,   and  $T=O\left(\frac{1}{K(1-\lambda)^4\epsilon^{3/2}}\right)$,  we can obtain
\begin{align} \label{eq:convergence-upper-bound-alt}
	\small
		& O\left(\frac{1}{\beta_1\eta T}\right) = O\left(\epsilon\right) \ , \quad 
		O\left(\frac{1}{\beta_2\eta T}\right) = O\left((1-\lambda)^2\epsilon\right) \ , \quad 
		O\left(\frac{1}{\beta_3\eta T}\right) = O\left(\epsilon\right) \ , \notag \\
		& O\left(\frac{1}{\eta T}\right) = O\left((1-\lambda)^4\epsilon\right) \ ,  \quad O\left(\frac{1}{\eta TB_0} \right) = O\left((1-\lambda)^4\epsilon^{3/2}\right)   \ ,  \notag \\
		& O\left(\frac{1}{\alpha_1 \eta^2 T KB_0}\right)  = O\left(\frac{(1-\lambda)^4\epsilon^{3/2}}{K}\right)  \ ,  \quad O\left(\frac{1}{\alpha_2 \eta^2 T KB_0}\right)  = O\left(\frac{(1-\lambda)^4\epsilon^{3/2}}{K}\right)  \ ,  \notag \\
		& O\left(\frac{1}{\alpha_3 \eta^2 T KB_0}\right)  = O\left(\frac{(1-\lambda)^4\epsilon^{3/2}}{K}\right)  \ , \notag \\
		& O\left( \frac{\alpha_1\eta^2}{ K}\right)  = O(\epsilon) \ , \quad  O\left( \frac{\alpha_2\eta^2}{ K}\right)  = O(\epsilon) \ , \quad  O\left( \frac{\alpha_3\eta^2}{ K}\right)  = O(\epsilon) \ , \notag \\ 
		& O\left(\alpha_1^2\eta^3\right) = O\left(K\epsilon^{3/2}\right) \ , \quad   O\left(\alpha_2^2\eta^3\right) = O\left(K\epsilon^{3/2}\right) \ , O\left(\alpha_3^2\eta^3\right)   = O\left(K\epsilon^{3/2}\right) \ .  \notag \\
		&   O\left(\frac{\eta\beta_2^2}{T}\right)  = O\left(K^2(1-\lambda)^8\epsilon^{2}\right)  \ , \quad O\left(\frac{\eta\beta_3^2}{T}\right)   = O\left(K^2(1-\lambda)^{12}\epsilon^{2}\right)  \ ,  \notag \\
		&  O\left(\frac{\eta^3\beta_2^2\beta_3^2}{T}\right)   =  O\left(K^4(1-\lambda)^{16}\epsilon^{3}\right)  \ ,   \quad  O\left(\frac{\eta\beta_2^2}{TB_0}\right)   = O\left(K^2(1-\lambda)^{8}\epsilon^{5/2}\right) \  , \notag \\
		& O\left( \frac{\eta^3\beta_2^2\beta_3^2}{ TB_0}\right)   = O\left(K^2(1-\lambda)^{8}\epsilon^{7/2}\right)  \ , \notag \\
		&  O\left(\frac{\beta_3^2}{\alpha_1 T}\right)  = O\left(K^2(1-\lambda)^{12}\epsilon^{3/2}\right) \ ,  \quad  O\left(\frac{\beta_2^2}{\alpha_1 T}\right)  = O\left(K^2(1-\lambda)^{8}\epsilon^{3/2}\right) \ ,  \notag \\
		& O\left( \frac{ \eta^2\beta_2^2\beta_3^2}{ \alpha_1T} \right) = O\left(K^4(1-\lambda)^{16}\epsilon^{5/2}\right) \ ,  \notag \\
		& O\left( \frac{ \beta_2^2}{ \alpha_1TB_0} \right)    = O\left(K^2(1-\lambda)^{8}\epsilon^2\right) \ , \quad O\left( \frac{ \eta^2\beta_2^2\beta_3^2}{ \alpha_1TB_0} \right) = O\left(K^4(1-\lambda)^{16}\epsilon^{3}\right) \ . 
\end{align}
It is worth noting that the last six lines of Eq.~(\ref{eq:convergence-upper-bound-alt}) include the extra terms of DSVRBGD-A's convergence rate compared with that of DSVRBGD-S, and these additional terms only affect the high order of $\epsilon$.

Finally, according to Eq.~(\ref{eq:convergence-upper-bound-alt}),  we have $\frac{1}{T}\sum_{t=0}^{T-1}\mathbb{E} [ \| \nabla F(\bar{x}_{t})\|^2 ]\leq \epsilon$.

\end{proof}

\section{Discussion on the Differences from SPARKLE \cite{zhu2024sparkle} and LoPA \cite{niu2025distributed}}

The key difference between our algorithm and SPARKLE \cite{zhu2024sparkle} and  LoPA \cite{niu2025distributed} lies in two aspects, which are depicted below. 
\begin{itemize}
	\item \textbf{Algorithm Design.} SPARKLE  and LoPA uses  stochastic gradient for updating the lower-level variable and the auxiliary variable, while  using  momentum to update the upper-level variable. Our two algorithms use the variance-reduced gradient estimators to update all variables.  
	\item \textbf{Proof Strategy.} When establishing the convergence rate, a key step is to bound the auxiliary variable $z$ (corresponding to $z$ in SPARKLE and $v$ in LoPA), i.e., the Hessian-inverse-vector product. The strategy used in our paper differs from that in SPARKLE and LoPA.
	\begin{itemize}
		\item Roughly speaking, our paper uses a projection step to guarantee the bound of $z$, while SPARKLE  and LoPA achieve it by constraining the step size. Specifically, in SPARKLE  and LoPA, to bound $\|z_t\|$, they consider its upper bound $\|z_t- z^*(x_t)\| + \|z^*(x_t)\|$. Here, the second term is naturally bounded according to its definition.  But the first term introduces additional constraints for the step size: the step size for $z$ should rely on the gradient variance $\sigma$. For example, on Page 33 of SPARKLE,  the second to last inequality shows how  $\|z_t\|$ is handled,  and the last inequality shows that the step size $\gamma$ relies on $\frac{1}{\sigma^2}$. Similarly, in LoPA, Eq.~(142) uses the same strategy to handle $v_t$. As a result, its step size $\lambda$ relies on $\frac{1}{\sigma^2}$, because $\lambda=O(\alpha)$ (See its Corollary 7) and $\alpha$ relies on $\frac{1}{\sigma^2}$ (See its Eq.~(62)). 
		The reason for this dependence on $\frac{1}{\sigma^2}$ is that $z$ is often coupled with the stochastic Hessian matrix, i.e.,  $\nabla^2_{yy} g(x_t, y_t; \xi)z_{t}$ (See  the second to last inequality in Page 33 of SPARKLE). When bounding $\|\nabla^2_{yy} g(x_t, y_t; \xi)z_{t}\|^2$, it can lead to $\|\nabla^2_{yy} g(x_t, y_t; \xi)z_{t}\|^2 \leq  2\|\nabla^2_{yy} g(x_t, y_t)z_{t}\|^2+ 2\|(\nabla^2_{yy} g(x_t, y_t; \xi)- \nabla^2_{yy} g(x_t, y_t))z_{t}\|^2\leq  2\|\nabla^2_{yy} g(x_t, y_t)z_{t}\|^2+ 2\sigma^2\|z_t\|^2\leq  2\|\nabla^2_{yy} g(x_t, y_t)z_{t}\|^2+ 4\sigma^2\|z_t-z^*(x_t)\|^2 + 4\sigma^2\|z^*(x_t)\|^2$.  Then, when bounding $\|z_t-z^*(x_t)\|^2$, its coefficient $\sigma^2$ will introduce the constraints for its step size. Note that the dependence on $\frac{1}{\sigma^2}$ could  have adverse effects in the heavy-tailed noise setting. In particular, $\sigma^2$ is infinity in the heavy-tailed noise setting, forcing the step size to approach zero.  Our algorithms directly uses a projection step to guarantee the bound of $\|z_t\|$. Therefore, our step size does not rely on $\frac{1}{\sigma^2}$. 
		\item The strategy for handling $\|z_t\|^2$ used in SPARKLE  and LoPA cannot be directly applied to our algorithms. This is because our algorithms use the STORM gradient estimator, which involves both $z_t$ and $z_{t-1}$ in this estimator. In particular, in Eq.~(C32), if using their strategy to handle $z_t$ and $z_{t-1}$, $\|z_t-z^*(x_t)\|^2 $ and $\|z_{t-1}-z^*(x_{t-1})\|^2$ could appear simultaneously in the upper bound, potentially complicating the convergence analysis.  On the contrary, SPARKLE  and LoPA just uses the standard stochastic gradient. Therefore, only $\|z_t-z^*(x_t)\|^2 $ appears in the upper bound. 
	\end{itemize}
\end{itemize}

\section{Discussion on the Differences from the Current Work SLDBO \cite{dong2023single}}
The first version of our paper was submitted to a conference in May 2023 and appeared on arXiv in November 19, 2023. SLDBO appeared on arXiv in November 15, 2023. Our paper significantly differs from SLDBO. Specifically, SLDBO focuses on the {the full gradient}  rather than the stochastic gradient.  Its initial version can only achieve an $O(1/\sqrt{T})$ convergence rate, which is slower than the $1/T$ of the full gradient descent method for nonconvex optimization. In addition, its initial version claims that its convergence rate does not require any heterogeneity assumption.
However, this claim is not grounded.
First, it assumes the upper-level loss function is Lipschitz continuous (See its Assumption 2.1 b)). Therefore, it shares the same strong assumption as DSBO \cite{chen2022decentralized}. As a result, it is very easy to bound the hypergradient (See Eq.~(27) in \cite{dong2023single}). Second,  in each iteration, it projects the variable $y$ to a Euclidean ball, whose diameter is $r>0$. As such,  it actually optimizes a different problem from ours. Specifically, its loss function should be 
\begin{equation} \label{loss_bilevel_dong}
	\begin{aligned}
		& \min_{x\in \mathbb{R}^{d_x}} \frac{1}{K}\sum_{k=1}^{K} f^{(k)}(x, y^*(x)), \quad s.t. \quad   \ y^*(x) =\arg\min_{{y\in \mathcal{D}}} \frac{1}{K}\sum_{k=1}^{K} g^{(k)}(x, y) \ , 
	\end{aligned}
\end{equation}
where ${\mathcal{D}=\{y: \|y\|\leq r\}}$. With such a constraint, the stochastic gradient of the lower-level loss function with respect to $y$ is bounded, i.e., $\|\nabla_{2} g^{(k)}(x, y)\|\leq C$ where $C>0$ is a constant (See the equation below Eq.~(46) in \cite{dong2023single}). However, for a general constrained problem, we cannot guarantee that $\nabla_2 g(x, y^*(x))= 0$, which is a fundamental condition for computing $\frac{\partial y^*(x)}{\partial x}$ in  nonconvex-strongly-convex bilevel optimization problems. But SLDBO still uses $\nabla_2 g(x, y^*(x))= 0$ to compute $\frac{\partial y^*(x)}{\partial x}$. 
In addition, its gradient upper bound depends on a constant $r_x$, which is actually not a constant because $r_x$ relies on the number of iterations. 
The third version of SLDBO, released on April 23, 2024, removed this projection step, but it appeared after we released our proof on February 14, 2024. In our paper, the independence from the heterogeneity assumption is achieved through the gradient tracking technique, which is also adopted in the third version of SLDBO in place of the bounded-gradient assumptions and the projection operation regarding $y$.

\end{document}